\documentclass{article}

\usepackage{microtype}
\usepackage{graphicx}
\usepackage{subfigure}
\usepackage{booktabs} 

\usepackage{amsmath}
\usepackage{amssymb}
\usepackage{algorithmic}
\usepackage{algorithm}
\usepackage{bbm}
\usepackage{booktabs}
\usepackage{todonotes}
\usepackage{url}

\usepackage{hyperref}

\usepackage[accepted]{mlsys2022}

\mlsystitlerunning{Improving the Accuracy-Memory Trade-Off of Random Forests Via Leaf-Refinement}

\begin{document}

\twocolumn[
\mlsystitle{Improving the Accuracy-Memory Trade-Off of Random Forests Via Leaf-Refinement}



\mlsyssetsymbol{equal}{*}

\begin{mlsysauthorlist}
\mlsysauthor{Sebastian Buschjäger}{tu}
\mlsysauthor{Katharina Morik}{tu}
\end{mlsysauthorlist}

\mlsysaffiliation{tu}{Chair for Artificial Intelligence, TU Dortmund University, Germany}

\mlsyscorrespondingauthor{Sebastian Buschjäger}{sebastian.buschjaeger@tu-dortmund.de}

\mlsyskeywords{Ensemble, Ensemble Pruning, Random Forest, Memory Management}

\pdfinfo{
Improving the Accuracy-Memory Trade-Off of Random Forests Via Leaf-Refinement
Sebastian Buschjaeger (sebastian.buschjaeger@tu-dortmund.de) and Katharina Morik 
}

\begin{abstract}

Random Forests (RF) are among the state-of-the-art in many machine learning applications. With the ongoing integration of ML models into everyday life, the deployment and continuous application of models becomes more and more an important issue. Hence, small models which offer good predictive performance but use small amounts of memory are required. 
Ensemble pruning is a standard technique to remove unnecessary classifiers from an ensemble to reduce the overall resource consumption and sometimes even improve the performance of the original ensemble. 
In this paper, we revisit ensemble pruning in the context of `modernly' trained Random Forests where trees are very large. We show that the improvement effects of pruning diminishes for ensembles of large trees but that pruning has an overall better accuracy-memory trade-off than RF. However, pruning does not offer fine-grained control over this trade-off because it removes entire trees from the ensemble.
To further improve the accuracy-memory trade-off we present a simple, yet surprisingly effective algorithm that refines the predictions in the leaf nodes in the forest via stochastic gradient descent. We evaluate our method against 7 state-of-the-art pruning methods and show that our method outperforms the other methods on 11 of 16 datasets with a statistically significant better accuracy-memory trade-off compared to most methods. We conclude our experimental evaluation with a case study showing that our method can be applied in a real-world setting.
\end{abstract}
]

\printAffiliationsAndNotice{} 

\section{Introduction}

Ensemble algorithms offer state-of-the-art performance in many applications and often outperform single classifiers by a large margin. With the ongoing integration of embedded systems and machine learning models into our everyday life, e.g in the form of the Internet of Things, the hardware platforms which execute ensembles must also be taken into account when training ensembles. 

From a hardware perspective, a small ensemble with minimal execution time and a small memory footprint is desired. Similar, learning theory indicates that ensembles of small models should generalize better which would make them ideal candidates for small, resource constraint devices\cite{Koltchinskii/Panchenko/2002,Cortes/etal/2014a}. Practical problems, on the other hand, often require ensembles of complex base learners to achieve good results. For some ensembling techniques such as Random Forest it is even desired that individual trees are as large as possible leading to overall large ensembles \cite{Breiman/2000,biau/2012,denil/2014,biau/Scornet/2016}.
Ensemble pruning is a standard technique for implementing ensembles on small devices  \cite{tsoumakas/etal/2009,zhang2006ensemble} by removing unnecessary classifiers from the ensemble. Remarkably, this removal can sometimes lead to a \emph{better} predictive performance \cite{margineantu/dietterich/1997,martinez/suarez/2006,li2012diversity}. In this paper, we revisit ensemble pruning and show that this improvement effect does not carry over to the modern-style of training individual trees as large as possible in Random Forests. Maybe even more frustrating, ensemble pruning does not seem to be necessary anymore to achieve the best accuracy if the original forest has a sufficient amount of large trees. If, however, one also considers the memory requirements of the individual trees the situation changes. We argue that, from a hardware perspective, the trade-off between memory and accuracy is what really matters. Although a Random Forest might produce a good model it might not be possible to deploy it onto a small device due to its memory requirements. As shown later, the best performing RF models are often larger than $5-10$ MB (see e.g. Fig. \ref{fig:auc_eeg} and Fig. \ref{fig:auc_chess}) while most available microcontroller units (MCU) only offer a few KB to a few MB of memory as depicted in Table \ref{tab:MUC_memory}. Hence, to deploy RF onto these small devices we require a good algorithm which gives accurate models for a variety of different memory constraints. 



\begin{table}[h]
\resizebox{\columnwidth}{!}{%
\begin{tabular}{@{}lllll@{}}
\toprule
MCU                        & Flash   & (S)RAM & Power \\ \midrule
Arduino Uno    & 32KB    & 2KB    & 12mA  \\
Arduino Mega   & 256KB   & 8KB    & 6mA   \\
Arduino Nano   & 26–32KB & 1–2KB  & 6mA   \\
STM32L0         & 192KB   & 20KB   & 7mA   \\
Arduino MKR1000  & 256KB   & 32KB   & 4mA   \\
Arduino Due    & 512KB   & 96KB   & 50mA  \\
STM32F2          & 1MB     & 128KB  & 21mA  \\
STM32F4         & 2MB     & 384KB  & 50mA  \\ \bottomrule
\end{tabular}
}
\caption{Available memory on different microcontroller units. Excerpt from \cite{branco/etal/2019}.}
\label{tab:MUC_memory}
\end{table}

We directly optimize the accuracy-memory trade-off by introducing a technique called leaf-refinement. Leaf-Refinement is a simple, but surprisingly effective method, which, instead of removing trees from the ensemble, further refines the predictions of small ensembles using gradient-descent. This way, we can refine any given tree-ensemble to optimize its accuracy thereby maximizing the accuracy-memory trade-off.
Our contributions are as follows:

\begin{itemize}
    \item \textbf{Revisiting ensemble pruning:} We revisit ensemble pruning in the context of modernly trained Random Forests in which individual trees are typically large. We show that pruning a Random Forest can improve the accuracy if individual trees are small, but this effect becomes neglectable for larger trees. Moreover, if we are only interested in the most accurate models where memory is no constraint we can simply train unpruned Random Forests which yields comparable results without the need for pruning.
    \item \textbf{Random Forest with Leaf Refinement:} We show that pruning exhibits a better accuracy-memory trade-off than RF does. To further optimize this trade-off we present a simple, yet surprisingly effective gradient-descent based algorithm called leaf-refinement (RF-LR) which refines the predictions of a pre-trained Random Forest. 
    \item \textbf{Experiments:} We show the performance of our algorithm on 16 datasets and compare it against 7 state-of-the-art pruning methods. We show that RF-LR outperforms the other methods on 11 of 16 datasets with a statistically significant better accuracy-memory trade-off compared to most methods. We conclude our experimental evaluation with a case study showing that our method can be applied on a real-world setting.
\end{itemize}

The paper is organized as the following. Section \ref{sec:rel-work} presents our notation and related work. In Section \ref{sec:RevisitingReducedErrorPruning} we revisit ensemble pruning in the context of `modern' Random Forests, whereas section \ref{sec:MaxMemTradeOff} discusses how to improve the accuracy-memory trade-off without ensemble pruning. In section \ref{sec:experiments} we experimentally evaluate our method and in section \ref{sec:conclusion} we conclude the paper. 

\section{Background and Notation}
\label{sec:rel-work}

We consider a supervised learning setting, in which we assume that training and test points are drawn i.i.d. according to some distribution $\mathcal D$ over the input space $\mathcal X$ and labels $\mathcal Y$. We assume that we have given a trained ensemble with $M$ classifiers $h_i \in \mathcal H$ of the following form:
\begin{equation}
f(x) = \frac{1}{M}\sum_{i=1}^M h_i(x)
\end{equation}
Additionally, we have given a labeled pruning sample $\mathcal{S} = \{(x_i,y_i)|i=1,\dots,N\}$ where $x_i \in \mathcal X \subseteq \mathbb R^d$ is a $d$-dimensional feature-vector and $y_i\in \mathcal Y \subseteq \mathbb R^C$ is the corresponding target vector. This sample can either be the original training data used to train $f$ or another pruning set not related to the training or test data. For classification problems with $C \ge 2$ classes we encode each label as a one-hot vector $y = (0,\dots,0,1,0,\dots,0)$ which contains a `$1$' at coordinate $c$ for label $c \in \{0,\dots,C-1\}$; for regression problems we have $C=1$ and $\mathcal Y = \mathbb R$. In this paper, we will focus on classification problems, but note that our approach is directly applicable for regression tasks, as well. Moreover we will focus on tree ensembles and specifically Random Forests, but note that most of our discussion directly translates to other tree ensembles such as Bagging \cite{breiman/1996}, ExtraTrees \cite{geurts/etal/2006}, Random Subspaces \cite{ho/1998} or Random Patches \cite{louppe/Geurts/2012}. 

The goal of ensemble pruning is to select a subset of $K$ classifier from $f$ which forms a small and accurate sub-ensemble. Formally, each classifier $h_i$ receives a corresponding pruning weight $w_i \in \{0,1\}$. Let
\begin{equation}
L(w) = \frac{1}{N}\sum_{(x,y) \in \mathcal S} \ell \left(\sum_{i=1}^M w_i h_i(x),y\right)
\end{equation}
be a loss function and let $\lVert w \rVert_0 = \sum_{i=1}^M 1\{w_i > 0\}$ be the $l_0$ norm which counts the number of nonzero entries in the weight vector $w = (w_1,w_2,\dots,w_M)$. Then the ensemble pruning problem is defined as:
\begin{equation}
    \label{eq:pruning}
    \arg\min_{w\in\{0,1\}^M} L(w) \text{~st.~} \lVert w \rVert_0 = K
\end{equation}



Many effective ensemble pruning methods have been proposed in literature. These methods usually differ in the specific loss function used to measure the performance of a sub-ensemble and the way this loss is minimized. Tsoumakas et al. give in \cite{tsoumakas/etal/2009} a detailed 
taxonomy of pruning methods which was later expanded in \cite{zhou2012ensemble} to which we refer interested readers.
Early works on ensemble pruning focus on ranking-based approaches which assign a rank to each classifier depending on their individual performance and then pick the top $K$ classifier from to the ranking. One of the first pruning methods in this direction was due to Margineantu and Dietterich which proposed to use the Cohen-Kappa statistic to rate the effectiveness of each classifier in \cite{margineantu/dietterich/1997}. More recent approaches also incorporate the ensemble's diversity into the selection such as \cite{lu2010ensemble,Jiang/etal/2017,guo2018margin}.
As an alternative to a simple ranking, Mixed Quadratic Integer Programming (MQIP) has also been proposed. Originally this approach was proposed by Zhang et al. in \cite{zhang2006ensemble} which uses the pairwise errors of each classifier to formulate an MQIP. Cavalcanti et al. expand on this idea in \cite{cavalcanti2016combining} which combines 5 different measures into the MQIP.
A third branch of pruning considers the clustering of ensemble members to promote diversity. 
The main idea is to cluster the classifiers into (diverse) groups and then to select one representative from each group \cite{giacinto/etal/2000,lazarevic/2001/effective}. 
Last, ordering-based pruning has been proposed. Ordering-based approaches order all ensemble members according to their overall contribution to the (sub-)ensemble and then pick the top $K$ classifier from this list. 
This approach was also first considered in \cite{margineantu/dietterich/1997} which proposed to greedily minimize the overall ensemble error. A series of works by Mart{\'\i}nez-Mu{\~n}oz, Su{\'a}rez and others \cite{martinez/suarez/2004,martinez/suarez/2006,martinez/etal/2008} add upon this work proposing different error measures. More recently, theoretical insights from PAC theory and the bias-variance decomposition were also transformed into greedy pruning approaches  \cite{li2012diversity,Jiang/etal/2017}.

Looking beyond ensemble pruning there are numerous, orthogonal methods to deploy ensembles to small devices. First, `classic' decision tree pruning algorithms (e.g. minimal cost complexity pruning or sample complexity pruning) already reduce the size of DTs while offering a better accuracy (c.f. \cite{Barros/etal/2015}). Second, in the context of model compression (see e.g. \cite{choudhary/etal/2020} for an overview) specific models such as Bonsai \cite{kumar/etal/2017} or Decision Jungles \cite{shotton/etal/2013} aim to find smaller tree ensembles already during training. Last, the optimal implementation of tree ensembles has also been studied, e.g. by optimizing the  memory layout for caching \cite{Buschjaeger/etal/2018} or changing the tree traversal to utilize SIMD instructions \cite{ye/etal/2018}. We find that all these methods are orthogonal to our approach and that they can be freely combined with one another, e.g. we may train a decision jungle, then perform ensemble pruning or leaf-refinement on it and finally find the optimal memory layout of the trees in the jungle for the best deployment.

\section{Revisiting Ensemble Pruning}
\label{sec:RevisitingReducedErrorPruning}

Before we discuss our method we first want to revisit Reduced Error Pruning (RE, \cite{margineantu/dietterich/1997}) and repeat some experiments performed with it. RE pruning is arguably one of the simplest pruning algorithms but often offers competitive performance. RE is a ordering-based pruning method. It starts with an empty ensemble and iteratively adds that tree which minimizes the overall ensemble error the most until $K$ members have been selected. Algorithm \ref{fig:RE} depicts this approach where $L$ is the $0-1$ loss and $e_i$ denotes the unit vector with a `1' entry at position $i$.

\begin{algorithm}[t]
  \begin{algorithmic}[1]
    \STATE{$w \gets (0,\dots,0)$}
    \STATE{$i \gets \arg\min\{L(w + \vec e_i) | i=1\dots,M\}$}
    \STATE{$w \gets w + \vec e_i$}
    \FOR{$j = 1, \dots, K-1$}
        \STATE{$i \gets \arg\min\{ L(w + \vec e_i) | i=1\dots,M, w_i \not= 1\}$}   
        \STATE{$w \gets w + \vec e_i$}
    \ENDFOR
  \end{algorithmic}
  \caption{Reduced Error Pruning (RE).}
  \label{fig:RE}
\end{algorithm}

We will now perform experiments in the spirit of \cite{martinez/suarez/2006}, but adapt a more modern approach to training the base ensembles. In the original experiments, the authors show that when pruning a Bagging Ensemble of 200 \emph{pruned} CART trees, that RE (among other methods) achieves a better accuracy with fewer trees compared to the original ensemble. 
This result has been empirically reproduced in various contexts (see e.g. \cite{margineantu/dietterich/1997,zhou/etal/2002,zhou2012ensemble}) and has been formalized in the \emph{Many-Could-Be-Better-Than-All-}Theorem \cite{zhou/etal/2002}. It shows that the error of an ensemble excluding the $k-$th classifier can be smaller than the error of the original ensemble if the bias $C_{k,k}$ is larger than its variance wrt. to the ensemble:
\begin{align}
\sum_{i=1, i \not= k }^M \sum_{j=1, i \not= k}^M \frac{C_{i,j}}{(M-1)^2} &\le \sum_{i=1}^M\sum_{h=1}^M\frac{C_{i,j}}{M^2} \\
\Leftrightarrow \quad  -2 \sum_{i=1, i \not= k }^M C_{i,k} &\le C_{k,k}
\end{align}
where
\begin{align}
C_{k,k} &= \mathbb E_{x,y \sim \mathcal D}\left[ (h_k(x) - y)^2 \right] \\
C_{k,i} &= \mathbb E_{x,y \sim \mathcal D}\left[ (h_k(x) - y) (h_i(x) - y) \right]
\end{align}

Recall that the bias of a DT rapidly decreases while the variance increases wrt. to the size of the tree \cite{domingos/etal/2000}. The original experiment used \emph{pruned} decision trees whereas the today's accepted standard is to train trees as large as possible for minimal errors (see \cite{Breiman/2000,biau/2012,denil/2014,biau/Scornet/2016} for more formal arguments on this). Hence, it is conceivable that ensemble pruning does not have the same beneficial effect on `modern' Random Forests compared to RF-like ensembles trained 20 years ago. We will now investigate this hypothesis experimentally. As an example we will consider the EEG dataset
which has $14~980$ datapoints with $14$ attributes and two classes (details for each dataset can be found in the appendix). By today's standards this dataset is small to medium size which allows us to quickly train and evaluate different configurations but it is roughly two times larger than the biggest dataset used in original experiments.
We perform experiments as follows: Oshiro et al. showed in \cite{oshiro/etal/2012} empirically on a variety of datasets that the prediction of a RF stabilizes between $128$ and $256$ trees and adding more trees to the ensemble does not yield significantly better results. Hence, we train the `base' Random Forests with $M = 256$ trees. To control the individual errors of trees we set the maximum number of leaf nodes $n_l$ to values between $n_l \in \{64,128,256,512,1024\}$. For ensemble pruning we use RE which is tasked to select $K \in \{2,4,8,16,32,64,128,256\}$ trees from the original RF. We compare this against a smaller RF with $K \in \{2,4,8,16,32,64,128,256\}$ trees, so that we recover the original RF for $K=M=256$ on both cases. For RE we use the training data as pruning set. Experiments with a dedicated pruning set can be found in the appendix. Figure \ref{fig:revisited_eeg} shows the average accuracy over the size of the ensemble for a $5$-fold cross-validation. The dashed lines depict the smaller RF and solid lines are the corresponding pruned ensemble. As expected, we find that ensemble pruning significantly improves the accuracy when smaller trees with $64-256$ leaf nodes are used. Moreover, the performance of the pruned forests approaches the performance of the original forests when more and more trees are added much like in the original experiments. However, the improvement in accuracy becomes negligible for trees with up to $1024$ leaf nodes. Here, the accuracy of the pruned and the unpruned forest are near identical for any given number of trees. Maybe even worse, if we are only interested in the most accurate model then there is no reason to prune the ensemble as an unpruned Random Forest already seems to achieves the best performance. 

\begin{figure}
\centering
\includegraphics[width=\columnwidth, keepaspectratio]{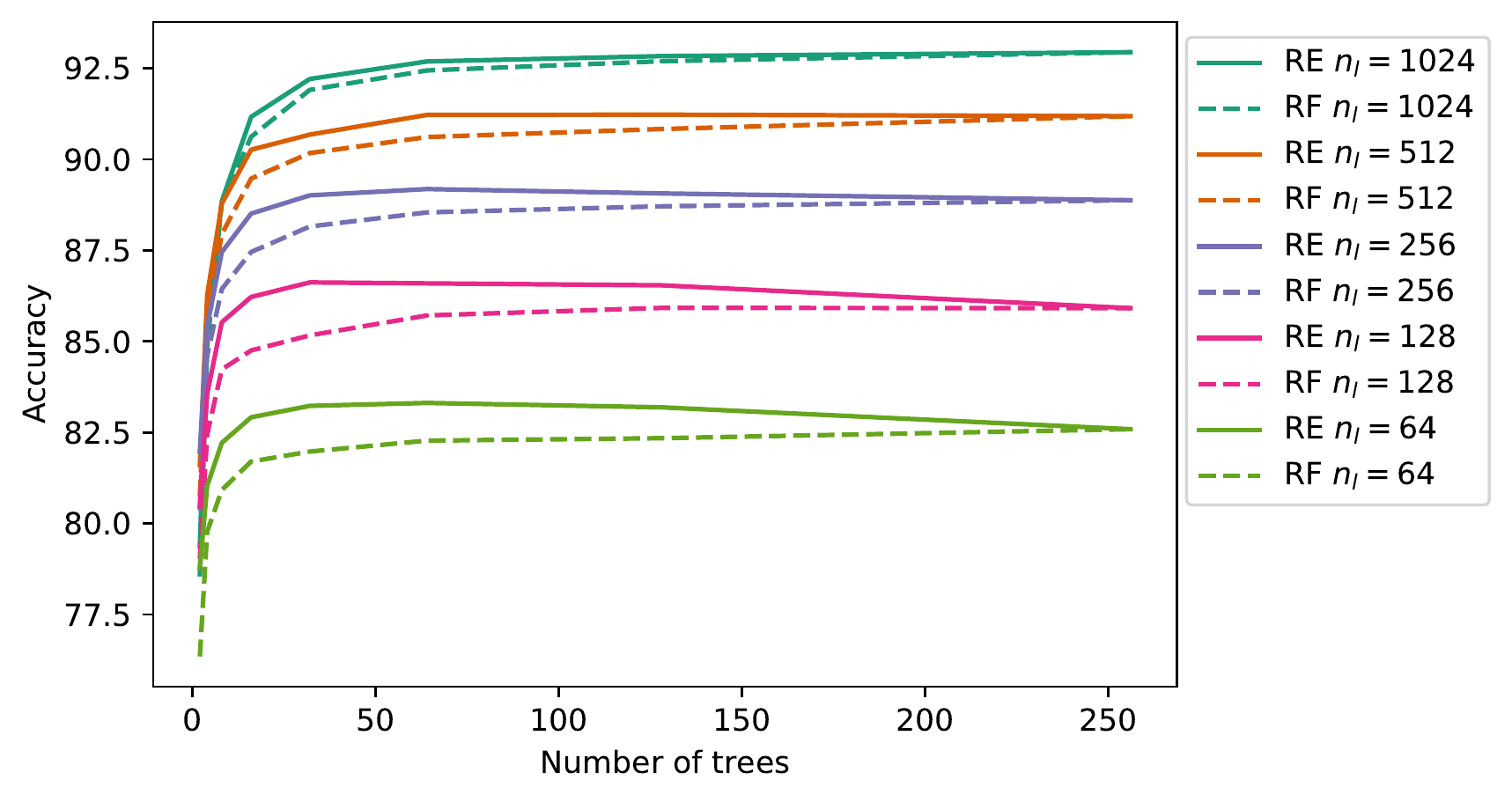}
\caption{5-fold cross-validation accuracy over the number of members in the ensemble for different $n_l$ parameters on the EEG dataset. Dashed lines depict the small RF and solid lines are the pruned ensemble via Reduced Error pruning. Best viewed in color.}
\label{fig:revisited_eeg}
\end{figure}

We acknowledge that this experiment is one-sided because we only use Reduced Error Pruning -- a nearly 25 year old method -- for comparison. Maybe the problem simply lies in RE itself and not pruning in general? To verify this hypothesis we also repeated the above experiment with 6 additional pruning algorithms from the three different categories. In total we compare two ranking-based methods namely IE \cite{Jiang/etal/2017} and IC \cite{lu2010ensemble}; the three ordering-based methods RE  \cite{margineantu/dietterich/1997}, DREP \cite{li2012diversity} and COMP \cite{martinez/suarez/2004} and the two clustering-based pruning CA \cite{lazarevic/2001/effective} and LMD \cite{giacinto/etal/2000}. We also experimented with MQIP pruning methods \cite{zhang2006ensemble,cavalcanti2016combining}, but unfortunately the MQIP solver (in our case Gurobi\footnote{\url{https://www.gurobi.com/}}) used during experiments would frequently fail or time-out. Thus we decided to not include any MQIP pruning methods in our evaluation.

\begin{table}[]
    \centering
    \resizebox{\columnwidth}{!}{%
    \begin{tabular}{llrrrrrrr}
    \toprule
    $n_l$ & K &   COMP &   DREP &     IC &     IE &    LMD &     RE &     RF \\
    \midrule
    64   & 8   &  81.86 &  80.79 &  81.71 &  81.34 &  81.46 &  \textbf{82.22} &  80.92 \\
         & 32  &  83.09 &  82.33 &  83.10 &  82.22 &  82.60 &  \textbf{83.23} &  81.98 \\
         & 128 &  83.17 &  82.38 &  83.10 &  82.49 &  82.87 &  \textbf{83.19} &  82.34\vspace{0.1cm} \\
    128  & 8   &  85.12 &  84.08 &  84.69 &  84.81 &  83.85 &  \textbf{85.53} &  84.23 \\
         & 32  &  86.34 &  85.49 &  86.38 &  85.76 &  85.65 &  \textbf{86.62} &  85.17 \\
         & 128 &  86.40 &  85.75 &  86.27 &  86.00 &  86.03 &  \textbf{86.54} &  85.92\vspace{0.1cm} \\
    256  & 8   &  87.37 &  86.24 &  87.14 &  87.16 &  86.36 &  \textbf{87.46} &  86.44 \\
         & 32  &  88.97 &  88.02 &  \textbf{89.07} &  88.70 &  88.37 &  89.01 &  88.16 \\
         & 128 &  88.97 &  88.70 &  \textbf{89.15} &  88.97 &  88.77 &  89.07 &  88.71\vspace{0.1cm} \\
    512  & 8   &  88.36 &  88.51 &  \textbf{88.96} &  88.78 &  87.44 &  88.79 &  87.95 \\
         & 32  &  91.11 &  90.30 &  \textbf{91.34} &  90.67 &  90.37 &  90.68 &  90.17 \\
         & 128 &  91.22 &  90.91 &  \textbf{91.41} &  91.30 &  90.87 &  91.23 &  90.83\vspace{0.1cm} \\
    1024 & 8   &  89.45 &  89.26 &  89.51 &  \textbf{89.70} &  88.30 &  88.85 &  88.83 \\
         & 32  &  92.25 &  92.30 &  \textbf{92.64} &  92.60 &  91.82 &  92.21 &  91.91 \\
         & 128 &  92.85 &  92.85 &  \textbf{93.17} &  92.98 &  92.70 &  92.84 &  92.70 \\
    \bottomrule
    \end{tabular}
    }
    
    \caption{5-fold cross-validation accuracy over the number of members $K\in\{8,32,128\}$ in the ensemble for different $n_l$ parameters and different methods on the EEG dataset. Rounded to the second decimal digit. Larger is better. The best method is depicted in bold.}
    \label{tab:revisited_eeg_raw}
\end{table}

Table \ref{tab:revisited_eeg_raw} shows the result of this experiment. For space reasons we only depict results for $K\in\{8,32,128\}$. As expected, all pruning methods manage to improve the performance of the original RF for smaller $n_l \le 256$ and keep this advantage to some degree for larger $n_l$. However, this advantage becomes smaller and smaller for larger $n_l$ until it is virtually non-existent for $n_l = 1024$ and the accuracies are near identical. Again, as expected setting $n_l$ to larger values leads to the overall best accuracy. 

For presentational purposes we highlighted this experiment on the EEG dataset, but we found that this behavior seems to hold universally across the other 15 datasets we experimented with. The detailed results for these experiments are given in the appendix. While the specific curves would differ we always found that the performance of a well-trained forest and its pruned counterpart would nearly match once the individual trees become large enough. For more plots with experiments on other datasets and other `base' ensembles please consult the appendix. 

\section{Improving the accuracy-memory trade-off of RF}
\label{sec:MaxMemTradeOff}

Clearly, the previous section shows that we cannot expect the accuracy of a pruned forest to improve much upon the performance of a well-trained Random Forest. On the one hand, this is a clear argument in favor of Random Forests -- why should we prune a pre-trained forest if we can directly train a similar forest in the first place? On the other hand, pruning shows clear superior performance for smaller $n_l$ compared to RF. While pruning and RF both converge against a very similar maximum accuracy, pruning shows a better trade-off between the model size (controlled by $n_l$ and $K$) and the accuracy. 

We argue that, from a hardware perspective, this trade-off is what really matters
and a good algorithm should produce accurate models for a variety of different model sizes. Ensemble pruning improves this trade-off by removing unnecessary trees from the ensemble thereby reducing the memory consumption while keeping (or improving) its predictive power. But, the removal of entire trees does not offer a very fine-grained control over this trade-off. 
For example, it could be better to train a large forest with many, but comparably small trees instead of having one small forest of large trees. Hence, we propose to directly evaluate the accuracy-memory trade-off and to optimize towards it. 

To do so, we present a simply and surprisingly effective method which refines the predictions of a given forest with Stochastic Gradient Descent (SGD). Our method trains a small initial Random Forest (e.g. by using small values for $n_l$ and $M$) and then refines the predictions of the individual trees to improve the overall performance: Recall that DTs use a series of axis-aligned splits of the form $\mathbbm{1}\{x_i \le t\}$ and $\mathbbm{1}\{x_i > t\}$ where $i$ is a pre-computed feature index and $t$ is a pre-computed threshold to determine the leaf nodes. Let $s_l(x) \colon \mathcal X \to \{0,1\}$ be the series of splits which is `1' if $x$ belongs to leaf $l$ and `0' if not, then the prediction of a tree is given by
\begin{equation}
h_i(x) = \sum_{l = 1}^{L_i} \widehat y_{i,l} s_{i,l}(x)
\end{equation}
where $\widehat y_{i,l} \in \mathbb R^C$ is the (constant) prediction value of leaf $l$ and $L_i$ is the total number of leaves in tree $h_i$. Let $\theta_i$ be the parameter vector of tree $h_i$ (e.g. containing split values, feature indices and leaf-predictions) and let $\theta = (\theta_1,\dots,\theta_M)$ be the parameter vector of the entire ensemble $f_{\theta}$. Then our goal is to solve
\begin{equation}
\arg\min_{\theta} \frac{1}{N}\sum_{(x,y) \in \mathcal S} \ell \left(f_{\theta}(x),y\right)
\end{equation}
for a given loss $\ell$. We propose to minimize this objective via stochastic gradient-descent. SGD is an iterative algorithm which takes a small step into the negative direction of the gradient in each iteration $t$ by using an estimation of the true gradient
\begin{equation}
\label{eq:sgd}
\theta^{t+1} \gets \theta^t - \alpha^t g_{\mathcal B}(\theta^t) 
\end{equation}
where 
\begin{equation}
g_{\mathcal B}(\theta^t) = \nabla_{\theta^t} \left( \sum_{(x,y) \in \mathcal B} \ell \left(f_{\theta^t}(x),y\right) \right)
\end{equation}
is the gradient of $\ell$ wrt. to $\theta^t$ computed on a mini-batch $\mathcal B$.

Unfortunately, the axis-aligned splits of a DT are not differentiable and thus it is difficult to refine them further with gradient-based approaches. However, the leaf predictions $\widehat y_{i,l}$ are simple constants that can easily be updated via SGD. 
Formally, we use $\theta_i = (\widehat y_{i,1},\widehat y_{i,2}, \dots)$ leading to
\begin{equation}
\label{eq:grad}
g_{\mathcal B}(\theta^t_i) = \frac{1}{|\mathcal B|} \left( \sum_{(x,y)\in\mathcal B} \frac{\partial \ell(f_{\theta^t}(x), y)}{\partial f_{\theta^t}(x)} w_i s_{i,l}(x)\right)_{l=1,2,\dots,L_i}
\end{equation}

Algorithm \ref{fig:LR} summarizes this approach. First, in \texttt{get\_forest} a forest with $K$ trees each containing at most $n_l$ leaf nodes is loaded. This forest can either be a pre-trained forest with $M$ trees from which we randomly sample $K$ trees or we may train an entirely new forest with $K$ trees directly. Once the forest has been obtained SGD is performed over the leaf-predictions of each tree using the step-size $\alpha^t \in \mathbb R_+$ to minimize the given loss $\ell$.

\begin{algorithm}
	\begin{algorithmic}[1]
	\STATE \COMMENT{Load forest and use constant weights}
	\STATE{$h \gets \texttt{get\_forest}(K, n_l)$} 
	\STATE{$w \gets (1/K, \dots, 1/K)$} 
	\STATE \COMMENT{Init. leaf predictions}
	\FOR{$i=1,\dots,K$} 
	    \STATE{$\theta_i \gets (\widehat y_{i,1},\widehat y_{i,2}, \dots)$}
	\ENDFOR
	\STATE\COMMENT{Perform SGD using Eq. \ref{eq:sgd} + Eq. \ref{eq:grad}}
	\FOR{receive batch $\mathcal B$} 
      \FOR{$i=1,\dots,K$}
        \STATE{$\theta^t_i \gets \theta^t_i - \alpha^t g_{\mathcal B}(\theta^t_i)$} 
      \ENDFOR
    \ENDFOR
	\end{algorithmic}
	\caption{RF with Leaf-Refinement (RF-LR).}
	\label{fig:LR}
\end{algorithm}

Leaf-Refinement is a flexible technique and can be used in combination with any tree ensemble such as Bagging \cite{breiman/1996}, ExtraTrees \cite{geurts/etal/2006}, Random Subspaces \cite{ho/1998} or Random Patches \cite{louppe/Geurts/2012}. Moreover, we can also refine the individual weights $w_i$ of the trees via SGD, although we did not find a meaningful improvement optimizing the weights and leafs simultaneously in our pre-experiments. For simplicity we will only focus on leaf-refinement in this paper without optimizing the individual weights and leave this for future research.

\section{Experiments}
\label{sec:experiments}

In this section we experimentally evaluate our method and compare its accuracy-memory trade-off with regular RF and pruned RF. As argued before, our main concern is the final model size as it determines the resource consumption, runtime, and energy of the model application during deployment\cite{buschjaeger/morik/2017,Buschjaeger/etal/2018}. The model size is computed as follows: A baseline implementation of DTs stores each node in an array and iterates over it \cite{Buschjaeger/etal/2018}. Each node inside the array requires a pointer to the left / right child (8 bytes in total), a boolean flag if it is a leaf-node (1 byte), the feature index as well as the threshold to compare the feature against (8 bytes). Last, entries for the class probabilities are required for the leaf nodes (4 bytes per class). Thus, in total, a single node requires $17+4\cdot C$ Bytes per node which we sum over all nodes in the entire ensemble. 

We follow a similar experimental protocol as before: As earlier we train various Random Forests with $M = 256$ trees using $n_l \in \{64, 128, 256, 512, 1024\}$. Again, we compare the aforementioned pruning methods COMP, DREP, IC, IE, LMD and RE with our leaf-refinement method (RF-LR) as well as a random selection of trees from the RF. Since our method shares some overlap with gradient boosted trees (GB, \cite{friedman/etal/2001}) we also include these in our evaluation. Each pruning method is tasked to select $K \in \{8, 16, 32, 64, 128\}$ trees from the `base' forest. For DREP, we additionally varied $\rho \in \{0.25,0.3,0.35,0.4,0.45,0.5\}$. For GB we use the deviance loss and train $\{8, 16, 32, 64, 128\}$ trees with the different $n_l$ values. For RF-LR we randomly sample $K \in \{8, 16, 32, 64, 128\}$ trees from the given forest and perform $50$ epochs\footnote{In one epoch we iterate once over the entire dataset.} of SGD with a constant step size $\alpha = 0.1$ and a batch size of $128$. We experimented with the mean-squared error (MSE) and the cross-entropy loss for minimization, but could not find meaningful differences between both losses. Hence, for these experiments we focus on the MSE loss. In all experiments we perform a $5$-fold cross validation except when the dataset comes with a given train/test split. We use the training set for both, training the initial forest and pruning it. For a fair comparison we made sure that each method receives the same forest in each cross-validation run. In all experiments, we use minimal pre-processing and encode categorical features as one-hot encoding. The base ensembles have been trained with Scikit-Learn \cite{Pedregosa/etal/2001} and the code for our experiments and all pruning methods are included in this submission. 
We implemented all pruning algorithm in a Python package for other researchers called PyPruning which is available under \url{https://github.com/sbuschjaeger/PyPruning}. The code for the experiments in this paper are available under \url{https://github.com/sbuschjaeger/leaf-refinement-experiments}. In total we performed $8~960$ experiments on $16$ different datasets which are detailed in the appendix. Additionally, more experiments with different `base' ensembles and a dedicated pruning set are shown in the appendix.

\subsection{Qualitative Analysis}

We are interested in the most accurate models with the smallest memory consumption. Clearly these two metrics can contradict each other. For a fair comparison we therefore use the best parameter configuration of each method across both dimensions. More specifically, we compute the Pareto front of each method which contains those parameter configurations which are not dominated across one or more dimensions. For space reasons we start with a qualitative analysis and focus the EEG and the chess dataset as they represent distinct behaviors we found during our experiments. 

\begin{figure}
\centering
\includegraphics[width=\columnwidth, keepaspectratio]{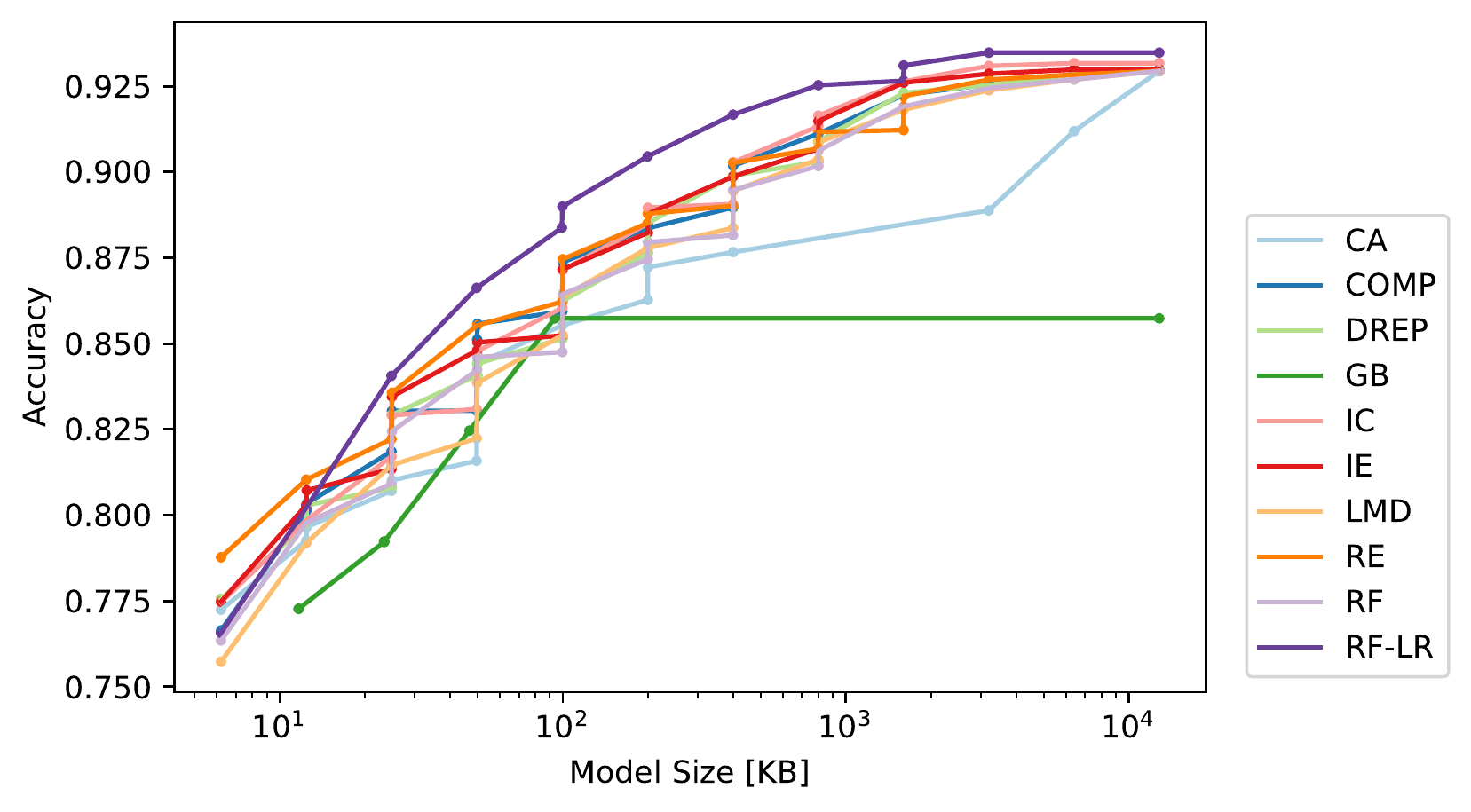}
\caption{5-fold cross-validation accuracy over the size of the ensemble on the EEG dataset. Single points are the individual parameter configurations whereas the solid line depicts the corresponding Pareto front. Best viewed in color.}
\label{fig:auc_eeg}
\end{figure}

Figure \ref{fig:auc_eeg} shows the results on the EEG dataset. As before, the accuracy ranges from $75 \%$ to $92.5 \%$ and the model size ranges from a few KB to roughly $12$MB (note the logarithmic scale on the x-axis). As before, larger models seem to generally perform better and all models seem to converge against a similar solution, expect GB which is stuck around $85\%$ accuracy. In the range of $0-4000$ KB, however, there are larger differences. For example, for roughly $1000$ KB, CA performs sub-optimal only reaching an accuracy around $87.5 \%$ whereas the other methods all seem to have a similar performance around $90 \%$ except RF-LR which has an accuracy around $91\%$. For smaller model sizes below $4000$ KB RF-LR seems to be the clear winner offering roughly up to $1 \%$ more accuracy compared to the other methods. Moreover, it shows a better overall accuracy-memory trade-off. 

\begin{figure}
\centering
\includegraphics[width=\columnwidth, keepaspectratio]{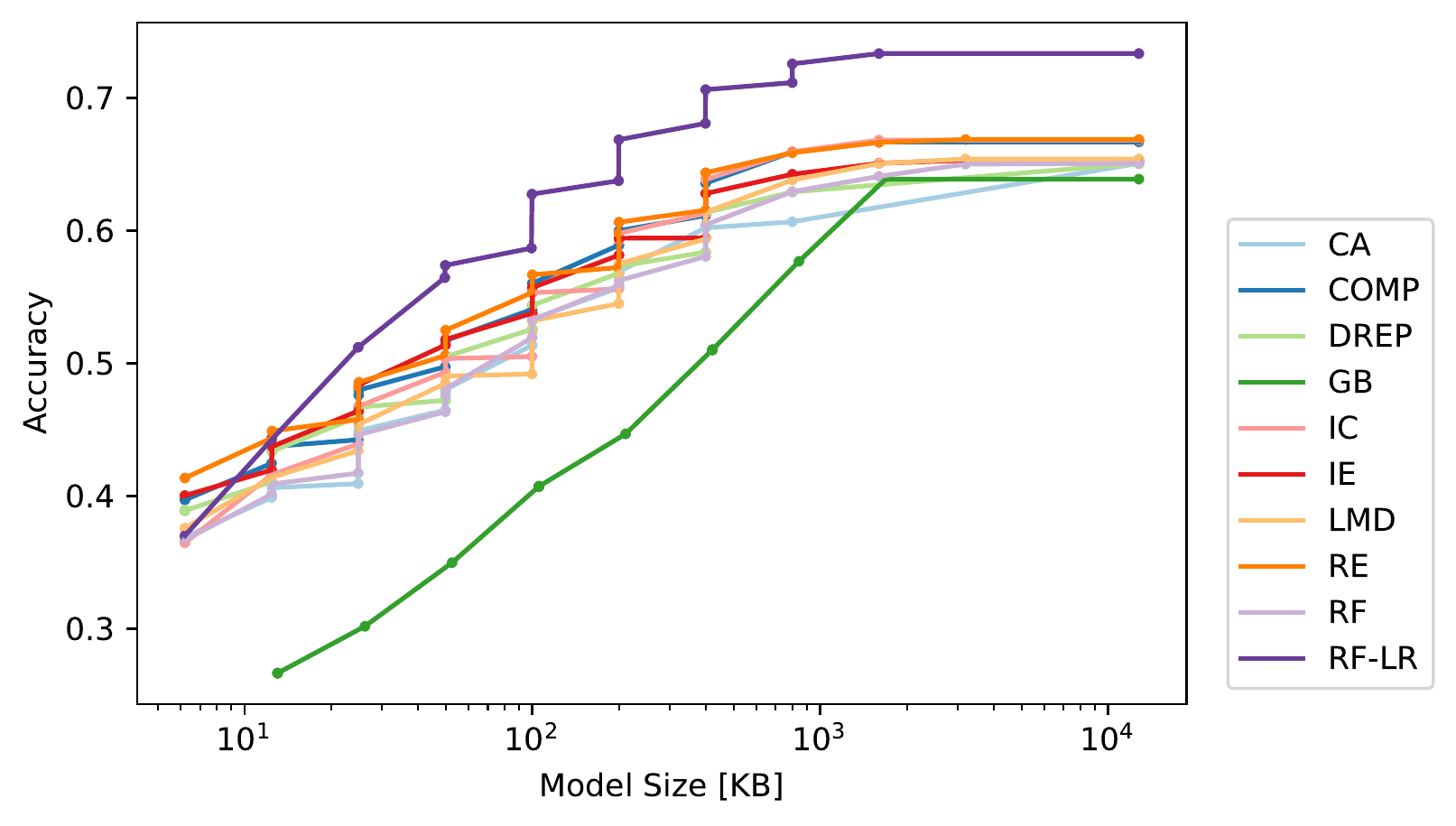}
\caption{5-fold cross-validation accuracy over the size of the ensemble on the chess dataset. Single points are the individual parameter configurations whereas the solid line depicts the corresponding Pareto front. Best viewed in color.}
\label{fig:auc_chess}
\end{figure}

Figure \ref{fig:auc_chess} shows the results on the chess dataset. Here the accuracy ranges from $28 \%$ to $75 \%$ with model sizes up to $12$ MB (again note the logarithmic scale on the x-axis). Similar to before, the pruning methods all converge against similar solutions just above $65 \%$. CA still seems to perform poorly for smaller model sizes, but not as bad as on the EEG data. Similar, GB also seems to struggle on this dataset. It is worth noting, that some pruning methods (e.g. IC or RE) have a better accuracy-memory trade-off compared to Random Forest and they outperform the original forest by about $2\%$. RF-LR offers the best performance on this dataset and outperforms the original forest by about $8\%$ accuracy across all model sizes. This effect is only present for RF-LR and cannot be seen for the other methods. Overall, RF-LR offers a much better accuracy-memory trade-off and offers the best overall accuracy. 

\noindent \textbf{Conclusion:} First we find that the well-performing RF models often require more than $1$ MB easily breaking the available memory on small MCUs (cf. Table \ref{tab:MUC_memory}). Second, we found two different behaviors of RF-LR: In many cases all methods converge against a similar accuracy when more memory is available, e.g. as seen in Figure \ref{fig:auc_eeg}. This can be expected since all methods derive their models from the same RF model. Here, RF-LR often has better models and hence offers a better accuracy-memory trade-off. In many other cases we found that RF-LR significantly outperforms the other methods and offers a much better accuracy across most model sizes, e.g. as depicted in Figure \ref{fig:auc_chess}. In these cases, RF-LR offers a much better accuracy-memory trade-off and the best overall accuracy. 

\subsection{Quantitative Analysis}

The previous section showed that RF-LR can offer substantial improvements on some datasets and smaller improvements in other cases. To give a more complete picture we will now look at the performance of each method under various memory constraints. Table \ref{tab:mem_64} shows the best accuracy of each method (across all hyperparameter configurations) with a final model size below $64$ KB. Such models could for example easily be deployed on an Arduino Due MCU (cf. Table \ref{tab:MUC_memory}). We find that RF-LR offers the best accuracy in $9$ out of $16$ cases followed by RE which is first in $4$ cases followed by GB with ranks first in $3$ cases. IE shares the first place with RE on the mozilla dataset. On some datasets such as ida2016 the differences are comparably small which can be expected since all models are derived from the same base Random Forest. However, on other datasets such as the eeg, chess, japanese-vowels or connect dataset we can find more substantial improvements where RF-LR offers up to $5\%$ better accuracy to the second ranking method.

A similar picture can be find in Table \ref{tab:mem_256} which shows the best accuracy of each method (across all hyperparameter configurations) with a final model size below $256$ KB. Such models could for example easily be deployed on an STM32F4 MCU (cf. Table \ref{tab:MUC_memory}). Now RF-LR offers the best accuracy in $10$ out of $16$ cases followed by RE which is first in only $1$ case followed by GB with ranks first in $2$ cases and IC which is now first in $3$ cases. IE now shares the first place with IC on the mozilla dataset. As before, the differences are comparably small on some datasets (e.g. the mozilla dataset) and more substantial on other datasets where RF-LR now offers up $6\%$ better accuracy against the second best method. 

\begin{table*}
\begin{tabular}{lrrrrrrrrrr}
\toprule
{model} & {CA} & {COMP} & {DREP} & {GB} & {IC} & {IE} & {LMD} & {RE} & {RF} & {RF-LR} \\
\midrule
adult & 85.455 & 85.777 & 85.618 & \textbf{86.616} & 85.799 & 85.882 & 85.378 & 86.128 & 85.464 & 86.241 \\
anura & 96.511 & 97.137 & 96.873 & 95.024 & 97.192 & 96.928 & 96.539 & 97.067 & 96.525 & \textbf{97.512} \\
avila & 91.930 & 96.957 & 95.318 & 67.724 & 96.760 & 96.689 & 86.965 & \textbf{97.048} & 92.534 & 94.278 \\
bank & 89.772 & 90.049 & 89.927 & \textbf{90.677} & 90.213 & 90.330 & 89.819 & 90.336 & 89.830 & 90.522 \\
chess & 48.025 & 51.714 & 50.492 & 34.969 & 50.356 & 51.793 & 49.041 & 52.495 & 48.090 & \textbf{57.382} \\
connect & 72.426 & 73.898 & 73.409 & 70.847 & 73.724 & 73.832 & 73.110 & 74.134 & 72.666 & \textbf{77.498} \\
eeg & 84.419 & 85.574 & 84.419 & 82.463 & 84.780 & 85.033 & 83.852 & 85.527 & 84.606 & \textbf{86.622} \\
elec & 83.523 & 83.620 & 84.075 & 83.878 & 82.749 & 84.556 & 82.817 & 84.680 & 83.391 & \textbf{84.894} \\
ida2016 & 99.044 & 99.119 & 99.025 & 99.094 & 99.106 & 99.100 & 99.031 & 99.125 & 99.075 & \textbf{99.219} \\
japanese-vowels & 90.061 & 91.487 & 91.045 & 87.341 & 91.517 & 90.332 & 90.794 & 91.587 & 90.503 & \textbf{93.173} \\
magic & 86.109 & 86.613 & 86.282 & \textbf{87.355} & 86.692 & 86.771 & 86.456 & 86.845 & 86.419 & 86.997 \\
mnist & 80.700 & 85.700 & 84.300 & 74.400 & 84.700 & 84.100 & 85.000 & 84.900 & 84.200 & \textbf{87.000} \\
mozilla & 94.590 & 94.764 & 94.661 & 94.590 & 94.815 & \textbf{94.860} & 94.468 & \textbf{94.860} & 94.545 & 94.699 \\
nomao & 95.508 & 95.804 & 95.575 & 95.958 & 95.749 & 95.819 & 95.358 & 95.802 & 95.633 & \textbf{96.063} \\
postures & 79.913 & 81.688 & 80.827 & 69.058 & 81.137 & 80.996 & 79.599 & \textbf{81.727} & 80.246 & 81.081 \\
satimage & 88.647 & 88.880 & 88.663 & 87.449 & 88.911 & 88.616 & 88.647 & \textbf{89.020} & 88.538 & 88.834 \\
\bottomrule
\end{tabular}
\caption{\label{tab:mem_64} Test accuracies for models with a memory consumption below $64$ KB for each method and each dataset averaged over a $5$ fold cross validation. Rounded to the third decimal digit. Larger is better. The best method is depicted in bold. More details on the experiments and datasets can be find in the appendix.}
\end{table*}

\begin{table*}
\begin{tabular}{lrrrrrrrrrr}
\toprule
{model} & {CA} & {COMP} & {DREP} & {GB} & {IC} & {IE} & {LMD} & {RE} & {RF} & {RF-LR} \\
\midrule
adult & 85.774 & 86.174 & 85.956 & \textbf{87.135} & 86.011 & 86.232 & 85.738 & 86.272 & 85.799 & 86.508 \\
anura & 96.511 & 97.818 & 97.748 & 97.429 & \textbf{98.040} & 97.929 & 97.582 & 97.818 & 97.735 & 98.013 \\
avila & 95.016 & 99.022 & 97.427 & 82.820 & 99.176 & 99.123 & 93.315 & \textbf{99.248} & 96.526 & 98.807 \\
bank & 89.967 & 90.297 & 90.082 & 90.737 & 90.290 & 90.443 & 89.952 & 90.471 & 90.038 & \textbf{90.874} \\
chess & 56.911 & 59.998 & 57.314 & 44.678 & 59.791 & 59.428 & 57.496 & 60.636 & 56.238 & \textbf{66.852} \\
connect & 74.522 & 76.270 & 75.119 & 76.170 & 76.239 & 76.333 & 75.100 & 76.303 & 74.997 & \textbf{80.340} \\
eeg & 87.223 & 88.364 & 88.511 & 85.734 & 88.959 & 88.778 & 87.784 & 88.785 & 87.951 & \textbf{90.454} \\
elec & 85.220 & 85.810 & 85.832 & 85.748 & 86.043 & 86.692 & 84.922 & 86.562 & 85.362 & \textbf{87.829} \\
ida2016 & 99.106 & 99.225 & 99.119 & 99.169 & \textbf{99.262} & 99.175 & 99.238 & 99.175 & 99.194 & 99.244 \\
japanese-vowels & 91.979 & 94.810 & 94.428 & 94.860 & 94.790 & 94.077 & 94.278 & 94.709 & 94.408 & \textbf{96.205} \\
magic & 86.655 & 87.281 & 87.218 & \textbf{87.733} & 87.234 & 87.239 & 87.302 & 87.365 & 86.950 & 87.570 \\
mnist & 86.900 & 90.100 & 89.300 & 84.600 & 90.600 & 89.700 & 88.800 & 89.600 & 89.300 & \textbf{91.800} \\
mozilla & 94.731 & 95.034 & 94.982 & 94.989 & \textbf{95.092} & \textbf{95.092} & 94.802 & 94.995 & 94.912 & 95.014 \\
nomao & 96.039 & 96.222 & 96.150 & 96.408 & 96.318 & 96.269 & 96.077 & 96.356 & 96.135 & \textbf{96.539} \\
postures & 88.587 & 89.596 & 88.747 & 81.100 & 89.231 & 89.390 & 88.085 & 89.633 & 88.281 & \textbf{90.504} \\
satimage & 88.802 & 90.218 & 89.891 & 89.782 & 89.705 & 89.891 & 90.000 & 90.156 & 90.016 & \textbf{90.715} \\
\bottomrule
\end{tabular}
\caption{\label{tab:mem_256} Test accuracies for models with a memory consumption below $256$ KB for each method and each dataset averaged over a $5$ fold cross validation. Rounded to the third decimal digit. Larger is better. The best method is depicted in bold. More details on the experiments and datasets can be find in the appendix.}
\end{table*}

\begin{table*}[ht]
\centering
\begin{tabular}{lrrrrrrrrrr}
\toprule
 &      CA &    COMP &    DREP &      GB &      IC &      IE &     LMD &      RE &      RF &   RF-LR \\
\midrule
chess           &  0.6251 &  0.6628 &  0.6363 &  0.6265 &  0.6638 &  0.6489 &  0.6486 &  0.6644 &  0.6441 &  \textbf{0.7290} \\
connect         &  0.7570 &  0.7712 &  0.7608 &  0.7780 &  0.7733 &  0.7726 &  0.7623 &  0.7737 &  0.7607 &  \textbf{0.8219} \\
eeg             &  0.9050 &  0.9242 &  0.9240 &  0.8570 &  0.9276 &  0.9258 &  0.9224 &  0.9242 &  0.9225 &  \textbf{0.9319} \\
elec            &  0.8667 &  0.8767 &  0.8722 &  0.8572 &  0.8787 &  0.8779 &  0.8714 &  0.8783 &  0.8720 &  \textbf{0.8974} \\
postures        &  0.9390 &  0.9497 &  0.9436 &  0.9105 &  0.9504 &  0.9486 &  0.9460 &  0.9501 &  0.9460 &\textbf{0.9688}\vspace{0.2cm}\\
anura           &  0.9710 &  0.9791 &  0.9790 &  0.9766 &  0.9795 &  0.9791 &  0.9780 &  0.9792 &  0.9779 &  \textbf{0.9800} \\
bank            &  0.9018 &  0.9050 &  0.9038 &  0.9073 &  0.9052 &  0.9050 &  0.9034 &  0.9052 &  0.9034 &  \textbf{0.9083} \\
japanese-vowels &  0.9568 &  0.9721 &  0.9717 &  0.9734 &  0.9731 &  0.9712 &  0.9719 &  0.9722 &  0.9707 &  \textbf{0.9741} \\
magic           &  0.8748 &  0.8783 &  0.8786 &  0.8772 &  0.8793 &  0.8788 &  0.8788 &  0.8795 &  0.8783 &  \textbf{0.8808} \\
mnist           &  0.9295 &  0.9393 &  0.9399 &  0.9377 &  0.9415 &  0.9400 &  0.9393 &  0.9403 &  0.9366 &  \textbf{0.9432} \\
nomao           &  0.9647 &  0.9678 &  0.9676 &  0.9640 &  0.9681 &  0.9678 &  0.9679 &  0.9677 &  0.9678 &  \textbf{0.9682}\vspace{0.2cm}\\
adult           &  0.8620 &  0.8638 &  0.8630 &  \textbf{0.8712} &  0.8642 &  0.8640 &  0.8627 &  0.8639 &  0.8631 &  0.8656\\
avila           &  0.9715 &  0.9924 &  0.9897 &  0.9909 &  \textbf{0.9965} &  0.9963 &  0.9750 &  0.9930 &  0.9886 &  0.9928 \\
ida2016         &  0.9901 &  \textbf{0.9916} &  0.9908 &  0.9915 &  0.9913 &  0.9909 &  0.9909 &  0.9908 &  0.9907 &  0.9912 \\
mozilla         &  0.9493 &  0.9520 &  0.9520 &  0.9498 &  0.9522 &  0.9525 &  0.9513 &  \textbf{0.9526} &  0.9519 &\textbf{0.9526}\\
satimage        &  0.9059 &  0.9135 &  0.9133 &  0.9119 &  0.9147 &  \textbf{0.9150} &  0.9140 &  0.9135 &  0.9138 &  0.9148 \\
\bottomrule
\end{tabular}
\caption{\label{tab:aucs} Normalized area under the Pareto front (APF) for each method and each dataset averaged over a $5$ fold cross validation. Rounded to the fourth decimal digit. Larger is better. The best method is depicted in bold.}
\end{table*}

To give a more complete picture across different memory constraints we will now summarize the performance of each method by the (normalized) area-under the Pareto front: Intuitively, we want to have an algorithm which gives small and accurate models and therefore places itself in the upper-left corner of the accuracy-memory plots. Similar to `regular' ROC-AUC curves we can compute the area under the Pareto front (APF) normalized by the biggest model to summarize the accuracy for different models on the same dataset. Table \ref{tab:aucs} depicts the normalized APF for the experiments. Looking at RF-LR, we see that it is the clear winner. In total, it is the best method on 11 of 14 datasets, shares the first place on 1 dataset (mozilla), is the second best method on 1 data-set (satimage), third best method (ida2016) and fourth best methond (avila) each on one dataset. In the first block of datasets (chess, connect, eeg, elec, postures) RF-LR achieves substantial improvements with $1\% - 8\%$ higher accuracies on average. Looking at the second block (adult, anura, bank, magic, mnist, nomao, japanese-vowels) RF-LR is still the best method, but the differences are smaller than before. Finally, in block three (ida2016,mozilla,satimage) RF-LR is not the best method alone anymore, but ranks among the best methods.

Table \ref{tab:aucs} implies that RF-LR can offer substantial improvements in some cases and moderate  improvement in many other cases. To give a statistical meaningful comparison we present the results in Table \ref{tab:aucs} as a CD diagram \cite{demvsar/2006}. A CD diagram ranks each method according to its performance on each dataset. Then, a Friedman-Test is performed to determine if there is a statistical difference between the average rank of each method. If this is the case, a  pairwise Wilcoxon-Test between all methods is used to check whether there is a statistical difference between two classifiers. CD diagrams visualize this evaluation by plotting the average rank of each method on the x-axis and connect all classifiers whose performances are statistically similar via a horizontal bar.

\begin{figure}
\centering
\includegraphics[width=\columnwidth, keepaspectratio]{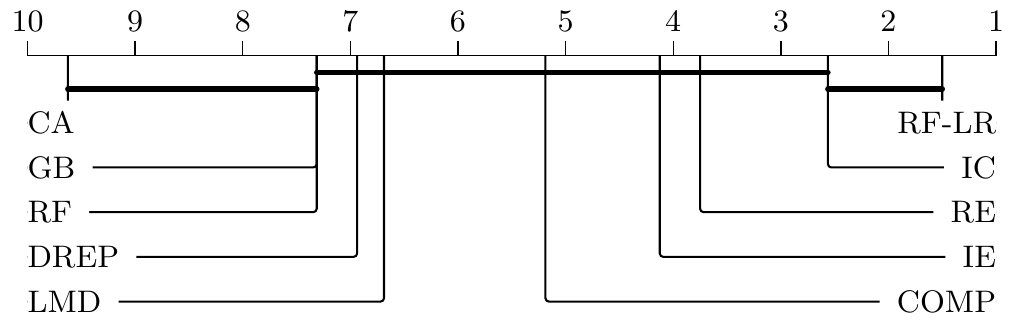}
\caption{CD-Diagram for the normalized area under the Pareto front for different methods over multiple datasets. For all statistical tests $p=0.95$ was used. More to the right (lower rank) is better. Methods in connected cliques are statistically similar.}
\label{fig:cd_auc}
\end{figure}

Figure \ref{fig:cd_auc} shows the CD diagram for the experiments, where $p=0.95$ was used for all statistical tests. RF-LR is the clear winner in this comparison. Its average rank is close to $1.5$ and it has some distance to the second best method IC with an average rank around $2.5$. It offers a \emph{statistically} better performance compared to $\{RE,IE,COMP,LMD,DREP,RF,GB,CA\}$. The second clique is given by $\{IC, RE, IE, COMP, LMD, DREP, RF, GB\}$ with ranks around $2.5-7$. Overall, RE places third with an average rank around $4$ which shows that a simple method can perform surprisingly well. We hypothesize that since RE minimizes the overall ensemble loss that it finds a good balance between the bias and the diversity of the ensemble as e.g. discussed in \cite{buschjager/etal/2020}. Next, $\{RF,GB,CA\}$ form the last clique with statistically similar performances which shows that a unpruned RF and GB doe not offer a good accuracy-memory trade-off. CA ranks last with some distance to the other methods. We are not sure why CA has such a bad performance and suspect a bug in our implementation which we could not find so far. 

\subsection{Case-Study On Raspberry Pi0}
\label{sec:casestudy}

To showcase the effectiveness of our approach we will now compare the performance of ensemble pruning and leaf-refinement on a Raspberry Pi0. The Raspbery Pi0 has 512 MB RAM and uses a \texttt{BCM 2835 SOC} CPU clocked at 1 GHz. This makes it considerably more powerful than the MCUs mentioned in Table \ref{tab:MUC_memory}, but also allows us to run a full Linux environment which simplifies the evaluation. Again we will now focus on the EEG dataset as our standard example. From our previous experiment we selected pruning configurations that resulted in an ensemble size below $256$ KB and generate ensemble-specific \texttt{C++} code as outline in \cite{Buschjaeger/etal/2018} which is then compiled on the Pi0 itself\footnote{For these experiments we excluded Gradient Boosting (GB) because scikit learn trains individual trees for each class instead of probability vectors which would have required substantial refactoring of our experiments.}. We compare the latency, the accuracy and the \emph{total} binary size of these implementations. Note that the total binary size may exceed 256 KB because the binary also contains additional functions from the standard library as well as a start routine and the corresponding ELF header. However, this overhead is the same for \emph{all} implementations. For simplicity we measure the accuracy as well as the latency using the first cross-validation set. To ensure a fair comparison we repeat each experiment 5 times. Table \ref{tab:pi0} contains the results for this evaluation. As one can see the binary sizes range from $268$ KB to roughly $800$ KB and each implementation requires $0.8$ to $1.6$ µs to classify a single observation. As expected, RF-LR offers the best accuracy around $91$ \% while ranking third in memory usage. Somewhat surprisingly, RF-LR has a comparably high latency. We conjecture that the structure of the trees in RF-LR is not very homogeneous which seems to be beneficial for the accuracy, but may hurt the caching behavior of the trees. A more thorough discussion can be found in \cite{Buschjaeger/etal/2018} on this topic and a combination of of both approaches should be considered in future research. Nevertheless, this evaluation shows that our approach can be applied in a real-world scenario and we believe that these results can be transferred to other hardware architectures as well.

\begin{table}[]
\begin{tabular}{@{}lrrr@{}}
\toprule
Method & Accuracy & Size [Bytes] & Latency [µs/obs.] \\ \midrule
CA & 86.2817 &  268~536     & 0.80107              \\
COMP & 89.8531 & 689~712 & 1.06809       \\
DREP & 88.7850 &  342~696 & 1.00134  \\
IC & 89.2523 &   793~280 & 1.06809  \\
IE &  88.8518  &  743~232 & 1.00134\\
LMD & 88.8518 &   784~896 & 1.06809\\
RE &  89.2523 &   792~456 & 1.13485\\
RF &  89.5194 &  588~336 & 1.60214  \\
RF-LR & \textbf{91.0881} & 588~088 & 1.46862\\ \bottomrule
\end{tabular}
\caption{\label{tab:pi0} Accuracy, Size and Latency on a Raspberry Pi0. Models are filtered so that the ensemble size does not exceed $256$ KB.}
\end{table}

\section{Conclusion}
\label{sec:conclusion}

Ensemble algorithms are among the state-of-the-art in many machine learning applications. With the ongoing integration of ML models into everyday life, the deployment and continuous application of models becomes more and more an important issue. By today's standard, Random Forests are trained with large trees for the best performance which can challenge the resources of small devices and sometimes make deployment impossible. Ensemble pruning is a standard technique to remove unnecessary classifiers from the ensemble to reduce the overall resource consumption while potentially improving its accuracy. This makes ensemble pruning ideal to bring accurate ensembles to small devices. While ensemble pruning improves the performance of ensembles of small trees we found that this improvement diminishes for ensembles of large trees. Moreover, it does not offer fine-grained control over this trade-off because it removes entire trees at once from the ensemble. We argue that, from a hardware perspective, the fine-grained control over the accuracy-memory trade-off is what really matters. We propose a simple and surprisingly effective algorithm which refines the predictions of the trees in a forest using SGD. We compared our Leaf-Refinement method against 7 state-of-the-art pruning methods on $16$ datasets. Leaf-Refinement outperforms the other methods on 11 of 16 datasets with a statistically significant better accuracy-memory trade-off compared to most methods. In a small study we showed that our approach can be applied in real-world scenarios. and we believe that our results can be transferred to other hardware architectures. Since our approach is orthogonal to existing approaches it can be freely combined with other methods for efficient deployment. Hence future research should include not only the combination of more diverse hardware, but also the combination of different methods. 

\bibliography{literature}

\begin{thebibliography}{3}
\providecommand{\natexlab}[1]{#1}
\providecommand{\url}[1]{\texttt{#1}}
\expandafter\ifx\csname urlstyle\endcsname\relax
  \providecommand{\doi}[1]{doi: #1}\else
  \providecommand{\doi}{doi: \begingroup \urlstyle{rm}\Url}\fi

\bibitem[Oshiro et~al.(2012)Oshiro, Perez, and Baranauskas]{oshiro/etal/2012}
Thais~Mayumi Oshiro, Pedro~Santoro Perez, and Jos{\'e}~Augusto Baranauskas.
\newblock How many trees in a random forest?
\newblock In \emph{International workshop on machine learning and data mining
  in pattern recognition}, pages 154--168. Springer, 2012.

\bibitem[Pedregosa et~al.(2011)Pedregosa, Varoquaux, Gramfort, Michel, Thirion,
  Grisel, Blondel, Prettenhofer, Weiss, Dubourg, Vanderplas, Passos,
  Cournapeau, Brucher, Perrot, and Duchesnay]{Pedregosa/etal/2001}
F.~Pedregosa, G.~Varoquaux, A.~Gramfort, V.~Michel, B.~Thirion, O.~Grisel,
  M.~Blondel, P.~Prettenhofer, R.~Weiss, V.~Dubourg, J.~Vanderplas, A.~Passos,
  D.~Cournapeau, M.~Brucher, M.~Perrot, and E.~Duchesnay.
\newblock Scikit-learn: Machine learning in {P}ython.
\newblock \emph{Journal of Machine Learning Research}, 12:\penalty0 2825--2830,
  2011.

\bibitem[Zhou et~al.(2002)Zhou, Wu, and Tang]{zhou/etal/2002}
Zhi-Hua Zhou, Jianxin Wu, and Wei Tang.
\newblock Ensembling neural networks: many could be better than all.
\newblock \emph{Artificial intelligence}, 137\penalty0 (1-2):\penalty0
  239--263, 2002.

\end{thebibliography}


\begin{thebibliography}{39}
\providecommand{\natexlab}[1]{#1}
\providecommand{\url}[1]{\texttt{#1}}
\expandafter\ifx\csname urlstyle\endcsname\relax
  \providecommand{\doi}[1]{doi: #1}\else
  \providecommand{\doi}{doi: \begingroup \urlstyle{rm}\Url}\fi

\bibitem[Barros et~al.(2015)Barros, de~Carvalho, and Freitas]{Barros/etal/2015}
Barros, R.~C., de~Carvalho, A. C. P. L.~F., and Freitas, A.~A.
\newblock \emph{Decision-Tree Induction}, pp.\  7--45.
\newblock Springer International Publishing, Cham, 2015.
\newblock ISBN 978-3-319-14231-9.
\newblock \doi{10.1007/978-3-319-14231-9_2}.
\newblock URL \url{https://doi.org/10.1007/978-3-319-14231-9_2}.

\bibitem[Biau(2012)]{biau/2012}
Biau, G.
\newblock Analysis of a random forests model.
\newblock \emph{Journal of Machine Learning Research}, 13\penalty0
  (Apr):\penalty0 1063--1095, 2012.

\bibitem[Biau \& Scornet(2016)Biau and Scornet]{biau/Scornet/2016}
Biau, G. and Scornet, E.
\newblock A random forest guided tour.
\newblock \emph{Test}, 25\penalty0 (2):\penalty0 197--227, 2016.

\bibitem[Branco et~al.(2019)Branco, Ferreira, and Cabral]{branco/etal/2019}
Branco, S., Ferreira, A.~G., and Cabral, J.
\newblock Machine learning in resource-scarce embedded systems, fpgas, and
  end-devices: A survey.
\newblock \emph{Electronics}, 8\penalty0 (11):\penalty0 1289, 2019.

\bibitem[Breiman(1996)]{breiman/1996}
Breiman, L.
\newblock Bagging predictors.
\newblock \emph{Machine learning}, 24\penalty0 (2):\penalty0 123--140, 1996.

\bibitem[Breiman(2000)]{Breiman/2000}
Breiman, L.
\newblock Some infinity theory for predictor ensembles.
\newblock Technical report, Technical Report 579, Statistics Dept. UCB, 2000.

\bibitem[Buschj{\"a}ger \& Morik(2017)Buschj{\"a}ger and
  Morik]{buschjaeger/morik/2017}
Buschj{\"a}ger, S. and Morik, K.
\newblock Decision tree and random forest implementations for fast filtering of
  sensor data.
\newblock \emph{IEEE Transactions on Circuits and Systems I: Regular Papers},
  65\penalty0 (1):\penalty0 209--222, 2017.

\bibitem[Buschj{\"a}ger et~al.(2020)Buschj{\"a}ger, Pfahler, and
  Morik]{buschjager/etal/2020}
Buschj{\"a}ger, S., Pfahler, L., and Morik, K.
\newblock Generalized negative correlation learning for deep ensembling.
\newblock \emph{arXiv preprint arXiv:2011.02952}, 2020.

\bibitem[{Buschjäger} et~al.(2018){Buschjäger}, {Chen}, {Chen}, and
  {Morik}]{Buschjaeger/etal/2018}
{Buschjäger}, S., {Chen}, K., {Chen}, J., and {Morik}, K.
\newblock Realization of random forest for real-time evaluation through tree
  framing.
\newblock In \emph{ICDM}, pp.\  19--28, 2018.
\newblock \doi{10.1109/ICDM.2018.00017}.

\bibitem[Cavalcanti et~al.(2016)Cavalcanti, Oliveira, Moura, and
  Carvalho]{cavalcanti2016combining}
Cavalcanti, G.~D., Oliveira, L.~S., Moura, T.~J., and Carvalho, G.~V.
\newblock Combining diversity measures for ensemble pruning.
\newblock \emph{Pattern Recognition Letters}, 74:\penalty0 38--45, 2016.

\bibitem[Choudhary et~al.(2020)Choudhary, Mishra, Goswami, and
  Sarangapani]{choudhary/etal/2020}
Choudhary, T., Mishra, V., Goswami, A., and Sarangapani, J.
\newblock A comprehensive survey on model compression and acceleration.
\newblock \emph{Artificial Intelligence Review}, 53\penalty0 (7):\penalty0
  5113--5155, 2020.

\bibitem[Cortes et~al.(2014)Cortes, Mohri, and Syed]{Cortes/etal/2014a}
Cortes, C., Mohri, M., and Syed, U.
\newblock Deep boosting.
\newblock In \emph{Proceedings of the Thirty-First International Conference on
  Machine Learning (ICML 2014)}, 2014.

\bibitem[Dem{\v{s}}ar(2006)]{demvsar/2006}
Dem{\v{s}}ar, J.
\newblock Statistical comparisons of classifiers over multiple data sets.
\newblock \emph{The Journal of Machine Learning Research}, 7:\penalty0 1--30,
  2006.

\bibitem[Denil et~al.(2014)Denil, Matheson, and De~Freitas]{denil/2014}
Denil, M., Matheson, D., and De~Freitas, N.
\newblock Narrowing the gap: Random forests in theory and in practice.
\newblock In \emph{International conference on machine learning (ICML)}, 2014.

\bibitem[Domingos(2000)]{domingos/etal/2000}
Domingos, P.
\newblock A unified bias-variance decomposition for zero-one and squared loss.
\newblock \emph{AAAI/IAAI}, 2000:\penalty0 564--569, 2000.

\bibitem[Friedman(2001)]{friedman/etal/2001}
Friedman, J.~H.
\newblock Greedy function approximation: a gradient boosting machine.
\newblock \emph{Annals of statistics}, pp.\  1189--1232, 2001.

\bibitem[Geurts et~al.(2006)Geurts, Ernst, and Wehenkel]{geurts/etal/2006}
Geurts, P., Ernst, D., and Wehenkel, L.
\newblock Extremely randomized trees.
\newblock \emph{Machine learning}, 63\penalty0 (1):\penalty0 3--42, 2006.

\bibitem[Giacinto et~al.(2000)Giacinto, Roli, and Fumera]{giacinto/etal/2000}
Giacinto, G., Roli, F., and Fumera, G.
\newblock Design of effective multiple classifier systems by clustering of
  classifiers.
\newblock In \emph{Proceedings 15th International Conference on Pattern
  Recognition. ICPR-2000}, volume~2, pp.\  160--163. IEEE, 2000.

\bibitem[Guo et~al.(2018)Guo, Liu, Li, Wu, Guo, and Xu]{guo2018margin}
Guo, H., Liu, H., Li, R., Wu, C., Guo, Y., and Xu, M.
\newblock Margin \& diversity based ordering ensemble pruning.
\newblock \emph{Neurocomputing}, 275:\penalty0 237--246, 2018.

\bibitem[Ho(1998)]{ho/1998}
Ho, T.~K.
\newblock The random subspace method for constructing decision forests.
\newblock \emph{IEEE transactions on pattern analysis and machine
  intelligence}, 20\penalty0 (8):\penalty0 832--844, 1998.

\bibitem[Jiang et~al.(2017)Jiang, Liu, Fu, and Wu]{Jiang/etal/2017}
Jiang, Z., Liu, H., Fu, B., and Wu, Z.
\newblock {Generalized ambiguity decompositions for classification with
  applications in active learning and unsupervised ensemble pruning}.
\newblock \emph{31st AAAI Conference on Artificial Intelligence, AAAI 2017},
  pp.\  2073--2079, 2017.

\bibitem[Koltchinskii et~al.(2002)]{Koltchinskii/Panchenko/2002}
Koltchinskii, V. et~al.
\newblock Empirical margin distributions and bounding the generalization error
  of combined classifiers.
\newblock \emph{The Annals of Statistics}, 30\penalty0 (1):\penalty0 1--50,
  2002.

\bibitem[Kumar et~al.(2017)Kumar, Goyal, and Varma]{kumar/etal/2017}
Kumar, A., Goyal, S., and Varma, M.
\newblock Resource-efficient machine learning in 2 kb ram for the internet of
  things.
\newblock In \emph{International Conference on Machine Learning}, pp.\
  1935--1944. PMLR, 2017.

\bibitem[Lazarevic \& Obradovic(2001)Lazarevic and
  Obradovic]{lazarevic/2001/effective}
Lazarevic, A. and Obradovic, Z.
\newblock Effective pruning of neural network classifier ensembles.
\newblock In \emph{IJCNN'01}, volume~2, pp.\  796--801. IEEE, 2001.

\bibitem[Li et~al.(2012)Li, Yu, and Zhou]{li2012diversity}
Li, N., Yu, Y., and Zhou, Z.-H.
\newblock Diversity regularized ensemble pruning.
\newblock In \emph{ECML PKDD}, pp.\  330--345. Springer, 2012.

\bibitem[Louppe \& Geurts(2012)Louppe and Geurts]{louppe/Geurts/2012}
Louppe, G. and Geurts, P.
\newblock Ensembles on random patches.
\newblock In \emph{Joint European Conference on Machine Learning and Knowledge
  Discovery in Databases}, pp.\  346--361. Springer, 2012.

\bibitem[Lu et~al.(2010)Lu, Wu, Zhu, and Bongard]{lu2010ensemble}
Lu, Z., Wu, X., Zhu, X., and Bongard, J.
\newblock Ensemble pruning via individual contribution ordering.
\newblock In \emph{Proc. of the ACM SIGKDD}, pp.\  871--880, 2010.

\bibitem[Margineantu \& Dietterich(1997)Margineantu and
  Dietterich]{margineantu/dietterich/1997}
Margineantu, D.~D. and Dietterich, T.~G.
\newblock Pruning adaptive boosting.
\newblock In \emph{ICML}, volume~97, pp.\  211--218, 1997.

\bibitem[Mart{\i}nez-Munoz \& Su{\'a}rez(2004)Mart{\i}nez-Munoz and
  Su{\'a}rez]{martinez/suarez/2004}
Mart{\i}nez-Munoz, G. and Su{\'a}rez, A.
\newblock Aggregation ordering in bagging.
\newblock In \emph{Proc. of the IASTED}, pp.\  258--263, 2004.

\bibitem[Mart{\'\i}nez-Mu{\~n}oz \& Su{\'a}rez(2006)Mart{\'\i}nez-Mu{\~n}oz and
  Su{\'a}rez]{martinez/suarez/2006}
Mart{\'\i}nez-Mu{\~n}oz, G. and Su{\'a}rez, A.
\newblock Pruning in ordered bagging ensembles.
\newblock In \emph{ICML}, pp.\  609--616, 2006.

\bibitem[Mart{\'\i}nez-Mu{\~n}oz et~al.(2008)Mart{\'\i}nez-Mu{\~n}oz,
  Hern{\'a}ndez-Lobato, and Su{\'a}rez]{martinez/etal/2008}
Mart{\'\i}nez-Mu{\~n}oz, G., Hern{\'a}ndez-Lobato, D., and Su{\'a}rez, A.
\newblock An analysis of ensemble pruning techniques based on ordered
  aggregation.
\newblock \emph{IEEE Transactions on Pattern Analysis and Machine
  Intelligence}, 31\penalty0 (2):\penalty0 245--259, 2008.

\bibitem[Oshiro et~al.(2012)Oshiro, Perez, and Baranauskas]{oshiro/etal/2012}
Oshiro, T.~M., Perez, P.~S., and Baranauskas, J.~A.
\newblock How many trees in a random forest?
\newblock In \emph{International workshop on machine learning and data mining
  in pattern recognition}, pp.\  154--168. Springer, 2012.

\bibitem[Pedregosa et~al.(2011)Pedregosa, Varoquaux, Gramfort, Michel, Thirion,
  Grisel, Blondel, Prettenhofer, Weiss, Dubourg, Vanderplas, Passos,
  Cournapeau, Brucher, Perrot, and Duchesnay]{Pedregosa/etal/2001}
Pedregosa, F., Varoquaux, G., Gramfort, A., Michel, V., Thirion, B., Grisel,
  O., Blondel, M., Prettenhofer, P., Weiss, R., Dubourg, V., Vanderplas, J.,
  Passos, A., Cournapeau, D., Brucher, M., Perrot, M., and Duchesnay, E.
\newblock Scikit-learn: Machine learning in {P}ython.
\newblock \emph{Journal of Machine Learning Research}, 12:\penalty0 2825--2830,
  2011.

\bibitem[Shotton et~al.(2013)Shotton, Sharp, Kohli, Nowozin, Winn, and
  Criminisi]{shotton/etal/2013}
Shotton, J., Sharp, T., Kohli, P., Nowozin, S., Winn, J., and Criminisi, A.
\newblock Decision jungles: Compact and rich models for classification.
\newblock In \emph{NIPS'13 Proceedings of the 26th International Conference on
  Neural Information Processing Systems}, pp.\  234--242, 2013.

\bibitem[Tsoumakas et~al.(2009)Tsoumakas, Partalas, and
  Vlahavas]{tsoumakas/etal/2009}
Tsoumakas, G., Partalas, I., and Vlahavas, I.
\newblock An ensemble pruning primer.
\newblock In \emph{Applications of supervised and unsupervised ensemble
  methods}. Springer, 2009.

\bibitem[Ye et~al.(2018)Ye, Zhou, Zou, Gao, and Zhang]{ye/etal/2018}
Ye, T., Zhou, H., Zou, W.~Y., Gao, B., and Zhang, R.
\newblock Rapidscorer: fast tree ensemble evaluation by maximizing compactness
  in data level parallelization.
\newblock In \emph{Proceedings of the 24th ACM SIGKDD International Conference
  on Knowledge Discovery \& Data Mining}, pp.\  941--950, 2018.

\bibitem[Zhang et~al.(2006)Zhang, Burer, and Street]{zhang2006ensemble}
Zhang, Y., Burer, S., and Street, W.~N.
\newblock Ensemble pruning via semi-definite programming.
\newblock \emph{Journal of machine learning research}, 7\penalty0
  (Jul):\penalty0 1315--1338, 2006.

\bibitem[Zhou(2012)]{zhou2012ensemble}
Zhou, Z.-H.
\newblock \emph{Ensemble methods: foundations and algorithms}.
\newblock CRC press, 2012.

\bibitem[Zhou et~al.(2002)Zhou, Wu, and Tang]{zhou/etal/2002}
Zhou, Z.-H., Wu, J., and Tang, W.
\newblock Ensembling neural networks: many could be better than all.
\newblock \emph{Artificial intelligence}, 137\penalty0 (1-2):\penalty0
  239--263, 2002.

\end{thebibliography}
\bibliographystyle{mlsys2022}
\end{document}


\title{APPENDIX: Improving the Accuracy-Memory Trade-Off of Random Forests Via Leaf-Refinement}

\author{\name Sebastian Buschjäger \email sebastian.buschjaeger@tu-dortmund.de \\ 
\name Katharina Morik \email katharina.morik@tu-dortmund.de 
}

\maketitle
 
\begin{abstract}
This appendix accompanies the paper `Improving the Accuracy-Memory Trade-Off of Random Forests Via Leaf-Refinement'. It provides results for more experiments which are not given in the paper due to space reasons. 
\end{abstract}

\section{Transformation of the Many-Could-Be-Better-Than-All-Theorem}

Let 
\begin{align}
C_{k,k} &= \mathbb E_{x,y \sim \mathcal D}\left[ (h_k(x) - y)^2 \right] \\
C_{k,i} &= \mathbb E_{x,y \sim \mathcal D}\left[ (h_k(x) - y) (h_i(x) - y) \right]
\end{align}
then from Eq. (14) and Eq. (15) in \cite{zhou/etal/2002} we have:
\begin{align*}
    \sum_{i=1, i \not= k }^M \sum_{j=1, i \not= k}^M \frac{C_{i,j}}{(M-1)^2} &\le \sum_{i=1}^M\sum_{h=1}^M\frac{C_{i,j}}{M^2} 
    = \sum_{i=1, i \not= k }^M \sum_{j=1, i \not= k}^M \frac{C_{i,j}}{M^2} + 2 \sum_{i=1, i \not= k}^M \frac{C_{i,k}}{M^2} + \frac{C_{k,k}}{M^2} \\
    &\le \frac{(M-1)^2}{M^2}\left( \sum_{i=1, i \not= k }^M \sum_{j=1, i \not= k}^M C_{i,j} + 2 \sum_{i=1, i \not= k}^M C_{i,k} + C_{k,k} \right) \\
    &\le \sum_{i=1, i \not= k }^M \sum_{j=1, i \not= k}^M C_{i,j} + 2 \sum_{i=1, i \not= k}^M C_{i,k} + C_{k,k} \\
    0 &\le 2 \sum_{i=1, i \not= k}^M C_{i,k} + C_{k,k} \\
    - 2 \sum_{i=1, i \not= k}^M C_{i,k} &\le C_{k,k}
\end{align*}

\section{Dataset}

Table \ref{tab:datasets} gives an overview of the datasets used for all experiments. All datasets are freely available online. The detailed download scripts for each dataset are provided in the anonymized version of the source code.

\begin{table}[h]
    \centering
    \caption{Summary of data sets for our experiments. All datasets are publicly available and download scripts are included in this submission. In all experiments we use a $5$-fold cross validation except for mnist and ida2016 in which we used the given test/train split. We use minimal pre-processing and by removing examples containing \texttt{NaN} values and computing a one-hot encoding for categorical features. $N$ depicts total number of datapoints (after removing \texttt{NaN}), $d$ is the dimensionality including all one-hot encoded features and $C$ is the number of classes.}
    \label{tab:datasets}
    \begin{tabular}{@{}crrr@{}}
    \toprule
    Dataset         & N        & d     & C \\ \midrule
    adult           & $32,561$  & 108 & 2 \\
    anura           & $7,195$   & 22 & 10 \\
    avila           & $20,867$  & 10 & 11 \\
    bank            & $45,211$  & 51  & 2 \\
    chess         & $28~056$  & 23  & 17 \\
    connect         & $67,557$  & 42  & 3 \\
    eeg             & $14,980$  & 14  & 2 \\
    elec            & $45,312$  & 14  & 2 \\ \bottomrule
    \end{tabular}~~~~
    \begin{tabular}{@{}crrr@{}}
    \toprule
    Dataset         & N        & d     & C \\ \midrule
    ida2016         & $76,000$  & 170 & 2 \\
    japanese-vowels & $9,961$   & 14  & 9 \\ 
    magic           & $19,019$  & 10  & 2 \\ 
    mnist           & $70,000$  & 784 & 10 \\
    mozilla         & $15,545$  & 5   & 2 \\
    nomao           & $34,465$  & 174 & 2 \\
    postures        & $78,095$  & 9  & 5 \\
    satimage        & $6,430$   & 36  & 6 \\ \bottomrule
    \end{tabular}
\end{table}

\section{Revisiting Ensemble Pruning on More Datasets}
The section `Revisiting Ensemble Pruning' showed results for the EEG dataset. In this section, we show the results for this experiment on the other dataset depicted in Table \ref{tab:datasets}. Recall the following experimental protocol: Oshiro et al. showed in \cite{oshiro/etal/2012} that the prediction of a RF stabilizes between $128$ and $256$ trees in the ensemble and adding more trees to the ensemble does not yield significantly better results. Hence, we train the `base' Random Forests with $M = 256$ trees. To control the individual errors of trees we set the maximum number of leaf nodes $n_l$ to values between $n_l \in \{64,128,256,512,1024\}$. For ensemble pruning we use RE and compare it against a random selection of trees from the original ensemble (which is the same a training a smaller forest directly). In both cases a sub-ensemble with $K \in \{2,4,8,16,32,64,128,256\}$ members is selected so that for $K=256$ the original RF is recovered. For RE we use the training data as pruning set. We report the average accuracy over a $5$-fold cross-validation. 

\begin{figure}[H]
\begin{minipage}{.49\textwidth}
    \centering
    \includegraphics[width=\textwidth,keepaspectratio]{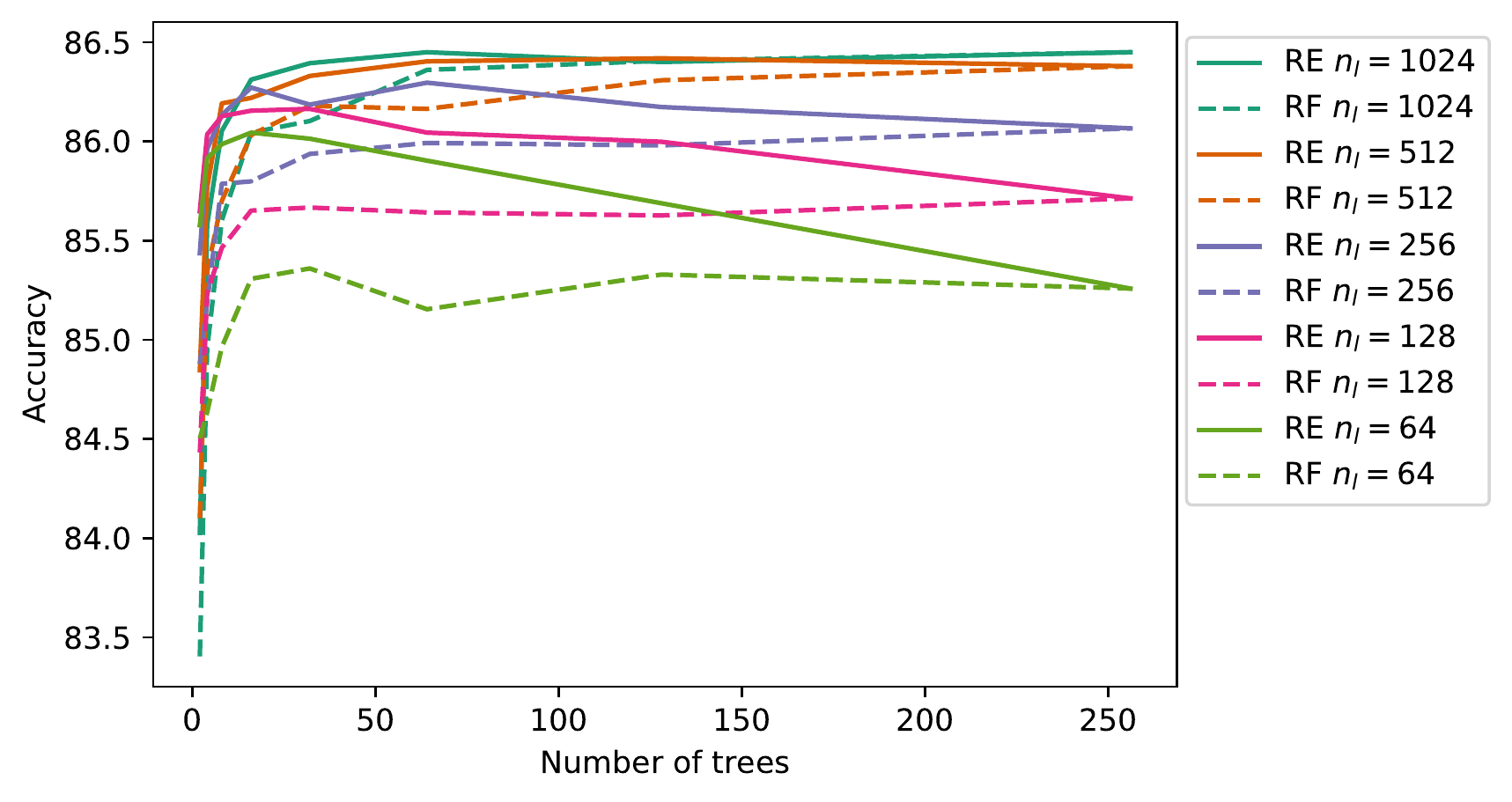}
\end{minipage}\hfill
\begin{minipage}{.49\textwidth}
    \centering 
    \resizebox{\textwidth}{!}{
        \begin{tabular}{llrrrrrr}
\toprule
     & model &   COMP &   DREP &     IC &     IE &    LMD &     RE \\
max\_leaf\_nodes & n\_estimators &        &        &        &        &        &        \\
\midrule
64   & 8.0   &  85.47 &  85.40 &  85.30 &  85.73 &  85.06 &  85.99 \\
     & 16.0  &  85.65 &  85.57 &  85.71 &  85.65 &  85.14 &  86.04 \\
     & 32.0  &  85.66 &  85.43 &  85.81 &  85.65 &  85.34 &  86.01 \\
     & 64.0  &  85.57 &  85.34 &  85.64 &  85.78 &  85.30 &  85.90 \\
     & 128.0 &  85.52 &  85.33 &  85.64 &  85.66 &  85.27 &  85.69 \\
128  & 8.0   &  85.78 &  85.62 &  85.80 &  85.88 &  85.38 &  86.13 \\
     & 16.0  &  85.92 &  85.74 &  85.98 &  85.91 &  85.74 &  86.16 \\
     & 32.0  &  85.89 &  85.76 &  86.01 &  85.96 &  85.59 &  86.16 \\
     & 64.0  &  85.98 &  85.80 &  86.05 &  85.84 &  85.69 &  86.04 \\
     & 128.0 &  85.92 &  85.70 &  86.00 &  85.88 &  85.70 &  86.00 \\
256  & 8.0   &  85.82 &  85.78 &  85.77 &  86.16 &  85.46 &  86.13 \\
     & 16.0  &  86.17 &  85.96 &  85.97 &  86.23 &  85.68 &  86.27 \\
     & 32.0  &  86.04 &  86.01 &  86.00 &  86.21 &  85.75 &  86.19 \\
     & 64.0  &  86.07 &  86.11 &  86.17 &  86.22 &  85.83 &  86.30 \\
     & 128.0 &  86.08 &  86.12 &  86.19 &  86.22 &  85.88 &  86.17 \\
512  & 8.0   &  85.78 &  85.90 &  85.93 &  86.09 &  85.35 &  86.19 \\
     & 16.0  &  86.08 &  86.00 &  86.11 &  86.20 &  85.88 &  86.22 \\
     & 32.0  &  86.15 &  86.19 &  86.28 &  86.29 &  86.02 &  86.33 \\
     & 64.0  &  86.40 &  86.17 &  86.38 &  86.41 &  86.22 &  86.40 \\
     & 128.0 &  86.36 &  86.26 &  86.38 &  86.47 &  86.26 &  86.42 \\
1024 & 8.0   &  85.83 &  85.77 &  85.91 &  86.13 &  85.36 &  86.05 \\
     & 16.0  &  86.19 &  86.06 &  86.21 &  86.29 &  85.73 &  86.31 \\
     & 32.0  &  86.30 &  86.16 &  86.46 &  86.34 &  85.90 &  86.39 \\
     & 64.0  &  86.31 &  86.35 &  86.45 &  86.45 &  86.26 &  86.45 \\
     & 128.0 &  86.47 &  86.40 &  86.52 &  86.47 &  86.33 &  86.40 \\
\bottomrule
\end{tabular}
    }
\end{minipage}
\caption{(Left) The error over the number of trees in the ensemble on the adult dataset. Dashed lines depict the Random Forest and solid lines are the corresponding pruned ensemble via Reduced Error pruning. (Right) The 5-fold cross-validation accuracy on the adult dataset. Rounded to the second decimal digit. Larger is better.}
\end{figure}

\begin{figure}[H]
\begin{minipage}{.49\textwidth}
    \centering
    \includegraphics[width=\textwidth,keepaspectratio]{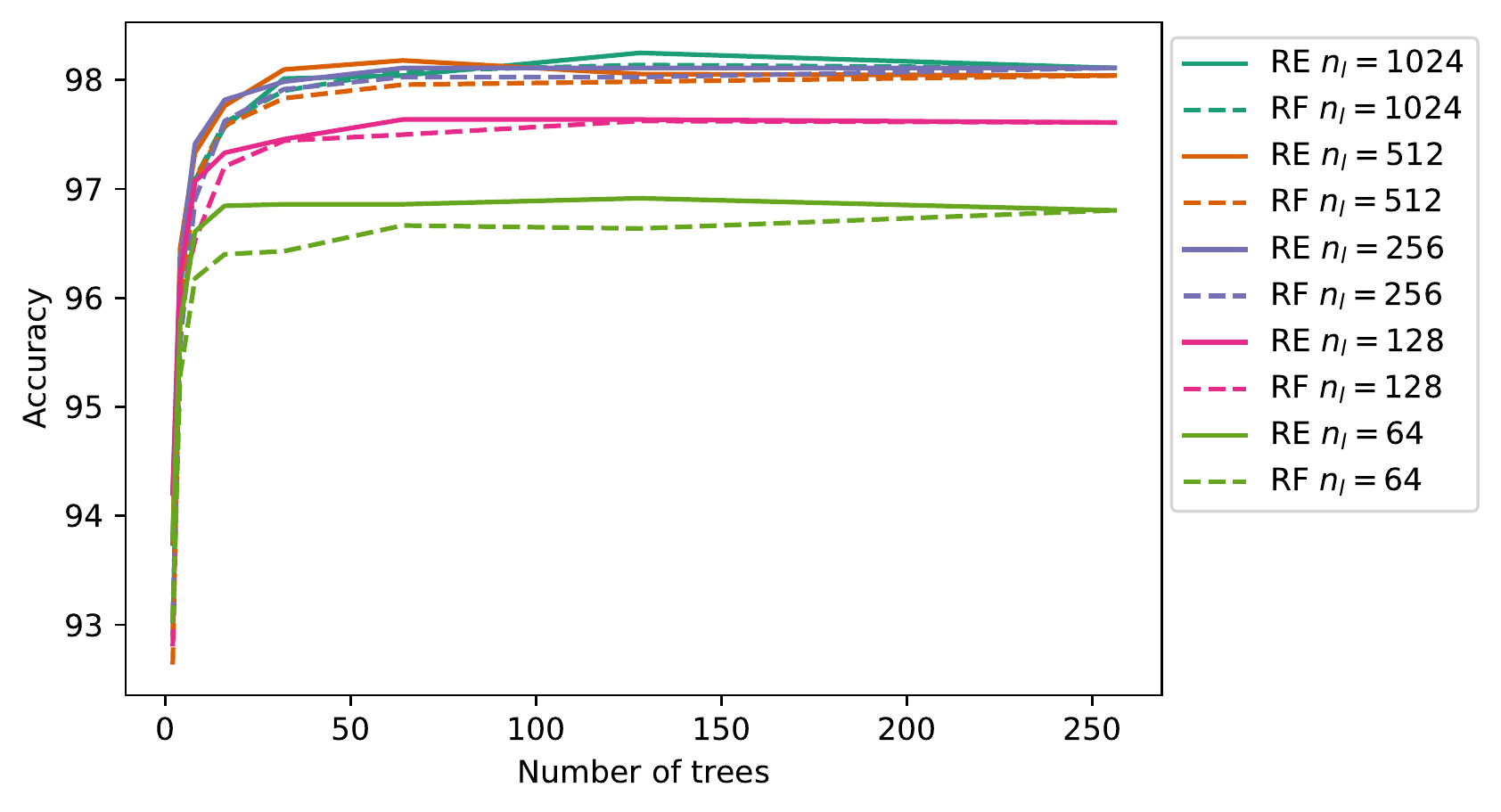}
\end{minipage}\hfill
\begin{minipage}{.49\textwidth}
    \centering 
    \resizebox{\textwidth}{!}{
        \begin{tabular}{llrrrrrr}
\toprule
     & model &   COMP &   DREP &     IC &     IE &    LMD &     RE \\
max\_leaf\_nodes & n\_estimators &        &        &        &        &        &        \\
\midrule
64   & 8.0   &  96.48 &  96.22 &  96.61 &  96.21 &  96.08 &  96.61 \\
     & 16.0  &  96.82 &  96.43 &  97.01 &  96.57 &  96.33 &  96.85 \\
     & 32.0  &  96.96 &  96.68 &  97.03 &  96.75 &  96.72 &  96.86 \\
     & 64.0  &  96.96 &  96.71 &  96.89 &  96.87 &  96.83 &  96.86 \\
     & 128.0 &  96.78 &  96.79 &  96.87 &  96.87 &  96.79 &  96.91 \\
128  & 8.0   &  97.14 &  96.87 &  97.19 &  96.93 &  96.54 &  97.07 \\
     & 16.0  &  97.57 &  97.14 &  97.65 &  97.44 &  97.14 &  97.33 \\
     & 32.0  &  97.72 &  97.36 &  97.72 &  97.72 &  97.37 &  97.46 \\
     & 64.0  &  97.69 &  97.53 &  97.98 &  97.90 &  97.55 &  97.64 \\
     & 128.0 &  97.76 &  97.62 &  97.82 &  97.73 &  97.62 &  97.64 \\
256  & 8.0   &  97.25 &  97.11 &  97.29 &  97.51 &  96.90 &  97.41 \\
     & 16.0  &  97.73 &  97.61 &  97.86 &  97.76 &  97.41 &  97.82 \\
     & 32.0  &  98.04 &  98.12 &  98.21 &  98.07 &  97.98 &  97.98 \\
     & 64.0  &  98.12 &  98.08 &  98.23 &  98.05 &  98.14 &  98.11 \\
     & 128.0 &  98.11 &  98.11 &  98.10 &  98.05 &  98.08 &  98.11 \\
512  & 8.0   &  96.96 &  97.04 &  97.68 &  97.21 &  97.08 &  97.33 \\
     & 16.0  &  97.50 &  97.57 &  98.04 &  97.80 &  97.53 &  97.76 \\
     & 32.0  &  97.93 &  97.93 &  97.96 &  97.78 &  97.90 &  98.10 \\
     & 64.0  &  98.19 &  98.17 &  98.04 &  97.98 &  98.08 &  98.18 \\
     & 128.0 &  98.07 &  98.04 &  98.12 &  98.04 &  98.11 &  98.05 \\
1024 & 8.0   &  97.23 &  97.21 &  97.28 &  97.36 &  96.89 &  97.08 \\
     & 16.0  &  97.82 &  97.75 &  97.69 &  97.93 &  97.54 &  97.57 \\
     & 32.0  &  98.05 &  98.00 &  98.03 &  97.98 &  97.89 &  98.01 \\
     & 64.0  &  98.03 &  97.98 &  98.18 &  98.07 &  98.04 &  98.04 \\
     & 128.0 &  98.23 &  98.23 &  98.17 &  98.25 &  98.01 &  98.25 \\
\bottomrule
\end{tabular}
    }
\end{minipage}
\caption{(Left) The error over the number of trees in the ensemble on the anura dataset. Dashed lines depict the Random Forest and solid lines are the corresponding pruned ensemble via Reduced Error pruning. (Right) The 5-fold cross-validation accuracy  on the anura dataset. Rounded to the second decimal digit. Larger is better.}
\end{figure}

\begin{figure}[H]
\begin{minipage}{.49\textwidth}
    \centering
    \includegraphics[width=\textwidth,keepaspectratio]{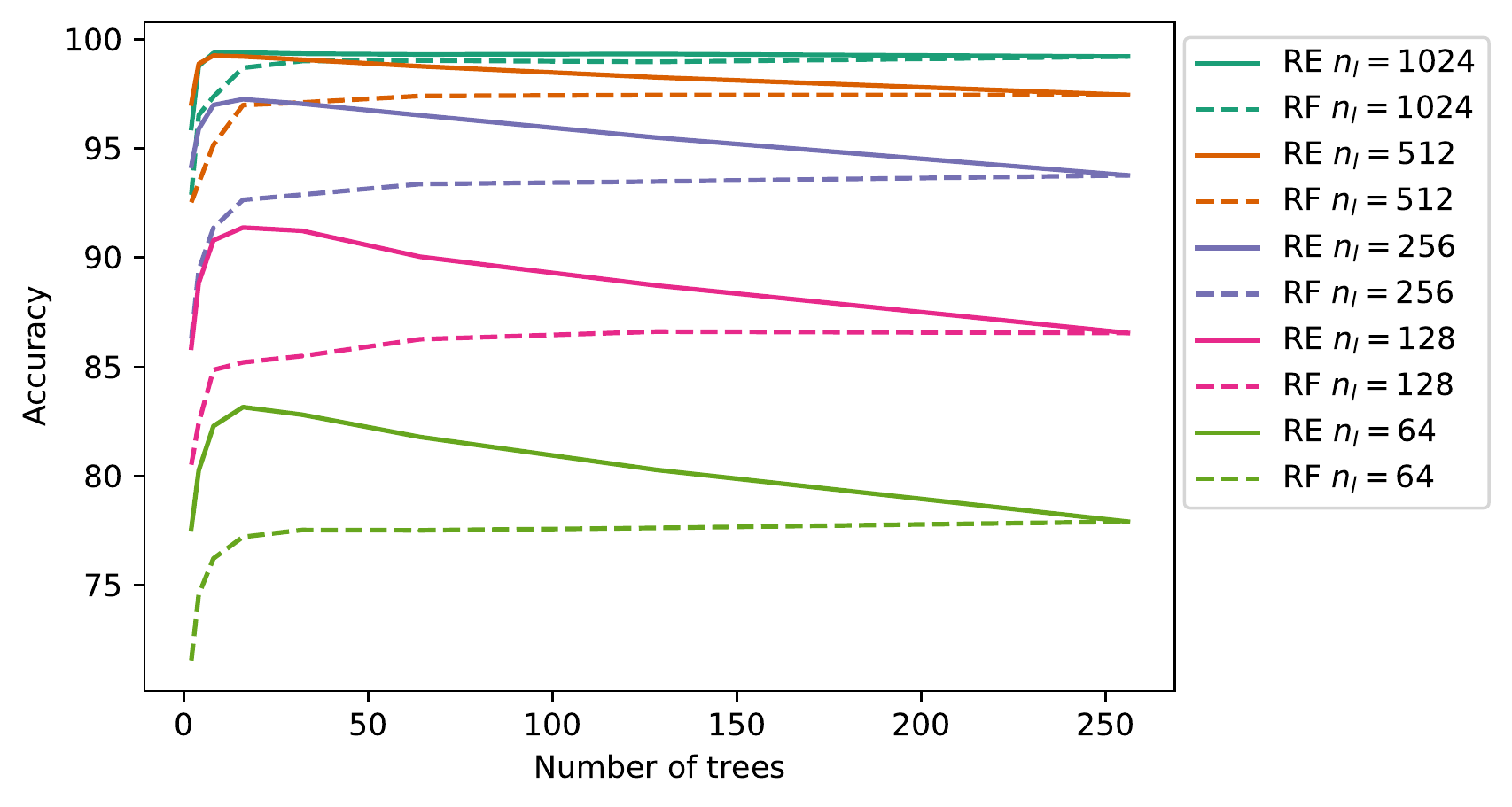}
\end{minipage}\hfill
\begin{minipage}{.49\textwidth}
    \centering 
    \resizebox{\textwidth}{!}{
        \begin{tabular}{llrrrrrr}
\toprule
     & model &   COMP &   DREP &     IC &     IE &    LMD &     RE \\
max\_leaf\_nodes & n\_estimators &        &        &        &        &        &        \\
\midrule
64   & 8.0   &  81.37 &  76.83 &  81.21 &  79.80 &  78.55 &  82.28 \\
     & 16.0  &  81.47 &  77.19 &  81.40 &  80.73 &  79.34 &  83.15 \\
     & 32.0  &  81.27 &  77.43 &  81.45 &  80.51 &  79.49 &  82.81 \\
     & 64.0  &  80.62 &  77.78 &  80.73 &  79.91 &  78.89 &  81.78 \\
     & 128.0 &  79.59 &    NaN &  79.72 &  79.19 &  78.52 &  80.28 \\
128  & 8.0   &  90.13 &  85.99 &  89.43 &  88.90 &  84.82 &  90.79 \\
     & 16.0  &  90.66 &  86.51 &  90.67 &  89.52 &  86.12 &  91.37 \\
     & 32.0  &  90.31 &  86.79 &  90.66 &  89.74 &  86.89 &  91.23 \\
     & 64.0  &  89.87 &  86.45 &  89.89 &  89.10 &  86.74 &  90.04 \\
     & 128.0 &  88.42 &    NaN &  88.55 &  88.18 &  86.77 &  88.72 \\
256  & 8.0   &  96.61 &  92.82 &  96.75 &  96.40 &  89.03 &  96.99 \\
     & 16.0  &  97.09 &  93.46 &  97.10 &  96.55 &  90.26 &  97.25 \\
     & 32.0  &  96.82 &  93.63 &  96.97 &  96.70 &  91.12 &  97.04 \\
     & 64.0  &  96.31 &    NaN &  96.55 &  96.12 &  91.63 &  96.52 \\
     & 128.0 &  95.50 &    NaN &  95.62 &  95.31 &  92.15 &  95.50 \\
512  & 8.0   &  99.02 &  96.35 &  99.18 &  99.12 &  93.31 &  99.25 \\
     & 16.0  &  99.21 &  96.56 &  99.32 &  99.15 &  93.61 &  99.20 \\
     & 32.0  &  99.06 &  97.04 &  99.13 &  98.97 &  94.23 &  99.06 \\
     & 64.0  &  98.81 &    NaN &  98.91 &  98.84 &  95.02 &  98.76 \\
     & 128.0 &  98.36 &    NaN &  98.53 &  98.46 &  96.04 &  98.25 \\
1024 & 8.0   &  99.14 &  98.41 &  99.49 &  99.54 &  94.41 &  99.37 \\
     & 16.0  &  99.28 &  98.62 &  99.70 &  99.72 &  95.42 &  99.39 \\
     & 32.0  &  99.30 &  98.88 &  99.76 &  99.74 &  96.50 &  99.33 \\
     & 64.0  &  99.30 &  99.16 &  99.74 &  99.73 &  97.22 &  99.30 \\
     & 128.0 &  99.36 &    NaN &  99.59 &  99.59 &  98.01 &  99.32 \\
\bottomrule
\end{tabular}
    }
\end{minipage}
\caption{(Left) The error over the number of trees in the ensemble on the avila dataset. Dashed lines depict the Random Forest and solid lines are the corresponding pruned ensemble via Reduced Error pruning. (Right) The 5-fold cross-validation accuracy  on the avila dataset. Rounded to the second decimal digit. Larger is better.}
\end{figure}

\begin{figure}[H]
\begin{minipage}{.49\textwidth}
    \centering
    \includegraphics[width=\textwidth,keepaspectratio]{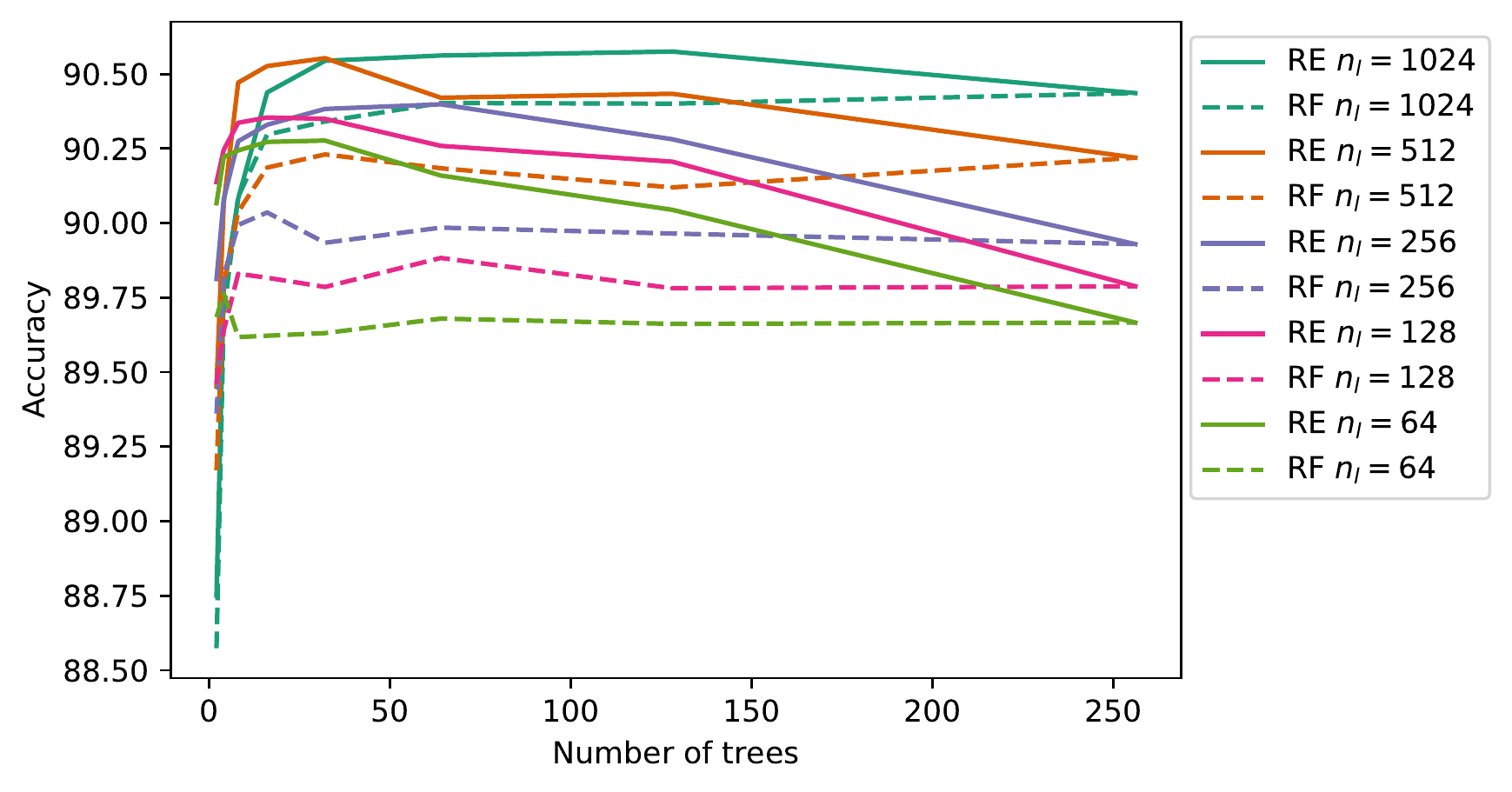}
\end{minipage}\hfill
\begin{minipage}{.49\textwidth}
    \centering 
    \resizebox{\textwidth}{!}{
        \begin{tabular}{llrrrrrr}
\toprule
     & model &   COMP &   DREP &     IC &     IE &    LMD &     RE \\
max\_leaf\_nodes & n\_estimators &        &        &        &        &        &        \\
\midrule
64   & 8.0   &  89.98 &  89.74 &  90.21 &  90.21 &  89.66 &  90.24 \\
     & 16.0  &  90.02 &  89.70 &  90.17 &  90.21 &  89.72 &  90.27 \\
     & 32.0  &  90.00 &  89.69 &  90.13 &  90.17 &  89.69 &  90.28 \\
     & 64.0  &  90.00 &  89.65 &  90.14 &  90.03 &  89.68 &  90.16 \\
     & 128.0 &  89.86 &  89.65 &  89.91 &  89.78 &  89.70 &  90.04 \\
128  & 8.0   &  90.05 &  89.85 &  90.15 &  90.29 &  89.82 &  90.34 \\
     & 16.0  &  90.12 &  89.83 &  90.22 &  90.27 &  89.80 &  90.35 \\
     & 32.0  &  90.16 &  89.84 &  90.22 &  90.32 &  89.90 &  90.35 \\
     & 64.0  &  90.21 &  89.83 &  90.29 &  90.21 &  89.86 &  90.26 \\
     & 128.0 &  90.03 &  89.79 &  90.05 &  90.00 &  89.84 &  90.21 \\
256  & 8.0   &  90.15 &  89.98 &  90.26 &  90.38 &  89.83 &  90.27 \\
     & 16.0  &  90.27 &  89.94 &  90.29 &  90.39 &  89.85 &  90.33 \\
     & 32.0  &  90.25 &  90.02 &  90.34 &  90.37 &  90.07 &  90.38 \\
     & 64.0  &  90.31 &  89.95 &  90.35 &  90.39 &  90.03 &  90.40 \\
     & 128.0 &  90.21 &  89.94 &  90.25 &  90.20 &  90.01 &  90.28 \\
512  & 8.0   &  90.30 &  90.08 &  90.27 &  90.44 &  89.95 &  90.47 \\
     & 16.0  &  90.32 &  90.28 &  90.40 &  90.53 &  90.16 &  90.53 \\
     & 32.0  &  90.41 &  90.13 &  90.51 &  90.50 &  90.29 &  90.55 \\
     & 64.0  &  90.45 &  90.20 &  90.43 &  90.47 &  90.28 &  90.42 \\
     & 128.0 &  90.40 &  90.18 &  90.49 &  90.49 &  90.29 &  90.43 \\
1024 & 8.0   &  90.30 &  90.12 &  90.36 &  90.28 &  90.12 &  90.08 \\
     & 16.0  &  90.50 &  90.43 &  90.51 &  90.47 &  90.28 &  90.44 \\
     & 32.0  &  90.54 &  90.39 &  90.55 &  90.44 &  90.34 &  90.54 \\
     & 64.0  &  90.56 &  90.38 &  90.58 &  90.48 &  90.40 &  90.56 \\
     & 128.0 &  90.55 &  90.39 &  90.59 &  90.55 &  90.42 &  90.58 \\
\bottomrule
\end{tabular}
    }
\end{minipage}
\caption{(Left) The error over the number of trees in the ensemble on the bank dataset. Dashed lines depict the Random Forest and solid lines are the corresponding pruned ensemble via Reduced Error pruning. (Right) The 5-fold cross-validation accuracy  on the bank dataset. Rounded to the second decimal digit. Larger is better.}
\end{figure}

\begin{figure}[H]
\begin{minipage}{.49\textwidth}
    \centering
    \includegraphics[width=\textwidth,keepaspectratio]{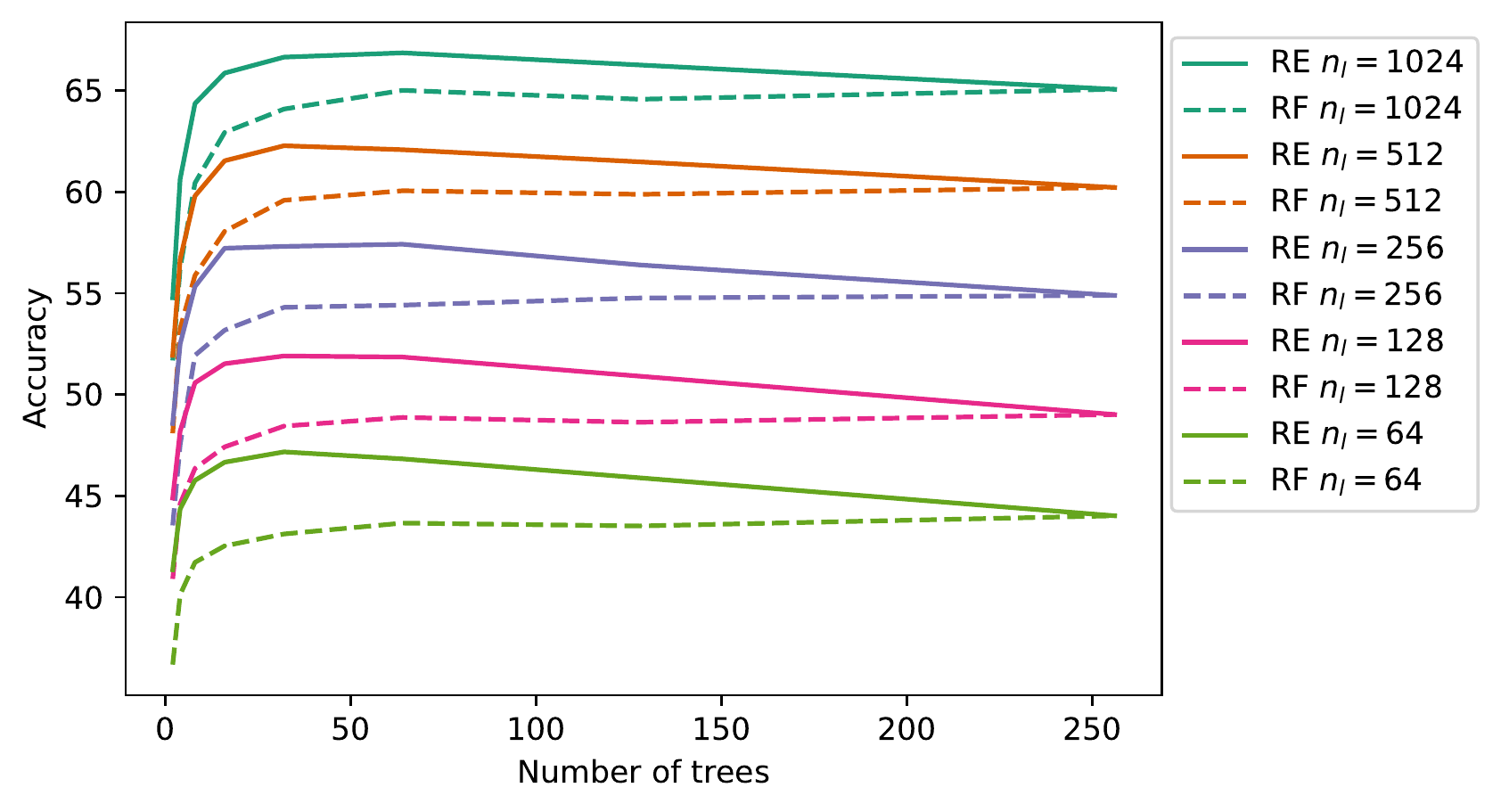}
\end{minipage}\hfill
\begin{minipage}{.49\textwidth}
    \centering 
    \resizebox{\textwidth}{!}{
        \begin{tabular}{llrrrrrr}
\toprule
     & model &   COMP &   DREP &     IC &     IE &    LMD &     RE \\
max\_leaf\_nodes & n\_estimators &        &        &        &        &        &        \\
\midrule
64   & 8.0   &  44.23 &  42.94 &  43.93 &  43.17 &  43.41 &  45.76 \\
     & 16.0  &  44.91 &  43.75 &  45.27 &  43.58 &  44.26 &  46.66 \\
     & 32.0  &  45.33 &  43.82 &  45.84 &  43.77 &  44.56 &  47.17 \\
     & 64.0  &  45.23 &    NaN &  45.25 &  43.82 &  44.36 &  46.82 \\
     & 128.0 &    NaN &    NaN &  44.77 &  44.01 &  44.41 &    NaN \\
128  & 8.0   &  49.74 &  47.21 &  49.33 &  48.18 &  48.47 &  50.58 \\
     & 16.0  &  50.92 &  48.56 &  50.51 &  48.31 &  49.19 &  51.52 \\
     & 32.0  &  50.79 &  49.00 &  50.95 &  48.94 &  49.93 &  51.90 \\
     & 64.0  &  50.53 &    NaN &  50.47 &  48.87 &  49.71 &  51.85 \\
     & 128.0 &  49.83 &    NaN &  50.14 &  48.96 &  49.49 &    NaN \\
256  & 8.0   &  54.06 &  52.57 &  53.66 &  53.77 &  52.41 &  55.33 \\
     & 16.0  &  55.65 &  53.46 &  55.61 &  54.48 &  54.50 &  57.22 \\
     & 32.0  &  56.04 &  54.09 &  56.49 &  54.89 &  55.10 &  57.31 \\
     & 64.0  &  56.43 &    NaN &  56.52 &  55.04 &  55.34 &  57.42 \\
     & 128.0 &    NaN &    NaN &  55.98 &  54.88 &  55.35 &  56.39 \\
512  & 8.0   &  58.90 &  56.79 &  59.24 &  58.15 &  56.89 &  59.80 \\
     & 16.0  &  61.14 &  58.38 &  61.31 &  59.44 &  59.37 &  61.54 \\
     & 32.0  &  61.79 &    NaN &  61.97 &  60.41 &  60.51 &  62.28 \\
     & 64.0  &  61.63 &    NaN &  62.08 &  60.77 &  60.89 &  62.08 \\
     & 128.0 &  61.28 &    NaN &  61.44 &  60.52 &  60.80 &  61.48 \\
1024 & 8.0   &  63.58 &  61.36 &  63.85 &  62.80 &  61.38 &  64.36 \\
     & 16.0  &  65.93 &  62.91 &  65.95 &  64.25 &  63.82 &  65.86 \\
     & 32.0  &  66.69 &    NaN &  66.82 &  65.07 &  65.05 &  66.65 \\
     & 64.0  &  66.48 &    NaN &  66.77 &  65.30 &  65.39 &  66.86 \\
     & 128.0 &    NaN &    NaN &  66.39 &  65.30 &  65.33 &    NaN \\
\bottomrule
\end{tabular}
    }
\end{minipage}
\caption{(Left) The error over the number of trees in the ensemble on the chess dataset. Dashed lines depict the Random Forest and solid lines are the corresponding pruned ensemble via Reduced Error pruning. (Right) The 5-fold cross-validation accuracy  on the chess dataset. Rounded to the second decimal digit. Larger is better.}
\end{figure}

\begin{figure}[H]
\begin{minipage}{.49\textwidth}
    \centering
    \includegraphics[width=\textwidth,keepaspectratio]{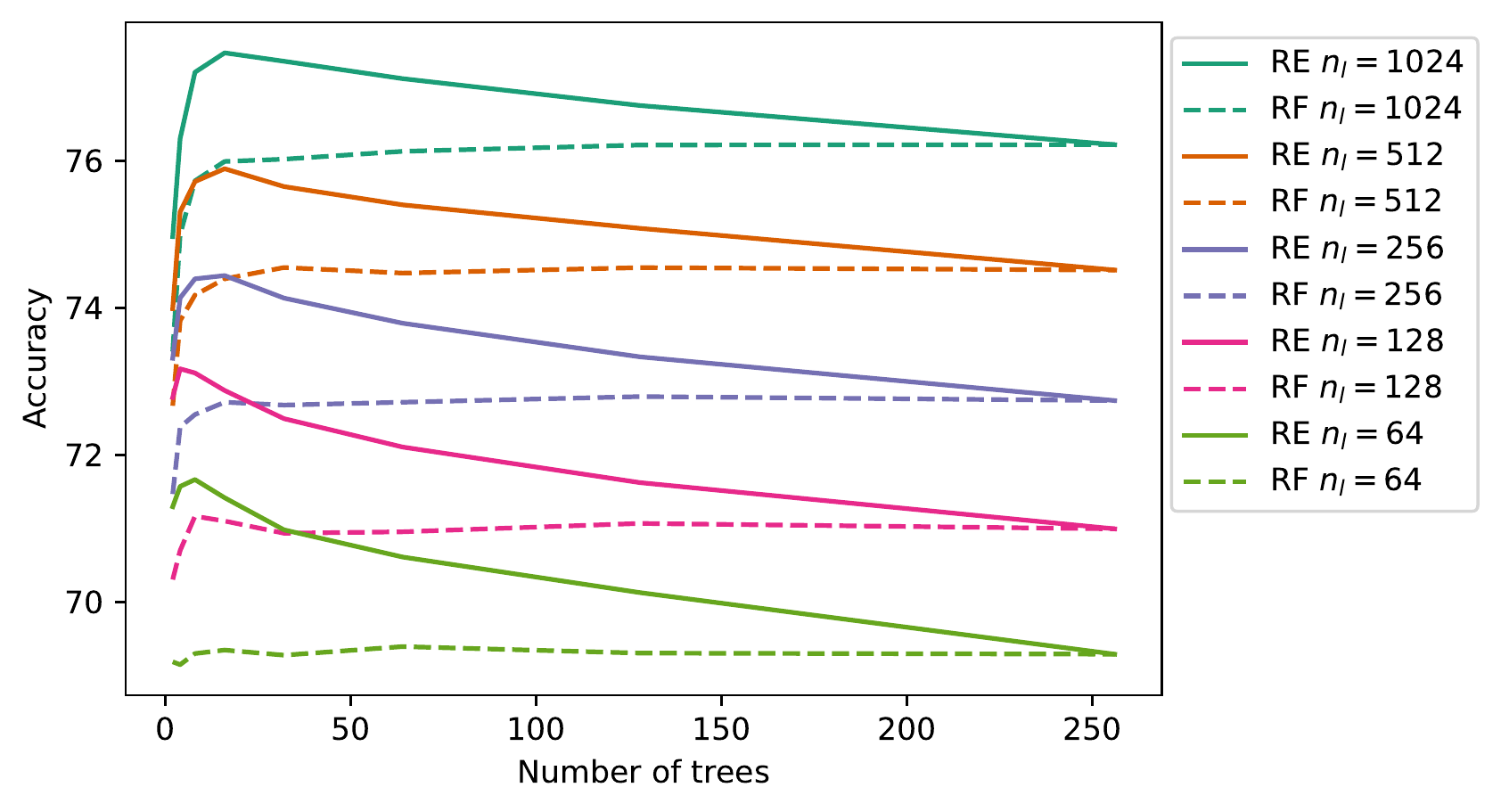}
\end{minipage}\hfill
\begin{minipage}{.49\textwidth}
    \centering 
    \resizebox{\textwidth}{!}{
        \begin{tabular}{llrrrrrr}
\toprule
     & model &   COMP &   DREP &     IC &     IE &    LMD &     RE \\
max\_leaf\_nodes & n\_estimators &        &        &        &        &        &        \\
\midrule
64   & 8.0   &  70.73 &  69.92 &  70.45 &  71.14 &  69.68 &  71.66 \\
     & 16.0  &  70.32 &  69.60 &  70.38 &  70.76 &  69.41 &  71.42 \\
     & 32.0  &  70.12 &  69.41 &  70.07 &  70.48 &  69.33 &  70.98 \\
     & 64.0  &  69.88 &    NaN &  69.87 &  70.30 &  69.31 &  70.61 \\
     & 128.0 &  69.67 &    NaN &  69.61 &  70.00 &  69.26 &  70.12 \\
128  & 8.0   &  72.32 &  71.45 &  72.54 &  72.62 &  71.43 &  73.11 \\
     & 16.0  &  72.15 &  71.36 &  72.17 &  72.34 &  71.17 &  72.87 \\
     & 32.0  &  71.89 &  71.13 &  72.00 &  72.05 &  71.13 &  72.49 \\
     & 64.0  &  71.66 &    NaN &  71.64 &  71.80 &  71.11 &  72.11 \\
     & 128.0 &  71.36 &    NaN &  71.36 &  71.47 &  71.07 &  71.62 \\
256  & 8.0   &  73.97 &  72.95 &  74.17 &  73.96 &  72.98 &  74.39 \\
     & 16.0  &  73.79 &  72.88 &  73.93 &  73.91 &  72.89 &  74.44 \\
     & 32.0  &  73.63 &  72.79 &  73.69 &  73.76 &  73.08 &  74.13 \\
     & 64.0  &  73.50 &    NaN &  73.53 &  73.46 &  72.93 &  73.79 \\
     & 128.0 &  73.16 &    NaN &  73.20 &  73.19 &  72.86 &  73.33 \\
512  & 8.0   &  75.46 &  74.31 &  75.51 &  75.40 &  74.73 &  75.72 \\
     & 16.0  &  75.44 &  74.30 &  75.56 &  75.47 &  74.70 &  75.89 \\
     & 32.0  &  75.48 &  74.54 &  75.50 &  75.42 &  74.71 &  75.65 \\
     & 64.0  &  75.30 &    NaN &  75.26 &  75.25 &  74.69 &  75.40 \\
     & 128.0 &  75.02 &    NaN &  75.02 &  74.96 &  74.63 &  75.08 \\
1024 & 8.0   &  76.93 &  75.87 &  76.98 &  77.08 &  75.96 &  77.20 \\
     & 16.0  &  77.14 &  75.94 &  77.41 &  77.36 &  76.23 &  77.47 \\
     & 32.0  &  77.22 &  76.15 &  77.44 &  77.27 &  76.33 &  77.35 \\
     & 64.0  &  77.06 &    NaN &  77.13 &  77.09 &  76.28 &  77.12 \\
     & 128.0 &  76.79 &    NaN &  76.82 &  76.75 &  76.27 &  76.75 \\
\bottomrule
\end{tabular}
    }
\end{minipage}
\caption{(Left) The error over the number of trees in the ensemble on the connect dataset. Dashed lines depict the Random Forest and solid lines are the corresponding pruned ensemble via Reduced Error pruning. (Right) The 5-fold cross-validation accuracy  on the connect dataset. Rounded to the second decimal digit. Larger is better.}
\end{figure}

\begin{figure}[H]
\begin{minipage}{.49\textwidth}
    \centering
    \includegraphics[width=\textwidth,keepaspectratio]{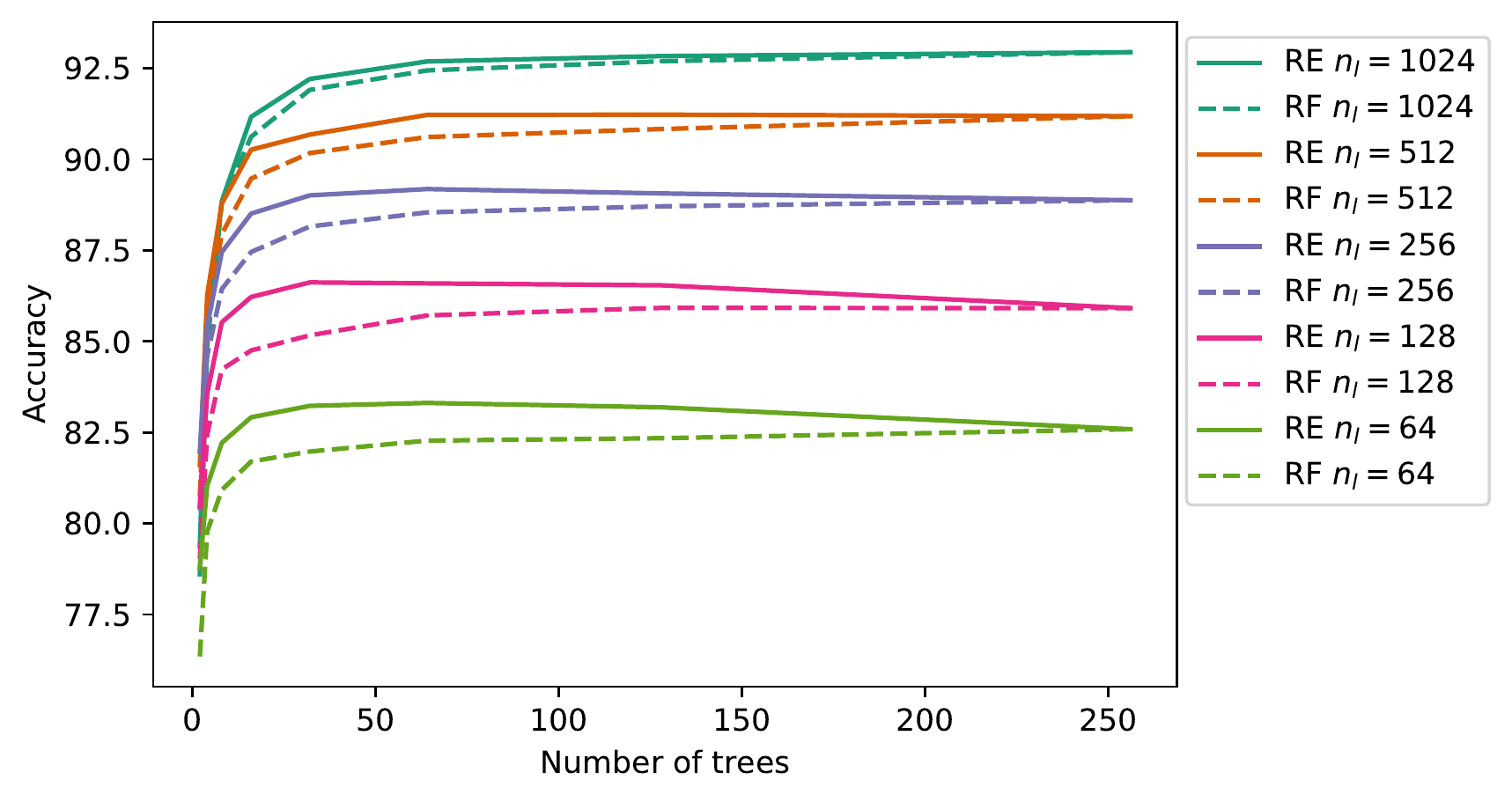}
\end{minipage}\hfill
\begin{minipage}{.49\textwidth}
    \centering 
    \resizebox{\textwidth}{!}{
        \begin{tabular}{llrrrrrr}
\toprule
     & model &   COMP &   DREP &     IC &     IE &    LMD &     RE \\
max\_leaf\_nodes & n\_estimators &        &        &        &        &        &        \\
\midrule
64   & 8.0   &  81.86 &  80.79 &  81.71 &  81.34 &  81.46 &  82.22 \\
     & 16.0  &  83.04 &  81.96 &  83.09 &  81.69 &  82.24 &  82.92 \\
     & 32.0  &  83.09 &  82.33 &  83.10 &  82.22 &  82.60 &  83.23 \\
     & 64.0  &  83.06 &  82.55 &  83.05 &  82.30 &  82.96 &  83.31 \\
     & 128.0 &  83.17 &  82.38 &  83.10 &  82.49 &  82.87 &  83.19 \\
128  & 8.0   &  85.12 &  84.08 &  84.69 &  84.81 &  83.85 &  85.53 \\
     & 16.0  &  85.93 &  85.15 &  86.04 &  85.23 &  85.25 &  86.22 \\
     & 32.0  &  86.34 &  85.49 &  86.38 &  85.76 &  85.65 &  86.62 \\
     & 64.0  &  86.39 &  85.55 &  86.49 &  85.96 &  86.04 &  86.60 \\
     & 128.0 &  86.40 &  85.75 &  86.27 &  86.00 &  86.03 &  86.54 \\
256  & 8.0   &  87.37 &  86.24 &  87.14 &  87.16 &  86.36 &  87.46 \\
     & 16.0  &  88.36 &  87.64 &  88.50 &  88.23 &  87.78 &  88.51 \\
     & 32.0  &  88.97 &  88.02 &  89.07 &  88.70 &  88.37 &  89.01 \\
     & 64.0  &  88.91 &  88.59 &  89.03 &  88.90 &  88.62 &  89.19 \\
     & 128.0 &  88.97 &  88.70 &  89.15 &  88.97 &  88.77 &  89.07 \\
512  & 8.0   &  88.36 &  88.51 &  88.96 &  88.78 &  87.44 &  88.79 \\
     & 16.0  &  90.18 &  89.89 &  90.28 &  89.87 &  89.45 &  90.27 \\
     & 32.0  &  91.11 &  90.30 &  91.34 &  90.67 &  90.37 &  90.68 \\
     & 64.0  &  91.04 &  90.75 &  91.56 &  91.16 &  90.79 &  91.22 \\
     & 128.0 &  91.22 &  90.91 &  91.41 &  91.30 &  90.87 &  91.23 \\
1024 & 8.0   &  89.45 &  89.26 &  89.51 &  89.70 &  88.30 &  88.85 \\
     & 16.0  &  90.77 &  90.89 &  91.64 &  91.48 &  90.86 &  91.17 \\
     & 32.0  &  92.25 &  92.30 &  92.64 &  92.60 &  91.82 &  92.21 \\
     & 64.0  &  92.55 &  92.54 &  93.09 &  92.86 &  92.38 &  92.69 \\
     & 128.0 &  92.85 &  92.85 &  93.17 &  92.98 &  92.70 &  92.84 \\
\bottomrule
\end{tabular}
    }
\end{minipage}
\caption{(Left) The error over the number of trees in the ensemble on the eeg dataset. Dashed lines depict the Random Forest and solid lines are the corresponding pruned ensemble via Reduced Error pruning. (Right) The 5-fold cross-validation accuracy  on the eeg dataset. Rounded to the second decimal digit. Larger is better.}
\end{figure}

\begin{figure}[H]
\begin{minipage}{.49\textwidth}
    \centering
    \includegraphics[width=\textwidth,keepaspectratio]{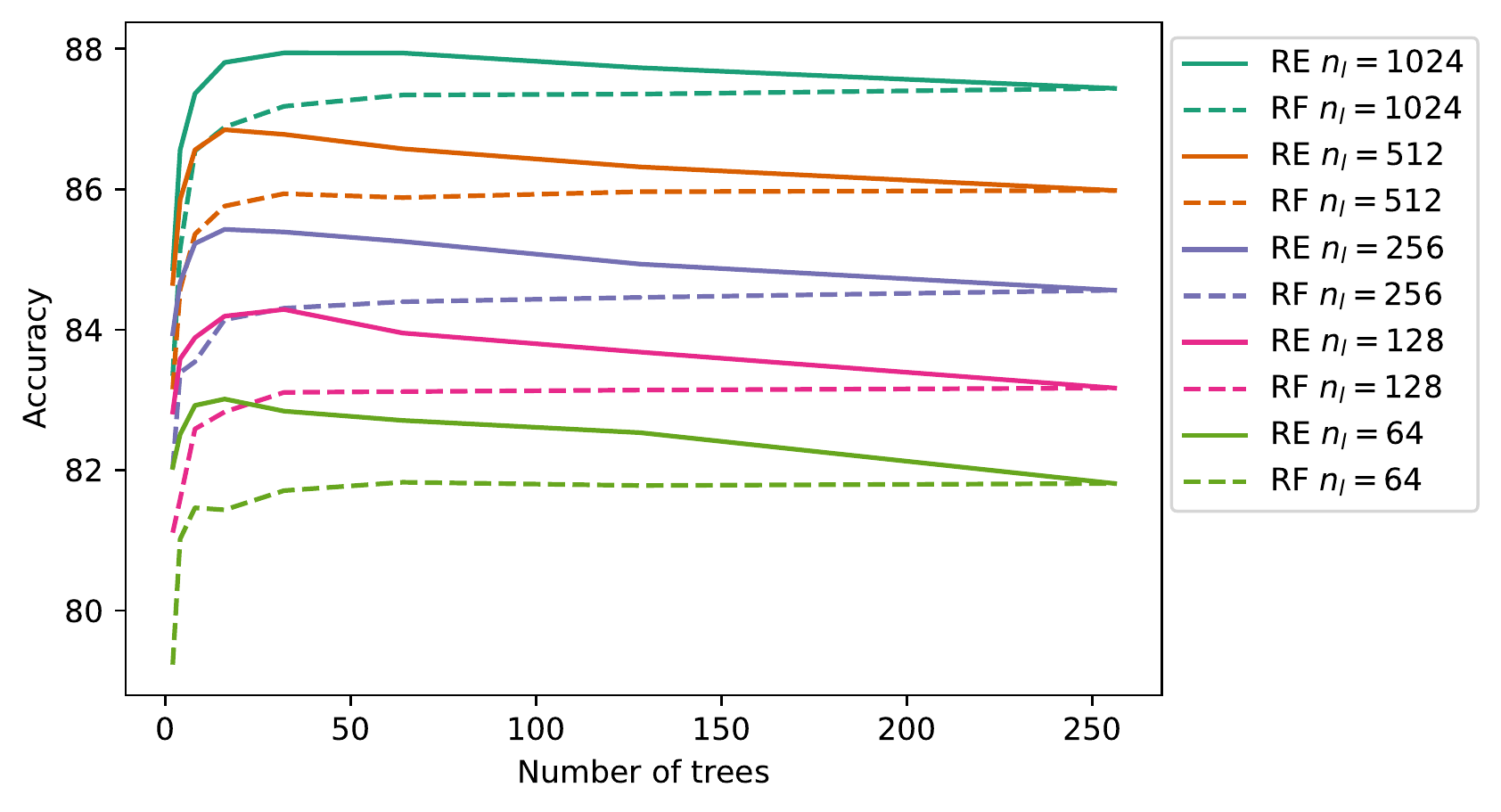}
\end{minipage}\hfill
\begin{minipage}{.49\textwidth}
    \centering 
    \resizebox{\textwidth}{!}{
        \begin{tabular}{llrrrrrr}
\toprule
     & model &   COMP &   DREP &     IC &     IE &    LMD &     RE \\
max\_leaf\_nodes & n\_estimators &        &        &        &        &        &        \\
\midrule
64   & 8.0   &  81.77 &  81.73 &  79.40 &  82.28 &  80.97 &  82.92 \\
     & 16.0  &  82.06 &  81.85 &  81.97 &  82.52 &  81.59 &  83.01 \\
     & 32.0  &  82.23 &  81.79 &  82.40 &  82.55 &  81.80 &  82.84 \\
     & 64.0  &  82.20 &  81.74 &  82.38 &  82.50 &  81.82 &  82.71 \\
     & 128.0 &  82.07 &  81.80 &  82.19 &  82.13 &  81.79 &  82.53 \\
128  & 8.0   &  83.42 &  82.75 &  82.75 &  83.58 &  82.82 &  83.89 \\
     & 16.0  &  83.50 &  82.97 &  83.51 &  83.80 &  83.30 &  84.19 \\
     & 32.0  &  83.75 &  82.97 &  83.75 &  83.66 &  83.20 &  84.29 \\
     & 64.0  &  83.65 &  83.09 &  83.74 &  83.52 &  83.14 &  83.95 \\
     & 128.0 &  83.48 &  83.10 &  83.56 &  83.41 &  83.19 &  83.68 \\
256  & 8.0   &  84.32 &  84.16 &  83.85 &  84.90 &  83.90 &  85.23 \\
     & 16.0  &  84.83 &  84.45 &  84.93 &  85.00 &  84.26 &  85.43 \\
     & 32.0  &  85.03 &  84.61 &  85.05 &  85.03 &  84.39 &  85.39 \\
     & 64.0  &  84.94 &  84.50 &  85.03 &  84.82 &  84.51 &  85.26 \\
     & 128.0 &  84.83 &  84.51 &  84.94 &  84.64 &  84.58 &  84.94 \\
512  & 8.0   &  85.81 &  85.48 &  86.04 &  86.35 &  84.78 &  86.56 \\
     & 16.0  &  86.45 &  85.72 &  86.58 &  86.46 &  85.49 &  86.85 \\
     & 32.0  &  86.35 &  85.82 &  86.68 &  86.47 &  85.64 &  86.78 \\
     & 64.0  &  86.28 &  85.96 &  86.56 &  86.37 &  85.69 &  86.58 \\
     & 128.0 &  86.22 &  85.93 &  86.28 &  86.21 &  85.86 &  86.32 \\
1024 & 8.0   &  86.95 &  86.45 &  87.19 &  87.44 &  86.18 &  87.36 \\
     & 16.0  &  87.46 &  86.90 &  87.85 &  87.77 &  86.74 &  87.81 \\
     & 32.0  &  87.74 &  87.18 &  88.04 &  87.91 &  87.13 &  87.94 \\
     & 64.0  &  87.84 &  87.32 &  87.94 &  87.75 &  87.21 &  87.94 \\
     & 128.0 &  87.65 &  87.38 &  87.80 &  87.64 &  87.36 &  87.73 \\
\bottomrule
\end{tabular}
    }
\end{minipage}
\caption{(Left) The error over the number of trees in the ensemble on the elec dataset. Dashed lines depict the Random Forest and solid lines are the corresponding pruned ensemble via Reduced Error pruning. (Right) The 5-fold cross-validation accuracy  on the elec dataset. Rounded to the second decimal digit. Larger is better.}
\end{figure}

\begin{figure}[H]
\begin{minipage}{.49\textwidth}
    \centering
    \includegraphics[width=\textwidth,keepaspectratio]{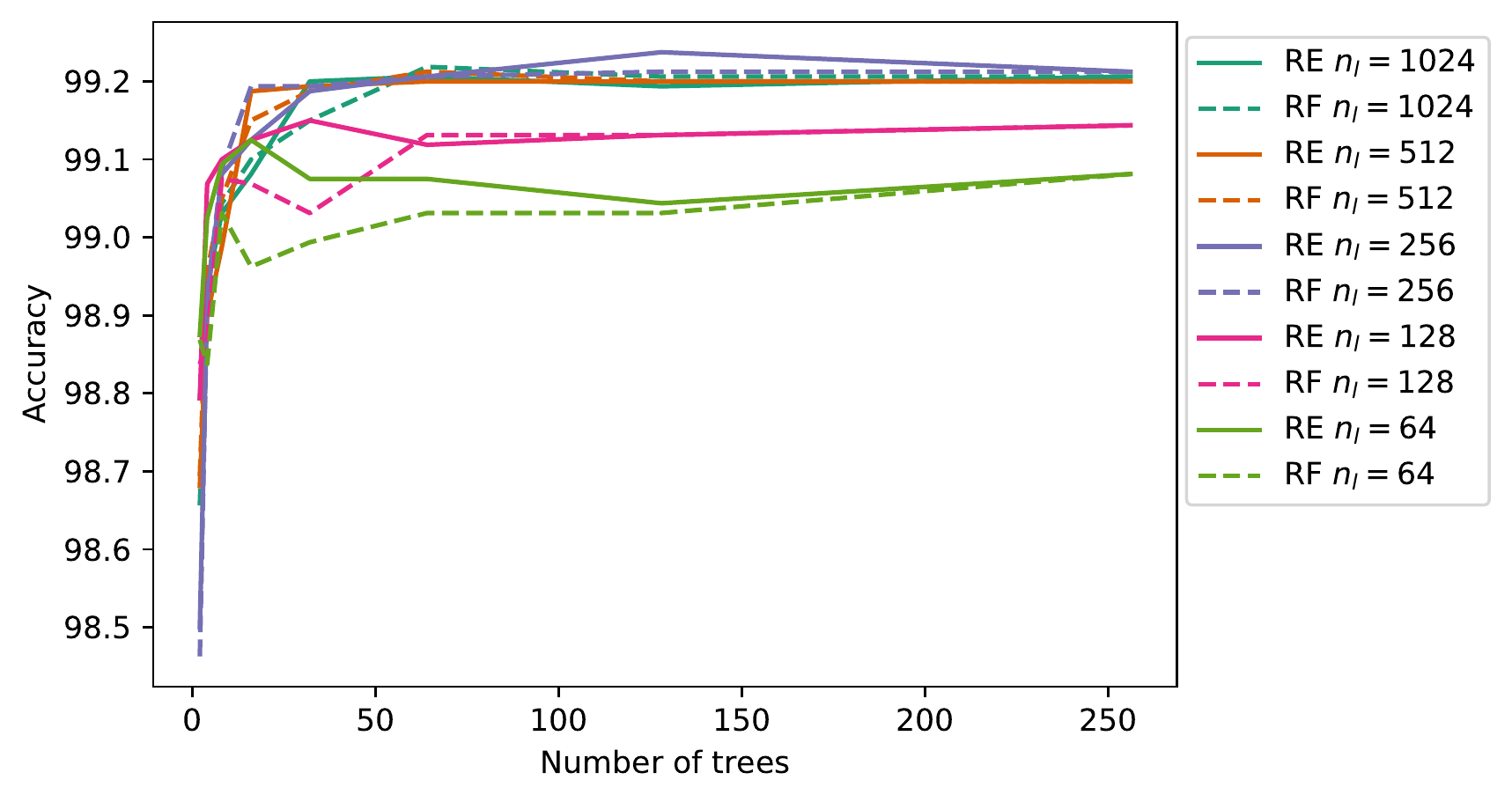}
\end{minipage}\hfill
\begin{minipage}{.49\textwidth}
    \centering 
    \resizebox{\textwidth}{!}{
        \begin{tabular}{llrrrrrr}
\toprule
     & model &   COMP &   DREP &     IC &     IE &    LMD &     RE \\
max\_leaf\_nodes & n\_estimators &        &        &        &        &        &        \\
\midrule
64   & 8.0   &  99.04 &  98.97 &  99.09 &  99.02 &  98.95 &  99.09 \\
     & 16.0  &  99.12 &  99.01 &  99.08 &  99.05 &  99.02 &  99.12 \\
     & 32.0  &  99.05 &  99.04 &  99.09 &  99.02 &  99.02 &  99.08 \\
     & 64.0  &  99.05 &  99.06 &  99.07 &  99.07 &  99.04 &  99.08 \\
     & 128.0 &  99.06 &  99.09 &  99.09 &  99.10 &  99.04 &  99.04 \\
128  & 8.0   &  99.06 &  99.02 &  99.11 &  99.10 &  99.03 &  99.10 \\
     & 16.0  &  99.08 &  99.09 &  99.12 &  99.13 &  99.13 &  99.12 \\
     & 32.0  &  99.08 &  99.04 &  99.09 &  99.14 &  99.15 &  99.15 \\
     & 64.0  &  99.06 &  99.11 &  99.09 &  99.14 &  99.15 &  99.12 \\
     & 128.0 &  99.11 &  99.16 &  99.16 &  99.14 &  99.16 &  99.13 \\
256  & 8.0   &  99.11 &  99.11 &  99.12 &  99.08 &  99.16 &  99.08 \\
     & 16.0  &  99.22 &  99.12 &  99.26 &  99.18 &  99.24 &  99.12 \\
     & 32.0  &  99.31 &  99.21 &  99.28 &  99.18 &  99.23 &  99.19 \\
     & 64.0  &  99.23 &  99.15 &  99.25 &  99.25 &  99.22 &  99.21 \\
     & 128.0 &  99.23 &  99.21 &  99.27 &  99.24 &  99.23 &  99.24 \\
512  & 8.0   &  99.07 &  99.03 &  99.12 &  99.01 &  99.02 &  98.99 \\
     & 16.0  &  99.14 &  99.12 &  99.18 &  99.13 &  99.14 &  99.19 \\
     & 32.0  &  99.19 &  99.18 &  99.18 &  99.19 &  99.21 &  99.19 \\
     & 64.0  &  99.20 &  99.21 &  99.21 &  99.20 &  99.18 &  99.20 \\
     & 128.0 &  99.19 &  99.19 &  99.21 &  99.23 &  99.22 &  99.20 \\
1024 & 8.0   &  99.04 &  99.07 &  99.14 &  99.03 &  99.01 &  99.03 \\
     & 16.0  &  99.12 &  99.22 &  99.16 &  99.16 &  99.15 &  99.08 \\
     & 32.0  &  99.17 &  99.21 &  99.22 &  99.17 &  99.20 &  99.20 \\
     & 64.0  &  99.21 &  99.23 &  99.24 &  99.20 &  99.22 &  99.21 \\
     & 128.0 &  99.21 &  99.22 &  99.21 &  99.18 &  99.21 &  99.19 \\
\bottomrule
\end{tabular}
    }
\end{minipage}
\caption{(Left) The error over the number of trees in the ensemble on the ida2016 dataset. Dashed lines depict the Random Forest and solid lines are the corresponding pruned ensemble via Reduced Error pruning. (Right) The 5-fold cross-validation accuracy  on the ida2016 dataset. Rounded to the second decimal digit. Larger is better. }
\end{figure}

\begin{figure}[H]
\begin{minipage}{.49\textwidth}
    \centering
    \includegraphics[width=\textwidth,keepaspectratio]{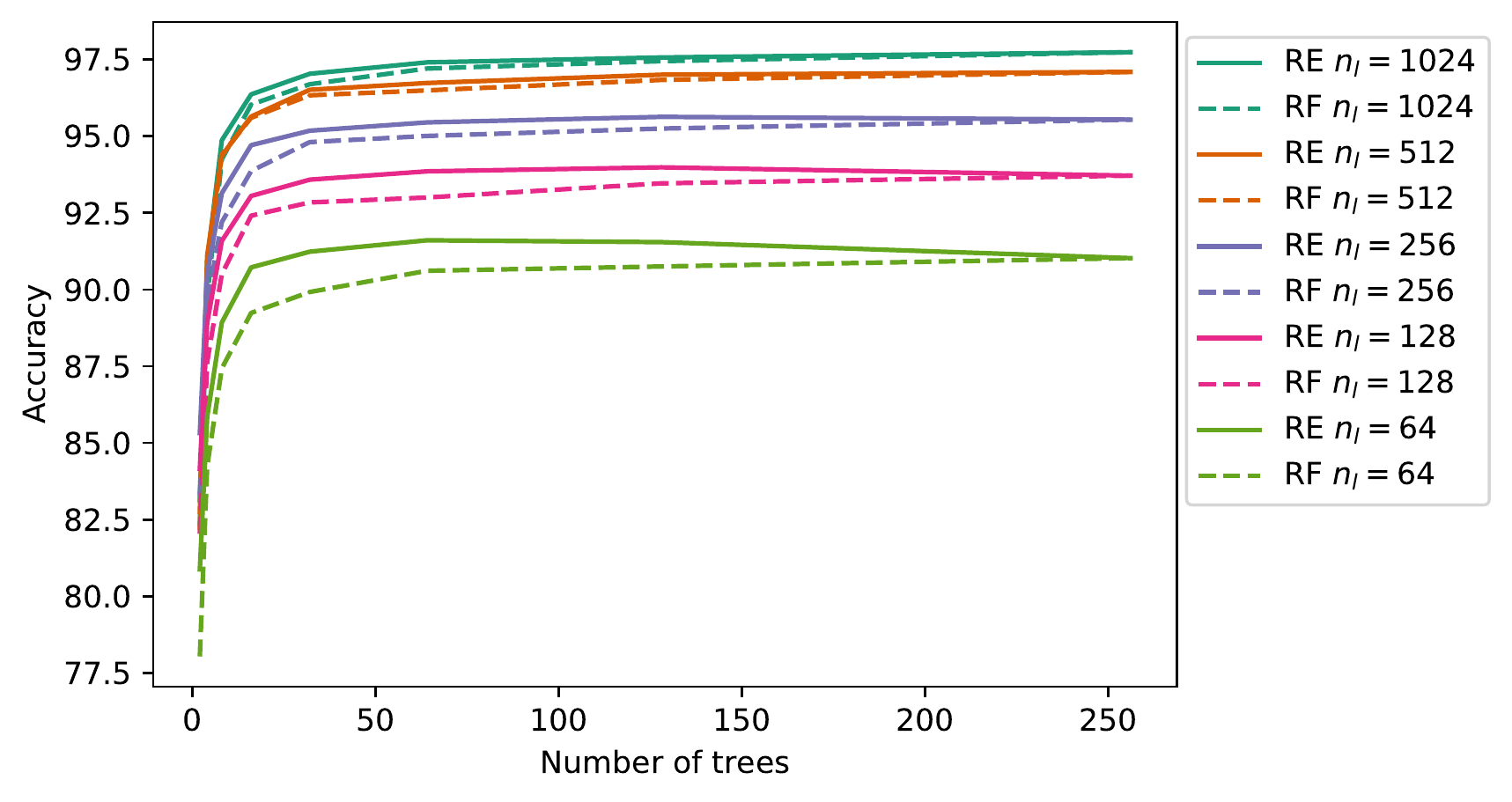}
\end{minipage}\hfill
\begin{minipage}{.49\textwidth}
    \centering 
    \resizebox{\textwidth}{!}{
        \begin{tabular}{llrrrrrr}
\toprule
     & model &   COMP &   DREP &     IC &     IE &    LMD &     RE \\
max\_leaf\_nodes & n\_estimators &        &        &        &        &        &        \\
\midrule
64   & 8.0   &  88.78 &  87.31 &  88.53 &  86.86 &  87.28 &  88.93 \\
     & 16.0  &  90.24 &  89.10 &  90.62 &  88.53 &  89.80 &  90.72 \\
     & 32.0  &  91.00 &  90.21 &  91.16 &  89.34 &  90.96 &  91.24 \\
     & 64.0  &  91.54 &  90.78 &  91.37 &  89.98 &  91.03 &  91.61 \\
     & 128.0 &  91.60 &  91.03 &  91.15 &  90.40 &  91.11 &  91.55 \\
128  & 8.0   &  91.49 &  91.05 &  91.52 &  90.33 &  90.79 &  91.59 \\
     & 16.0  &  92.70 &  92.06 &  92.97 &  91.62 &  92.37 &  93.05 \\
     & 32.0  &  93.10 &  92.99 &  93.71 &  92.22 &  93.16 &  93.59 \\
     & 64.0  &  93.76 &  93.15 &  93.91 &  92.76 &  93.60 &  93.86 \\
     & 128.0 &  93.76 &  93.54 &  93.71 &  93.10 &  93.89 &  93.99 \\
256  & 8.0   &  93.57 &  92.91 &  93.44 &  92.65 &  92.63 &  93.14 \\
     & 16.0  &  94.81 &  94.43 &  94.79 &  93.87 &  94.13 &  94.71 \\
     & 32.0  &  95.23 &  95.12 &  95.32 &  94.50 &  95.13 &  95.18 \\
     & 64.0  &  95.63 &  95.23 &  95.50 &  95.16 &  95.48 &  95.45 \\
     & 128.0 &  95.59 &  95.54 &  95.59 &  95.26 &  95.74 &  95.63 \\
512  & 8.0   &  94.34 &  93.96 &  94.51 &  94.08 &  94.28 &  94.37 \\
     & 16.0  &  95.76 &  95.68 &  95.79 &  95.45 &  95.65 &  95.64 \\
     & 32.0  &  96.39 &  96.39 &  96.69 &  96.36 &  96.26 &  96.52 \\
     & 64.0  &  96.77 &  96.71 &  96.91 &  96.72 &  96.70 &  96.74 \\
     & 128.0 &  97.14 &  96.98 &  97.11 &  96.89 &  96.86 &  97.01 \\
1024 & 8.0   &  94.45 &  94.24 &  95.07 &  94.77 &  94.32 &  94.87 \\
     & 16.0  &  96.19 &  95.80 &  96.68 &  96.02 &  96.26 &  96.37 \\
     & 32.0  &  96.94 &  96.86 &  97.33 &  96.83 &  96.99 &  97.04 \\
     & 64.0  &  97.32 &  97.37 &  97.48 &  97.18 &  97.39 &  97.41 \\
     & 128.0 &  97.60 &  97.57 &  97.66 &  97.51 &  97.61 &  97.57 \\
\bottomrule
\end{tabular}
    }
\end{minipage}
\caption{(Left) The error over the number of trees in the ensemble on the japanese-vowels dataset. Dashed lines depict the Random Forest and solid lines are the corresponding pruned ensemble via Reduced Error pruning. (Right) The 5-fold cross-validation accuracy  on the japanese-vowels dataset. Rounded to the second decimal digit. Larger is better.}
\end{figure}

\begin{figure}[H]
\begin{minipage}{.49\textwidth}
    \centering
    \includegraphics[width=\textwidth,keepaspectratio]{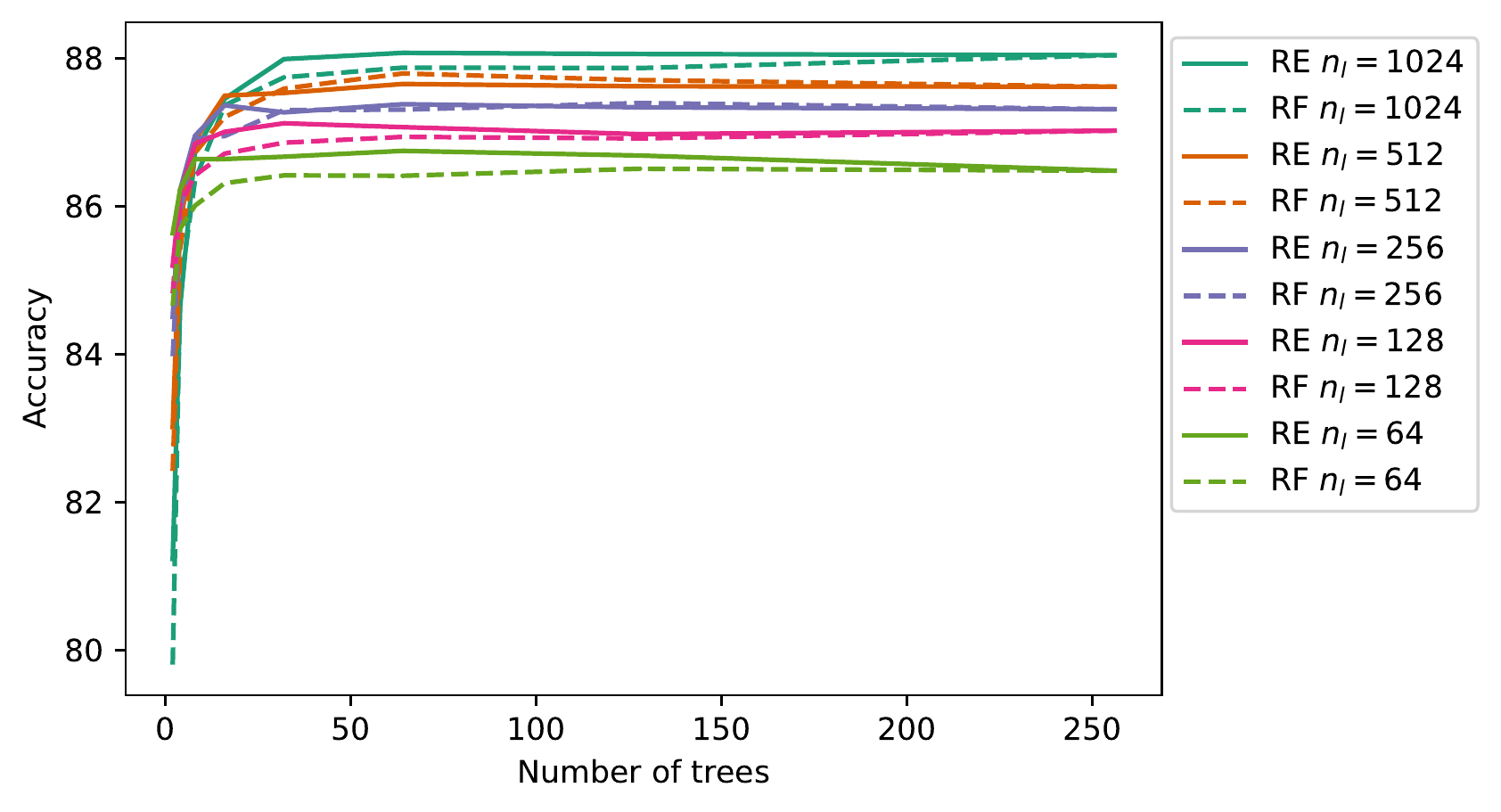}
\end{minipage}\hfill
\begin{minipage}{.49\textwidth}
    \centering 
    \resizebox{\textwidth}{!}{
        \begin{tabular}{llrrrrrr}
\toprule
     & model &   COMP &   DREP &     IC &     IE &    LMD &     RE \\
max\_leaf\_nodes & n\_estimators &        &        &        &        &        &        \\
\midrule
64   & 8.0   &  86.37 &  86.14 &  86.07 &  86.10 &  86.11 &  86.64 \\
     & 16.0  &  86.60 &  86.11 &  86.66 &  86.35 &  86.45 &  86.64 \\
     & 32.0  &  86.81 &  86.29 &  86.83 &  86.41 &  86.55 &  86.67 \\
     & 64.0  &  86.79 &  86.35 &  86.80 &  86.42 &  86.57 &  86.75 \\
     & 128.0 &  86.64 &  86.47 &  86.72 &  86.43 &  86.62 &  86.69 \\
128  & 8.0   &  86.61 &  86.28 &  86.69 &  86.77 &  86.46 &  86.84 \\
     & 16.0  &  86.88 &  86.63 &  87.07 &  86.99 &  87.00 &  87.01 \\
     & 32.0  &  87.19 &  86.68 &  87.18 &  86.87 &  87.04 &  87.12 \\
     & 64.0  &  87.13 &  86.81 &  87.17 &  87.04 &  86.98 &  87.07 \\
     & 128.0 &  87.12 &  86.90 &  87.11 &  87.03 &  87.13 &  86.98 \\
256  & 8.0   &  86.74 &  86.66 &  86.93 &  86.92 &  87.09 &  86.96 \\
     & 16.0  &  87.28 &  87.22 &  87.23 &  87.24 &  87.30 &  87.37 \\
     & 32.0  &  87.62 &  87.32 &  87.54 &  87.16 &  87.27 &  87.27 \\
     & 64.0  &  87.50 &  87.16 &  87.44 &  87.37 &  87.46 &  87.38 \\
     & 128.0 &  87.42 &  87.41 &  87.41 &  87.34 &  87.40 &  87.34 \\
512  & 8.0   &  87.07 &  86.68 &  87.02 &  87.00 &  86.72 &  86.89 \\
     & 16.0  &  87.59 &  87.10 &  87.65 &  87.57 &  87.20 &  87.50 \\
     & 32.0  &  87.76 &  87.49 &  87.84 &  87.75 &  87.35 &  87.53 \\
     & 64.0  &  87.76 &  87.49 &  87.73 &  87.74 &  87.63 &  87.65 \\
     & 128.0 &  87.72 &  87.62 &  87.73 &  87.77 &  87.61 &  87.62 \\
1024 & 8.0   &  86.26 &  86.29 &  86.69 &  86.49 &  86.38 &  86.73 \\
     & 16.0  &  87.40 &  87.33 &  87.50 &  87.38 &  87.36 &  87.45 \\
     & 32.0  &  87.56 &  87.74 &  87.90 &  87.76 &  87.80 &  87.99 \\
     & 64.0  &  87.86 &  87.96 &  88.01 &  87.96 &  87.91 &  88.08 \\
     & 128.0 &  87.87 &  87.94 &  87.91 &  87.99 &  88.04 &  88.06 \\
\bottomrule
\end{tabular}
    }
\end{minipage}
\caption{(Left) The error over the number of trees in the ensemble on the magic dataset. Dashed lines depict the Random Forest and solid lines are the corresponding pruned ensemble via Reduced Error pruning. (Right) The 5-fold cross-validation accuracy  on the magic dataset. Rounded to the second decimal digit. Larger is better}
\end{figure}

\begin{figure}[H]
\begin{minipage}{.49\textwidth}
    \centering
    \includegraphics[width=\textwidth,keepaspectratio]{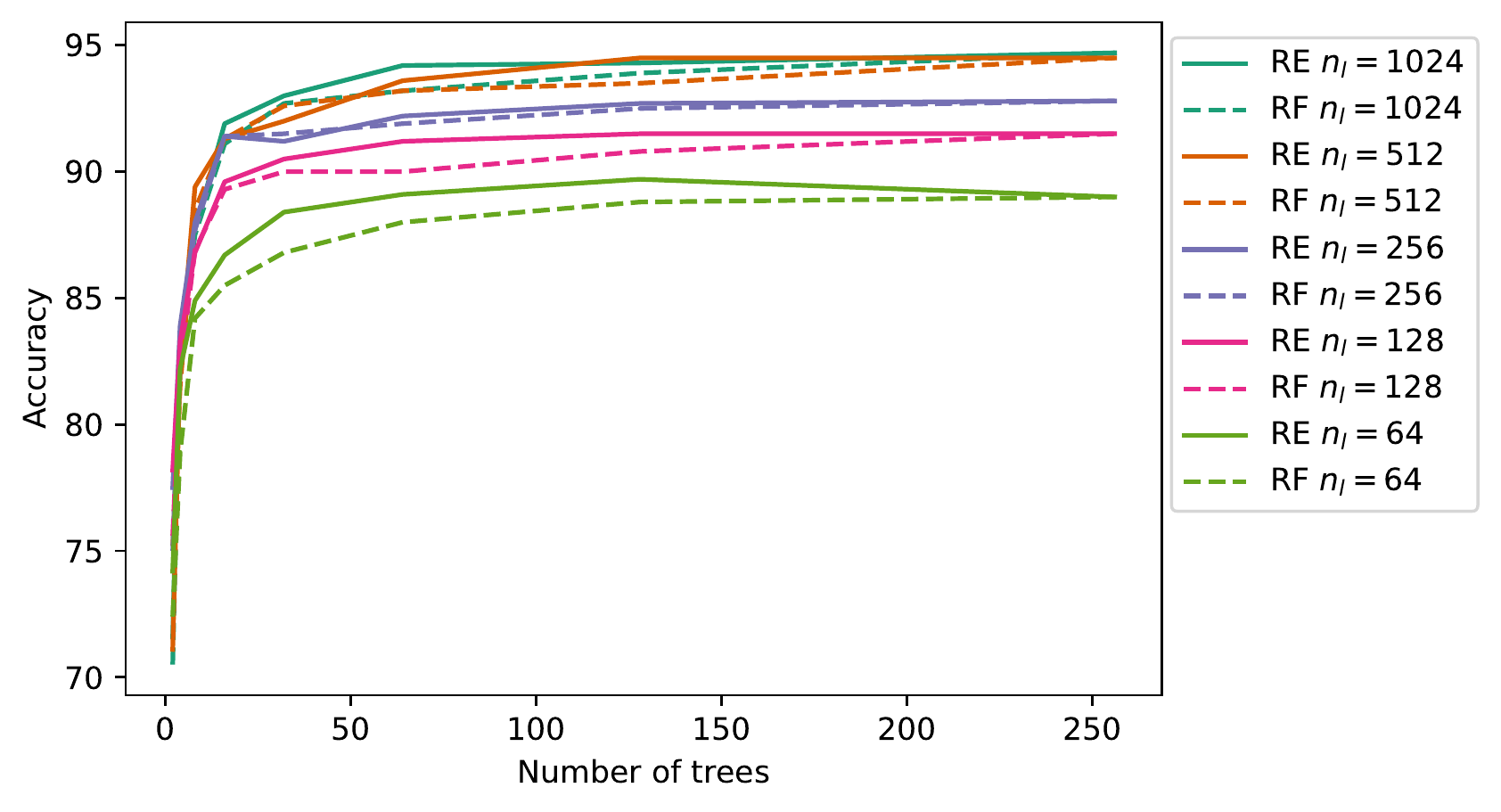}
\end{minipage}\hfill
\begin{minipage}{.49\textwidth}
    \centering 
    \resizebox{\textwidth}{!}{
        \begin{tabular}{llrrrrrr}
\toprule
     & model &  COMP &  DREP &    IC &    IE &   LMD &    RE \\
max\_leaf\_nodes & n\_estimators &       &       &       &       &       &       \\
\midrule
64   & 8.0   &  85.7 &  84.3 &  84.7 &  84.1 &  85.0 &  84.9 \\
     & 16.0  &  87.4 &  87.6 &  88.5 &  86.6 &  87.2 &  86.7 \\
     & 32.0  &  89.4 &  88.5 &  89.4 &  87.6 &  87.9 &  88.4 \\
     & 64.0  &  89.8 &  89.6 &  89.8 &  88.6 &  88.5 &  89.1 \\
     & 128.0 &  89.0 &  89.4 &  89.5 &  88.8 &  89.0 &  89.7 \\
128  & 8.0   &  89.0 &  87.1 &  87.9 &  87.4 &  86.6 &  86.8 \\
     & 16.0  &  90.1 &  89.3 &  90.6 &  89.7 &  88.8 &  89.6 \\
     & 32.0  &  90.0 &  90.5 &  90.4 &  89.7 &  91.0 &  90.5 \\
     & 64.0  &  91.0 &  90.6 &  90.8 &  90.5 &  91.1 &  91.2 \\
     & 128.0 &  91.1 &  91.3 &  91.1 &  90.6 &  92.0 &  91.5 \\
256  & 8.0   &  88.1 &  87.6 &  89.1 &  87.9 &  87.7 &  88.0 \\
     & 16.0  &  91.1 &  90.8 &  90.9 &  91.1 &  89.9 &  91.4 \\
     & 32.0  &  91.9 &  91.5 &  91.5 &  91.9 &  92.2 &  91.2 \\
     & 64.0  &  92.3 &  92.3 &  91.8 &  91.2 &  93.4 &  92.2 \\
     & 128.0 &  92.9 &  92.8 &  93.0 &  92.6 &  93.0 &  92.7 \\
512  & 8.0   &  86.7 &  89.2 &  88.6 &  88.7 &  88.1 &  89.4 \\
     & 16.0  &  91.4 &  90.3 &  92.2 &  91.6 &  90.8 &  91.3 \\
     & 32.0  &  92.6 &  93.1 &  93.3 &  93.1 &  92.8 &  92.0 \\
     & 64.0  &  93.6 &  93.8 &  93.7 &  93.6 &  93.7 &  93.6 \\
     & 128.0 &  94.2 &  94.3 &  93.8 &  93.8 &  93.9 &  94.5 \\
1024 & 8.0   &  87.2 &  88.0 &  87.7 &  88.7 &  87.9 &  87.7 \\
     & 16.0  &  91.5 &  91.3 &  91.6 &  92.1 &  91.5 &  91.9 \\
     & 32.0  &  92.6 &  93.2 &  93.9 &  93.3 &  93.1 &  93.0 \\
     & 64.0  &  93.9 &  94.0 &  94.2 &  94.0 &  93.3 &  94.2 \\
     & 128.0 &  94.4 &  94.5 &  94.6 &  94.5 &  93.9 &  94.3 \\
\bottomrule
\end{tabular}
    }
\end{minipage}
\caption{(Left) The error over the number of trees in the ensemble on the mnist dataset. Dashed lines depict the Random Forest and solid lines are the corresponding pruned ensemble via Reduced Error pruning. (Right) The 5-fold cross-validation accuracy  on the mnist dataset. Rounded to the second decimal digit. Larger is better.}
\end{figure}

\begin{figure}[H]
\begin{minipage}{.49\textwidth}
    \centering
    \includegraphics[width=\textwidth,keepaspectratio]{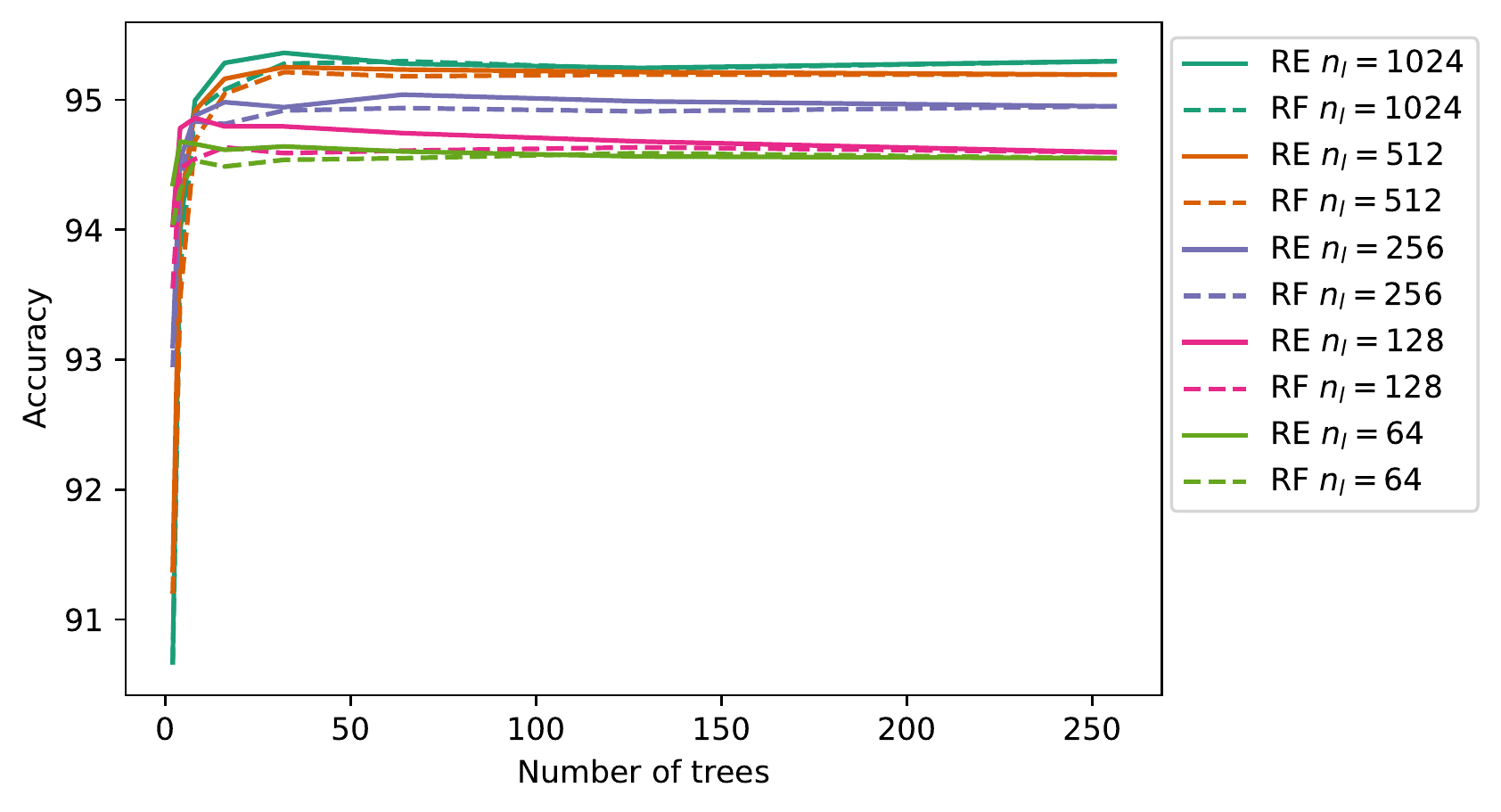}
\end{minipage}\hfill
\begin{minipage}{.49\textwidth}
    \centering 
    \resizebox{\textwidth}{!}{
        \begin{tabular}{llrrrrrr}
\toprule
     & model &   COMP &   DREP &     IC &     IE &    LMD &     RE \\
max\_leaf\_nodes & n\_estimators &        &        &        &        &        &        \\
\midrule
64   & 8.0   &  94.63 &  94.47 &  94.82 &  94.68 &  94.27 &  94.66 \\
     & 16.0  &  94.62 &  94.50 &  94.76 &  94.66 &  94.35 &  94.62 \\
     & 32.0  &  94.65 &  94.51 &  94.69 &  94.67 &  94.47 &  94.64 \\
     & 64.0  &  94.60 &  94.54 &  94.63 &  94.65 &  94.53 &  94.60 \\
     & 128.0 &  94.56 &  94.54 &  94.56 &  94.58 &  94.55 &  94.56 \\
128  & 8.0   &  94.71 &  94.65 &  94.76 &  94.86 &  94.47 &  94.86 \\
     & 16.0  &  94.71 &  94.69 &  94.81 &  94.80 &  94.53 &  94.80 \\
     & 32.0  &  94.72 &  94.67 &  94.79 &  94.73 &  94.60 &  94.80 \\
     & 64.0  &  94.72 &  94.66 &  94.76 &  94.73 &  94.61 &  94.74 \\
     & 128.0 &  94.69 &  94.64 &  94.71 &  94.67 &  94.62 &  94.68 \\
256  & 8.0   &  95.00 &  94.74 &  95.00 &  94.96 &  94.78 &  94.88 \\
     & 16.0  &  95.03 &  94.88 &  95.09 &  95.05 &  94.80 &  94.98 \\
     & 32.0  &  95.07 &  94.85 &  95.09 &  95.03 &  94.86 &  94.94 \\
     & 64.0  &  95.05 &  94.95 &  95.08 &  95.05 &  94.84 &  95.04 \\
     & 128.0 &  94.96 &  94.92 &  95.05 &  95.02 &  94.90 &  94.99 \\
512  & 8.0   &  95.02 &  94.85 &  95.05 &  95.09 &  94.71 &  94.92 \\
     & 16.0  &  95.14 &  95.09 &  95.27 &  95.14 &  94.91 &  95.16 \\
     & 32.0  &  95.19 &  95.21 &  95.26 &  95.28 &  95.03 &  95.25 \\
     & 64.0  &  95.22 &  95.13 &  95.30 &  95.34 &  95.10 &  95.23 \\
     & 128.0 &  95.18 &  95.13 &  95.29 &  95.30 &  95.12 &  95.21 \\
1024 & 8.0   &  94.72 &  94.98 &  94.94 &  94.85 &  94.50 &  95.00 \\
     & 16.0  &  95.20 &  95.29 &  95.25 &  95.19 &  94.92 &  95.28 \\
     & 32.0  &  95.30 &  95.26 &  95.25 &  95.29 &  95.19 &  95.36 \\
     & 64.0  &  95.27 &  95.20 &  95.30 &  95.31 &  95.21 &  95.28 \\
     & 128.0 &  95.27 &  95.26 &  95.31 &  95.36 &  95.21 &  95.25 \\
\bottomrule
\end{tabular}
    }
\end{minipage}
\caption{(Left) The error over the number of trees in the ensemble on the mozilla dataset. Dashed lines depict the Random Forest and solid lines are the corresponding pruned ensemble via Reduced Error pruning. (Right) The 5-fold cross-validation accuracy  on the mozilla dataset. Rounded to the second decimal digit. Larger is better.}
\end{figure}

\begin{figure}[H]
\begin{minipage}{.49\textwidth}
    \centering
    \includegraphics[width=\textwidth,keepaspectratio]{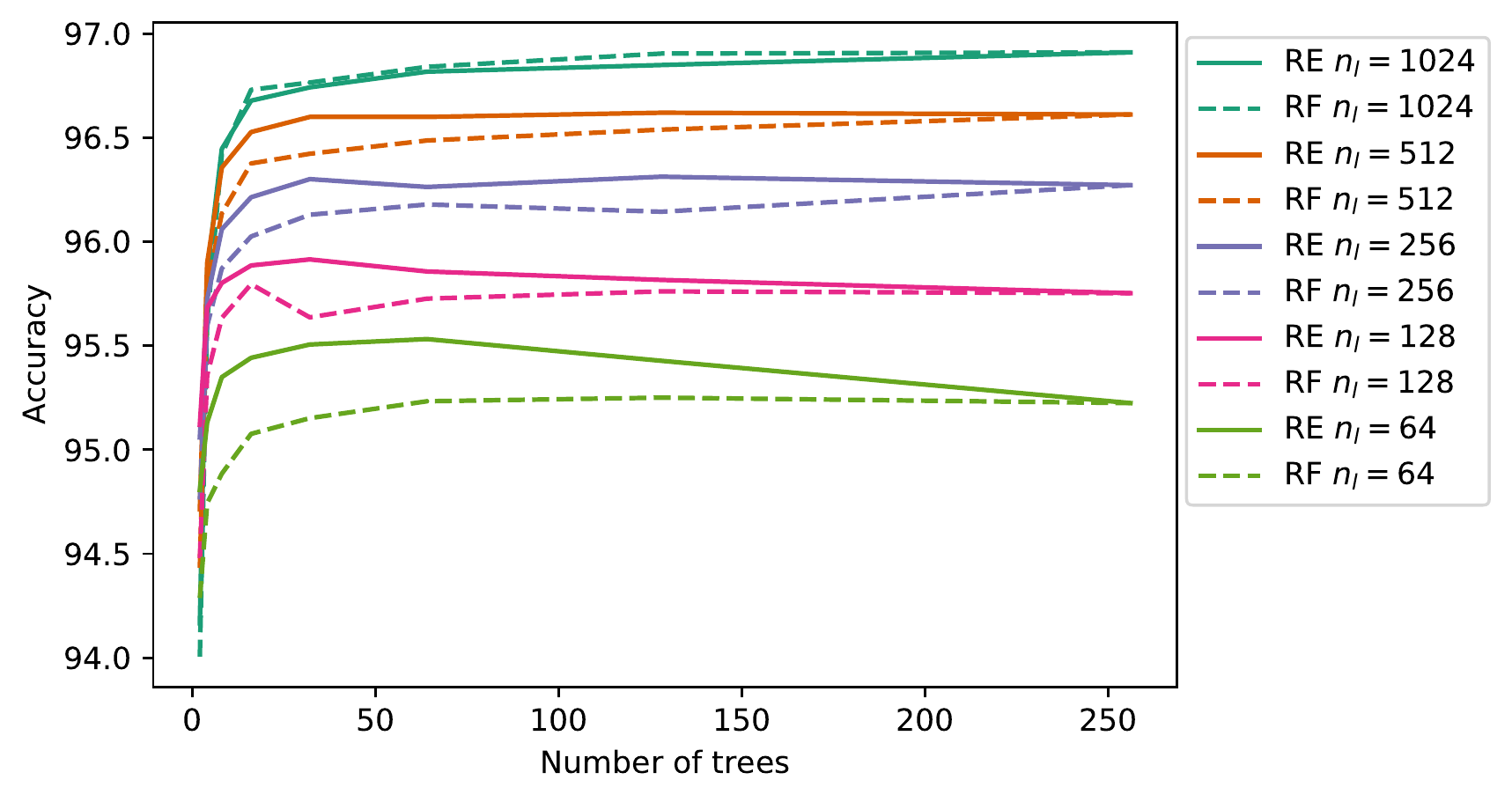}
\end{minipage}\hfill
\begin{minipage}{.49\textwidth}
    \centering 
    \resizebox{\textwidth}{!}{
        \begin{tabular}{llrrrrrr}
\toprule
     & model &   COMP &   DREP &     IC &     IE &    LMD &     RE \\
max\_leaf\_nodes & n\_estimators &        &        &        &        &        &        \\
\midrule
64   & 8.0   &  95.37 &  95.11 &  95.19 &  95.19 &  94.97 &  95.35 \\
     & 16.0  &  95.39 &  95.29 &  95.39 &  95.29 &  95.08 &  95.44 \\
     & 32.0  &  95.45 &  95.29 &  95.50 &  95.29 &  95.13 &  95.51 \\
     & 64.0  &  95.41 &  95.20 &  95.49 &  95.31 &  95.19 &  95.53 \\
     & 128.0 &  95.40 &    NaN &  95.40 &  95.29 &  95.21 &  95.43 \\
128  & 8.0   &  95.63 &  95.47 &  95.75 &  95.75 &  95.36 &  95.80 \\
     & 16.0  &  95.75 &  95.69 &  95.87 &  95.89 &  95.52 &  95.89 \\
     & 32.0  &  95.83 &  95.75 &  95.91 &  95.93 &  95.67 &  95.91 \\
     & 64.0  &  95.89 &  95.75 &  95.90 &  95.91 &  95.76 &  95.86 \\
     & 128.0 &  95.85 &    NaN &  95.82 &  95.82 &  95.76 &  95.82 \\
256  & 8.0   &  96.09 &  95.90 &  95.97 &  96.06 &  95.92 &  96.06 \\
     & 16.0  &  96.19 &  96.12 &  96.23 &  96.18 &  96.08 &  96.21 \\
     & 32.0  &  96.30 &  96.15 &  96.35 &  96.24 &  96.17 &  96.30 \\
     & 64.0  &  96.31 &  96.21 &  96.34 &  96.25 &  96.19 &  96.26 \\
     & 128.0 &  96.29 &  96.24 &  96.29 &  96.23 &  96.22 &  96.31 \\
512  & 8.0   &  96.22 &  96.15 &  96.32 &  96.27 &  96.04 &  96.36 \\
     & 16.0  &  96.39 &  96.37 &  96.55 &  96.38 &  96.41 &  96.53 \\
     & 32.0  &  96.56 &  96.47 &  96.56 &  96.52 &  96.53 &  96.60 \\
     & 64.0  &  96.57 &  96.53 &  96.66 &  96.58 &  96.56 &  96.60 \\
     & 128.0 &  96.61 &  96.55 &  96.61 &  96.65 &  96.62 &  96.62 \\
1024 & 8.0   &  96.29 &  96.33 &  96.25 &  96.36 &  96.31 &  96.45 \\
     & 16.0  &  96.64 &  96.66 &  96.63 &  96.70 &  96.69 &  96.68 \\
     & 32.0  &  96.74 &  96.74 &  96.80 &  96.76 &  96.85 &  96.74 \\
     & 64.0  &  96.86 &  96.84 &  96.85 &  96.81 &  96.86 &  96.82 \\
     & 128.0 &  96.89 &  96.87 &  96.95 &  96.91 &  96.90 &  96.85 \\
\bottomrule
\end{tabular}
    }
\end{minipage}
\caption{(Left) The error over the number of trees in the ensemble on the nomao dataset. Dashed lines depict the Random Forest and solid lines are the corresponding pruned ensemble via Reduced Error pruning. (Right) The 5-fold cross-validation accuracy  on the nomao dataset. Rounded to the second decimal digit. Larger is better.}
\end{figure}

\begin{figure}[H]
\begin{minipage}{.49\textwidth}
    \centering
    \includegraphics[width=\textwidth,keepaspectratio]{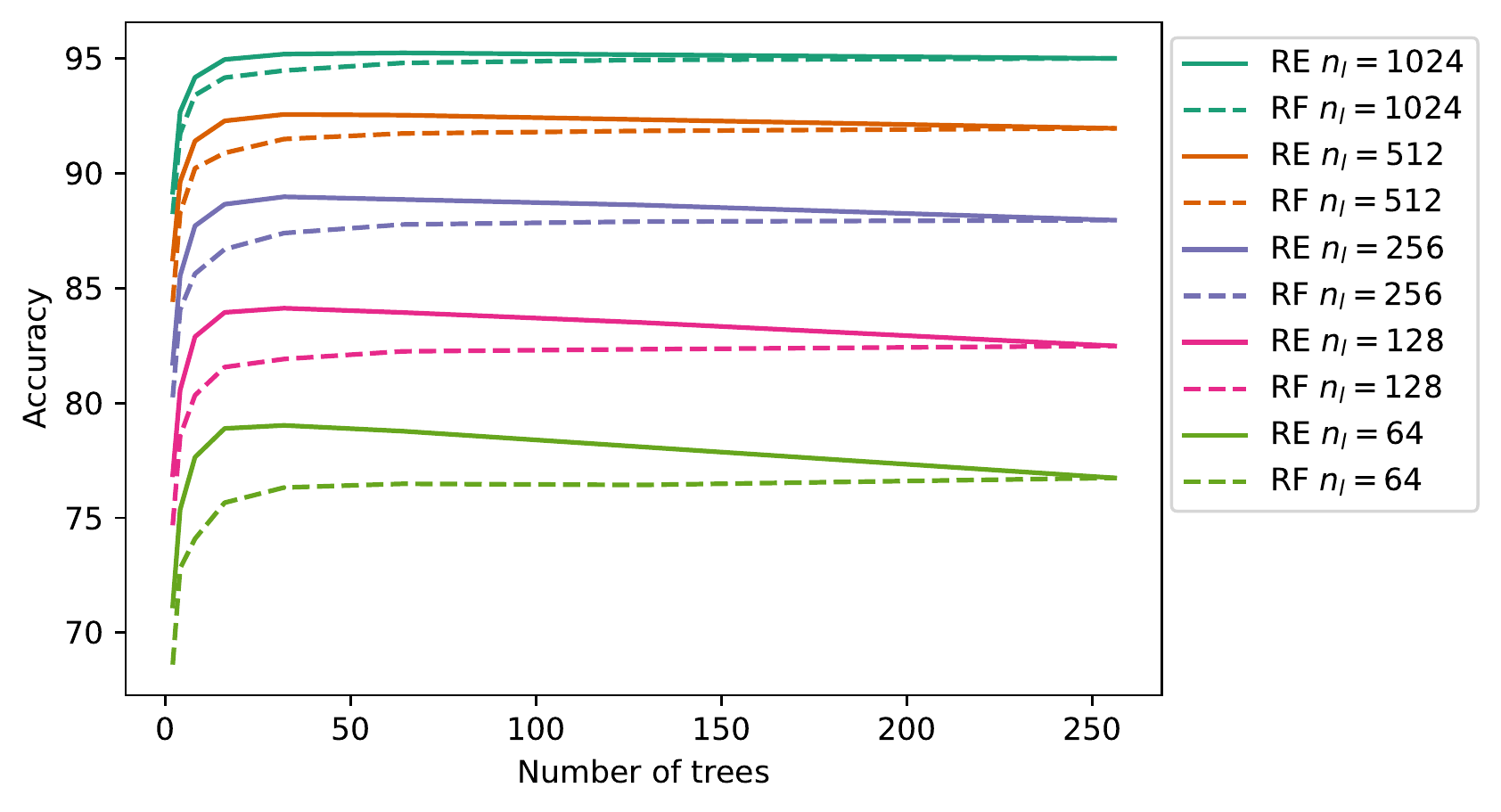}
\end{minipage}\hfill
\begin{minipage}{.49\textwidth}
    \centering 
    \resizebox{\textwidth}{!}{
        \begin{tabular}{llrrrrrr}
\toprule
     & model &   COMP &   DREP &     IC &     IE &    LMD &     RE \\
max\_leaf\_nodes & n\_estimators &        &        &        &        &        &        \\
\midrule
64   & 8.0   &  77.19 &  74.43 &  76.87 &  75.15 &  76.41 &  77.64 \\
     & 16.0  &  78.05 &  75.57 &  78.01 &  75.75 &  77.33 &  78.89 \\
     & 32.0  &  78.30 &    NaN &  78.19 &  76.16 &  77.77 &  79.02 \\
     & 64.0  &  78.06 &    NaN &  78.08 &  76.36 &  77.63 &  78.77 \\
     & 128.0 &    NaN &    NaN &  77.76 &  76.51 &  77.36 &    NaN \\
128  & 8.0   &  82.54 &  80.47 &  82.43 &  81.23 &  81.53 &  82.88 \\
     & 16.0  &  83.41 &  81.64 &  83.45 &  81.93 &  82.66 &  83.95 \\
     & 32.0  &  83.69 &    NaN &  83.65 &  82.53 &  82.99 &  84.13 \\
     & 64.0  &  83.62 &    NaN &  83.62 &  82.74 &  83.04 &  83.94 \\
     & 128.0 &    NaN &    NaN &  83.30 &  82.68 &  82.86 &  83.51 \\
256  & 8.0   &  87.44 &  85.67 &  87.20 &  86.36 &  86.04 &  87.71 \\
     & 16.0  &  88.45 &  86.70 &  88.31 &  87.50 &  87.29 &  88.65 \\
     & 32.0  &  88.69 &    NaN &  88.68 &  87.94 &  87.83 &  88.98 \\
     & 64.0  &  88.65 &    NaN &  88.68 &  88.16 &  87.99 &  88.86 \\
     & 128.0 &  88.40 &    NaN &  88.45 &  88.11 &  88.13 &  88.62 \\
512  & 8.0   &  91.36 &  90.50 &  91.22 &  90.96 &  90.12 &  91.41 \\
     & 16.0  &  92.15 &  91.31 &  92.18 &  91.76 &  91.22 &  92.29 \\
     & 32.0  &  92.46 &    NaN &  92.46 &  92.22 &  91.71 &  92.57 \\
     & 64.0  &  92.46 &    NaN &  92.48 &  92.29 &  91.91 &  92.53 \\
     & 128.0 &  92.31 &    NaN &  92.40 &  92.18 &  91.99 &    NaN \\
1024 & 8.0   &  94.14 &  93.20 &  94.11 &  93.78 &  93.17 &  94.18 \\
     & 16.0  &  94.88 &  94.17 &  94.89 &  94.66 &  94.14 &  94.96 \\
     & 32.0  &  95.16 &    NaN &  95.20 &  95.00 &  94.68 &  95.19 \\
     & 64.0  &  95.20 &    NaN &  95.30 &  95.12 &  94.80 &  95.25 \\
     & 128.0 &  95.17 &    NaN &  95.20 &  95.12 &  94.92 &    NaN \\
\bottomrule
\end{tabular}
    }
\end{minipage}
\caption{(Left) The error over the number of trees in the ensemble on the postures dataset. Dashed lines depict the Random Forest and solid lines are the corresponding pruned ensemble via Reduced Error pruning. (Right) The 5-fold cross-validation accuracy  on the postures dataset. Rounded to the second decimal digit. Larger is better.}
\end{figure}

\begin{figure}[H]
\begin{minipage}{.49\textwidth}
    \centering
    \includegraphics[width=\textwidth,keepaspectratio]{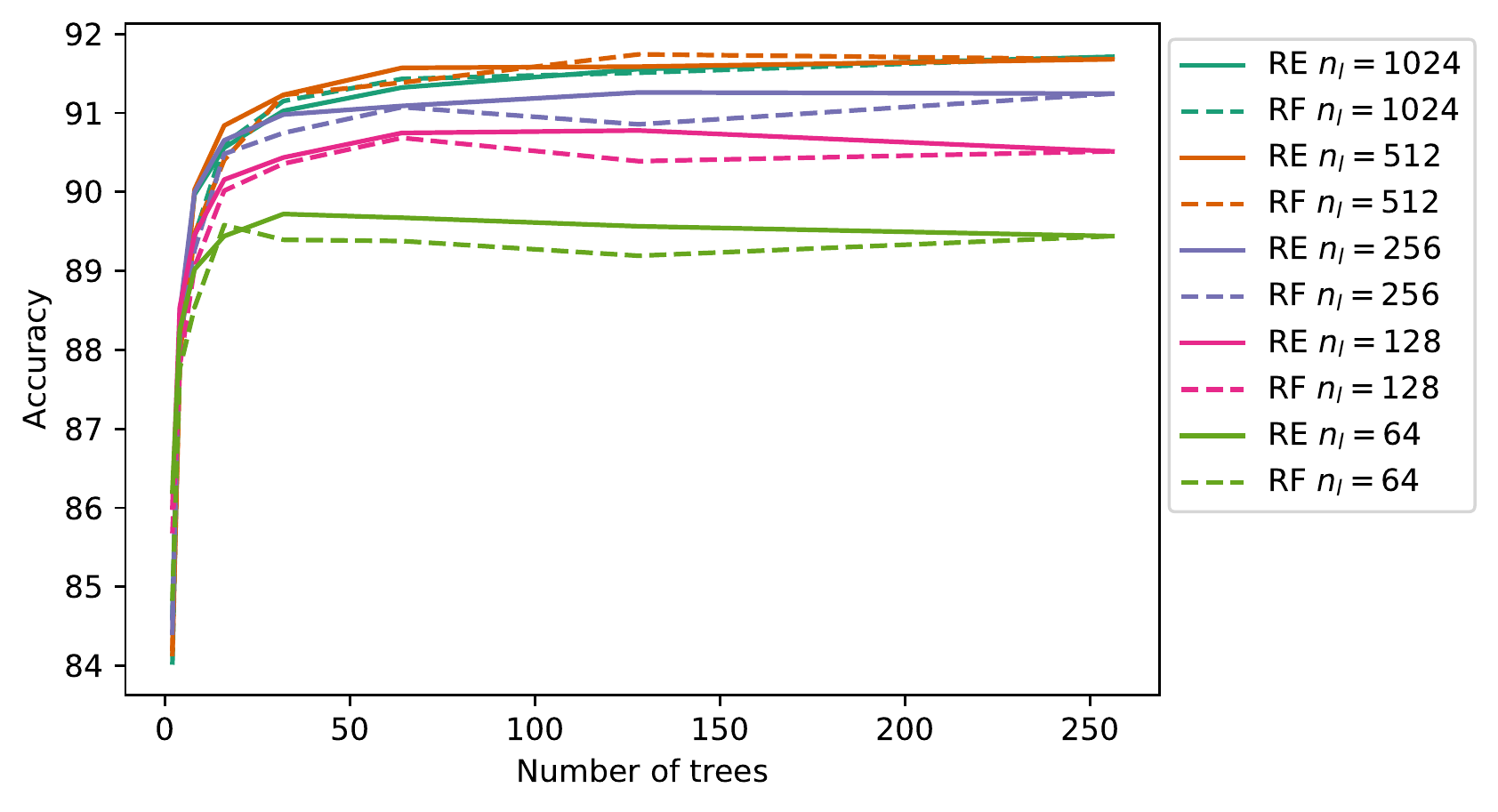}
\end{minipage}\hfill
\begin{minipage}{.49\textwidth}
    \centering 
    \resizebox{\textwidth}{!}{
        \begin{tabular}{llrrrrrr}
\toprule
     & model &   COMP &   DREP &     IC &     IE &    LMD &     RE \\
max\_leaf\_nodes & n\_estimators &        &        &        &        &        &        \\
\midrule
64   & 8.0   &  88.88 &  88.66 &  88.91 &  88.62 &  88.65 &  89.02 \\
     & 16.0  &  89.28 &  89.00 &  89.42 &  89.10 &  89.24 &  89.44 \\
     & 32.0  &  89.77 &  89.32 &  89.56 &  89.36 &  89.46 &  89.72 \\
     & 64.0  &  89.61 &  89.42 &  89.52 &  89.41 &  89.52 &  89.67 \\
     & 128.0 &  89.64 &  89.35 &  89.61 &  89.47 &  89.53 &  89.56 \\
128  & 8.0   &  89.38 &  89.30 &  89.11 &  89.39 &  89.52 &  89.46 \\
     & 16.0  &  90.22 &  89.89 &  89.67 &  89.89 &  90.00 &  90.16 \\
     & 32.0  &  90.16 &  90.09 &  90.31 &  90.36 &  90.16 &  90.44 \\
     & 64.0  &  90.54 &  90.44 &  90.36 &  90.45 &  90.34 &  90.75 \\
     & 128.0 &  90.65 &  90.44 &  90.64 &  90.47 &  90.51 &  90.78 \\
256  & 8.0   &  90.14 &  89.32 &  89.70 &  89.38 &  89.94 &  90.00 \\
     & 16.0  &  90.72 &  89.98 &  90.34 &  90.45 &  90.45 &  90.65 \\
     & 32.0  &  90.96 &  90.68 &  91.06 &  90.84 &  90.84 &  90.98 \\
     & 64.0  &  91.23 &  91.10 &  91.09 &  91.01 &  91.20 &  91.09 \\
     & 128.0 &  91.24 &  91.45 &  91.17 &  90.98 &  91.20 &  91.26 \\
512  & 8.0   &  89.52 &  90.19 &  89.52 &  89.94 &  89.38 &  90.03 \\
     & 16.0  &  90.53 &  90.76 &  90.72 &  90.93 &  90.25 &  90.84 \\
     & 32.0  &  91.31 &  91.12 &  91.35 &  91.45 &  91.04 &  91.23 \\
     & 64.0  &  91.46 &  91.56 &  91.60 &  91.49 &  91.54 &  91.57 \\
     & 128.0 &  91.60 &  91.60 &  91.59 &  91.70 &  91.74 &  91.59 \\
1024 & 8.0   &  89.83 &  89.49 &  89.25 &  90.19 &  89.53 &  89.97 \\
     & 16.0  &  90.58 &  90.67 &  90.65 &  91.04 &  90.56 &  90.56 \\
     & 32.0  &  91.17 &  91.06 &  91.56 &  91.66 &  91.06 &  91.03 \\
     & 64.0  &  91.46 &  91.42 &  91.59 &  91.60 &  91.68 &  91.32 \\
     & 128.0 &  91.49 &  91.59 &  91.79 &  91.80 &  91.57 &  91.56 \\
\bottomrule
\end{tabular}
    }
\end{minipage}
\caption{(Left) The error over the number of trees in the ensemble on the satimage dataset. Dashed lines depict the Random Forest and solid lines are the corresponding pruned ensemble via Reduced Error pruning. (Right) The 5-fold cross-validation accuracy  on the satimage dataset. Rounded to the second decimal digit. Larger is better.}
\end{figure}

\section{Plotting the Pareto Front For More Datasets}

\begin{figure}[H]
\begin{minipage}{.49\textwidth}
    \centering
    \includegraphics[width=\textwidth,keepaspectratio]{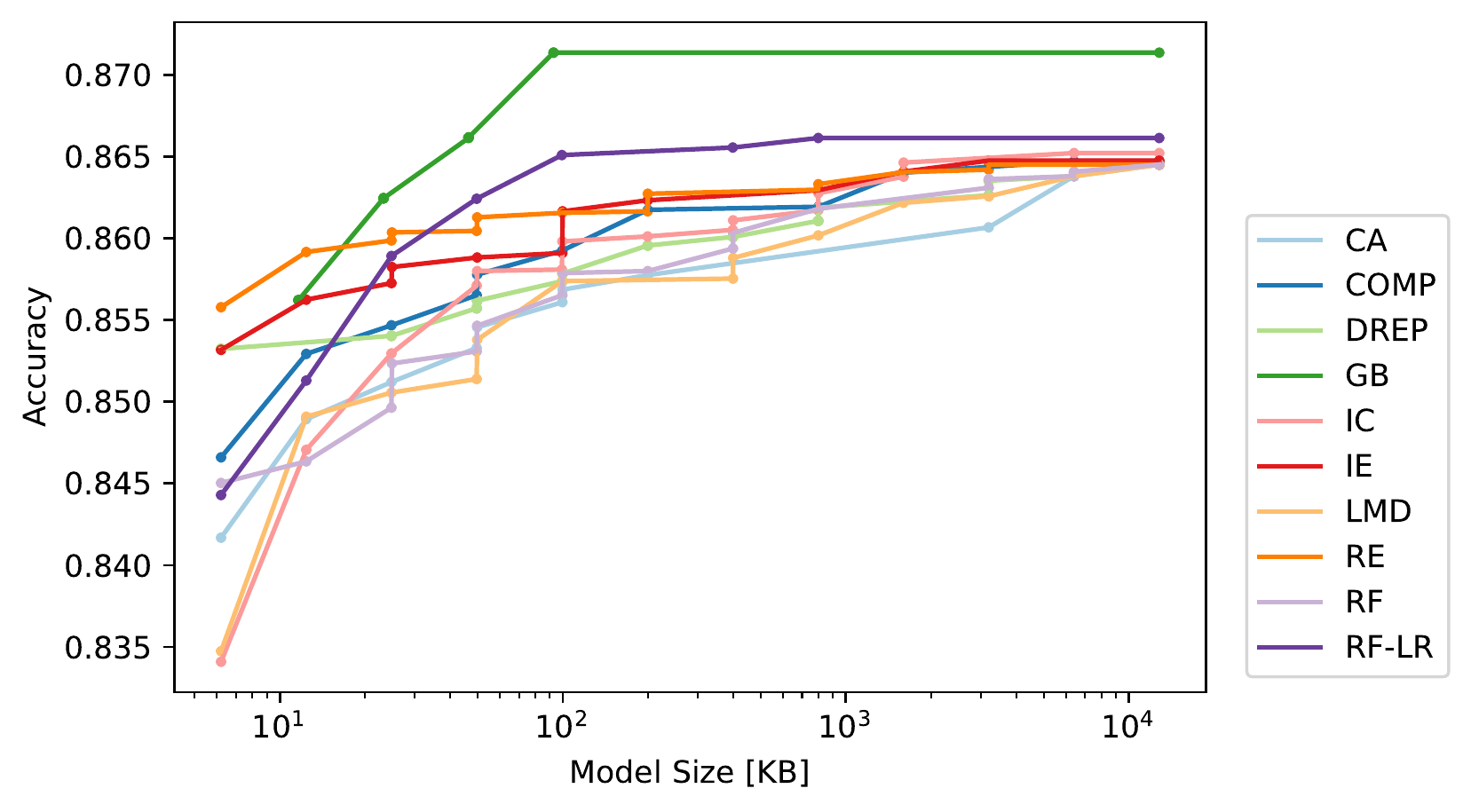}
\end{minipage}\hfill
\begin{minipage}{.49\textwidth}
    \centering 
    \includegraphics[width=\textwidth,keepaspectratio]{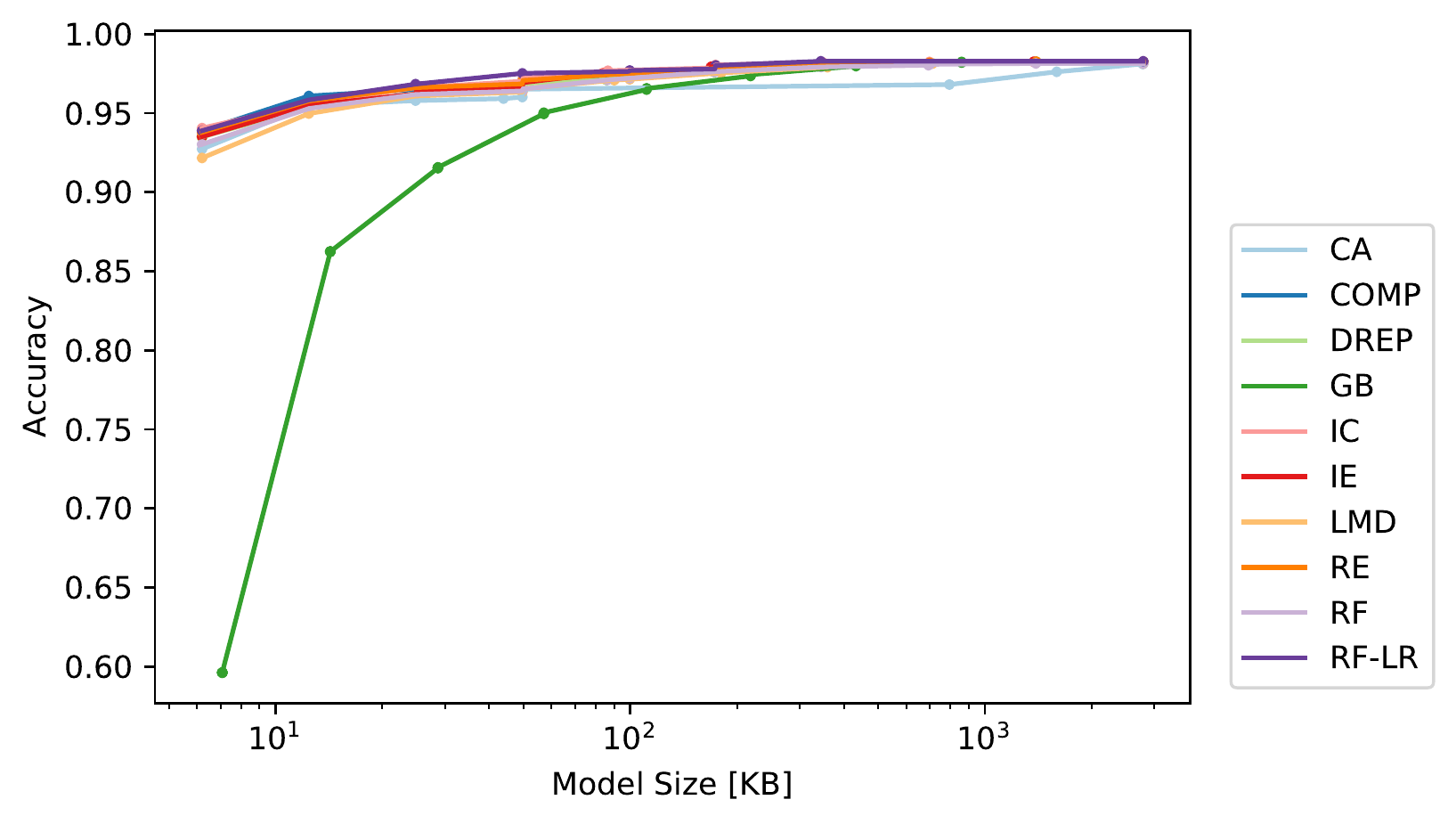}
\end{minipage}
\caption{5-fold cross-validation accuracy over the size of the ensemble for different $n_l$ and different $M$ on the chess dataset. Single points are the individual parameter configurations whereas the solid line depicts the corresponding Pareto Front. Left side show the adult dataset, right side shows the anura dataset.}
\end{figure}

\begin{figure}[H]
\begin{minipage}{.49\textwidth}
    \centering
    \includegraphics[width=\textwidth,keepaspectratio]{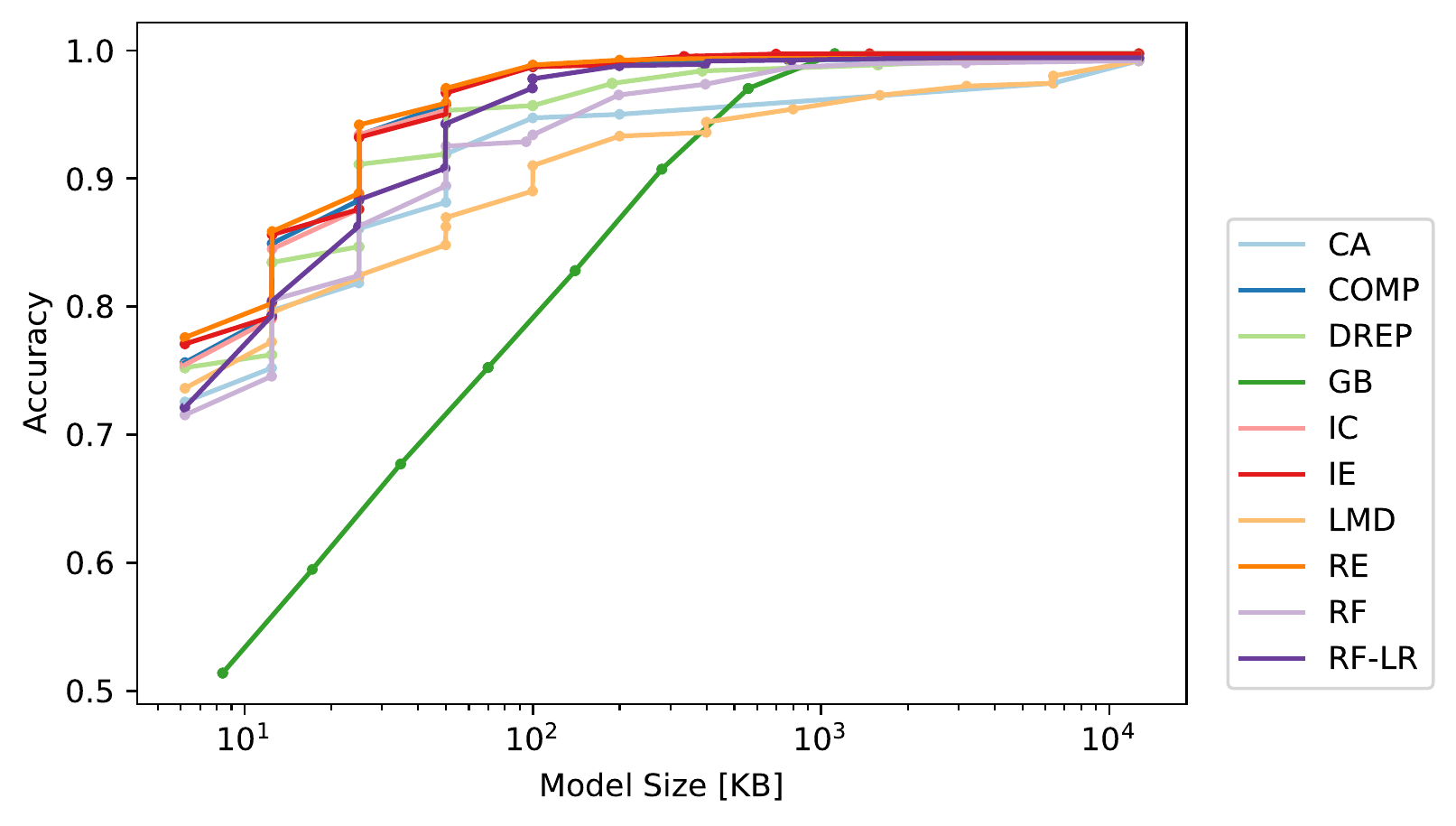}
\end{minipage}\hfill
\begin{minipage}{.49\textwidth}
    \centering 
    \includegraphics[width=\textwidth,keepaspectratio]{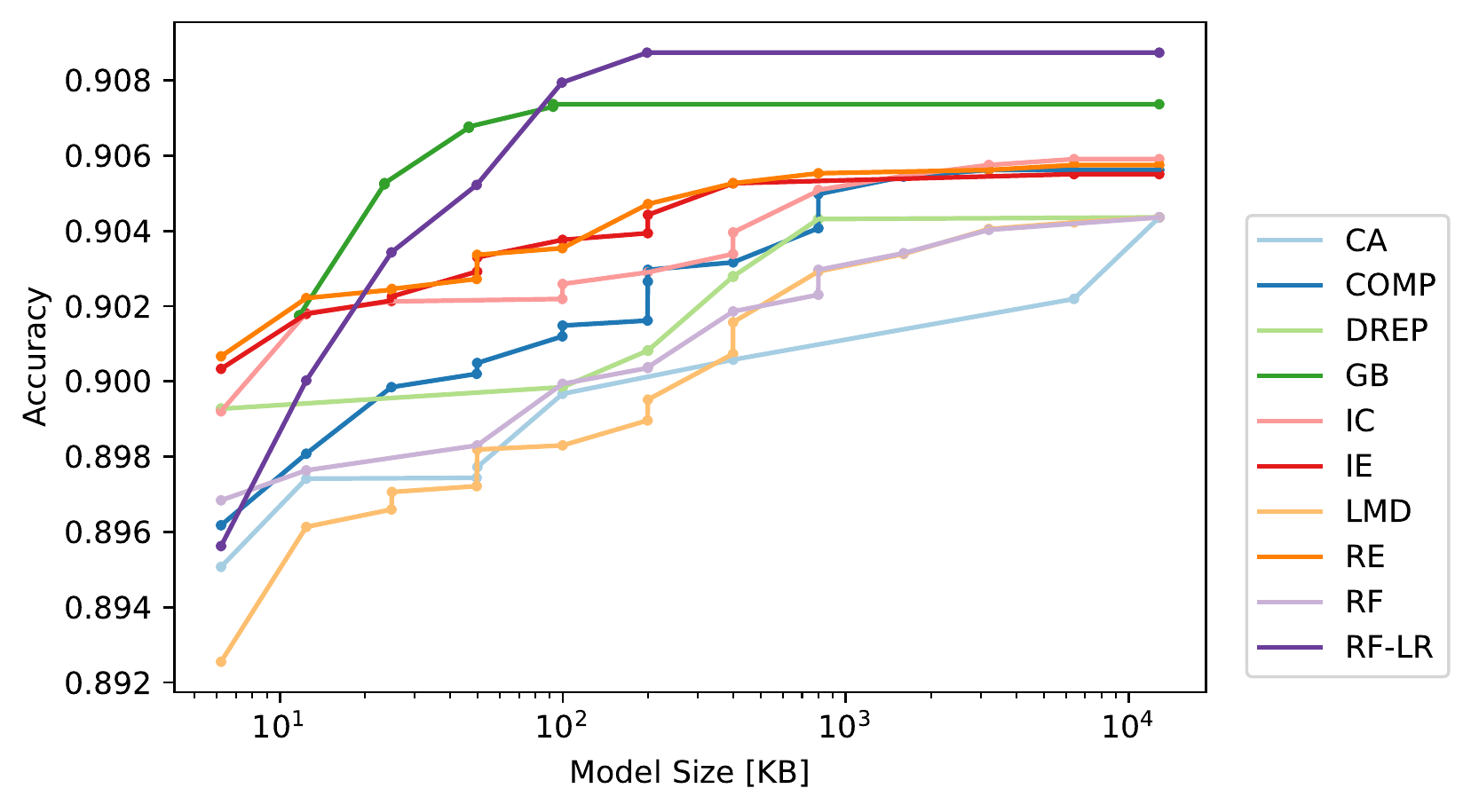}
\end{minipage}
\caption{5-fold cross-validation accuracy over the size of the ensemble for different $n_l$ and different $M$ on the chess dataset. Single points are the individual parameter configurations whereas the solid line depicts the corresponding Pareto Front. Left side show the avila dataset, right side shows the bank dataset.}
\end{figure}

\begin{figure}[H]
\begin{minipage}{.49\textwidth}
    \centering
    \includegraphics[width=\textwidth,keepaspectratio]{figures/RandomForestClassifier_chess_paretofront.pdf}
\end{minipage}\hfill
\begin{minipage}{.49\textwidth}
    \centering 
    \includegraphics[width=\textwidth,keepaspectratio]{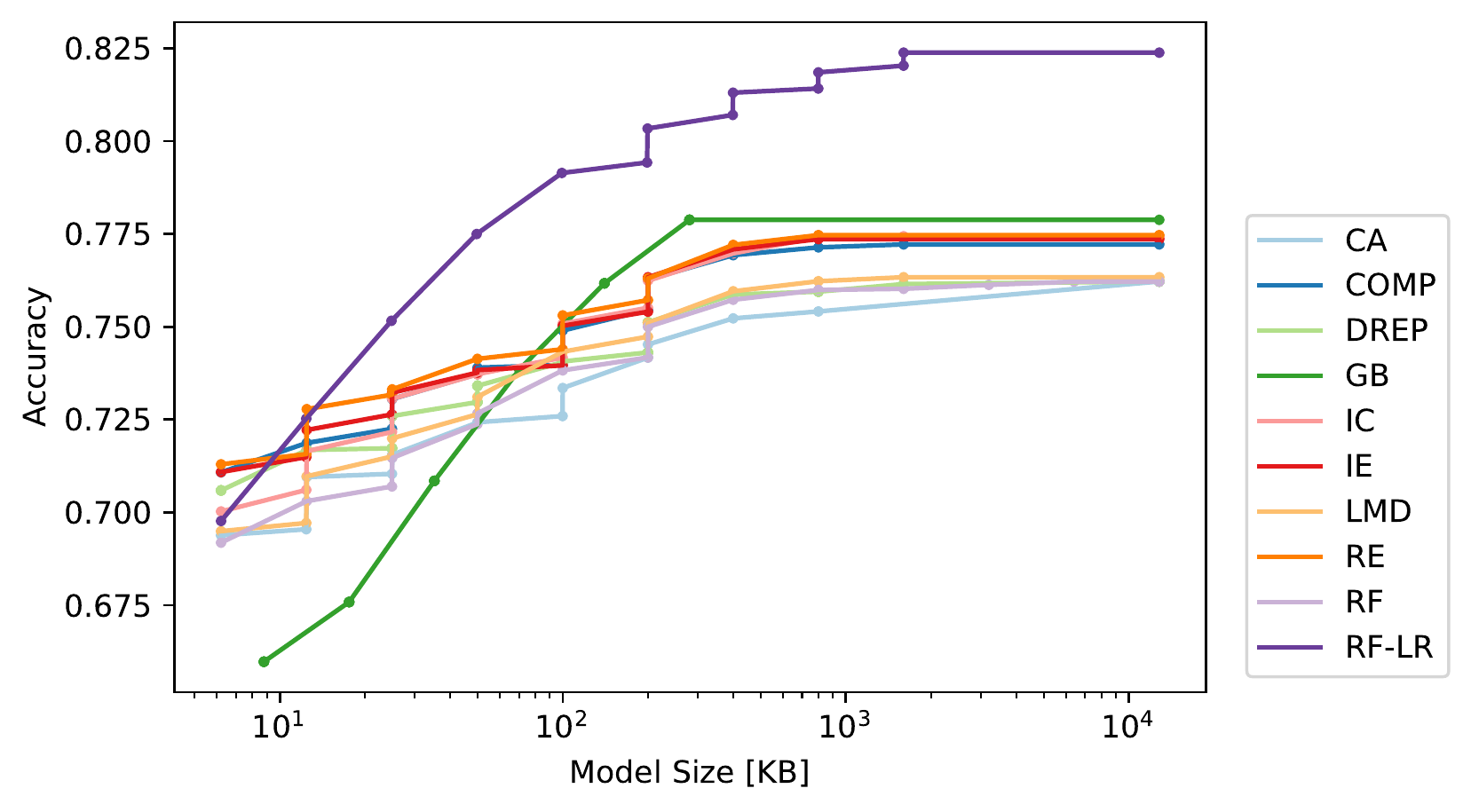}
\end{minipage}
\caption{5-fold cross-validation accuracy over the size of the ensemble for different $n_l$ and different $M$ on the chess dataset. Single points are the individual parameter configurations whereas the solid line depicts the corresponding Pareto Front. Left side show the chess dataset, right side shows the connect dataset.}
\end{figure}

\begin{figure}[H]
\begin{minipage}{.49\textwidth}
    \centering
    \includegraphics[width=\textwidth,keepaspectratio]{figures/RandomForestClassifier_eeg_paretofront.pdf}
\end{minipage}\hfill
\begin{minipage}{.49\textwidth}
    \centering 
    \includegraphics[width=\textwidth,keepaspectratio]{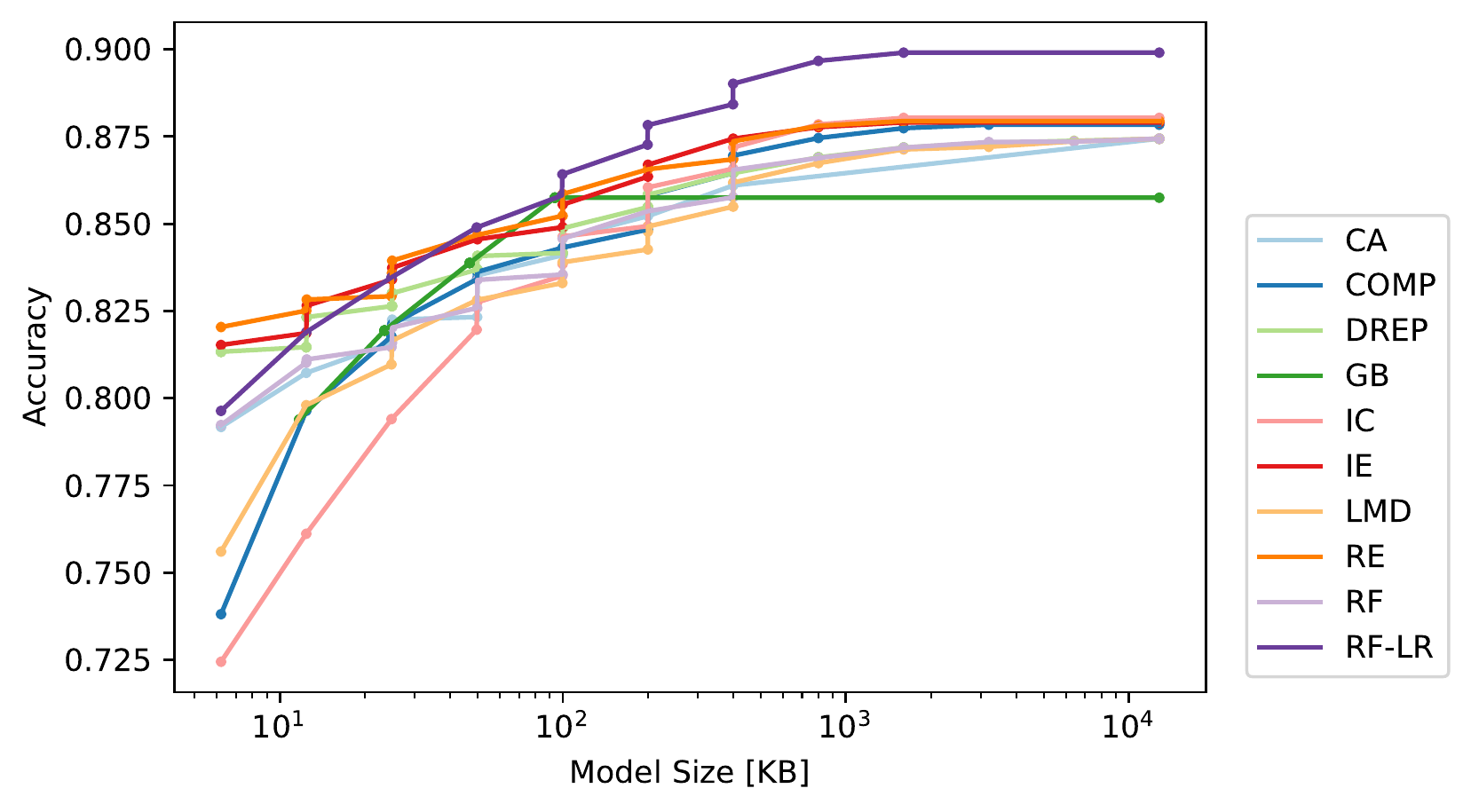}
\end{minipage}
\caption{5-fold cross-validation accuracy over the size of the ensemble for different $n_l$ and different $M$ on the chess dataset. Single points are the individual parameter configurations whereas the solid line depicts the corresponding Pareto Front. Left side show the eeg dataset, right side shows the elec dataset.}
\end{figure}

\begin{figure}[H]
\begin{minipage}{.49\textwidth}
    \centering
    \includegraphics[width=\textwidth,keepaspectratio]{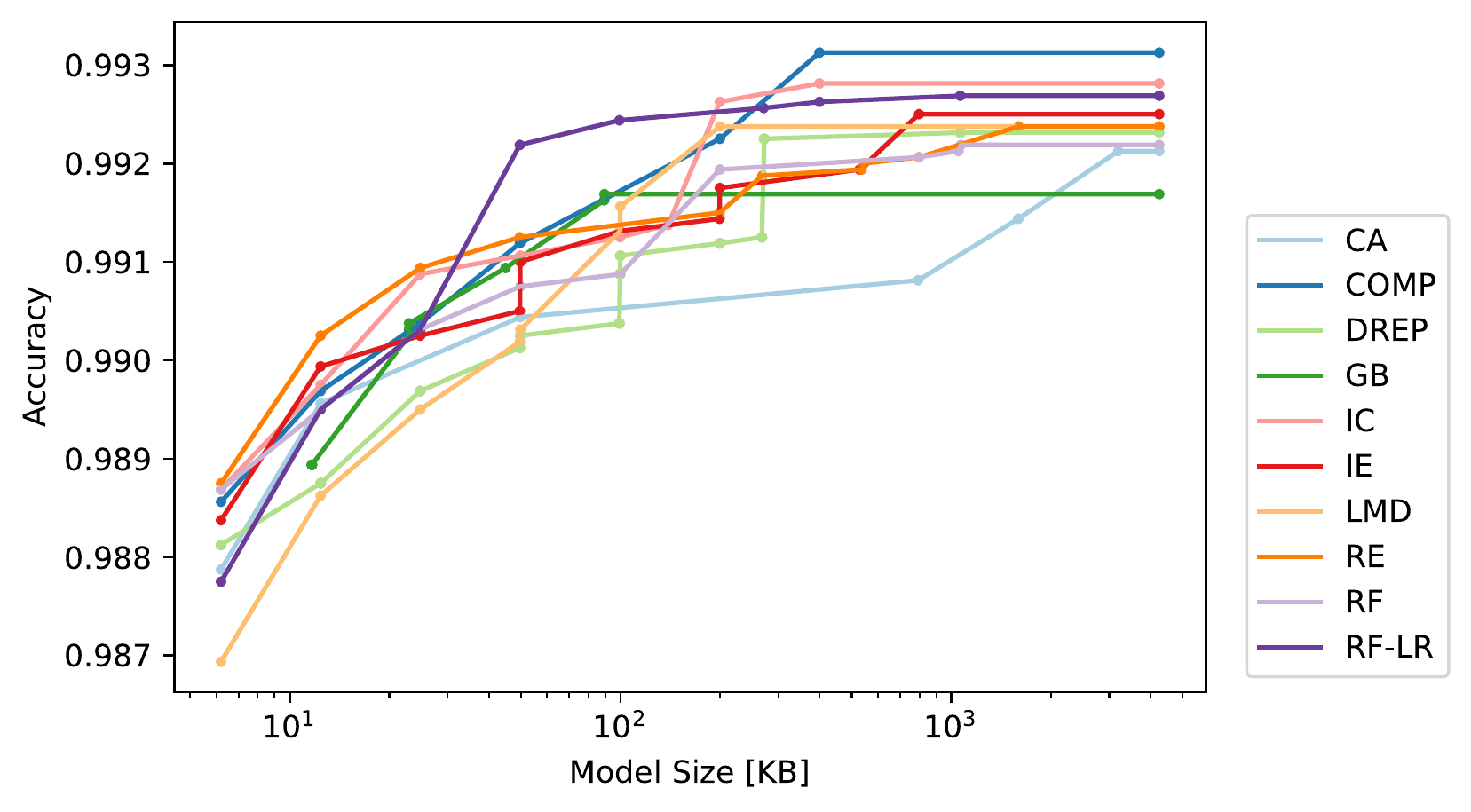}
\end{minipage}\hfill
\begin{minipage}{.49\textwidth}
    \centering 
    \includegraphics[width=\textwidth,keepaspectratio]{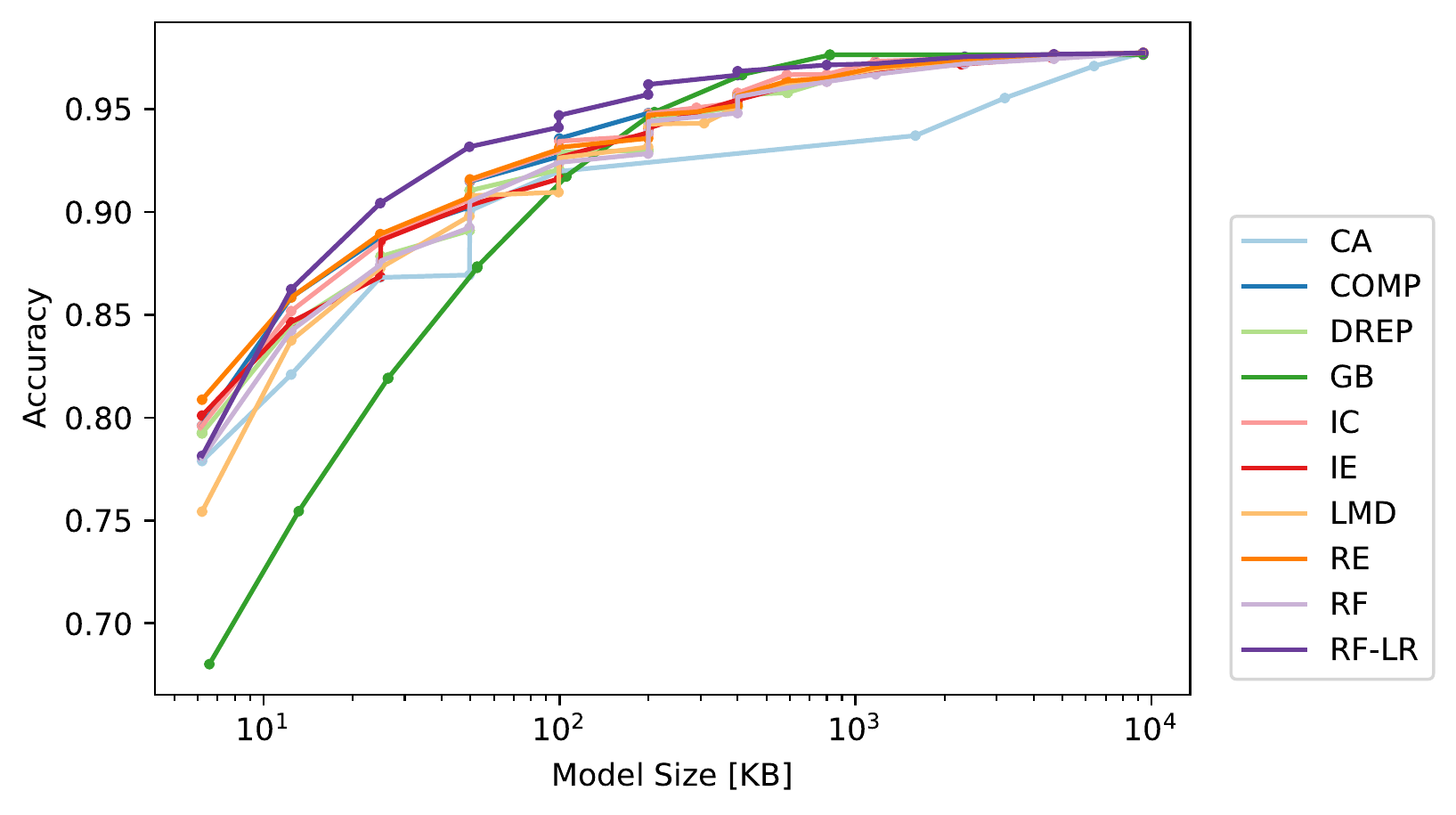}
\end{minipage}
\caption{5-fold cross-validation accuracy over the size of the ensemble for different $n_l$ and different $M$ on the chess dataset. Single points are the individual parameter configurations whereas the solid line depicts the corresponding Pareto Front. Left side show the ida2016 dataset, right side shows the japanese-vowels dataset.}
\end{figure}

\begin{figure}[H]
\begin{minipage}{.49\textwidth}
    \centering
    \includegraphics[width=\textwidth,keepaspectratio]{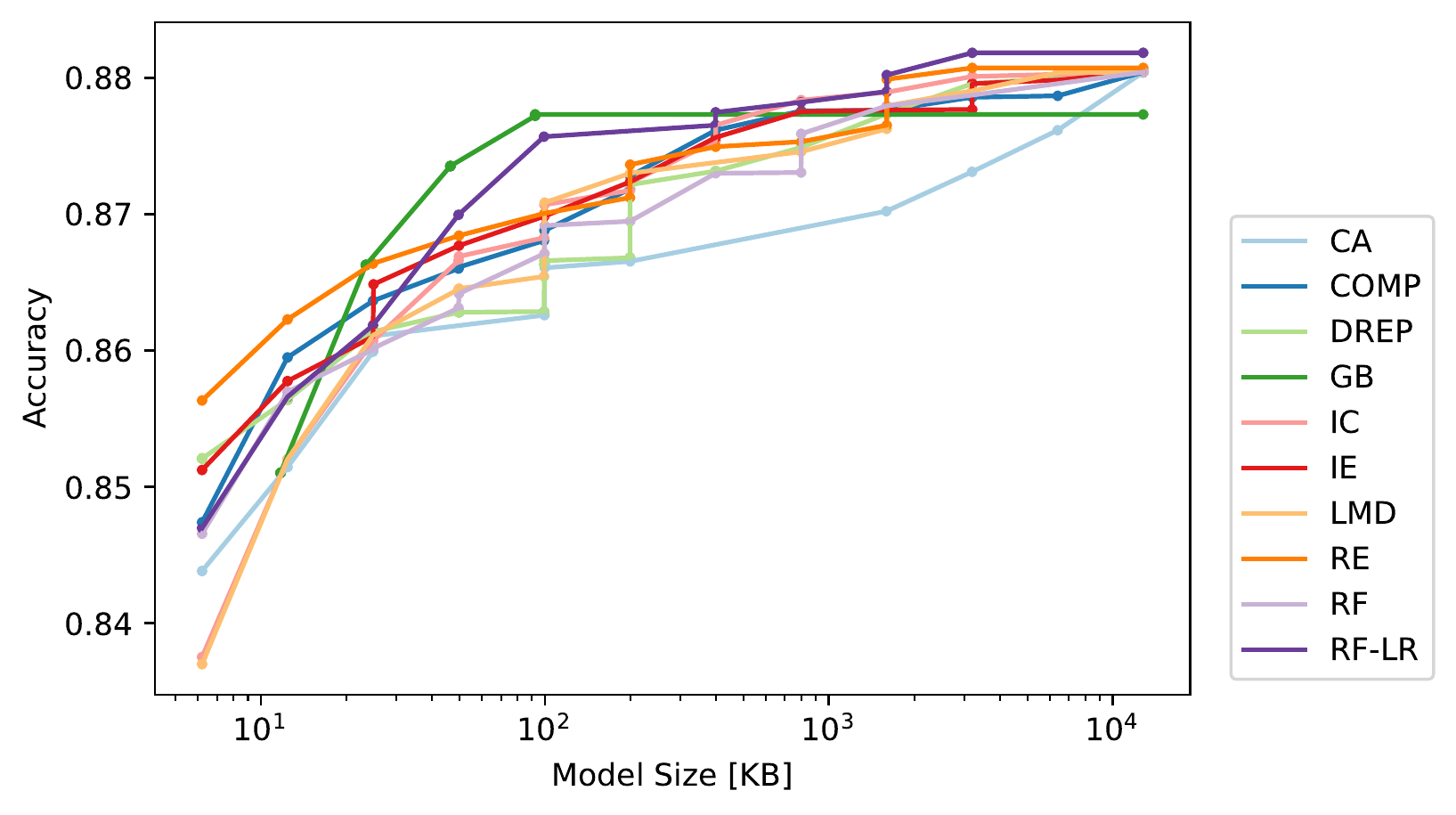}
\end{minipage}\hfill
\begin{minipage}{.49\textwidth}
    \centering 
    \includegraphics[width=\textwidth,keepaspectratio]{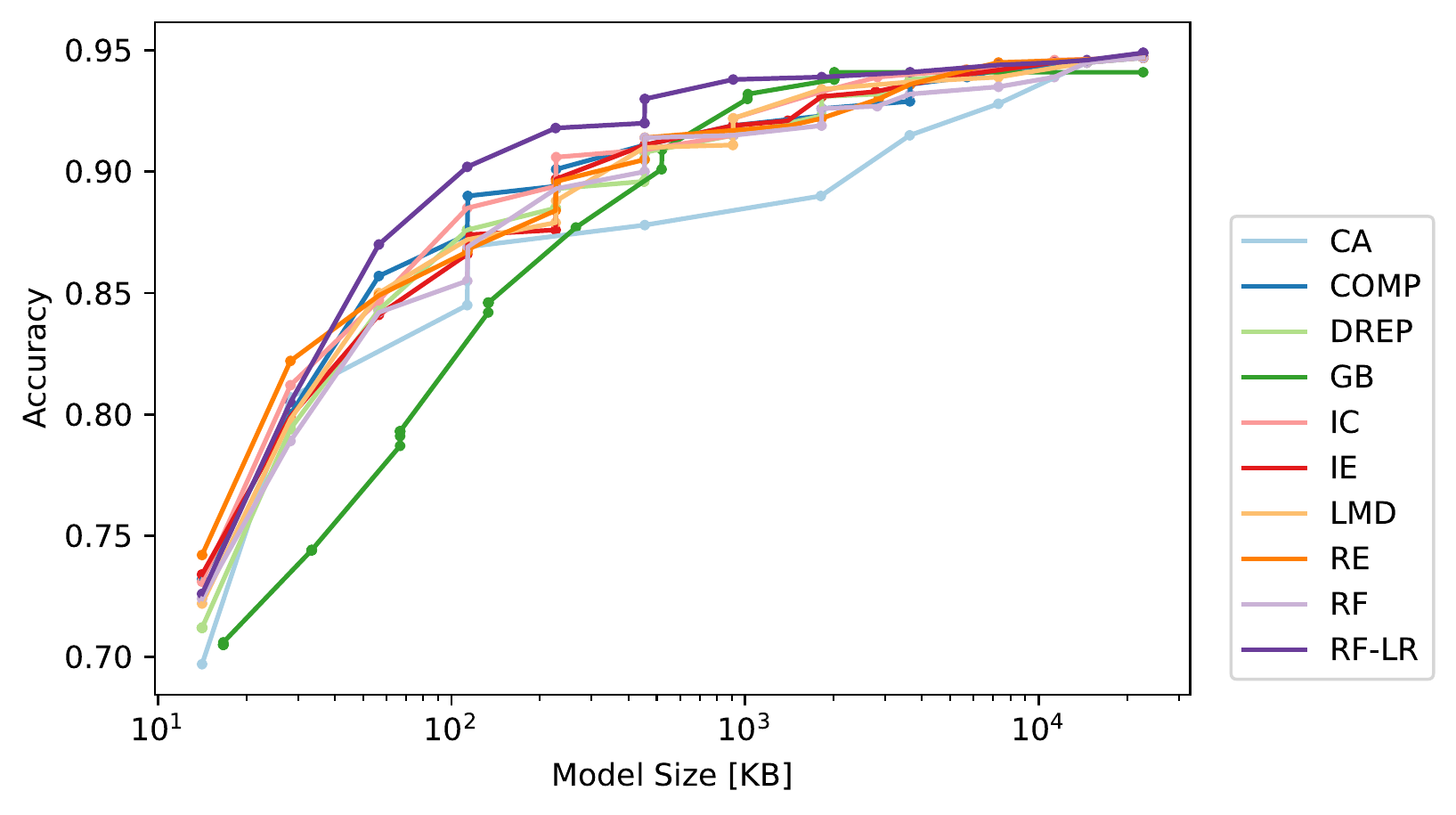}
\end{minipage}
\caption{5-fold cross-validation accuracy over the size of the ensemble for different $n_l$ and different $M$ on the chess dataset. Single points are the individual parameter configurations whereas the solid line depicts the corresponding Pareto Front. Left side show the magic dataset, right side shows the mnist dataset.}
\end{figure}

\begin{figure}[H]
\begin{minipage}{.49\textwidth}
    \centering
    \includegraphics[width=\textwidth,keepaspectratio]{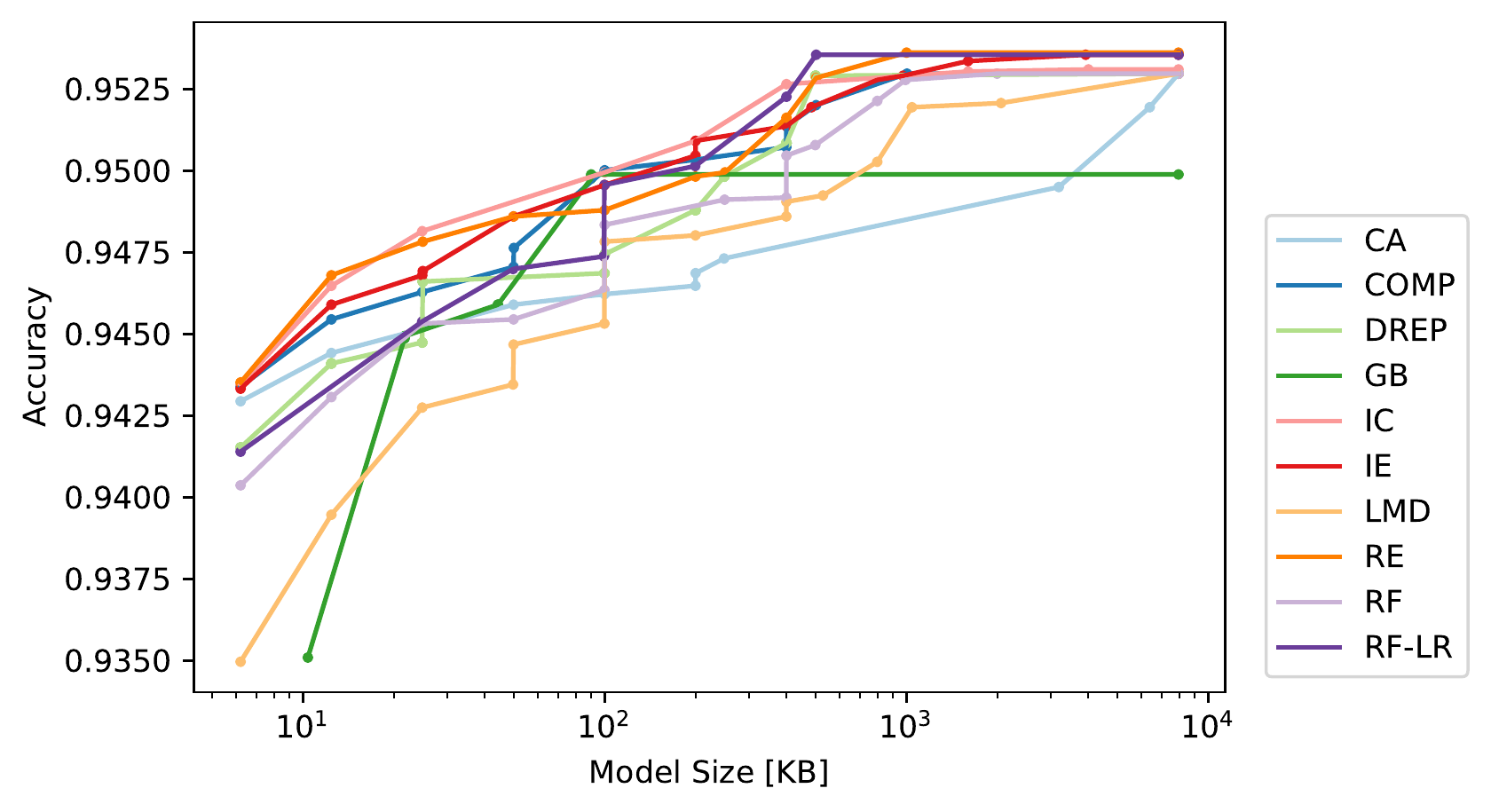}
\end{minipage}\hfill
\begin{minipage}{.49\textwidth}
    \centering 
    \includegraphics[width=\textwidth,keepaspectratio]{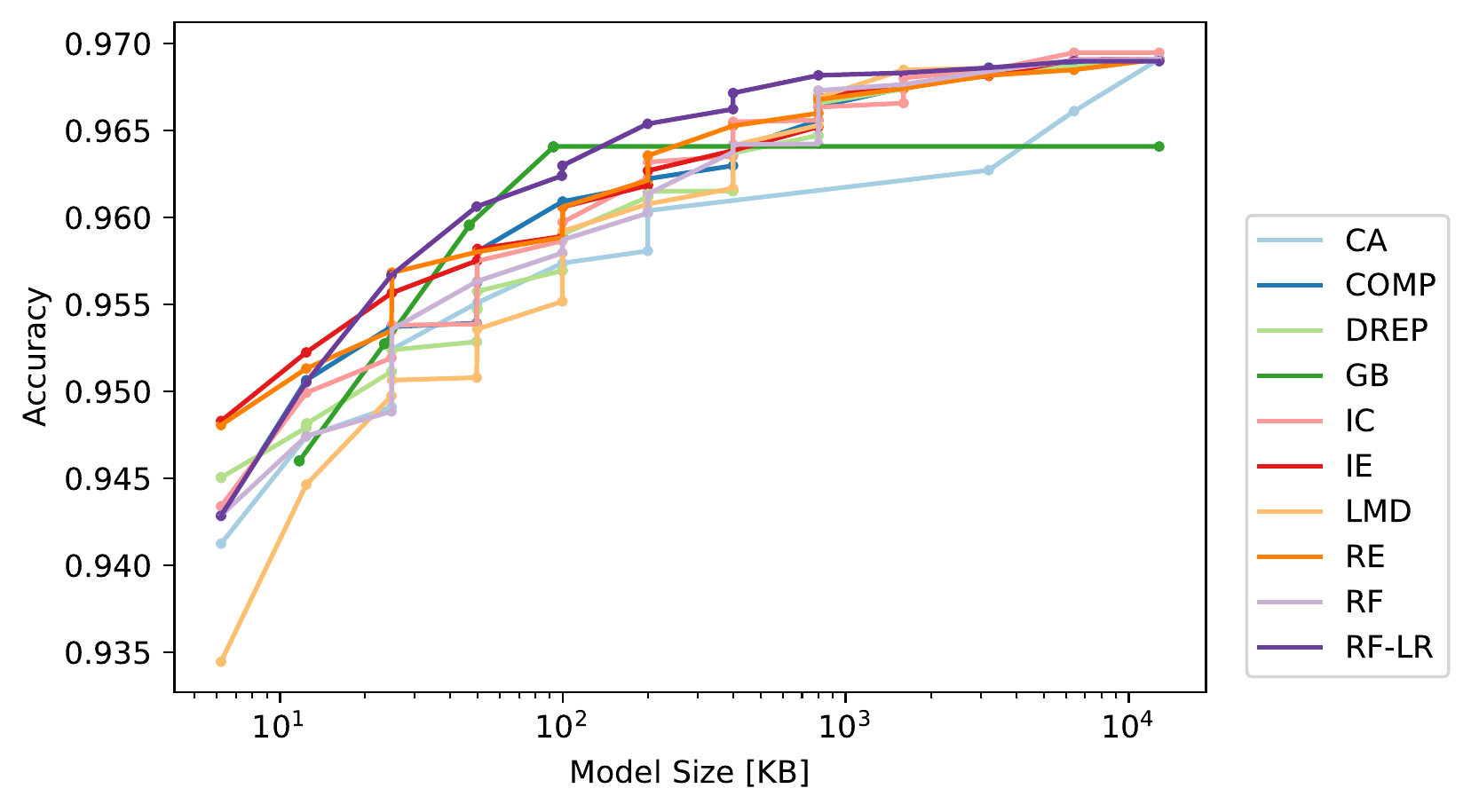}
\end{minipage}
\caption{5-fold cross-validation accuracy over the size of the ensemble for different $n_l$ and different $M$ on the chess dataset. Single points are the individual parameter configurations whereas the solid line depicts the corresponding Pareto Front. Left side show the mozilla dataset, right side shows the nomao dataset.}
\end{figure}

\begin{figure}[H]
\begin{minipage}{.49\textwidth}
    \centering
    \includegraphics[width=\textwidth,keepaspectratio]{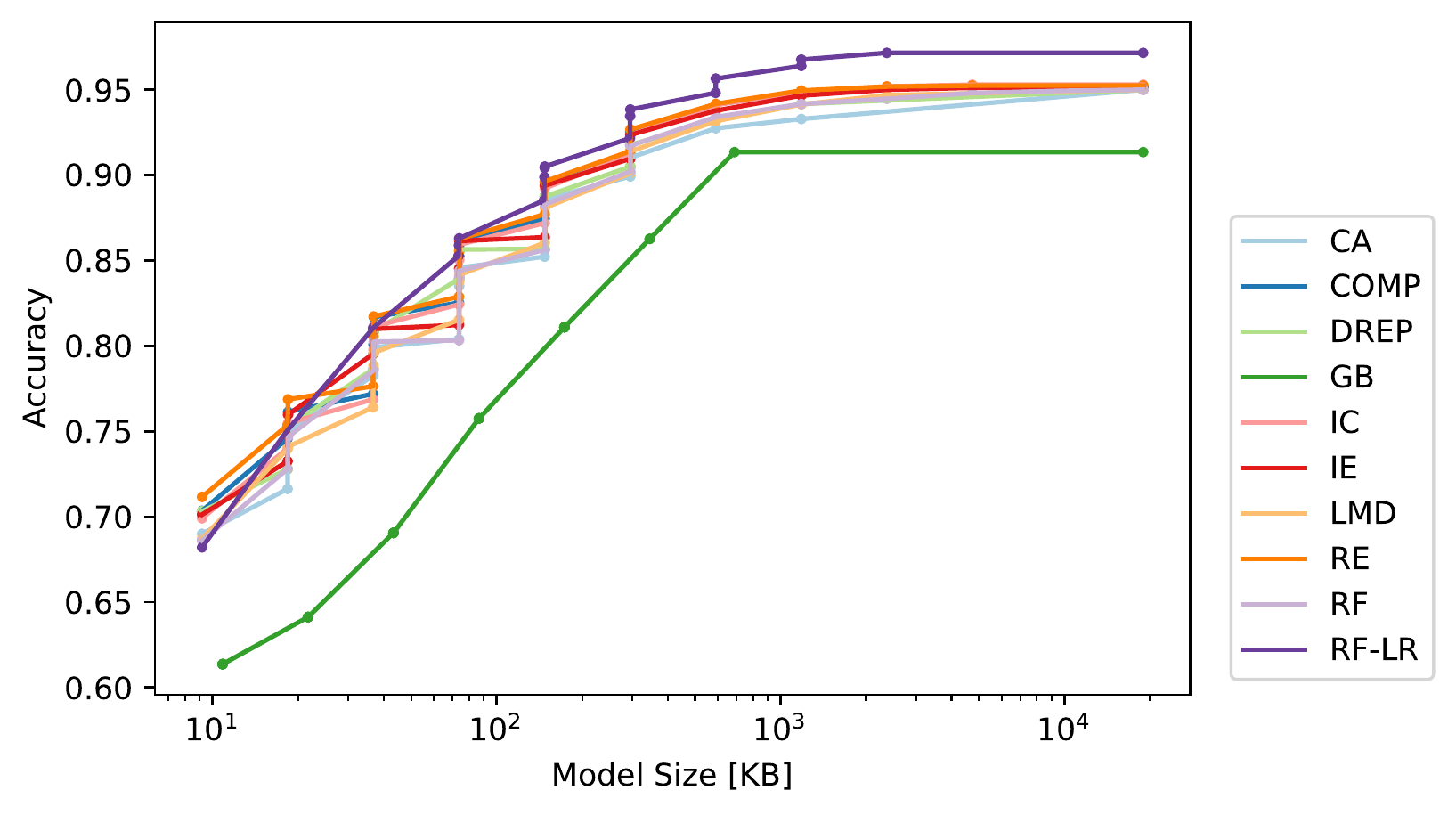}
\end{minipage}\hfill
\begin{minipage}{.49\textwidth}
    \centering 
    \includegraphics[width=\textwidth,keepaspectratio]{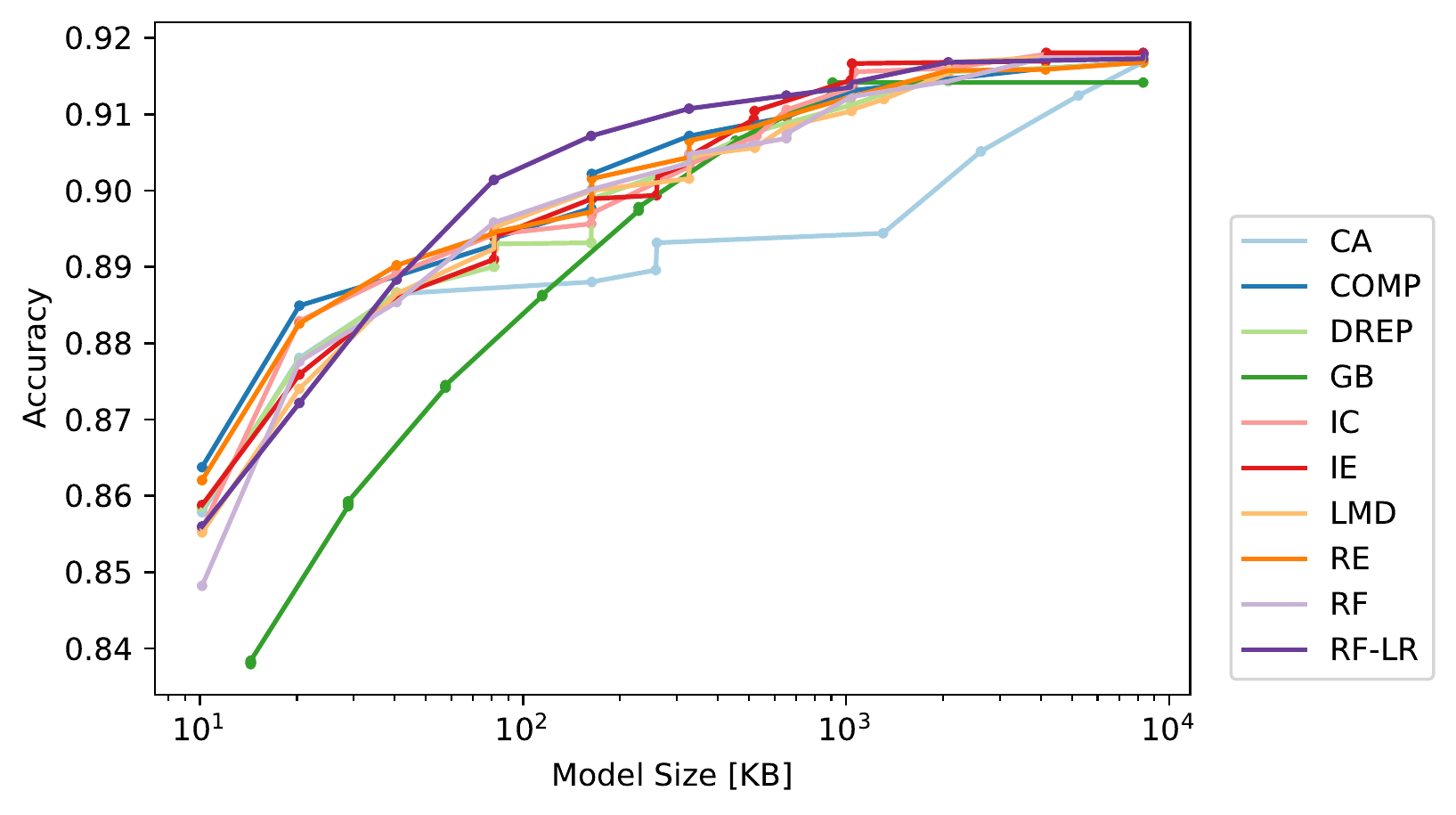}
\end{minipage}
\caption{5-fold cross-validation accuracy over the size of the ensemble for different $n_l$ and different $M$ on the chess dataset. Single points are the individual parameter configurations whereas the solid line depicts the corresponding Pareto Front. Left side show the postures dataset, right side shows the satimage dataset.}
\end{figure}

\subsection{Accuracies under various resource constraints}

\begin{table}
    \centering
    \resizebox{\textwidth}{!}{
    \begin{tabular}{lrrrrrrrrrr}
\toprule
{model} & {CA} & {COMP} & {DREP} & {GB} & {IC} & {IE} & {LMD} & {RE} & {RF} & {RF-LR} \\
\midrule
adult & 85.120 & 85.467 & 85.403 & \textbf{86.244} & 85.295 & 85.824 & 85.056 & 86.035 & 85.234 & 85.891 \\
anura & 95.789 & 96.484 & 96.220 & 91.564 & 96.609 & 96.372 & 96.081 & 96.609 & 96.178 & \textbf{96.831} \\
avila & 86.102 & 93.382 & 91.115 & 59.496 & 93.435 & 93.219 & 82.446 & \textbf{94.197} & 86.285 & 88.355 \\
bank & 89.741 & 89.985 & 89.927 & \textbf{90.527} & 90.213 & 90.226 & 89.706 & 90.246 & 89.764 & 90.343 \\
chess & 44.914 & 47.986 & 46.696 & 30.175 & 46.753 & 48.371 & 45.391 & 48.571 & 44.614 & \textbf{51.205} \\
connect & 71.551 & 73.052 & 72.595 & 67.584 & 73.057 & 73.223 & 71.993 & 73.313 & 71.466 & \textbf{75.160} \\
eeg & 81.015 & 83.037 & 82.917 & 79.226 & 82.911 & 83.451 & 81.455 & 83.565 & 82.443 & \textbf{84.059} \\
elec & 82.252 & 82.122 & 83.009 & 81.941 & 79.401 & 83.737 & 81.649 & \textbf{83.942} & 82.027 & 83.464 \\
ida2016 & 98.956 & 99.038 & 98.969 & 99.038 & 99.087 & 99.025 & 98.950 & \textbf{99.094} & 99.031 & 99.031 \\
japanese-vowels & 86.819 & 88.796 & 87.843 & 81.930 & 88.716 & 88.626 & 87.280 & 88.927 & 87.632 & \textbf{90.433} \\
magic & 86.109 & 86.366 & 86.140 & 86.629 & 86.072 & 86.487 & 86.114 & \textbf{86.640} & 86.019 & 86.187 \\
mnist & 80.700 & 80.000 & 79.400 & 70.600 & 81.200 & 79.900 & 79.800 & \textbf{82.200} & 78.900 & 80.500 \\
mozilla & 94.442 & 94.628 & 94.661 & 94.493 & \textbf{94.815} & 94.693 & 94.275 & 94.783 & 94.532 & 94.538 \\
nomao & 95.242 & 95.372 & 95.239 & 95.273 & 95.381 & 95.567 & 95.065 & \textbf{95.683} & 95.358 & 95.668 \\
postures & 74.960 & 76.138 & 75.261 & 64.126 & 75.477 & 75.978 & 74.101 & \textbf{76.873} & 74.673 & 75.069 \\
satimage & 87.807 & \textbf{88.491} & 87.776 & 85.925 & 88.289 & 87.589 & 87.403 & 88.258 & 87.760 & 87.216 \\
\bottomrule
\end{tabular}
    }
    \caption{Test accuracies for models with a memory consumption below 32 KB for each method and each dataset averaged over a 5 fold cross validation. Rounded to the third decimal digit. Larger is better. The best method is depicted in bold.}
\end{table}

\begin{table}
    \centering
    \resizebox{\textwidth}{!}{
    \begin{tabular}{lrrrrrrrrrr}
\toprule
{model} & {CA} & {COMP} & {DREP} & {GB} & {IC} & {IE} & {LMD} & {RE} & {RF} & {RF-LR} \\
\midrule
adult & 85.455 & 85.777 & 85.618 & \textbf{86.616} & 85.799 & 85.882 & 85.378 & 86.128 & 85.464 & 86.241 \\
anura & 96.511 & 97.137 & 96.873 & 95.024 & 97.192 & 96.928 & 96.539 & 97.067 & 96.525 & \textbf{97.512} \\
avila & 91.930 & 96.957 & 95.318 & 67.724 & 96.760 & 96.689 & 86.965 & \textbf{97.048} & 92.534 & 94.278 \\
bank & 89.772 & 90.049 & 89.927 & \textbf{90.677} & 90.213 & 90.330 & 89.819 & 90.336 & 89.830 & 90.522 \\
chess & 48.025 & 51.714 & 50.492 & 34.969 & 50.356 & 51.793 & 49.041 & 52.495 & 48.090 & \textbf{57.382} \\
connect & 72.426 & 73.898 & 73.409 & 70.847 & 73.724 & 73.832 & 73.110 & 74.134 & 72.666 & \textbf{77.498} \\
eeg & 84.419 & 85.574 & 84.419 & 82.463 & 84.780 & 85.033 & 83.852 & 85.527 & 84.606 & \textbf{86.622} \\
elec & 83.523 & 83.620 & 84.075 & 83.878 & 82.749 & 84.556 & 82.817 & 84.680 & 83.391 & \textbf{84.894} \\
ida2016 & 99.044 & 99.119 & 99.025 & 99.094 & 99.106 & 99.100 & 99.031 & 99.125 & 99.075 & \textbf{99.219} \\
japanese-vowels & 90.061 & 91.487 & 91.045 & 87.341 & 91.517 & 90.332 & 90.794 & 91.587 & 90.503 & \textbf{93.173} \\
magic & 86.109 & 86.613 & 86.282 & \textbf{87.355} & 86.692 & 86.771 & 86.456 & 86.845 & 86.419 & 86.997 \\
mnist & 80.700 & 85.700 & 84.300 & 74.400 & 84.700 & 84.100 & 85.000 & 84.900 & 84.200 & \textbf{87.000} \\
mozilla & 94.590 & 94.764 & 94.661 & 94.590 & 94.815 & \textbf{94.860} & 94.468 & \textbf{94.860} & 94.545 & 94.699 \\
nomao & 95.508 & 95.804 & 95.575 & 95.958 & 95.749 & 95.819 & 95.358 & 95.802 & 95.633 & \textbf{96.063} \\
postures & 79.913 & 81.688 & 80.827 & 69.058 & 81.137 & 80.996 & 79.599 & \textbf{81.727} & 80.246 & 81.081 \\
satimage & 88.647 & 88.880 & 88.663 & 87.449 & 88.911 & 88.616 & 88.647 & \textbf{89.020} & 88.538 & 88.834 \\
\bottomrule
\end{tabular}
    }
    \caption{Test accuracies for models with a memory consumption below 64 KB for each method and each dataset averaged over a 5 fold cross validation. Rounded to the third decimal digit. Larger is better. The best method is depicted in bold.}
\end{table}

\begin{table}
    \centering
    \resizebox{\textwidth}{!}{
    \begin{tabular}{lrrrrrrrrrr}
\toprule
{model} & {CA} & {COMP} & {DREP} & {GB} & {IC} & {IE} & {LMD} & {RE} & {RF} & {RF-LR} \\
\midrule
adult & 85.685 & 85.925 & 85.784 & \textbf{87.135} & 85.980 & 86.164 & 85.738 & 86.155 & 85.787 & 86.508 \\
anura & 96.511 & 97.568 & 97.206 & 96.553 & 97.679 & 97.512 & 97.137 & 97.415 & 97.206 & \textbf{97.693} \\
avila & 94.738 & 98.720 & 95.697 & 75.262 & 98.845 & 98.730 & 91.015 & \textbf{98.869} & 93.411 & 97.791 \\
bank & 89.967 & 90.148 & 89.985 & 90.737 & 90.259 & 90.376 & 89.830 & 90.354 & 89.994 & \textbf{90.794} \\
chess & 53.315 & 56.027 & 54.370 & 40.719 & 55.318 & 55.703 & 53.201 & 56.672 & 53.261 & \textbf{62.749} \\
connect & 73.348 & 74.897 & 74.065 & 73.850 & 75.083 & 75.026 & 74.327 & 75.305 & 73.825 & \textbf{79.142} \\
eeg & 85.534 & 87.370 & 86.242 & 85.734 & 87.136 & 87.163 & 86.355 & 87.457 & 86.442 & \textbf{88.992} \\
elec & 84.589 & 84.318 & 84.872 & 85.748 & 84.635 & 85.549 & 83.896 & 85.851 & 84.574 & \textbf{86.416} \\
ida2016 & 99.044 & 99.119 & 99.106 & 99.169 & 99.125 & 99.131 & 99.156 & 99.125 & 99.087 & \textbf{99.244} \\
japanese-vowels & 91.979 & 93.575 & 92.912 & 91.748 & 93.444 & 92.651 & 92.631 & 93.143 & 92.410 & \textbf{94.699} \\
magic & 86.608 & 86.882 & 86.661 & \textbf{87.733} & 87.071 & 86.987 & 87.087 & 87.008 & 86.918 & 87.570 \\
mnist & 86.900 & 89.000 & 87.600 & 79.300 & 88.500 & 87.400 & 87.200 & 86.800 & 86.900 & \textbf{90.200} \\
mozilla & 94.622 & \textbf{95.002} & 94.744 & 94.989 & 94.995 & 94.957 & 94.783 & 94.879 & 94.834 & 94.957 \\
nomao & 95.738 & 96.092 & 95.903 & \textbf{96.408} & 95.973 & 96.060 & 95.920 & 96.060 & 95.871 & 96.298 \\
postures & 84.582 & 86.166 & 85.655 & 75.760 & 85.933 & 86.149 & 84.167 & 86.267 & 84.400 & \textbf{86.300} \\
satimage & 88.647 & 89.378 & 89.300 & 88.631 & 89.425 & 89.393 & 89.518 & 89.456 & 89.580 & \textbf{90.140} \\
\bottomrule
\end{tabular}
    }
    \caption{Test accuracies for models with a memory consumption below 128 KB for each method and each dataset averaged over a 5 fold cross validation. Rounded to the third decimal digit. Larger is better. The best method is depicted in bold.}
\end{table}

\begin{table}
    \centering
    \resizebox{\textwidth}{!}{
    \begin{tabular}{lrrrrrrrrrr}
\toprule
{model} & {CA} & {COMP} & {DREP} & {GB} & {IC} & {IE} & {LMD} & {RE} & {RF} & {RF-LR} \\
\midrule
adult & 85.774 & 86.174 & 85.956 & \textbf{87.135} & 86.011 & 86.232 & 85.738 & 86.272 & 85.799 & 86.508 \\
anura & 96.511 & 97.818 & 97.748 & 97.429 & \textbf{98.040} & 97.929 & 97.582 & 97.818 & 97.735 & 98.013 \\
avila & 95.016 & 99.022 & 97.427 & 82.820 & 99.176 & 99.123 & 93.315 & \textbf{99.248} & 96.526 & 98.807 \\
bank & 89.967 & 90.297 & 90.082 & 90.737 & 90.290 & 90.443 & 89.952 & 90.471 & 90.038 & \textbf{90.874} \\
chess & 56.911 & 59.998 & 57.314 & 44.678 & 59.791 & 59.428 & 57.496 & 60.636 & 56.238 & \textbf{66.852} \\
connect & 74.522 & 76.270 & 75.119 & 76.170 & 76.239 & 76.333 & 75.100 & 76.303 & 74.997 & \textbf{80.340} \\
eeg & 87.223 & 88.364 & 88.511 & 85.734 & 88.959 & 88.778 & 87.784 & 88.785 & 87.951 & \textbf{90.454} \\
elec & 85.220 & 85.810 & 85.832 & 85.748 & 86.043 & 86.692 & 84.922 & 86.562 & 85.362 & \textbf{87.829} \\
ida2016 & 99.106 & 99.225 & 99.119 & 99.169 & \textbf{99.262} & 99.175 & 99.238 & 99.175 & 99.194 & 99.244 \\
japanese-vowels & 91.979 & 94.810 & 94.428 & 94.860 & 94.790 & 94.077 & 94.278 & 94.709 & 94.408 & \textbf{96.205} \\
magic & 86.655 & 87.281 & 87.218 & \textbf{87.733} & 87.234 & 87.239 & 87.302 & 87.365 & 86.950 & 87.570 \\
mnist & 86.900 & 90.100 & 89.300 & 84.600 & 90.600 & 89.700 & 88.800 & 89.600 & 89.300 & \textbf{91.800} \\
mozilla & 94.731 & 95.034 & 94.982 & 94.989 & \textbf{95.092} & \textbf{95.092} & 94.802 & 94.995 & 94.912 & 95.014 \\
nomao & 96.039 & 96.222 & 96.150 & 96.408 & 96.318 & 96.269 & 96.077 & 96.356 & 96.135 & \textbf{96.539} \\
postures & 88.587 & 89.596 & 88.747 & 81.100 & 89.231 & 89.390 & 88.085 & 89.633 & 88.281 & \textbf{90.504} \\
satimage & 88.802 & 90.218 & 89.891 & 89.782 & 89.705 & 89.891 & 90.000 & 90.156 & 90.016 & \textbf{90.715} \\
\bottomrule
\end{tabular}
    }
    \caption{Test accuracies for models with a memory consumption below 256 KB for each method and each dataset averaged over a 5 fold cross validation. Rounded to the third decimal digit. Larger is better. The best method is depicted in bold.}
\end{table}

\section{Revisiting Ensemble Pruning on More Datasets with a dedicated pruning set}

Some authors use a dedicated pruning set (see e.g. \cite{}) for ensemble pruning which was not used for training the ensemble. For completeness, we adapt this approach into the experimental protocol. We now split the training data into two sets with $2/3$ and $1/3$ of the original training data. The $2/3$ of the training data is used to train the base ensemble, and the $1/3$ of the data is used for pruning. As before, we either use a 5-fold cross validation or the given test/train split. For reference, recall our experimental protocol: Oshiro et al. showed in \cite{oshiro/etal/2012} that the prediction of a RF stabilizes between $128$ and $256$ trees in the ensemble and adding more trees to the ensemble does not yield significantly better results. Hence, we train the `base' Random Forests with $M = 256$ trees. To control the individual errors of trees we set the maximum number of leaf nodes $n_l$ to values between $n_l \in \{64,128,256,512,1024\}$. For ensemble pruning we use RE and compare it against a random selection of trees from the original ensemble (which is the same a training a smaller forest directly). In both cases a sub-ensemble with $K \in \{2,4,8,16,32,64,128,256\}$ members is selected so that for $K=256$ the original RF is recovered. 

\begin{figure}[H]
\begin{minipage}{.49\textwidth}
    \centering
    \includegraphics[width=\textwidth,keepaspectratio]{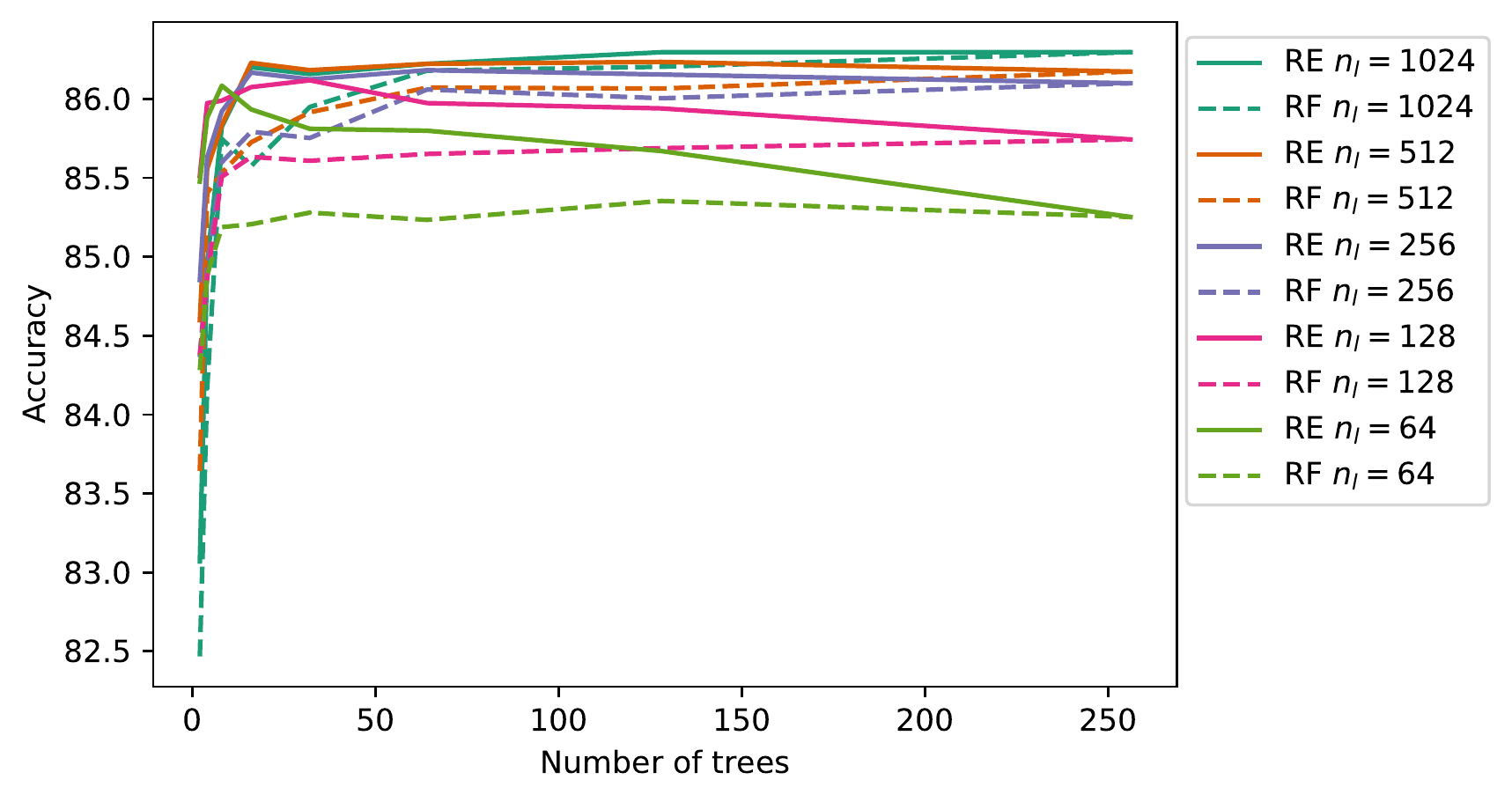}
\end{minipage}\hfill
\begin{minipage}{.49\textwidth}
    \centering 
    \resizebox{\textwidth}{!}{
        \begin{tabular}{llrrrrrr}
\toprule
     & model &   COMP &   DREP &     IC &     IE &    LMD &     RE \\
max\_leaf\_nodes & n\_estimators &        &        &        &        &        &        \\
\midrule
64   & 8.0   &  85.59 &  85.14 &  85.38 &  85.53 &  85.07 &  86.08 \\
     & 16.0  &  85.64 &  85.39 &  85.61 &  85.66 &  85.17 &  85.93 \\
     & 32.0  &  85.64 &  85.43 &  85.80 &  85.71 &  85.35 &  85.81 \\
     & 64.0  &  85.71 &  85.28 &  85.83 &  85.68 &  85.40 &  85.80 \\
     & 128.0 &  85.57 &  85.23 &  85.65 &  85.62 &  85.34 &  85.67 \\
128  & 8.0   &  85.55 &  85.51 &  85.51 &  85.93 &  85.16 &  85.99 \\
     & 16.0  &  85.85 &  85.63 &  85.84 &  85.97 &  85.29 &  86.08 \\
     & 32.0  &  85.91 &  85.65 &  85.92 &  85.93 &  85.51 &  86.12 \\
     & 64.0  &  85.88 &  85.68 &  86.06 &  86.01 &  85.72 &  85.97 \\
     & 128.0 &  85.91 &  85.69 &  85.97 &  85.99 &  85.70 &  85.94 \\
256  & 8.0   &  85.77 &  85.71 &  85.65 &  86.06 &  85.43 &  85.92 \\
     & 16.0  &  85.87 &  85.97 &  86.05 &  86.04 &  85.63 &  86.17 \\
     & 32.0  &  86.14 &  85.92 &  86.15 &  86.18 &  85.71 &  86.12 \\
     & 64.0  &  86.12 &  85.97 &  86.17 &  86.16 &  85.92 &  86.18 \\
     & 128.0 &  86.15 &  86.00 &  86.19 &  86.20 &  85.95 &  86.16 \\
512  & 8.0   &  85.90 &  85.72 &  85.68 &  85.95 &  85.42 &  85.84 \\
     & 16.0  &  85.97 &  85.99 &  86.02 &  86.08 &  85.63 &  86.23 \\
     & 32.0  &  86.13 &  86.11 &  86.12 &  86.18 &  85.91 &  86.18 \\
     & 64.0  &  86.18 &  86.19 &  86.19 &  86.31 &  86.05 &  86.22 \\
     & 128.0 &  86.19 &  86.24 &  86.23 &  86.29 &  86.04 &  86.24 \\
1024 & 8.0   &  85.56 &  85.55 &  85.43 &  85.79 &  85.23 &  85.82 \\
     & 16.0  &  85.90 &  85.87 &  85.92 &  86.03 &  85.81 &  86.20 \\
     & 32.0  &  86.07 &  86.03 &  86.24 &  86.17 &  86.06 &  86.16 \\
     & 64.0  &  86.24 &  86.15 &  86.33 &  86.26 &  86.12 &  86.22 \\
     & 128.0 &  86.30 &  86.20 &  86.27 &  86.32 &  86.24 &  86.30 \\
\bottomrule
\end{tabular}
    }
\end{minipage}
\caption{(Left) The error over the number of trees in the ensemble on the adult dataset. Dashed lines depict the Random Forest and solid lines are the corresponding pruned ensemble via Reduced Error pruning. (Right) The 5-fold cross-validation accuracy  on the adult dataset. Rounded to the second decimal digit. Larger is better.}
\end{figure}

\begin{figure}[H]
\begin{minipage}{.49\textwidth}
    \centering
    \includegraphics[width=\textwidth,keepaspectratio]{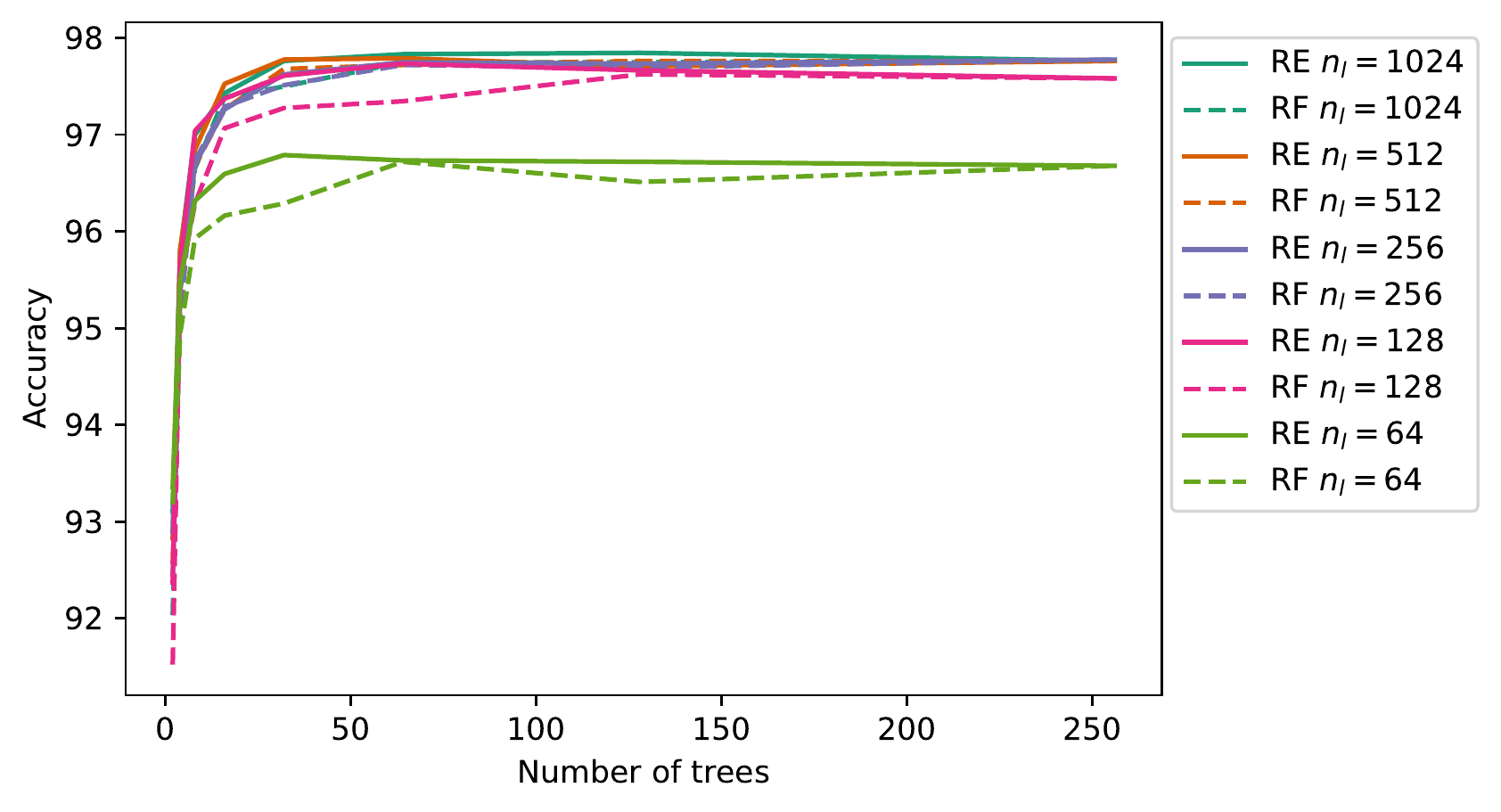}
\end{minipage}\hfill
\begin{minipage}{.49\textwidth}
    \centering 
    \resizebox{\textwidth}{!}{
        \begin{tabular}{llrrrrrr}
\toprule
     & model &   COMP &   DREP &     IC &     IE &    LMD &     RE \\
max\_leaf\_nodes & n\_estimators &        &        &        &        &        &        \\
\midrule
64   & 8.0   &  96.50 &  96.26 &  96.64 &  96.65 &  95.89 &  96.32 \\
     & 16.0  &  96.69 &  96.44 &  96.85 &  96.76 &  96.23 &  96.59 \\
     & 32.0  &  96.86 &  96.55 &  97.07 &  96.80 &  96.47 &  96.79 \\
     & 64.0  &  96.83 &  96.58 &  96.89 &  96.76 &  96.55 &  96.73 \\
     & 128.0 &  96.82 &  96.75 &  96.79 &  96.76 &  96.71 &  96.72 \\
128  & 8.0   &  96.62 &  96.69 &  96.79 &  96.48 &  96.50 &  97.04 \\
     & 16.0  &  97.26 &  97.10 &  97.33 &  97.18 &  97.16 &  97.37 \\
     & 32.0  &  97.53 &  97.44 &  97.68 &  97.47 &  97.29 &  97.61 \\
     & 64.0  &  97.69 &  97.57 &  97.72 &  97.48 &  97.58 &  97.73 \\
     & 128.0 &  97.62 &  97.50 &  97.68 &  97.51 &  97.58 &  97.67 \\
256  & 8.0   &  96.97 &  96.71 &  96.94 &  96.86 &  96.76 &  96.69 \\
     & 16.0  &  97.54 &  97.26 &  97.64 &  97.58 &  97.43 &  97.26 \\
     & 32.0  &  97.80 &  97.54 &  97.78 &  97.60 &  97.62 &  97.62 \\
     & 64.0  &  97.98 &  97.78 &  97.89 &  97.72 &  97.73 &  97.75 \\
     & 128.0 &  97.79 &  97.79 &  97.79 &  97.73 &  97.85 &  97.73 \\
512  & 8.0   &  96.61 &  96.55 &  96.91 &  96.69 &  96.82 &  96.85 \\
     & 16.0  &  97.37 &  97.10 &  97.48 &  97.39 &  97.46 &  97.53 \\
     & 32.0  &  97.72 &  97.61 &  97.68 &  97.64 &  97.73 &  97.78 \\
     & 64.0  &  97.83 &  97.71 &  97.92 &  97.65 &  97.73 &  97.79 \\
     & 128.0 &  97.85 &  97.80 &  97.87 &  97.83 &  97.82 &  97.71 \\
1024 & 8.0   &  96.79 &  96.62 &  96.79 &  96.65 &  96.75 &  97.00 \\
     & 16.0  &  97.53 &  97.16 &  97.53 &  97.54 &  97.32 &  97.43 \\
     & 32.0  &  97.80 &  97.43 &  97.79 &  97.75 &  97.69 &  97.76 \\
     & 64.0  &  97.89 &  97.64 &  97.94 &  97.87 &  97.64 &  97.83 \\
     & 128.0 &  97.76 &  97.62 &  97.85 &  97.85 &  97.83 &  97.85 \\
\bottomrule
\end{tabular}
    }
\end{minipage}
\caption{(Left) The error over the number of trees in the ensemble on the anura dataset. Dashed lines depict the Random Forest and solid lines are the corresponding pruned ensemble via Reduced Error pruning. (Right) The 5-fold cross-validation accuracy  on the anura dataset. Rounded to the second decimal digit. Larger is better.}
\end{figure}

\begin{figure}[H]
\begin{minipage}{.49\textwidth}
    \centering
    \includegraphics[width=\textwidth,keepaspectratio]{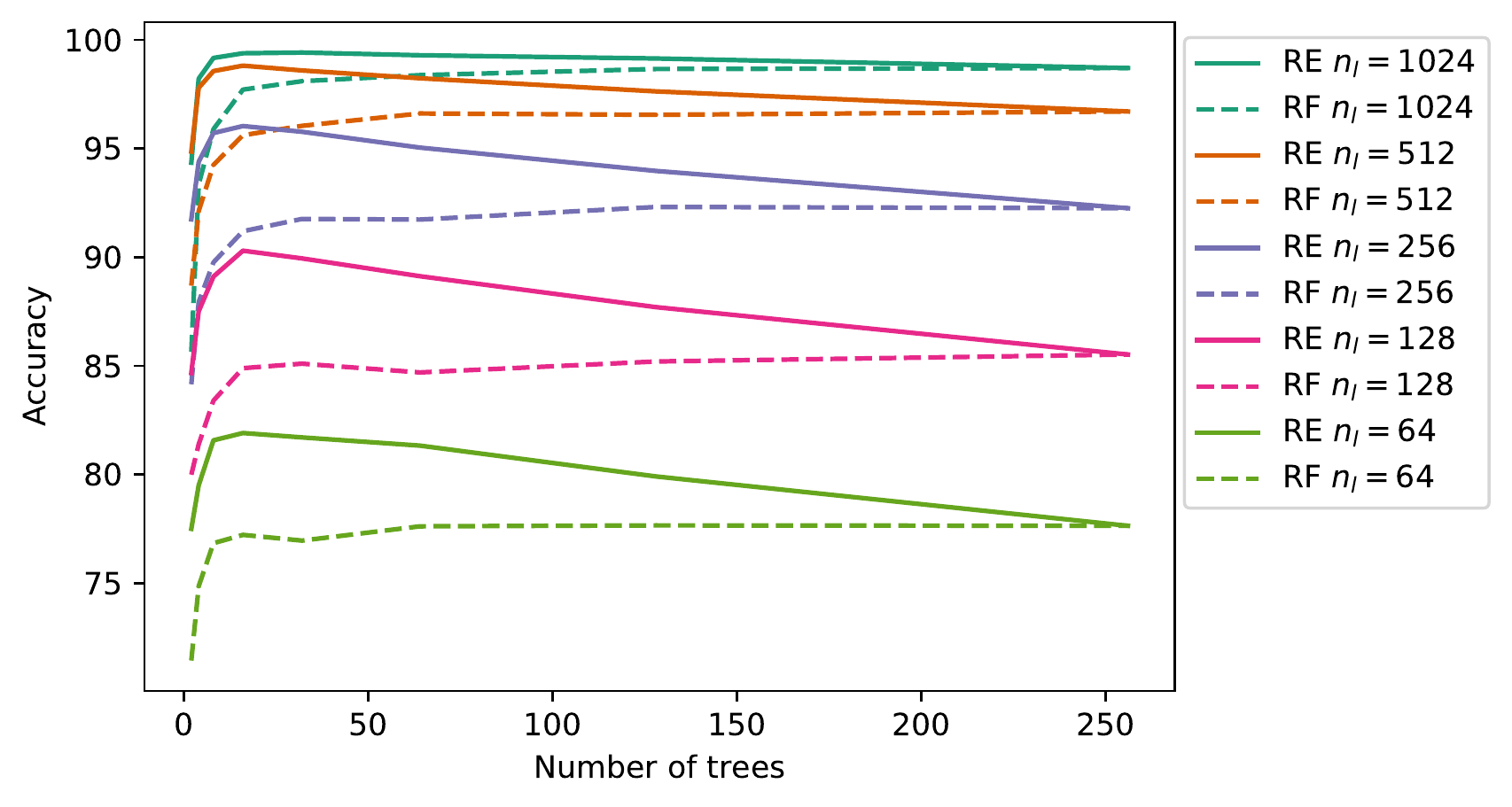}
\end{minipage}\hfill
\begin{minipage}{.49\textwidth}
    \centering 
    \resizebox{\textwidth}{!}{
        \begin{tabular}{llrrrrrr}
\toprule
     & model &   COMP &   DREP &     IC &     IE &    LMD &     RE \\
max\_leaf\_nodes & n\_estimators &        &        &        &        &        &        \\
\midrule
64   & 8.0   &  80.60 &  77.01 &  80.67 &  79.33 &  78.23 &  81.57 \\
     & 16.0  &  81.63 &  77.53 &  81.21 &  79.88 &  79.14 &  81.91 \\
     & 32.0  &  81.02 &  77.64 &  80.84 &  80.02 &  78.76 &  81.71 \\
     & 64.0  &  80.19 &  77.93 &  80.49 &  79.90 &  78.48 &  81.33 \\
     & 128.0 &  79.49 &  77.78 &  79.62 &  79.29 &  78.30 &  79.91 \\
128  & 8.0   &  88.80 &  83.96 &  88.93 &  88.21 &  84.01 &  89.11 \\
     & 16.0  &  89.85 &  84.39 &  90.00 &  88.85 &  85.40 &  90.30 \\
     & 32.0  &  89.57 &  84.84 &  89.80 &  89.19 &  85.75 &  89.95 \\
     & 64.0  &  88.98 &  85.08 &  89.07 &  88.37 &  85.97 &  89.13 \\
     & 128.0 &  87.66 &  85.46 &  87.92 &  87.41 &  85.95 &  87.70 \\
256  & 8.0   &  95.44 &  90.01 &  95.50 &  94.91 &  87.79 &  95.72 \\
     & 16.0  &  96.03 &  91.13 &  96.07 &  95.69 &  89.23 &  96.04 \\
     & 32.0  &  95.80 &  91.71 &  96.06 &  95.52 &  90.33 &  95.77 \\
     & 64.0  &  95.26 &  92.09 &  95.35 &  95.08 &  90.71 &  95.04 \\
     & 128.0 &  94.15 &  92.10 &  94.35 &  94.04 &  91.13 &  93.97 \\
512  & 8.0   &  98.64 &  95.35 &  98.62 &  98.44 &  90.87 &  98.57 \\
     & 16.0  &  98.68 &  96.05 &  98.84 &  98.71 &  92.28 &  98.82 \\
     & 32.0  &  98.68 &  96.49 &  98.75 &  98.67 &  93.37 &  98.60 \\
     & 64.0  &  98.37 &  96.77 &  98.55 &  98.37 &  93.89 &  98.24 \\
     & 128.0 &  97.83 &  96.70 &  97.97 &  97.88 &  94.86 &  97.63 \\
1024 & 8.0   &  99.13 &  97.30 &  99.20 &  99.21 &  93.93 &  99.17 \\
     & 16.0  &  99.46 &  98.01 &  99.54 &  99.51 &  94.75 &  99.39 \\
     & 32.0  &  99.45 &  98.44 &  99.57 &  99.51 &  95.43 &  99.42 \\
     & 64.0  &  99.37 &  98.63 &  99.48 &  99.46 &  96.28 &  99.30 \\
     & 128.0 &  99.22 &  98.67 &  99.30 &  99.28 &  97.33 &  99.15 \\
\bottomrule
\end{tabular}
    }
\end{minipage}
\caption{(Left) The error over the number of trees in the ensemble on the avila dataset. Dashed lines depict the Random Forest and solid lines are the corresponding pruned ensemble via Reduced Error pruning. (Right) The 5-fold cross-validation accuracy  on the avila dataset. Rounded to the second decimal digit. Larger is better.}
\end{figure}

\begin{figure}[H]
\begin{minipage}{.49\textwidth}
    \centering
    \includegraphics[width=\textwidth,keepaspectratio]{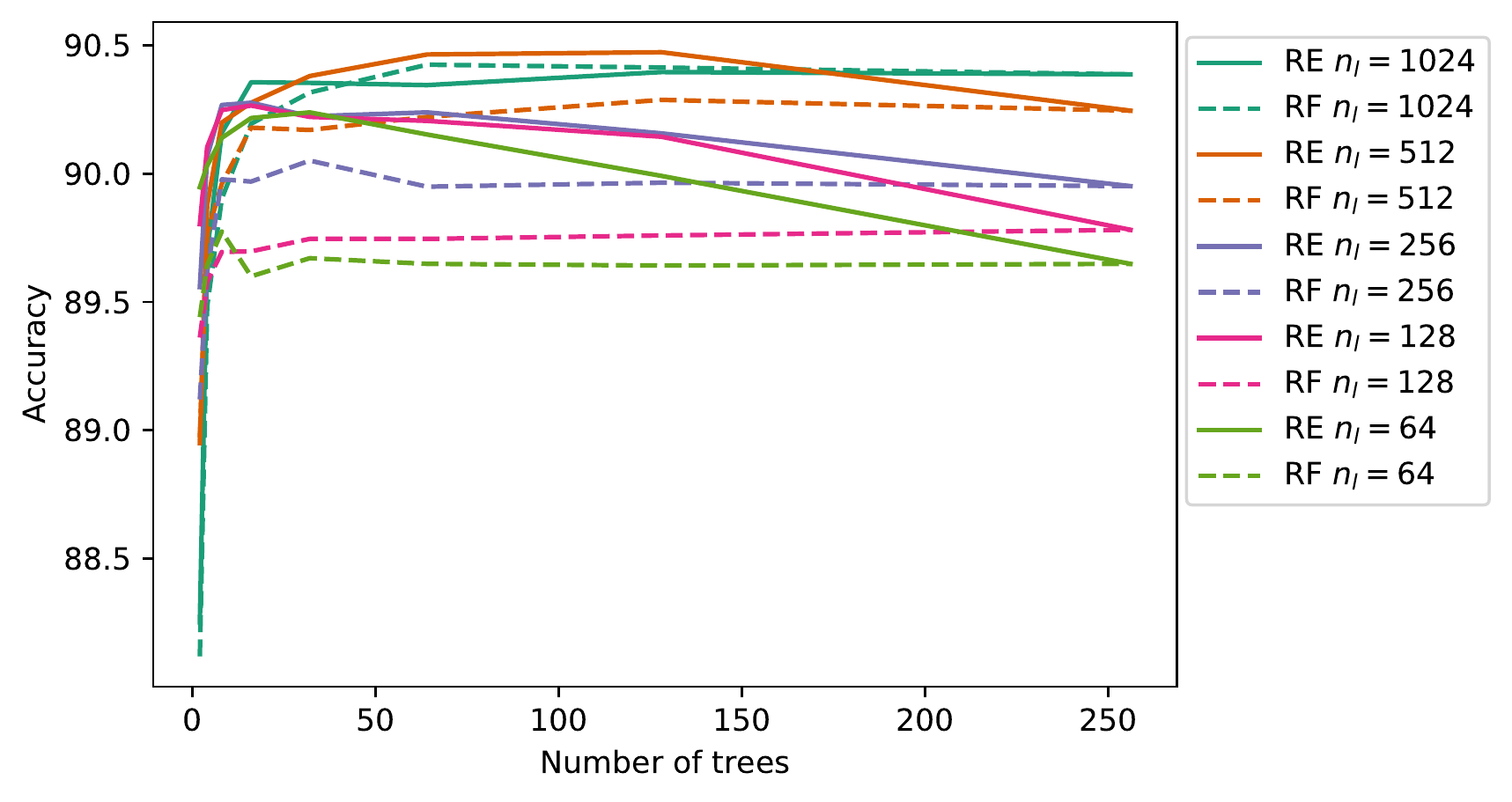}
\end{minipage}\hfill
\begin{minipage}{.49\textwidth}
    \centering 
    \resizebox{\textwidth}{!}{
        \begin{tabular}{llrrrrrr}
\toprule
     & model &   COMP &   DREP &     IC &     IE &    LMD &     RE \\
max\_leaf\_nodes & n\_estimators &        &        &        &        &        &        \\
\midrule
64   & 8.0   &  89.89 &  89.67 &  90.05 &  90.17 &  89.69 &  90.14 \\
     & 16.0  &  89.99 &  89.57 &  90.17 &  90.07 &  89.73 &  90.22 \\
     & 32.0  &  90.08 &  89.66 &  90.15 &  90.04 &  89.71 &  90.24 \\
     & 64.0  &  90.01 &  89.66 &  90.11 &  89.86 &  89.68 &  90.15 \\
     & 128.0 &  89.89 &  89.65 &  89.87 &  89.69 &  89.69 &  89.99 \\
128  & 8.0   &  90.11 &  89.94 &  90.23 &  90.21 &  89.69 &  90.25 \\
     & 16.0  &  90.13 &  89.75 &  90.25 &  90.08 &  89.83 &  90.27 \\
     & 32.0  &  90.06 &  89.73 &  90.21 &  90.13 &  89.90 &  90.22 \\
     & 64.0  &  90.07 &  89.71 &  90.08 &  89.97 &  89.82 &  90.21 \\
     & 128.0 &  89.95 &  89.81 &  89.96 &  89.82 &  89.79 &  90.14 \\
256  & 8.0   &  90.18 &  89.98 &  90.26 &  90.19 &  89.95 &  90.27 \\
     & 16.0  &  90.26 &  90.00 &  90.28 &  90.31 &  90.05 &  90.28 \\
     & 32.0  &  90.22 &  90.02 &  90.29 &  90.26 &  90.09 &  90.22 \\
     & 64.0  &  90.25 &  89.99 &  90.33 &  90.14 &  89.98 &  90.24 \\
     & 128.0 &  90.11 &  89.95 &  90.17 &  90.02 &  90.02 &  90.16 \\
512  & 8.0   &  90.07 &  90.10 &  90.10 &  90.25 &  90.01 &  90.20 \\
     & 16.0  &  90.19 &  90.24 &  90.28 &  90.40 &  90.13 &  90.28 \\
     & 32.0  &  90.43 &  90.21 &  90.45 &  90.44 &  90.24 &  90.38 \\
     & 64.0  &  90.51 &  90.21 &  90.54 &  90.44 &  90.30 &  90.46 \\
     & 128.0 &  90.46 &  90.27 &  90.50 &  90.39 &  90.40 &  90.47 \\
1024 & 8.0   &  90.07 &  90.04 &  90.06 &  90.15 &  89.87 &  90.16 \\
     & 16.0  &  90.27 &  90.22 &  90.27 &  90.33 &  90.34 &  90.36 \\
     & 32.0  &  90.42 &  90.27 &  90.47 &  90.39 &  90.40 &  90.35 \\
     & 64.0  &  90.53 &  90.41 &  90.54 &  90.42 &  90.45 &  90.35 \\
     & 128.0 &  90.42 &  90.37 &  90.42 &  90.47 &  90.38 &  90.40 \\
\bottomrule
\end{tabular}
    }
\end{minipage}
\caption{(Left) The error over the number of trees in the ensemble on the bank dataset. Dashed lines depict the Random Forest and solid lines are the corresponding pruned ensemble via Reduced Error pruning. (Right) The 5-fold cross-validation accuracy  on the bank dataset. Rounded to the second decimal digit. Larger is better.}
\end{figure}

\begin{figure}[H]
\begin{minipage}{.49\textwidth}
    \centering
    \includegraphics[width=\textwidth,keepaspectratio]{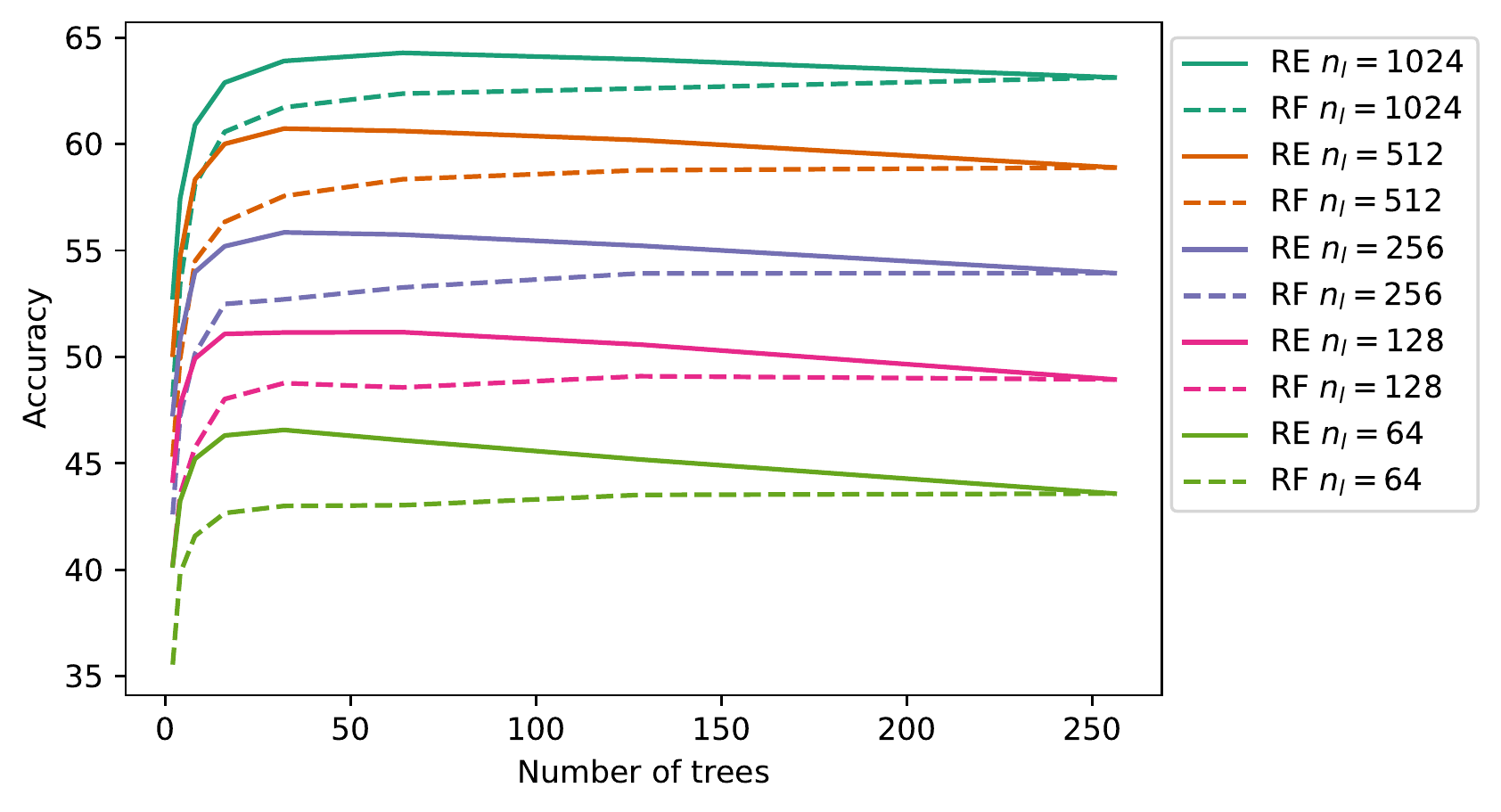}
\end{minipage}\hfill
\begin{minipage}{.49\textwidth}
    \centering 
    \resizebox{\textwidth}{!}{
        \begin{tabular}{llrrrrrr}
\toprule
     & model &   COMP &   DREP &     IC &     IE &    LMD &     RE \\
max\_leaf\_nodes & n\_estimators &        &        &        &        &        &        \\
\midrule
64   & 8.0   &  44.04 &  41.52 &  44.13 &  42.74 &  43.03 &  45.21 \\
     & 16.0  &  44.98 &  42.83 &  44.76 &  43.41 &  44.07 &  46.31 \\
     & 32.0  &  45.11 &  43.35 &  45.04 &  43.59 &  44.55 &  46.57 \\
     & 64.0  &  45.03 &  43.47 &  45.03 &  43.41 &  44.56 &  46.08 \\
     & 128.0 &  44.44 &    NaN &  44.43 &  43.38 &  44.11 &  45.18 \\
128  & 8.0   &  48.71 &  46.29 &  49.12 &  47.63 &  47.48 &  49.93 \\
     & 16.0  &  50.35 &  47.80 &  50.31 &  48.66 &  48.82 &  51.08 \\
     & 32.0  &  50.54 &  48.15 &  50.32 &  49.14 &  49.75 &  51.15 \\
     & 64.0  &  50.48 &  48.69 &  50.65 &  49.01 &  49.76 &  51.16 \\
     & 128.0 &  50.22 &    NaN &  50.10 &  49.08 &  49.65 &  50.58 \\
256  & 8.0   &  53.39 &  51.18 &  53.68 &  52.42 &  52.15 &  53.98 \\
     & 16.0  &  54.88 &  52.18 &  55.04 &  53.60 &  54.12 &  55.20 \\
     & 32.0  &  55.33 &  53.09 &  55.76 &  54.20 &  54.55 &  55.85 \\
     & 64.0  &  55.18 &  53.36 &  55.19 &  54.22 &  54.48 &  55.75 \\
     & 128.0 &  54.65 &    NaN &  54.73 &  54.07 &  54.10 &  55.23 \\
512  & 8.0   &  58.07 &  54.25 &  57.23 &  56.90 &  56.01 &  58.33 \\
     & 16.0  &  59.36 &  56.66 &  59.50 &  58.22 &  58.48 &  60.00 \\
     & 32.0  &  60.23 &  57.85 &  60.58 &  59.17 &  59.18 &  60.72 \\
     & 64.0  &  60.41 &  58.25 &  60.62 &  59.36 &  59.44 &  60.61 \\
     & 128.0 &  60.07 &    NaN &  59.99 &  59.01 &  59.35 &  60.18 \\
1024 & 8.0   &  60.55 &  58.56 &  61.14 &  59.81 &  59.11 &  60.90 \\
     & 16.0  &  62.86 &  60.20 &  63.14 &  62.12 &  62.01 &  62.88 \\
     & 32.0  &  64.01 &  61.51 &  64.23 &  63.31 &  63.01 &  63.90 \\
     & 64.0  &  64.50 &  62.40 &  64.26 &  63.60 &  63.63 &  64.28 \\
     & 128.0 &  64.02 &    NaN &  64.20 &  63.26 &  63.71 &  63.98 \\
\bottomrule
\end{tabular}
    }
\end{minipage}
\caption{(Left) The error over the number of trees in the ensemble on the chess dataset. Dashed lines depict the Random Forest and solid lines are the corresponding pruned ensemble via Reduced Error pruning. (Right) The 5-fold cross-validation accuracy  on the chess dataset. Rounded to the second decimal digit. Larger is better.}
\end{figure}

\begin{figure}[H]
\begin{minipage}{.49\textwidth}
    \centering
    \includegraphics[width=\textwidth,keepaspectratio]{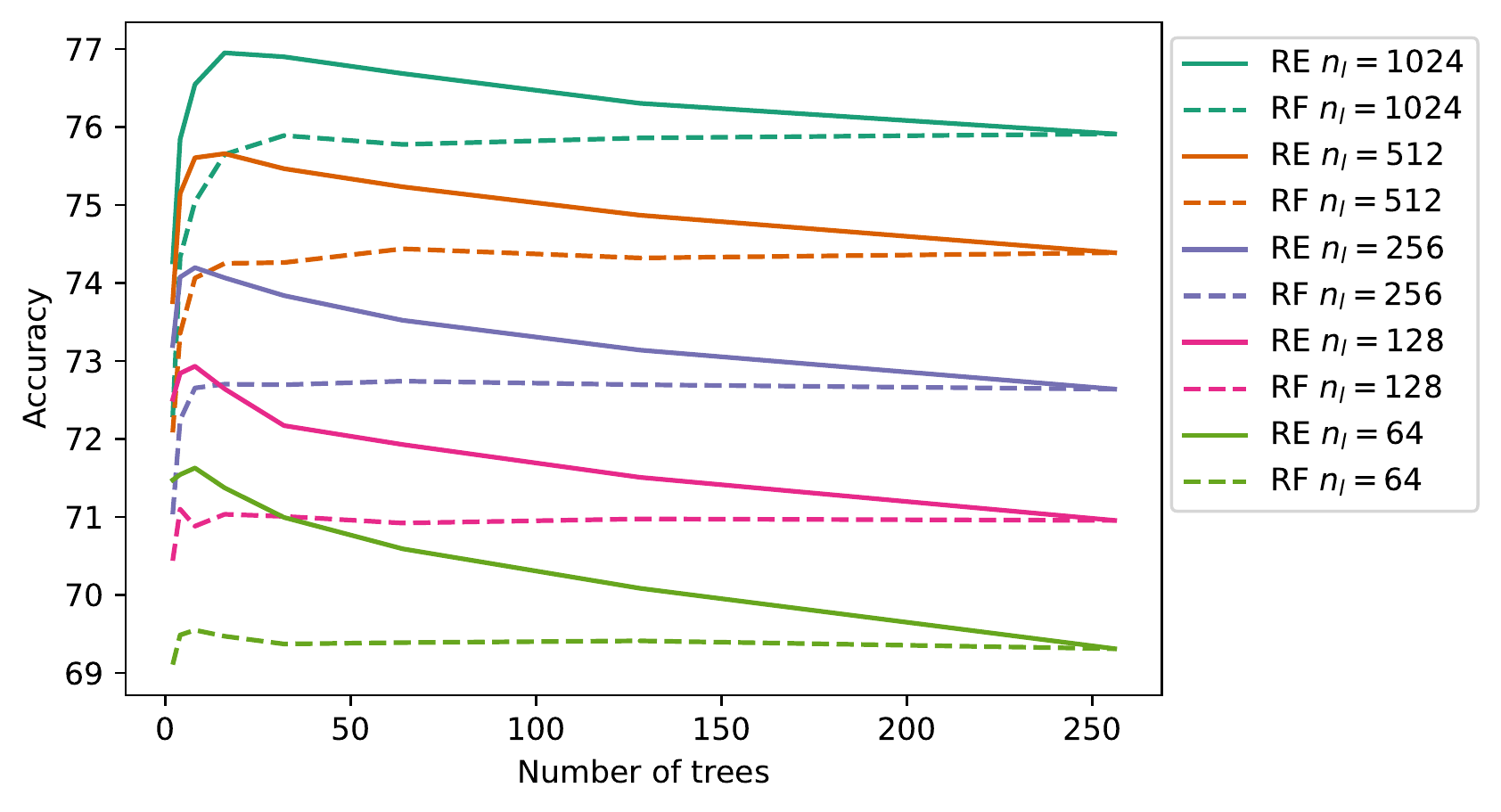}
\end{minipage}\hfill
\begin{minipage}{.49\textwidth}
    \centering 
    \resizebox{\textwidth}{!}{
        \input{figures/RandomForestClassifier_with_prune_connect_table}
    }
\end{minipage}
\caption{(Left) The error over the number of trees in the ensemble on the connect dataset. Dashed lines depict the Random Forest and solid lines are the corresponding pruned ensemble via Reduced Error pruning. (Right) The 5-fold cross-validation accuracy  on the connect dataset. Rounded to the second decimal digit. Larger is better.}
\end{figure}

\begin{figure}[H]
\begin{minipage}{.49\textwidth}
    \centering
    \includegraphics[width=\textwidth,keepaspectratio]{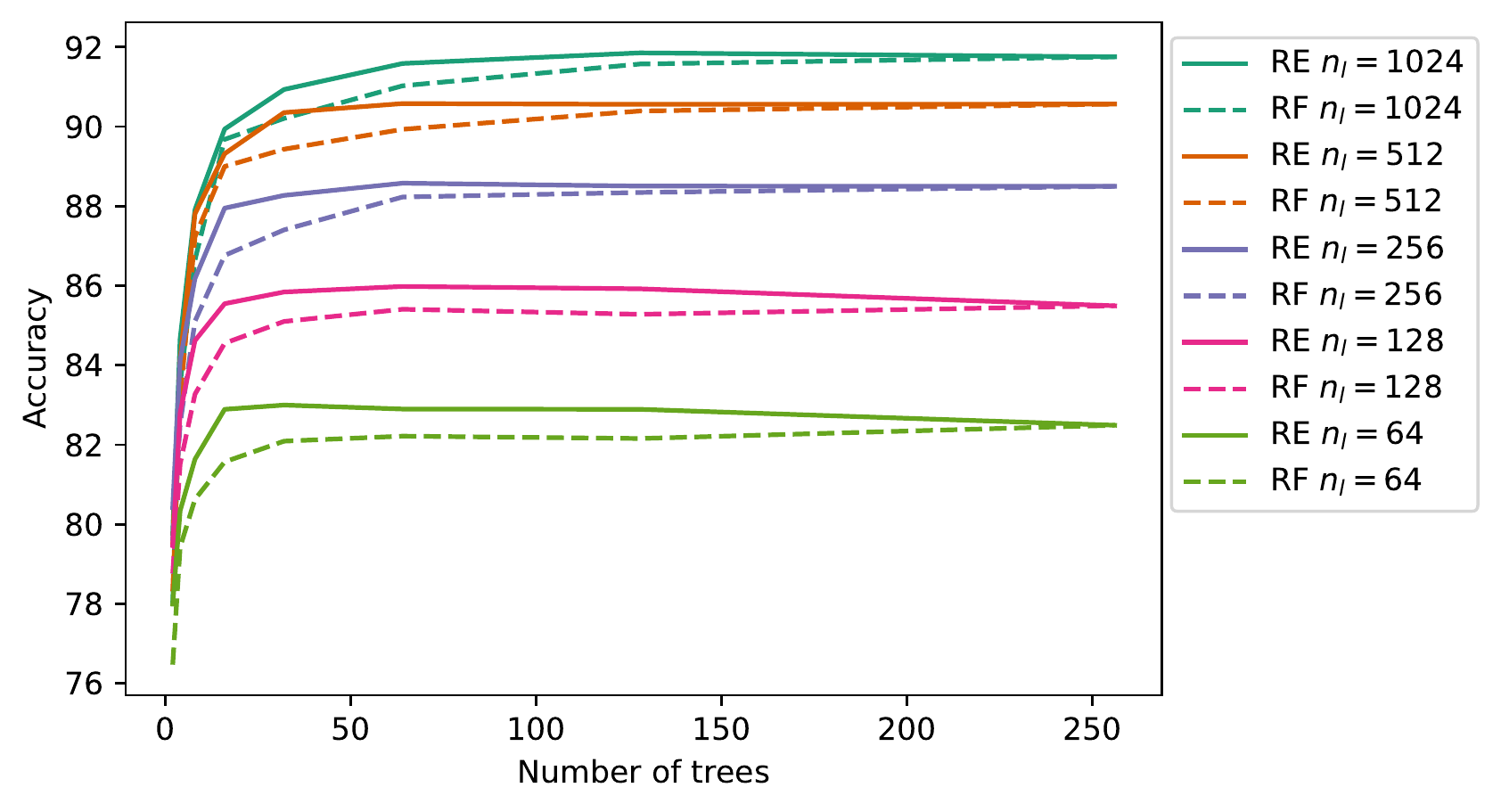}
\end{minipage}\hfill
\begin{minipage}{.49\textwidth}
    \centering 
    \resizebox{\textwidth}{!}{
        \input{figures/RandomForestClassifier_with_prune_eeg_table}
    }
\end{minipage}
\caption{(Left) The error over the number of trees in the ensemble on the eeg dataset. Dashed lines depict the Random Forest and solid lines are the corresponding pruned ensemble via Reduced Error pruning. (Right) The 5-fold cross-validation accuracy  on the eeg dataset. Rounded to the second decimal digit. Larger is better.}
\end{figure}

\begin{figure}[H]
\begin{minipage}{.49\textwidth}
    \centering
    \includegraphics[width=\textwidth,keepaspectratio]{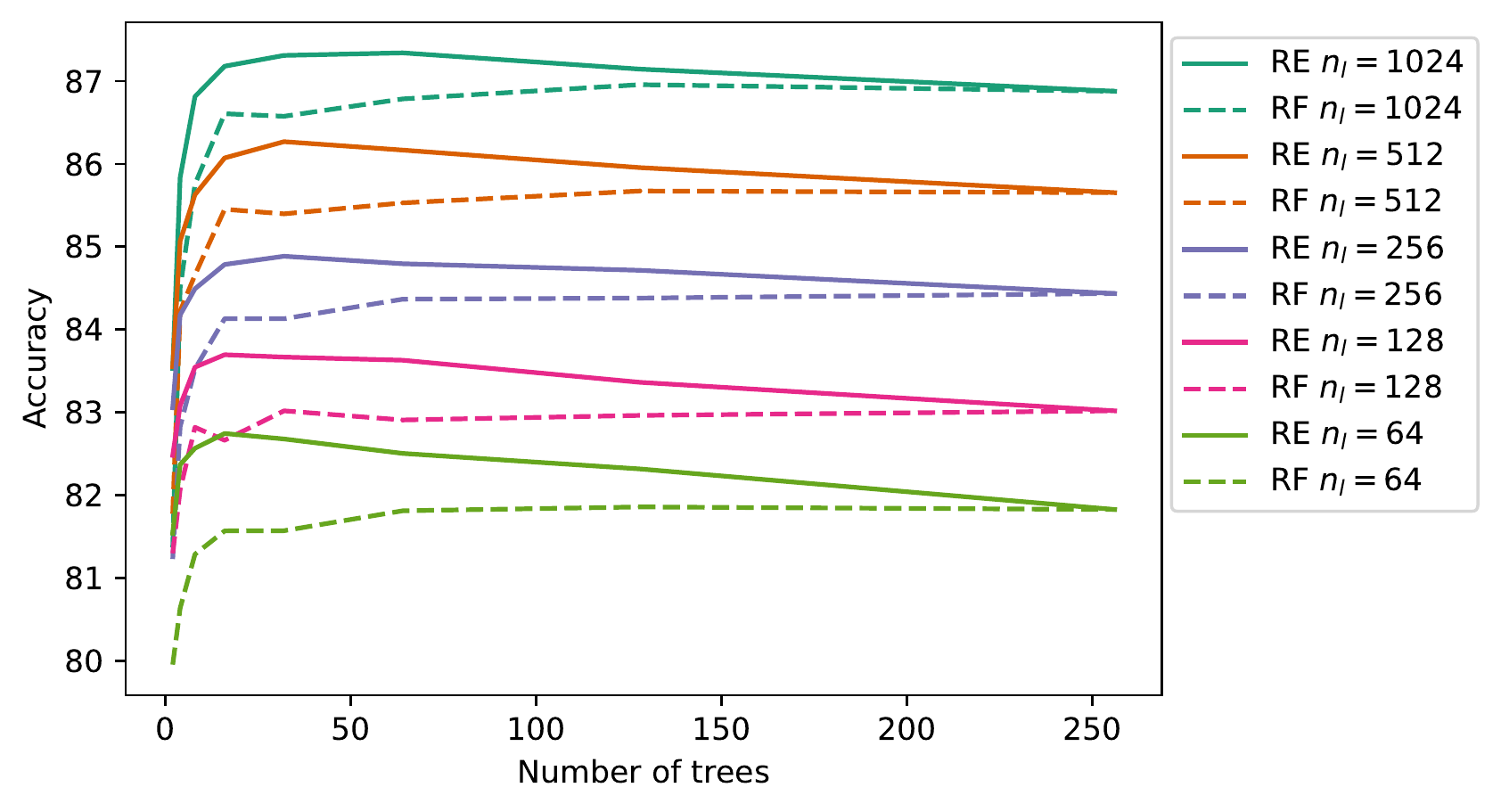}
\end{minipage}\hfill
\begin{minipage}{.49\textwidth}
    \centering 
    \resizebox{\textwidth}{!}{
        \input{figures/RandomForestClassifier_with_prune_elec_table}
    }
\end{minipage}
\caption{(Left) The error over the number of trees in the ensemble on the elec dataset. Dashed lines depict the Random Forest and solid lines are the corresponding pruned ensemble via Reduced Error pruning. (Right) The 5-fold cross-validation accuracy  on the elec dataset. Rounded to the second decimal digit. Larger is better.}
\end{figure}

\begin{figure}[H]
\begin{minipage}{.49\textwidth}
    \centering
    \includegraphics[width=\textwidth,keepaspectratio]{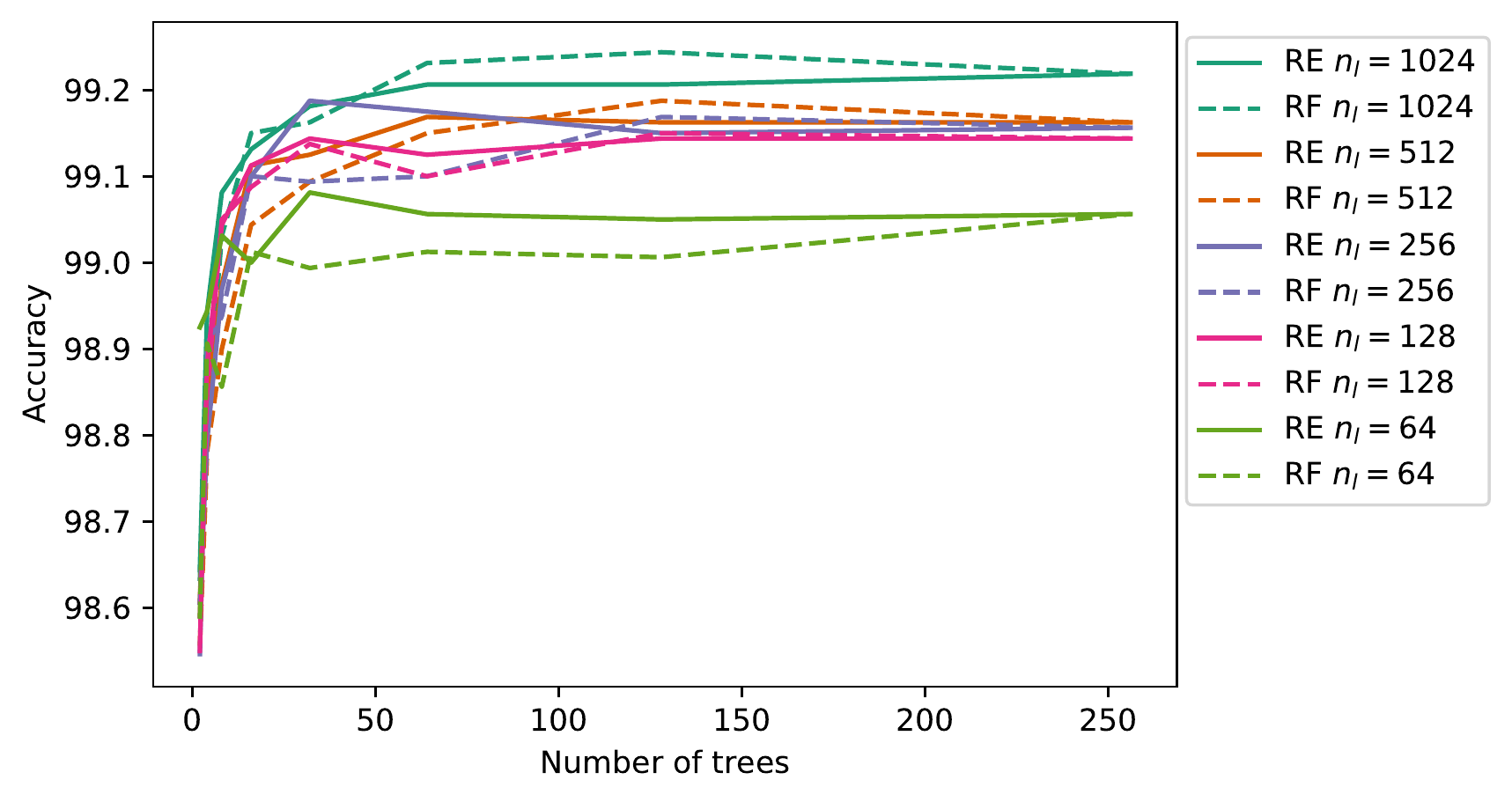}
\end{minipage}\hfill
\begin{minipage}{.49\textwidth}
    \centering 
    \resizebox{\textwidth}{!}{
        \input{figures/RandomForestClassifier_with_prune_ida2016_table}
    }
\end{minipage}
\caption{(Left) The error over the number of trees in the ensemble on the ida2016 dataset. Dashed lines depict the Random Forest and solid lines are the corresponding pruned ensemble via Reduced Error pruning. (Right) The 5-fold cross-validation accuracy  on the ida2016 dataset. Rounded to the second decimal digit. Larger is better. }
\end{figure}

\begin{figure}[H]
\begin{minipage}{.49\textwidth}
    \centering
    \includegraphics[width=\textwidth,keepaspectratio]{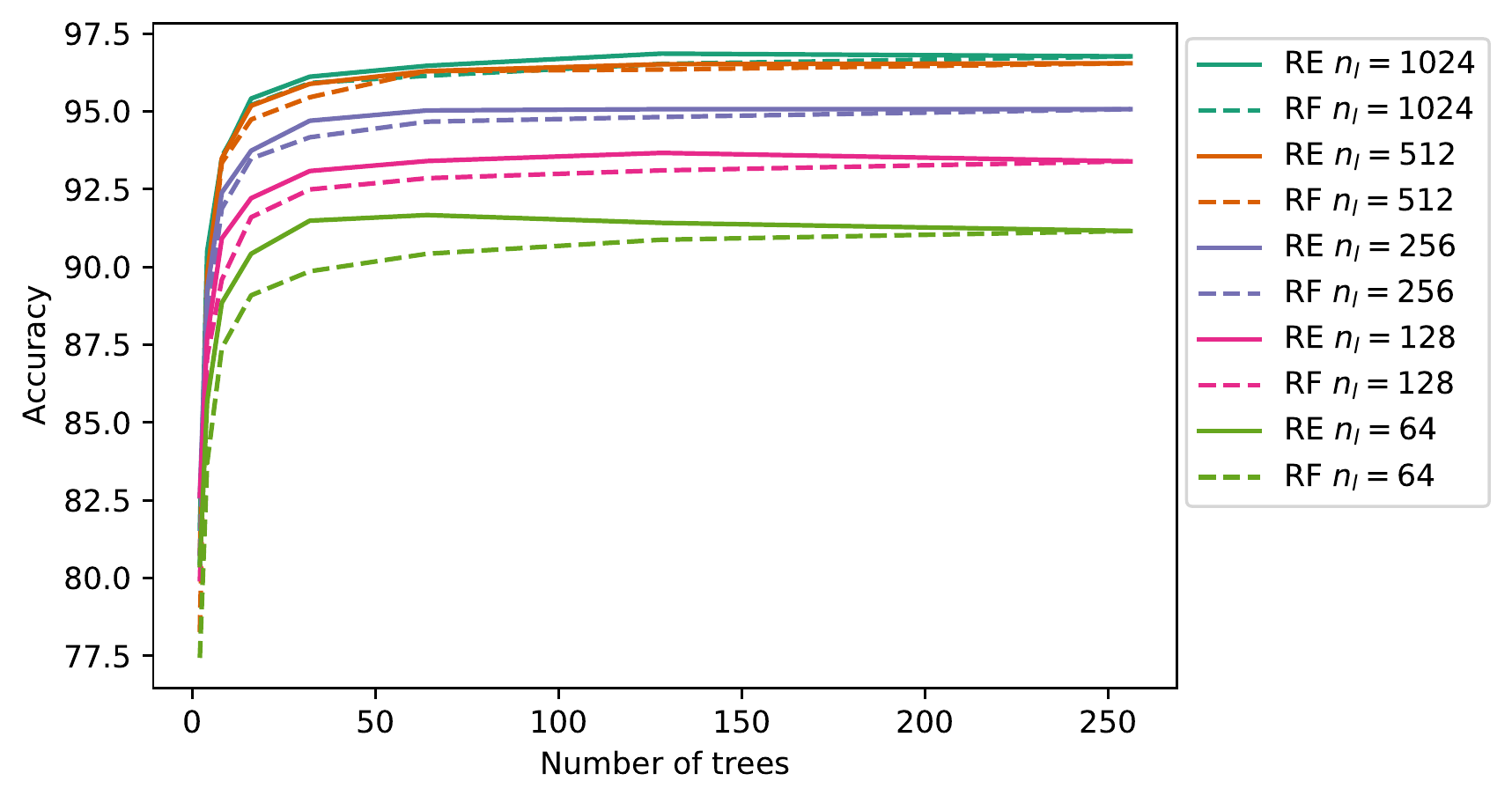}
\end{minipage}\hfill
\begin{minipage}{.49\textwidth}
    \centering 
    \resizebox{\textwidth}{!}{
        \input{figures/RandomForestClassifier_with_prune_japanese-vowels_table}
    }
\end{minipage}
\caption{(Left) The error over the number of trees in the ensemble on the japanese-vowels dataset. Dashed lines depict the Random Forest and solid lines are the corresponding pruned ensemble via Reduced Error pruning. (Right) The 5-fold cross-validation accuracy  on the japanese-vowels dataset. Rounded to the second decimal digit. Larger is better.}
\end{figure}

\begin{figure}[H]
\begin{minipage}{.49\textwidth}
    \centering
    \includegraphics[width=\textwidth,keepaspectratio]{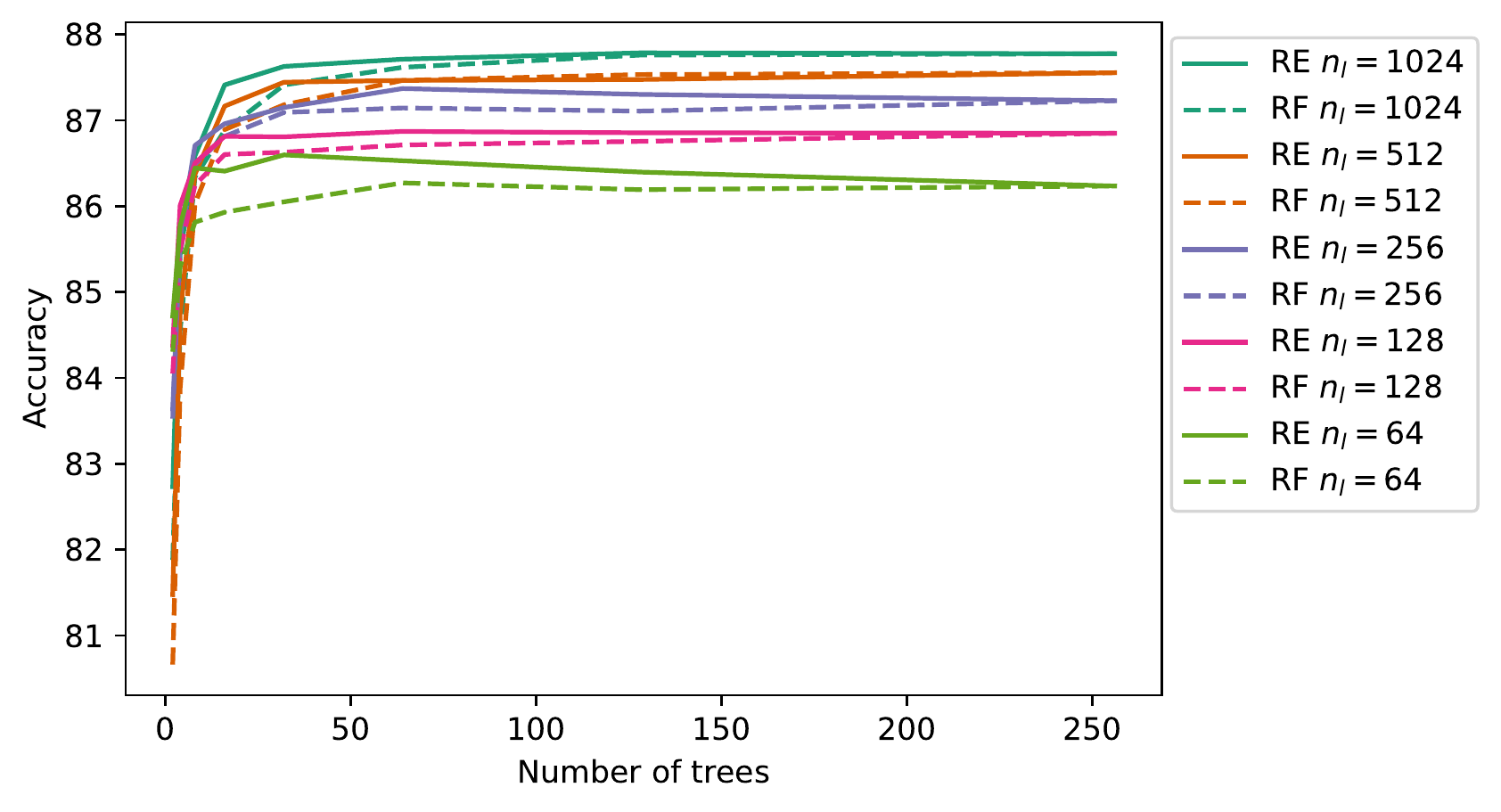}
\end{minipage}\hfill
\begin{minipage}{.49\textwidth}
    \centering 
    \resizebox{\textwidth}{!}{
        \input{figures/RandomForestClassifier_with_prune_magic_table}
    }
\end{minipage}
\caption{(Left) The error over the number of trees in the ensemble on the magic dataset. Dashed lines depict the Random Forest and solid lines are the corresponding pruned ensemble via Reduced Error pruning. (Right) The 5-fold cross-validation accuracy  on the magic dataset. Rounded to the second decimal digit. Larger is better}
\end{figure}

\begin{figure}[H]
\begin{minipage}{.49\textwidth}
    \centering
    \includegraphics[width=\textwidth,keepaspectratio]{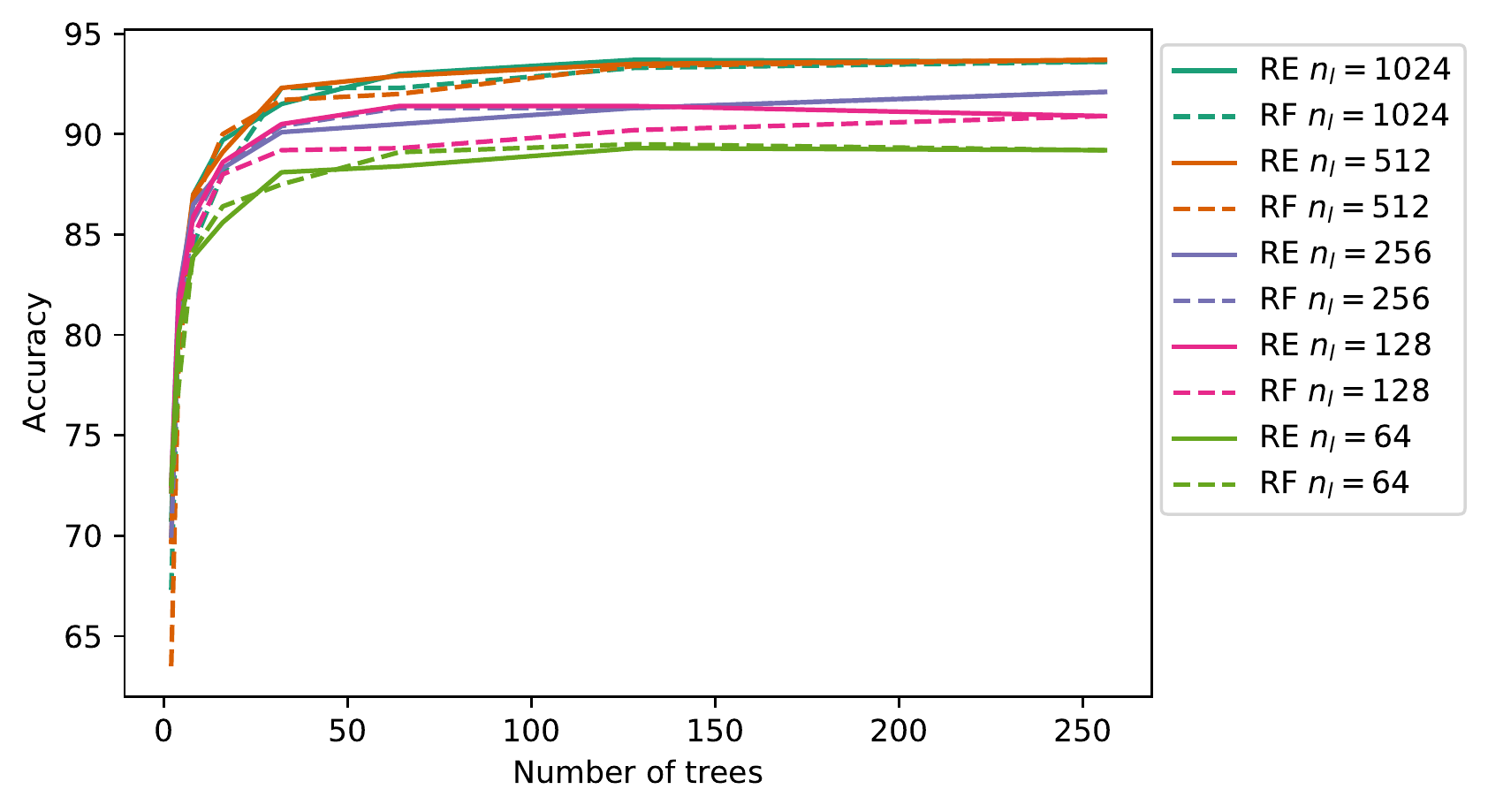}
\end{minipage}\hfill
\begin{minipage}{.49\textwidth}
    \centering 
    \resizebox{\textwidth}{!}{
        \input{figures/RandomForestClassifier_with_prune_mnist_table}
    }
\end{minipage}
\caption{(Left) The error over the number of trees in the ensemble on the mnist dataset. Dashed lines depict the Random Forest and solid lines are the corresponding pruned ensemble via Reduced Error pruning. (Right) The 5-fold cross-validation accuracy  on the mnist dataset. Rounded to the second decimal digit. Larger is better.}
\end{figure}

\begin{figure}[H]
\begin{minipage}{.49\textwidth}
    \centering
    \includegraphics[width=\textwidth,keepaspectratio]{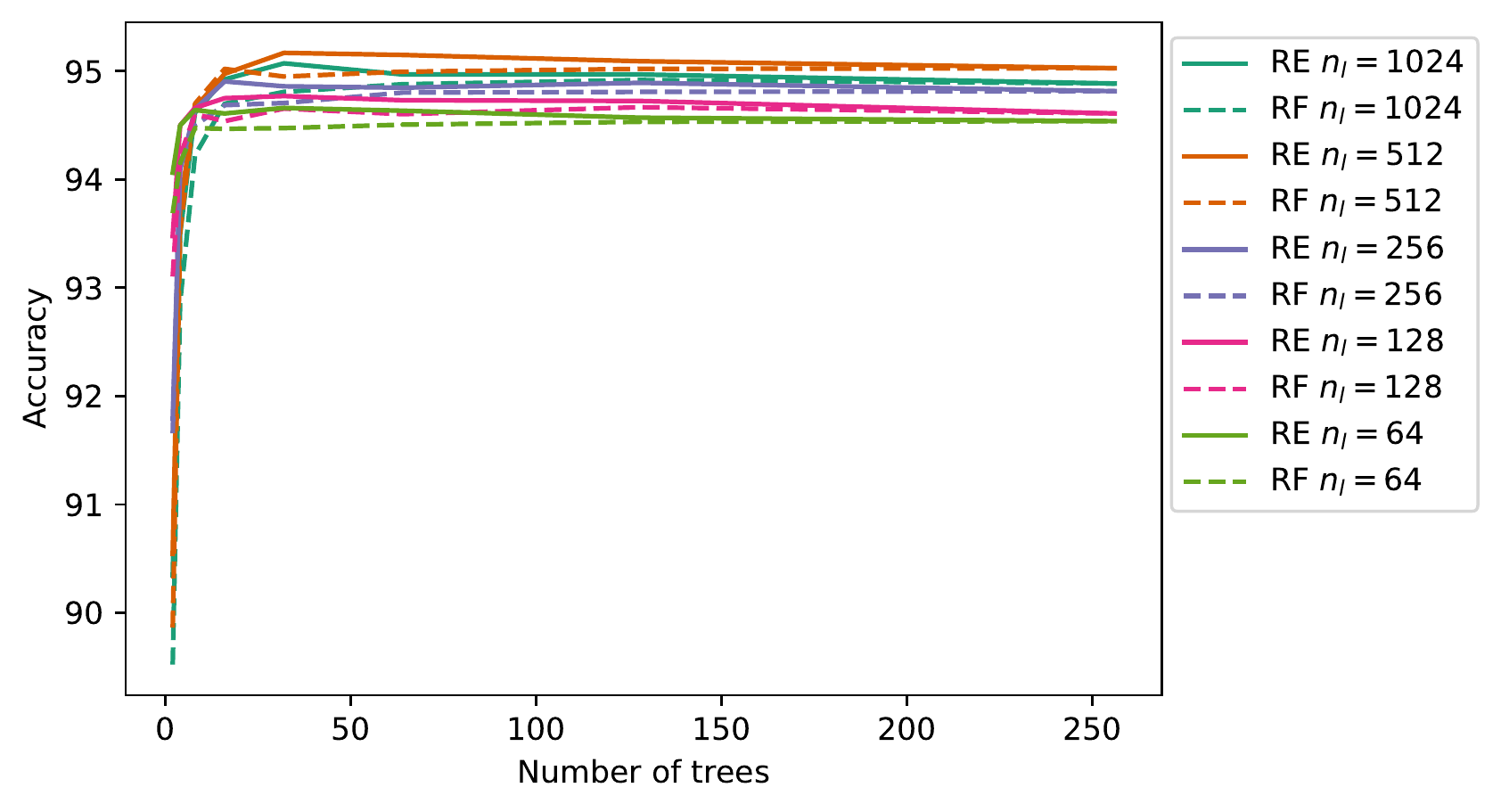}
\end{minipage}\hfill
\begin{minipage}{.49\textwidth}
    \centering 
    \resizebox{\textwidth}{!}{
        \input{figures/RandomForestClassifier_with_prune_mozilla_table}
    }
\end{minipage}
\caption{(Left) The error over the number of trees in the ensemble on the mozilla dataset. Dashed lines depict the Random Forest and solid lines are the corresponding pruned ensemble via Reduced Error pruning. (Right) The 5-fold cross-validation accuracy  on the mozilla dataset. Rounded to the second decimal digit. Larger is better.}
\end{figure}

\begin{figure}[H]
\begin{minipage}{.49\textwidth}
    \centering
    \includegraphics[width=\textwidth,keepaspectratio]{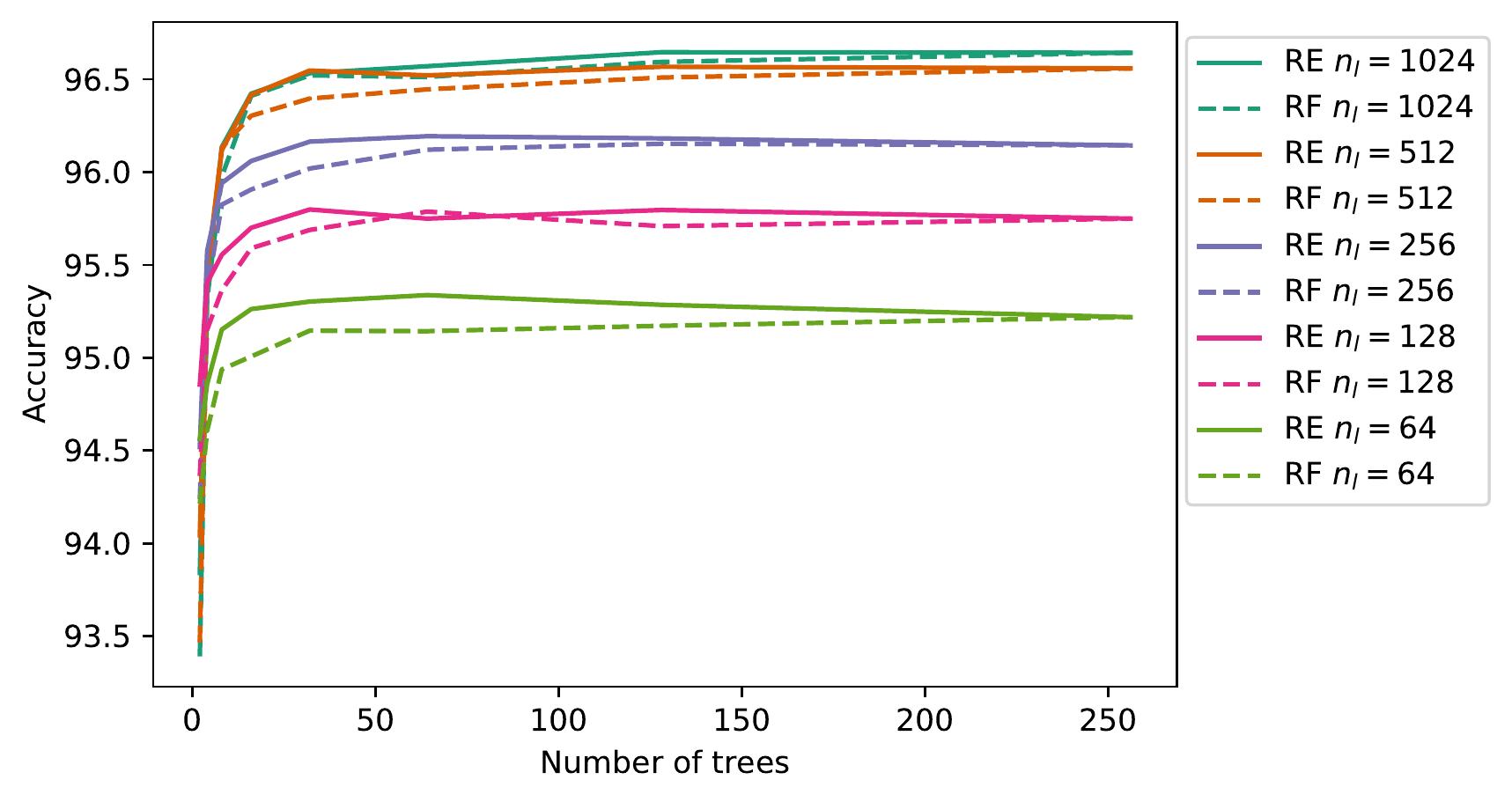}
\end{minipage}\hfill
\begin{minipage}{.49\textwidth}
    \centering 
    \resizebox{\textwidth}{!}{
        \input{figures/RandomForestClassifier_with_prune_nomao_table}
    }
\end{minipage}
\caption{(Left) The error over the number of trees in the ensemble on the nomao dataset. Dashed lines depict the Random Forest and solid lines are the corresponding pruned ensemble via Reduced Error pruning. (Right) The 5-fold cross-validation accuracy  on the nomao dataset. Rounded to the second decimal digit. Larger is better.}
\end{figure}

\begin{figure}[H]
\begin{minipage}{.49\textwidth}
    \centering
    \includegraphics[width=\textwidth,keepaspectratio]{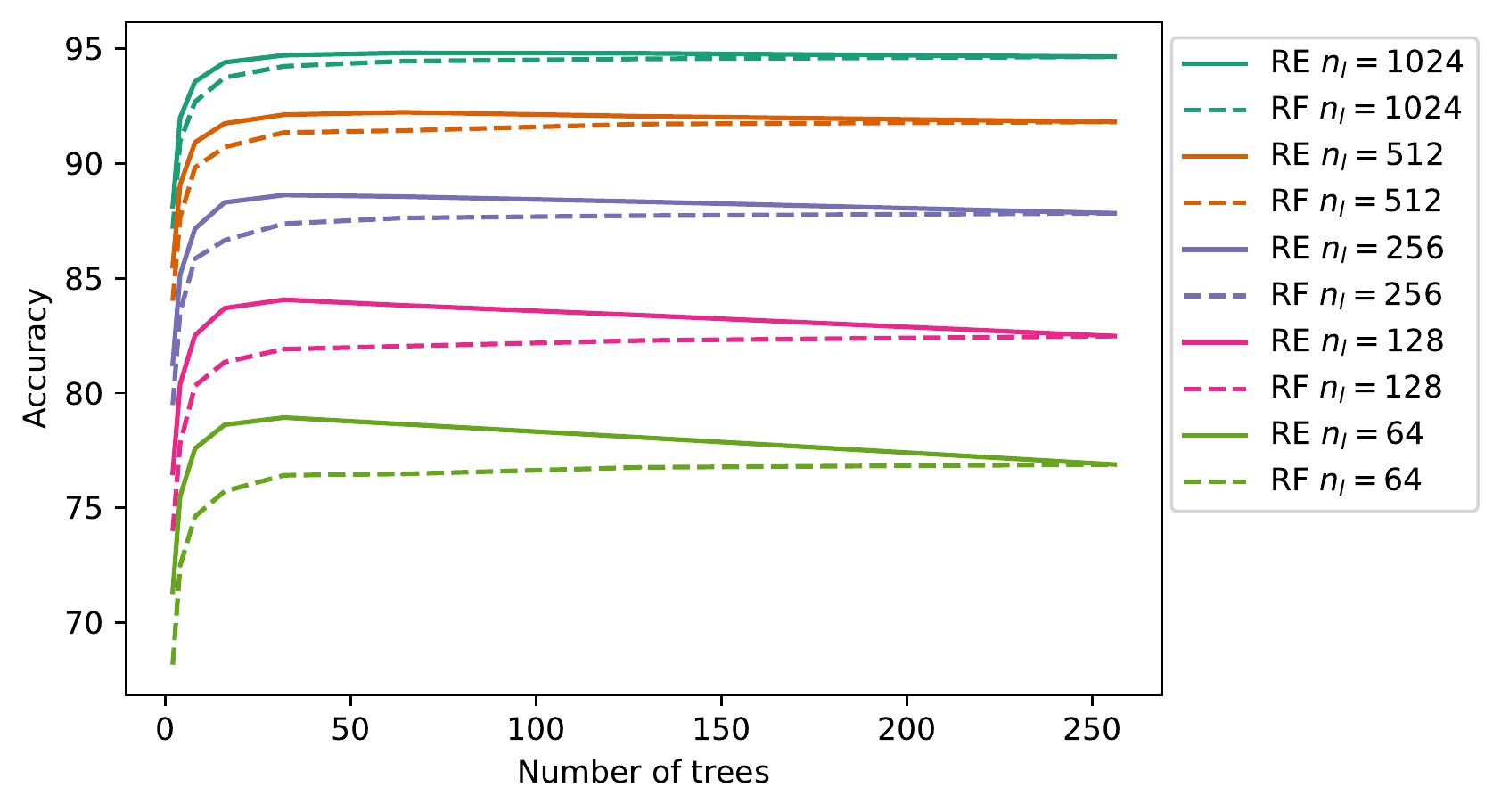}
\end{minipage}\hfill
\begin{minipage}{.49\textwidth}
    \centering 
    \resizebox{\textwidth}{!}{
        \input{figures/RandomForestClassifier_with_prune_postures_table}
    }
\end{minipage}
\caption{(Left) The error over the number of trees in the ensemble on the postures dataset. Dashed lines depict the Random Forest and solid lines are the corresponding pruned ensemble via Reduced Error pruning. (Right) The 5-fold cross-validation accuracy  on the postures dataset. Rounded to the second decimal digit. Larger is better.}
\end{figure}

\begin{figure}[H]
\begin{minipage}{.49\textwidth}
    \centering
    \includegraphics[width=\textwidth,keepaspectratio]{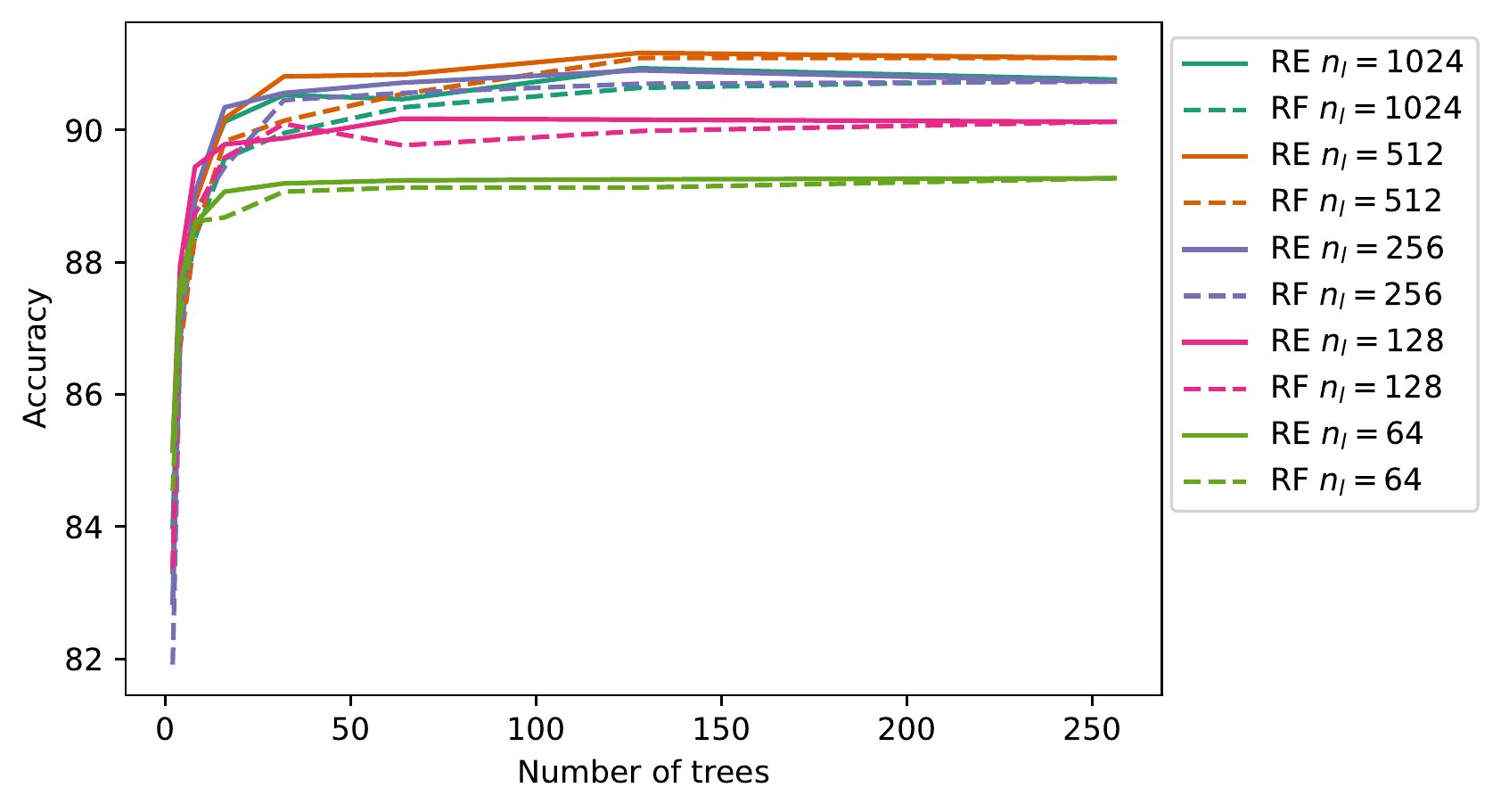}
\end{minipage}\hfill
\begin{minipage}{.49\textwidth}
    \centering 
    \resizebox{\textwidth}{!}{
        \input{figures/RandomForestClassifier_with_prune_satimage_table}
    }
\end{minipage}
\caption{(Left) The error over the number of trees in the ensemble on the satimage dataset. Dashed lines depict the Random Forest and solid lines are the corresponding pruned ensemble via Reduced Error pruning. (Right) The 5-fold cross-validation accuracy  on the satimage dataset. Rounded to the second decimal digit. Larger is better.}
\end{figure}

\subsection{Plotting the Pareto Front For More Datasets with Dedicated Pruning Set}

\begin{figure}[H]
\begin{minipage}{.49\textwidth}
    \centering
    \includegraphics[width=\textwidth,keepaspectratio]{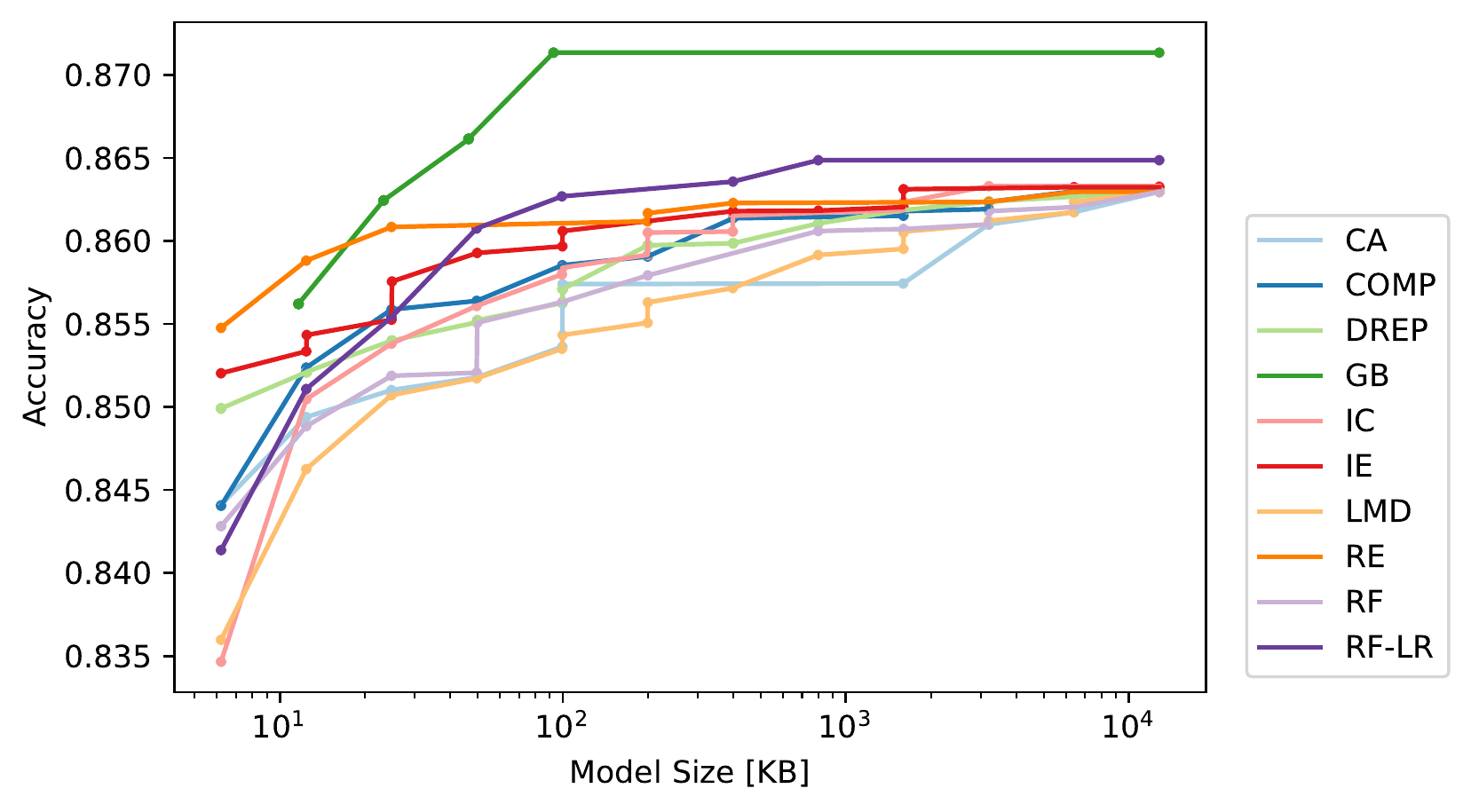}
\end{minipage}\hfill
\begin{minipage}{.49\textwidth}
    \centering 
    \includegraphics[width=\textwidth,keepaspectratio]{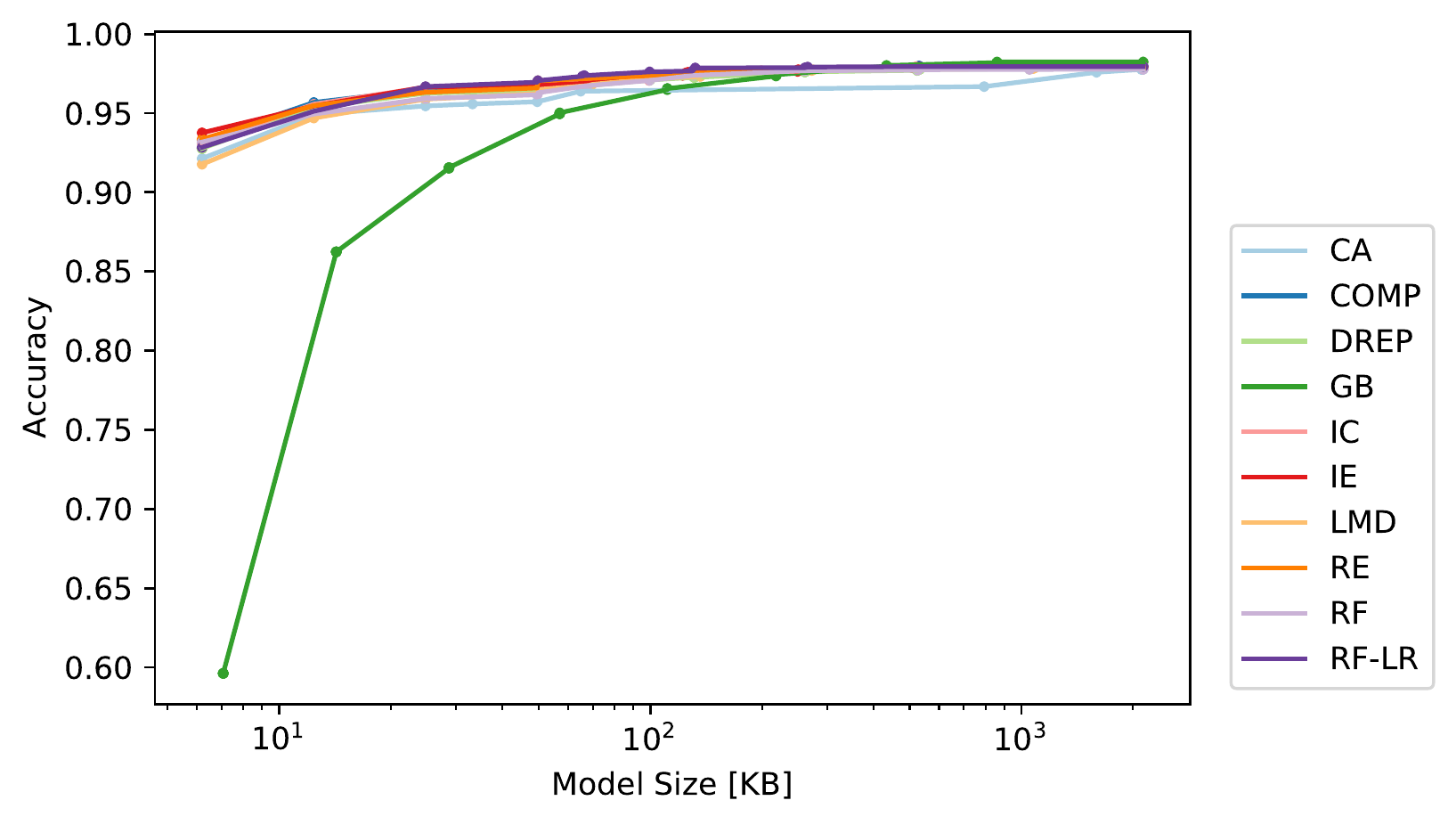}
\end{minipage}
\caption{(left) 5-fold cross-validation accuracy on the adult dataset. (right) 5-fold cross-validation accuracy on the anura dataset.}
\end{figure}

\begin{figure}[H]
\begin{minipage}{.49\textwidth}
    \centering
    \includegraphics[width=\textwidth,keepaspectratio]{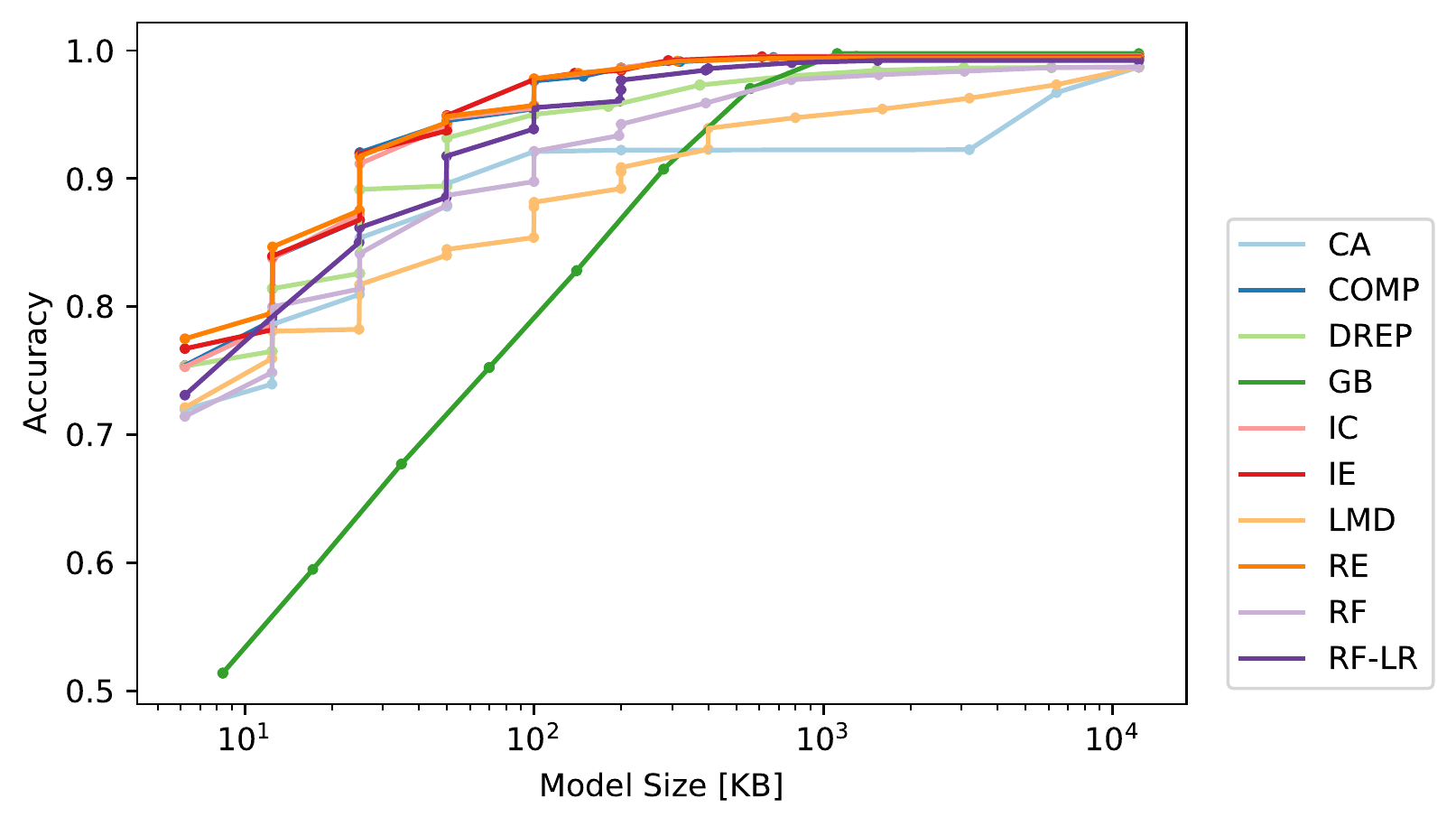}
\end{minipage}\hfill
\begin{minipage}{.49\textwidth}
    \centering 
    \includegraphics[width=\textwidth,keepaspectratio]{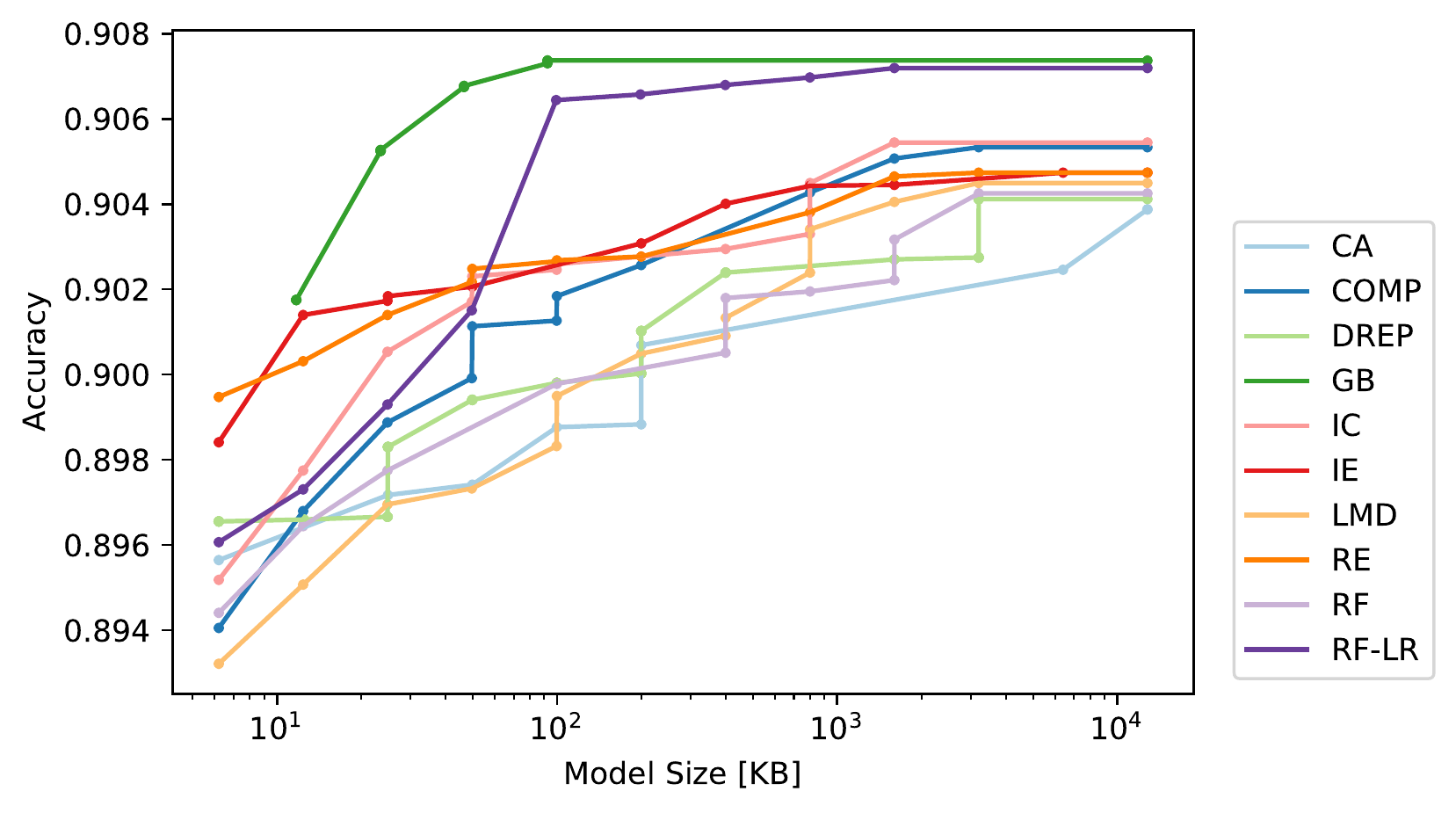}
\end{minipage}
\caption{(left) 5-fold cross-validation accuracy on the avila dataset. (right) 5-fold cross-validation accuracy on the bank dataset.}
\end{figure}

\begin{figure}[H]
\begin{minipage}{.49\textwidth}
    \centering
    \includegraphics[width=\textwidth,keepaspectratio]{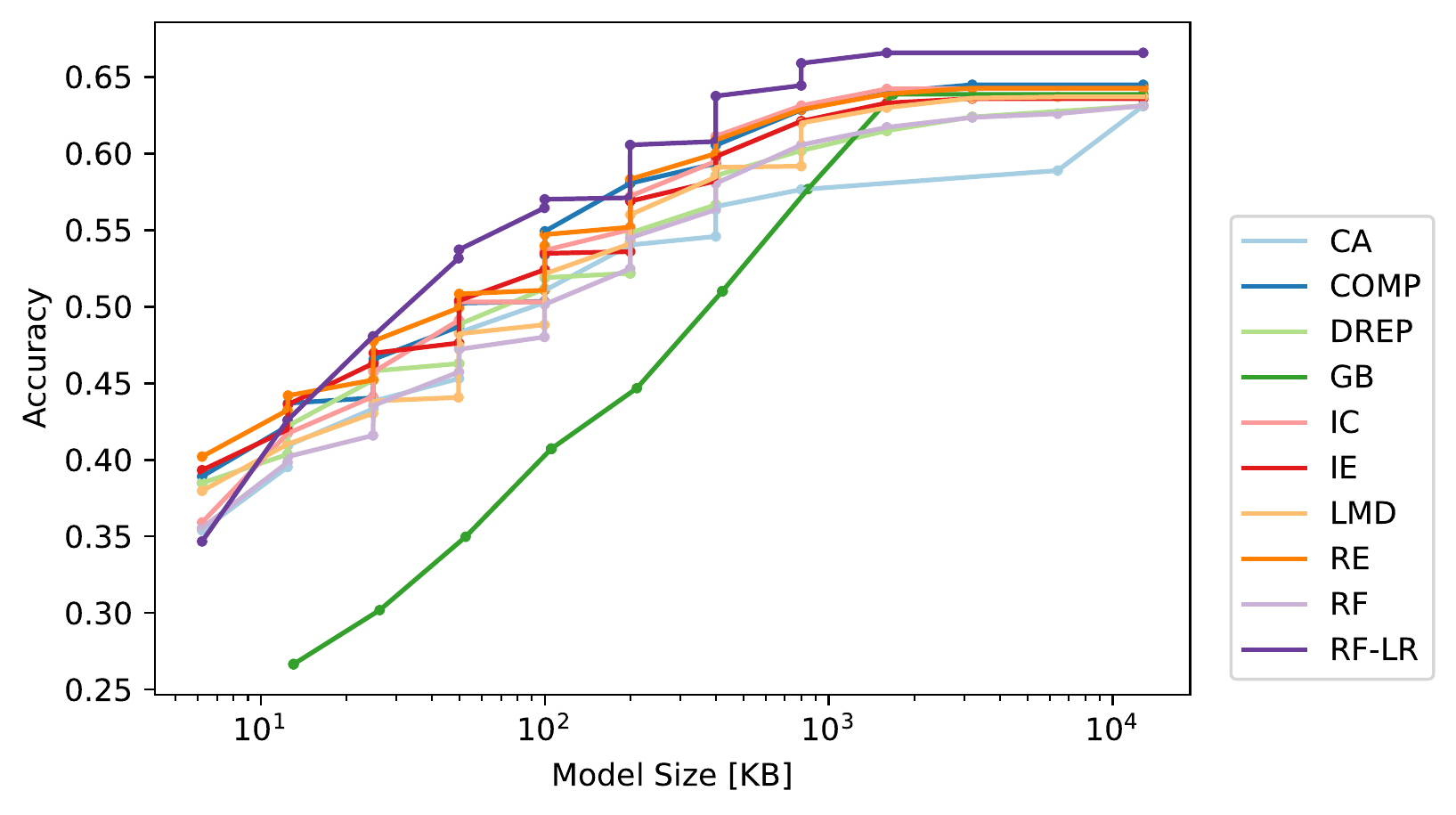}
\end{minipage}\hfill
\begin{minipage}{.49\textwidth}
    \centering 
    \includegraphics[width=\textwidth,keepaspectratio]{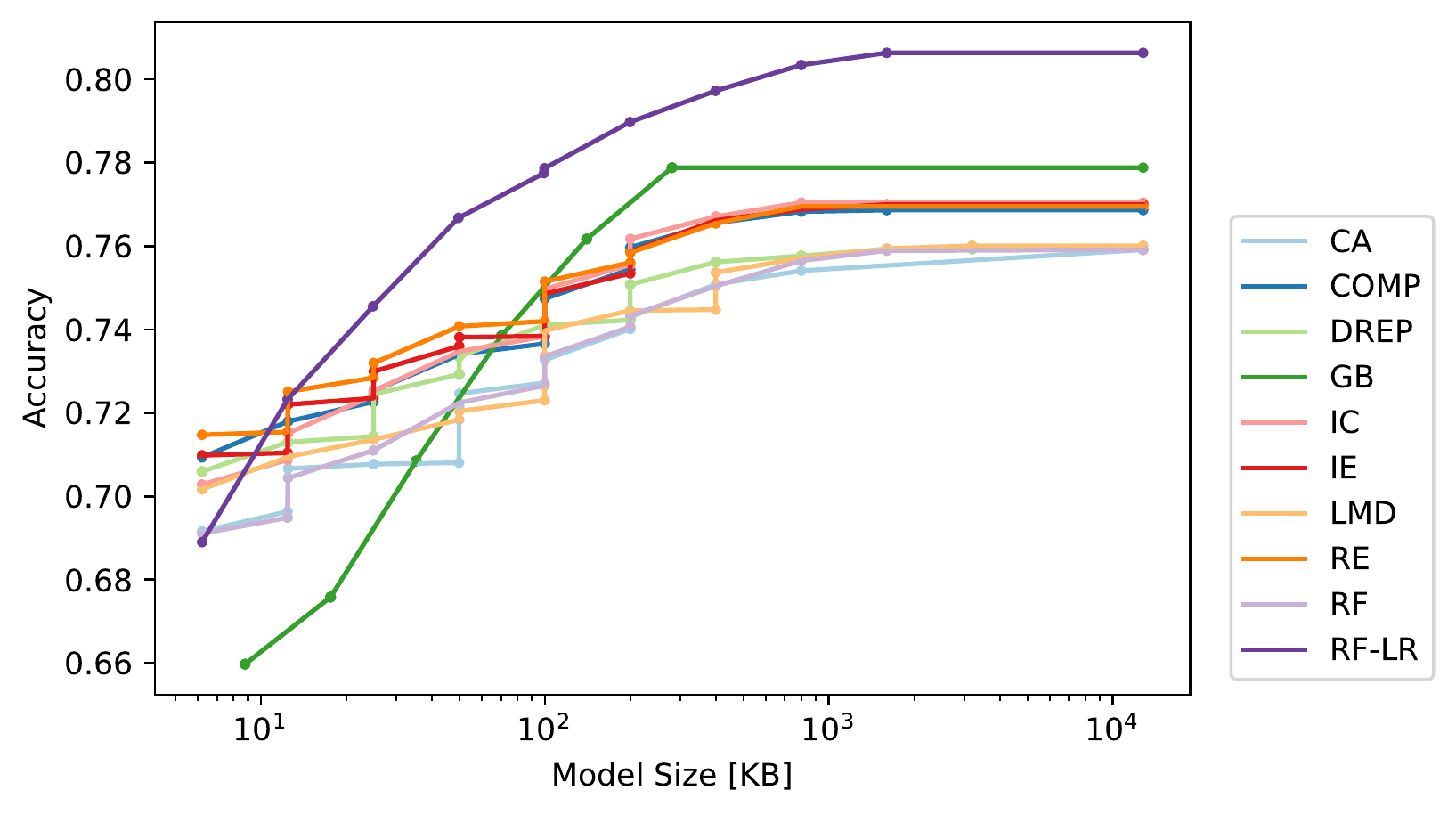}
\end{minipage}
\caption{(left) 5-fold cross-validation accuracy on the chess dataset. (right) 5-fold cross-validation accuracy on the connect dataset.}
\end{figure}

\begin{figure}[H]
\begin{minipage}{.49\textwidth}
    \centering
    \includegraphics[width=\textwidth,keepaspectratio]{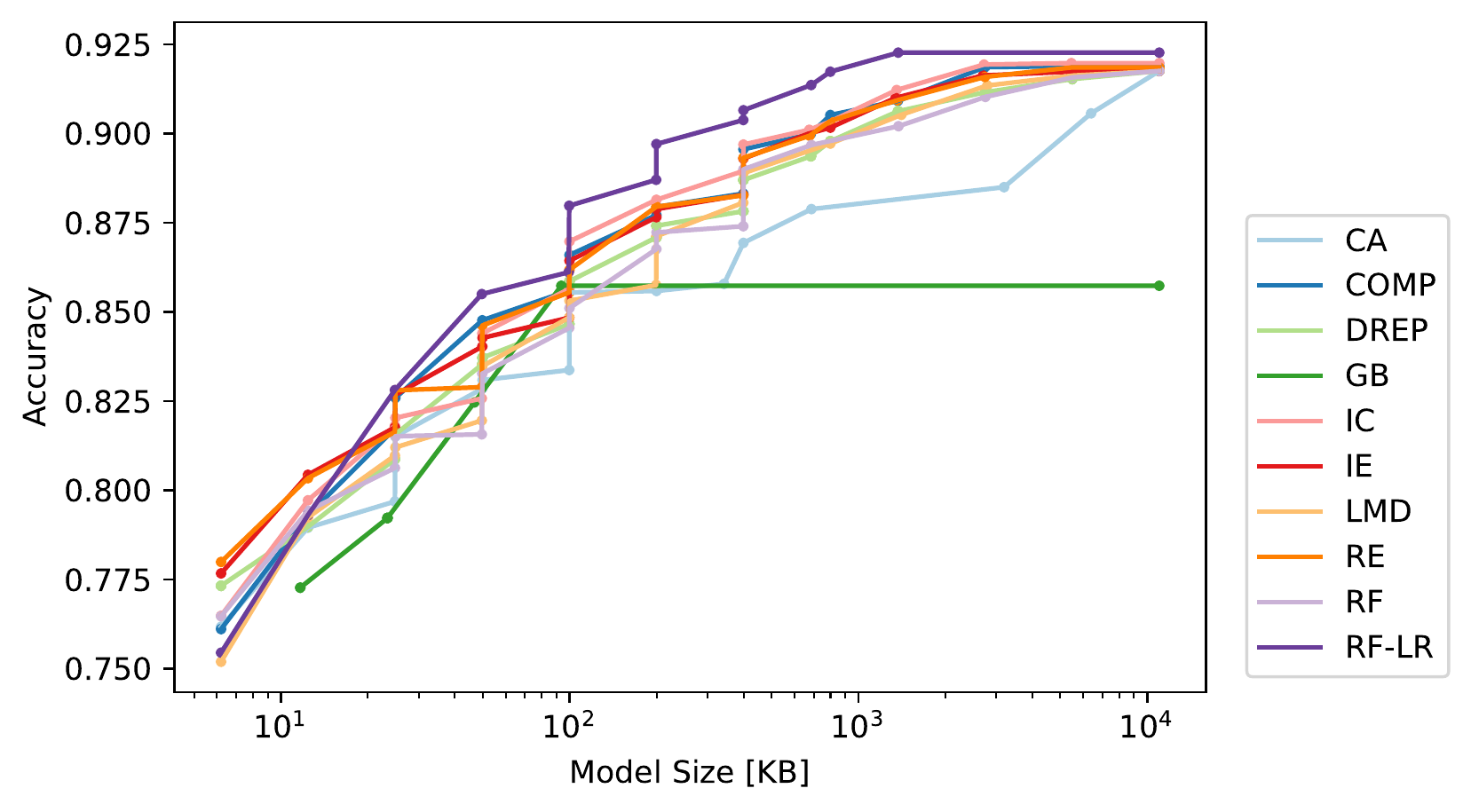}
\end{minipage}\hfill
\begin{minipage}{.49\textwidth}
    \centering 
    \includegraphics[width=\textwidth,keepaspectratio]{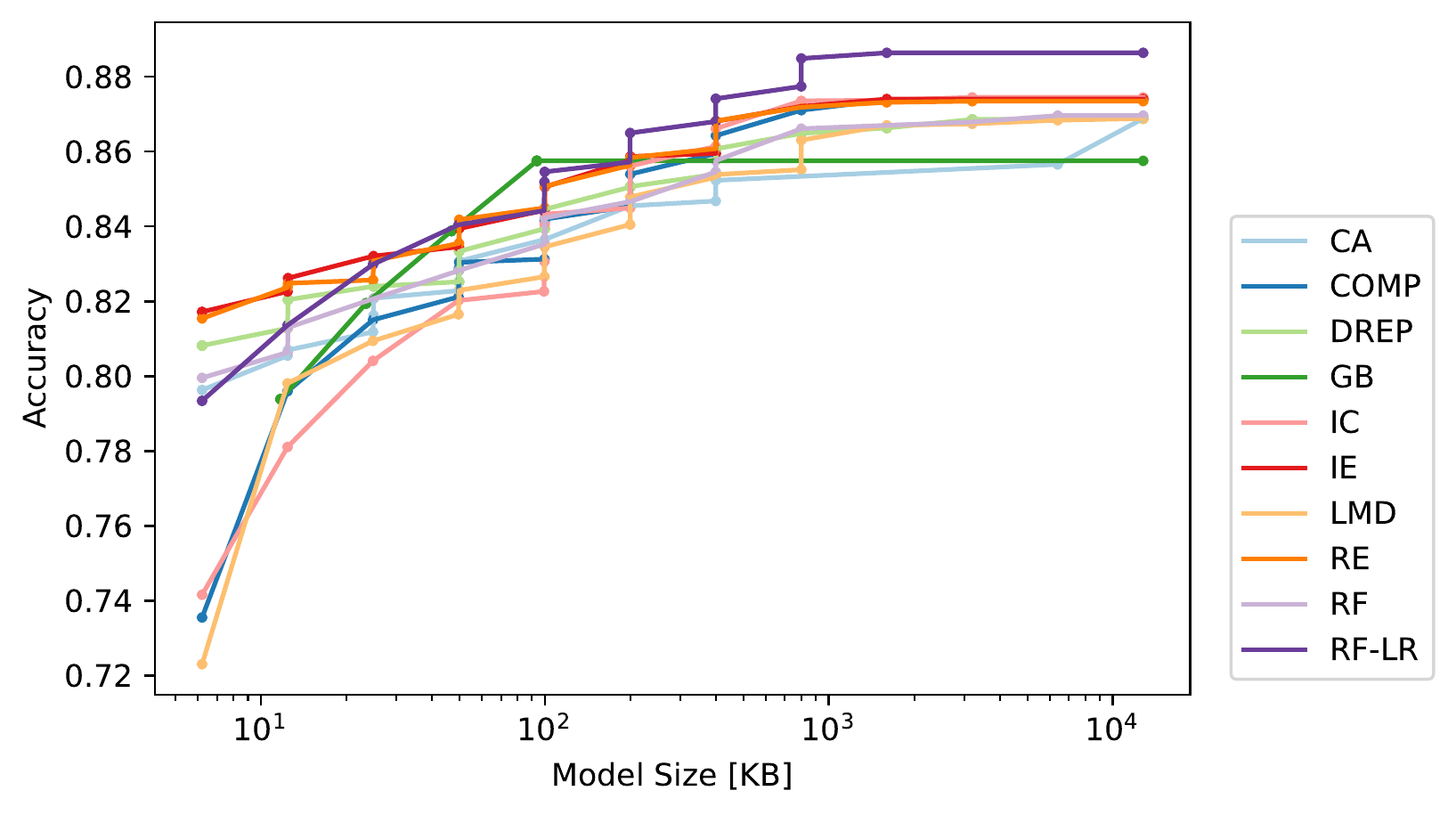}
\end{minipage}
\caption{(left) 5-fold cross-validation accuracy on the eeg dataset. (right) 5-fold cross-validation accuracy on the elec dataset.}
\end{figure}

\begin{figure}[H]
\begin{minipage}{.49\textwidth}
    \centering
    \includegraphics[width=\textwidth,keepaspectratio]{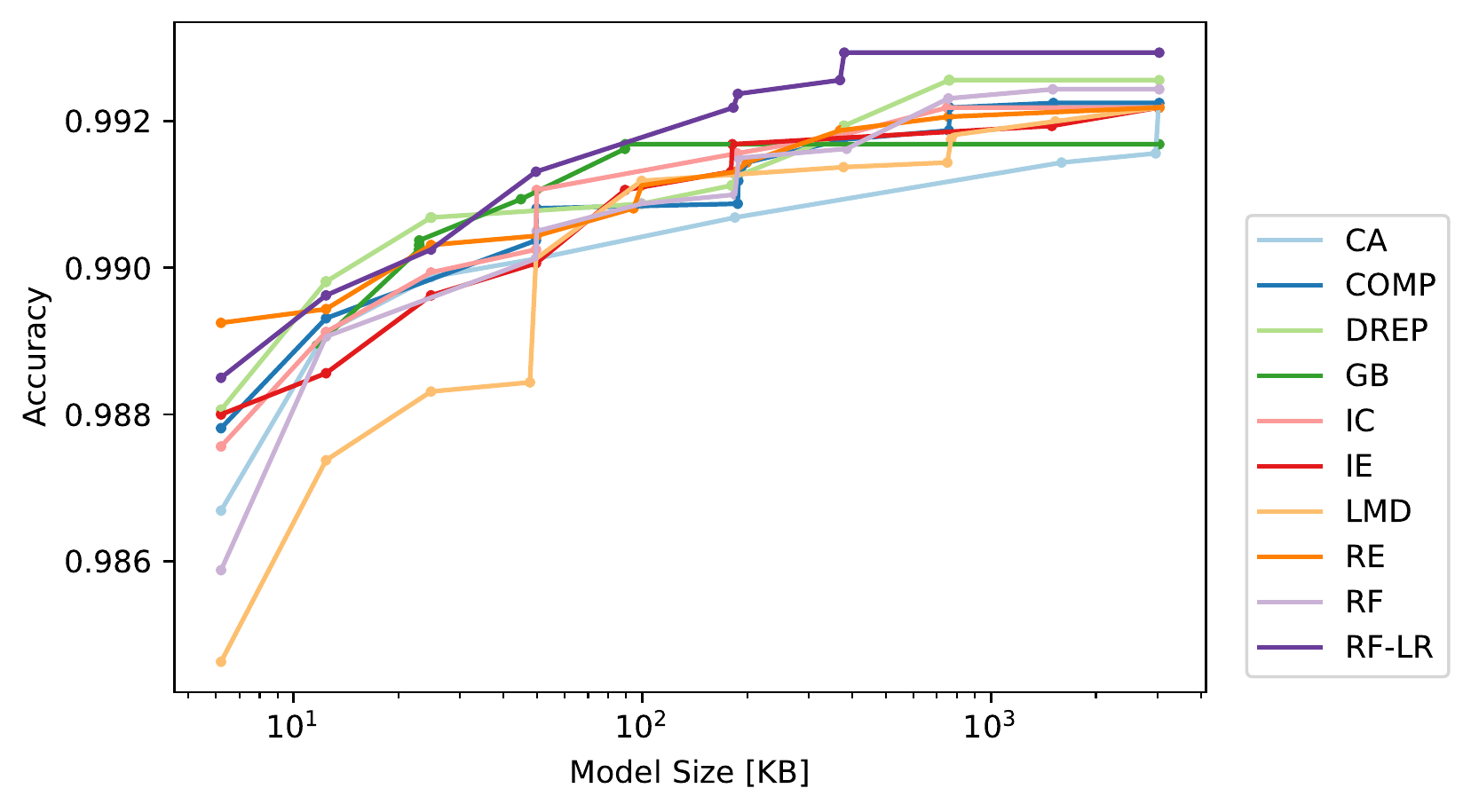}
\end{minipage}\hfill
\begin{minipage}{.49\textwidth}
    \centering 
    \includegraphics[width=\textwidth,keepaspectratio]{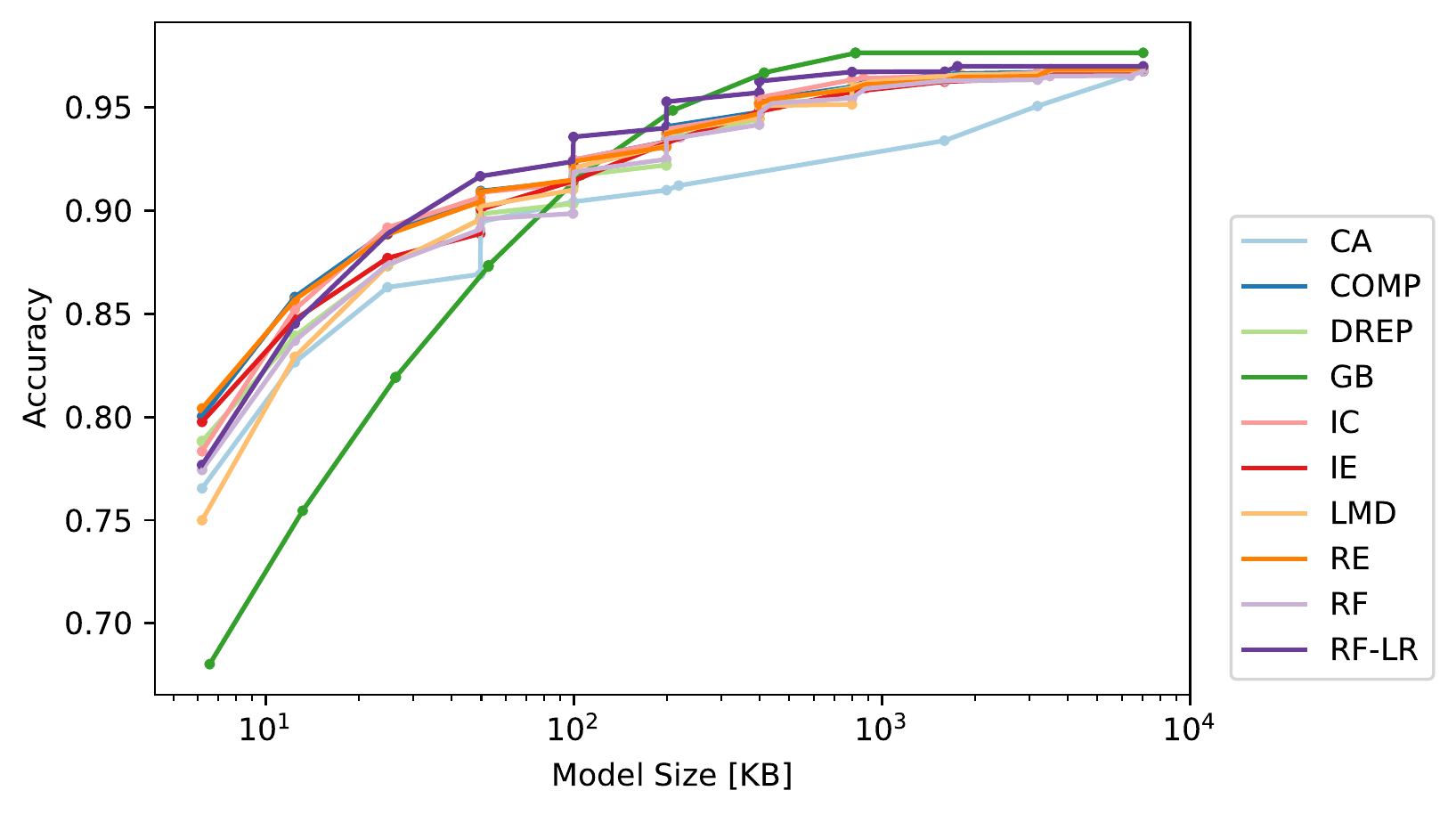}
\end{minipage}
\caption{(left) 5-fold cross-validation accuracy on the ida2016 dataset. (right) 5-fold cross-validation accuracy on the japanese-vowels dataset.}
\end{figure}

\begin{figure}[H]
\begin{minipage}{.49\textwidth}
    \centering
    \includegraphics[width=\textwidth,keepaspectratio]{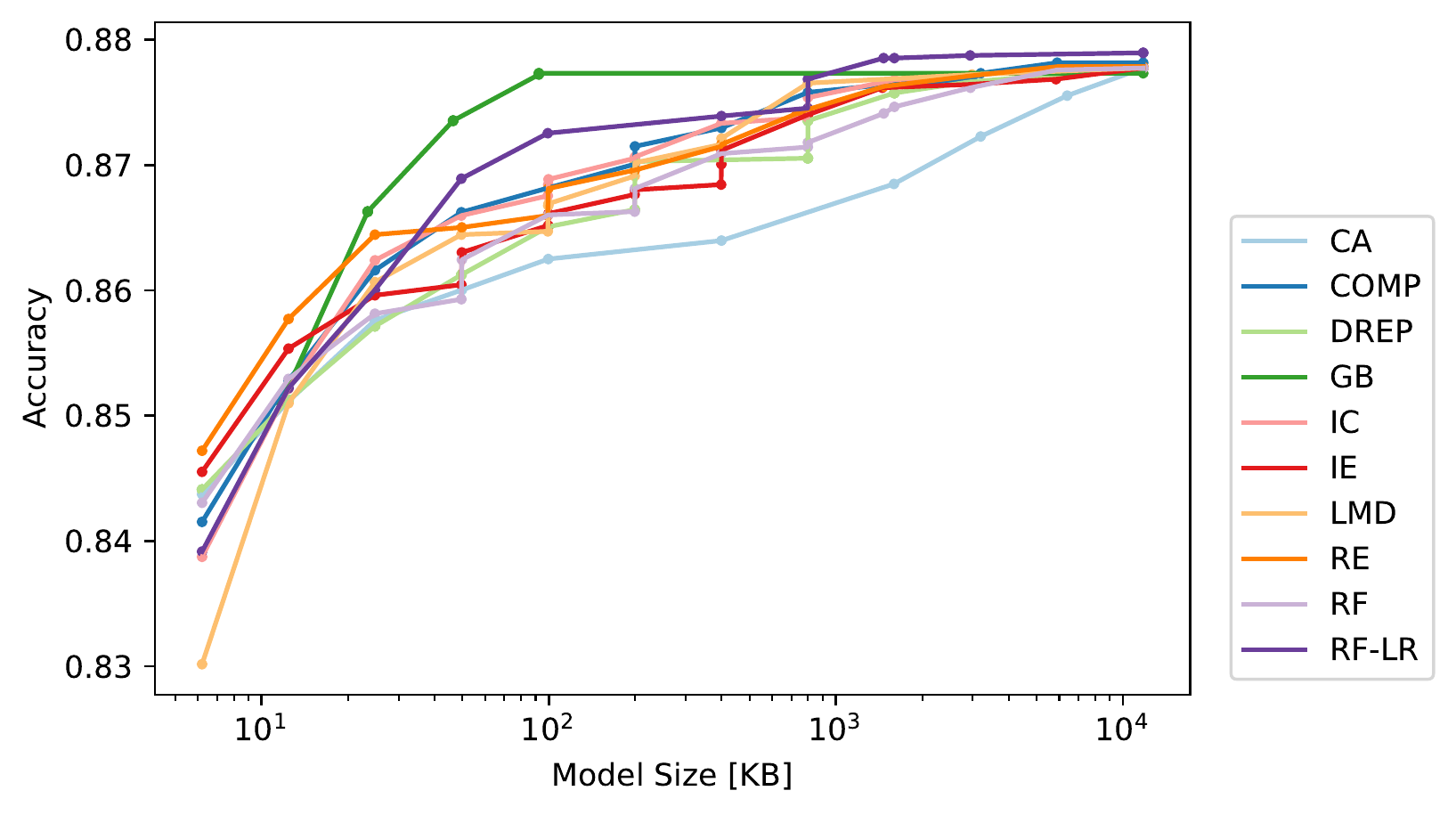}
\end{minipage}\hfill
\begin{minipage}{.49\textwidth}
    \centering 
    \includegraphics[width=\textwidth,keepaspectratio]{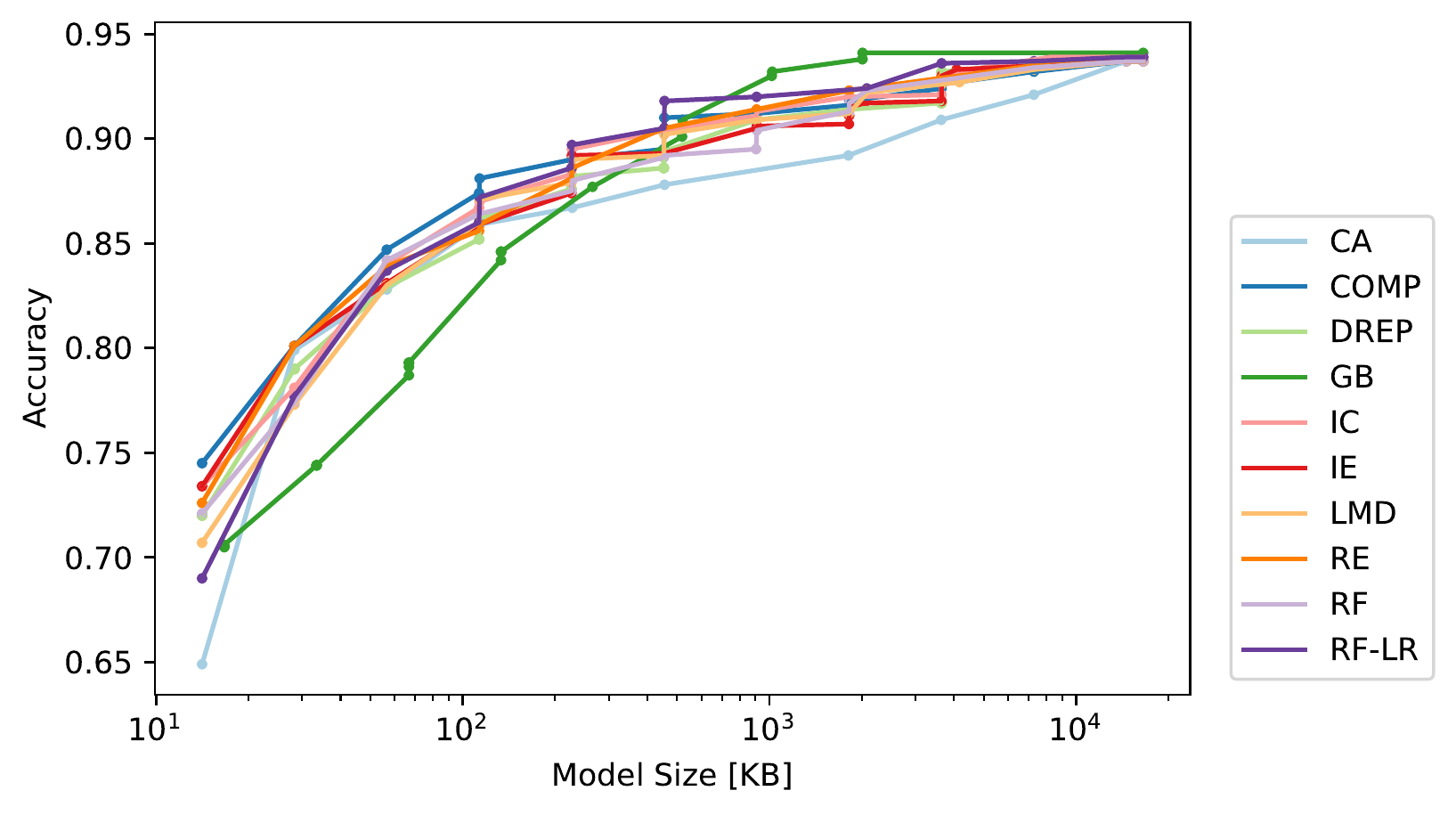}
\end{minipage}
\caption{(left) 5-fold cross-validation accuracy on the magic dataset. (right) 5-fold cross-validation accuracy on the mnist dataset.}
\end{figure}

\begin{figure}[H]
\begin{minipage}{.49\textwidth}
    \centering
    \includegraphics[width=\textwidth,keepaspectratio]{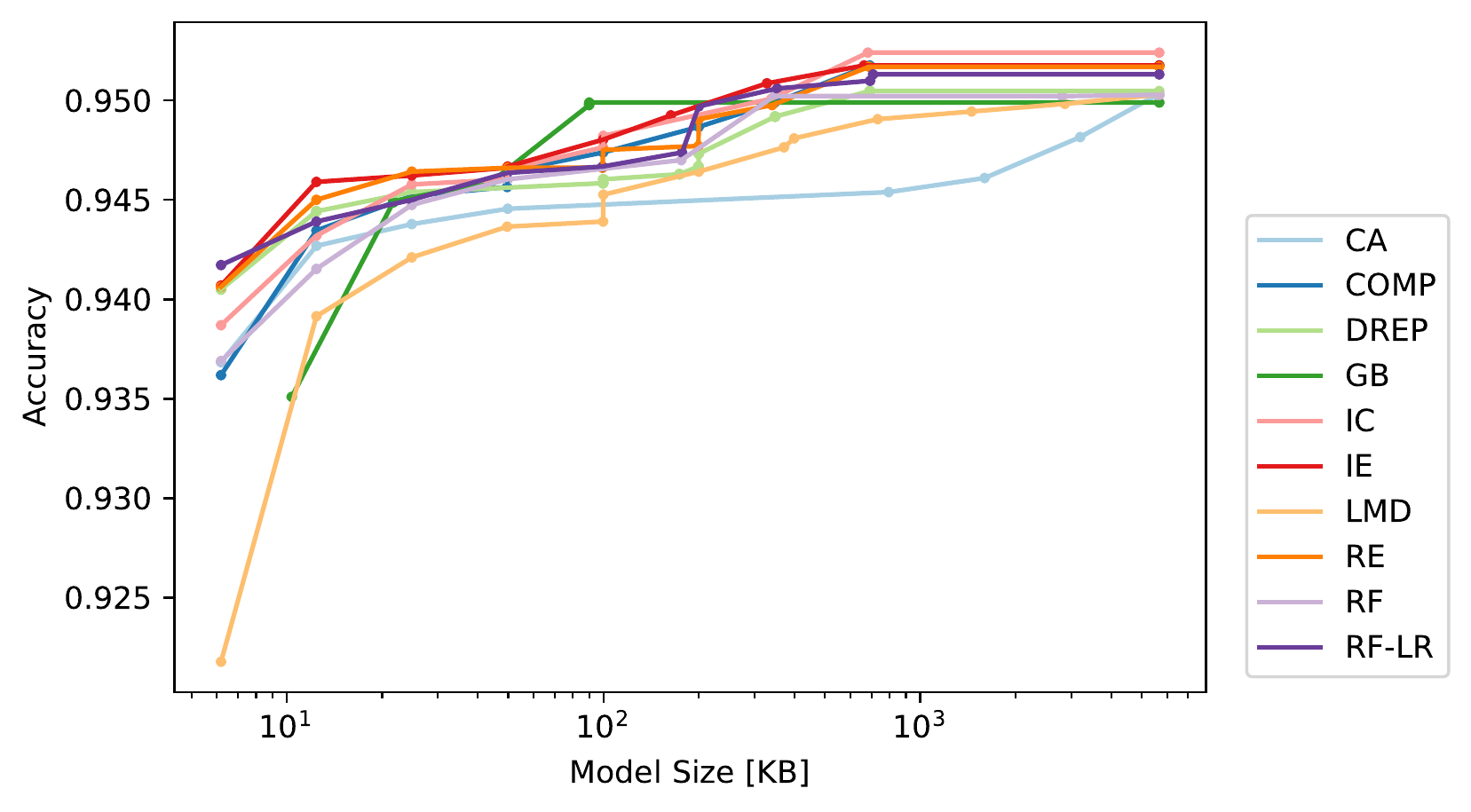}
\end{minipage}\hfill
\begin{minipage}{.49\textwidth}
    \centering 
    \includegraphics[width=\textwidth,keepaspectratio]{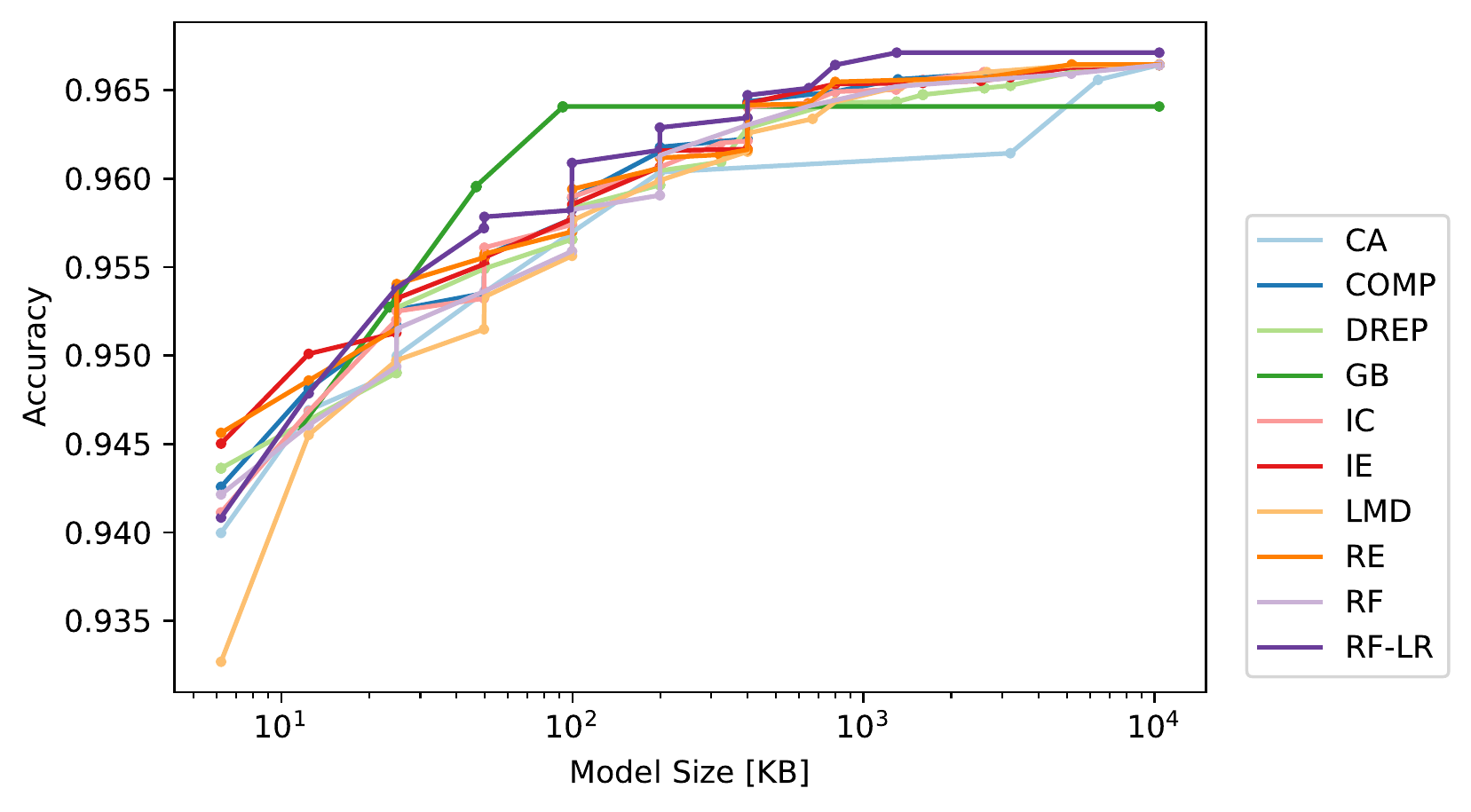}
\end{minipage}
\caption{(left) 5-fold cross-validation accuracy on the mozilla dataset. (right) 5-fold cross-validation accuracy on the nomao dataset.}
\end{figure}

\begin{figure}[H]
\begin{minipage}{.49\textwidth}
    \centering
    \includegraphics[width=\textwidth,keepaspectratio]{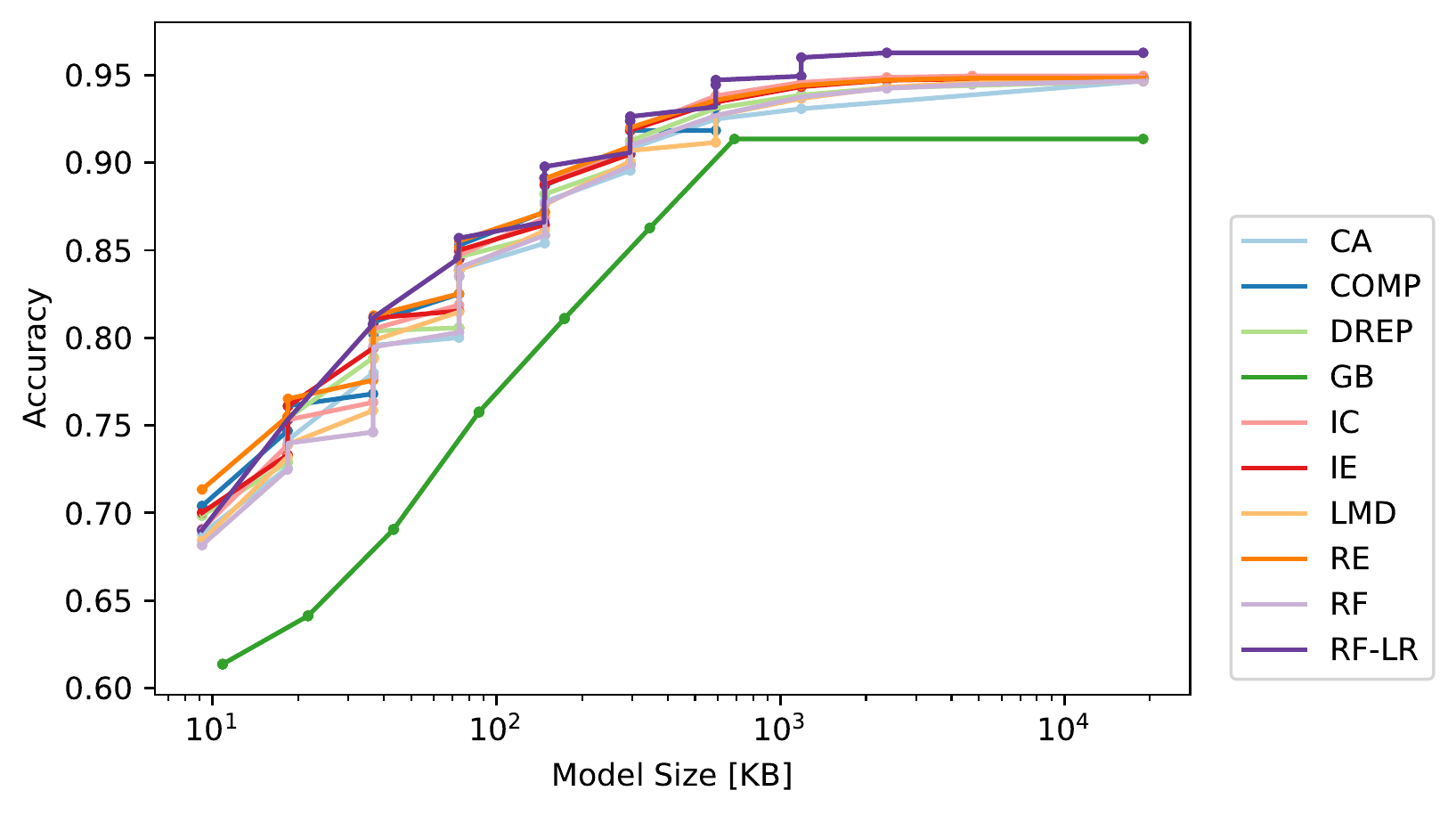}
\end{minipage}\hfill
\begin{minipage}{.49\textwidth}
    \centering 
    \includegraphics[width=\textwidth,keepaspectratio]{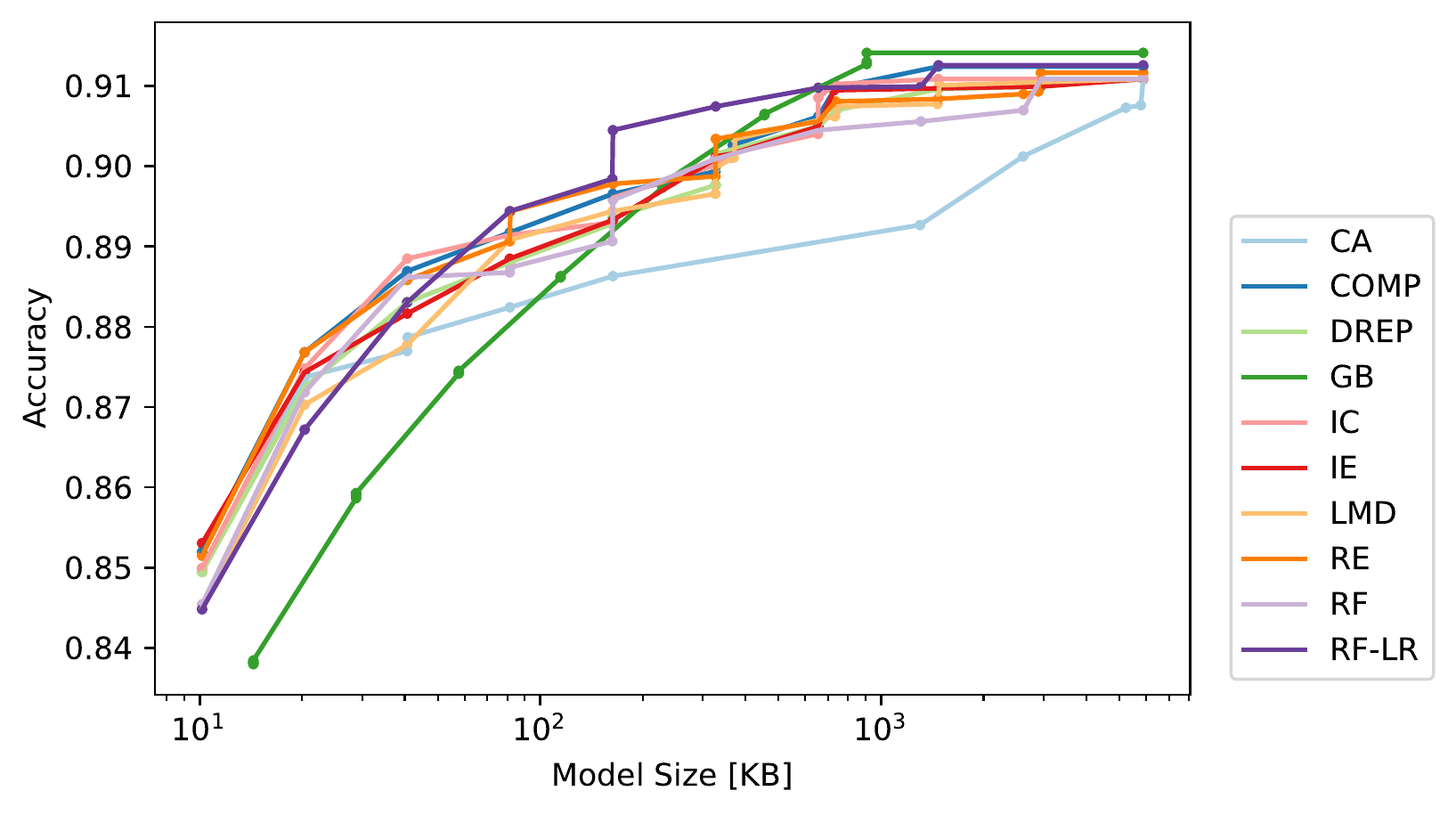}
\end{minipage}
\caption{(left) 5-fold cross-validation accuracy on the postures dataset. (right) 5-fold cross-validation accuracy on the satimage dataset.}
\end{figure}

\subsection{Accuracies under various resource constraints with Dedicated Pruning Set}

\begin{table}
    \centering
    \resizebox{\textwidth}{!}{
    \input{figures/raw_RandomForestClassifier_with_prune_32}
    }
    \caption{Test accuracies for models with a memory consumption below 32 KB for each method and each dataset averaged over a 5 fold cross validation using a dedicated pruning set. Rounded to the third decimal digit. Larger is better. The best method is depicted in bold.}
\end{table}

\begin{table}
    \centering
    \resizebox{\textwidth}{!}{
    \input{figures/raw_RandomForestClassifier_with_prune_64}
    }
    \caption{Test accuracies for models with a memory consumption below 64 KB for each method and each dataset averaged over a 5 fold cross validation using a dedicated pruning set. Rounded to the third decimal digit. Larger is better. The best method is depicted in bold.}
\end{table}

\begin{table}
    \centering
    \resizebox{\textwidth}{!}{
    \input{figures/raw_RandomForestClassifier_with_prune_128}
    }
    \caption{Test accuracies for models with a memory consumption below 128 KB for each method and each dataset averaged over a 5 fold cross validation using a dedicated pruning set. Rounded to the third decimal digit. Larger is better. The best method is depicted in bold.}
\end{table}

\begin{table}
    \centering
    \resizebox{\textwidth}{!}{
    \input{figures/raw_RandomForestClassifier_with_prune_256}
    }
    \caption{Test accuracies for models with a memory consumption below 256 KB for each method and each dataset averaged over a 5 fold cross validation using a dedicated pruning set. Rounded to the third decimal digit. Larger is better. The best method is depicted in bold.}
\end{table}

\subsection{Area Under the Pareto Front with Dedicated Pruning Set}

\resizebox{\textwidth}{!}{
    \begin{tabular}{lrrrrrrrrrr}
\toprule
{} & \multicolumn{10}{l}{AUC} \\
model &      CA &    COMP &    DREP &      GB &      IC &      IE &     LMD &      RE &      RF &   RF-LR \\
dataset         &         &         &         &         &         &         &         &         &         &         \\
\midrule
adult           &  0.8606 &  0.8621 &  0.8619 &  0.8705 &  0.8625 &  0.8626 &  0.8611 &  0.8623 &  0.8614 &  0.8643 \\
anura           &  0.9673 &  0.9759 &  0.9740 &  0.9749 &  0.9757 &  0.9750 &  0.9743 &  0.9748 &  0.9738 &  0.9760 \\
avila           &  0.9536 &  0.9933 &  0.9843 &  0.9907 &  0.9944 &  0.9938 &  0.9677 &  0.9930 &  0.9823 &  0.9903 \\
bank            &  0.9019 &  0.9047 &  0.9032 &  0.9065 &  0.9048 &  0.9042 &  0.9037 &  0.9042 &  0.9034 &  0.9067 \\
chess           &  0.5939 &  0.6401 &  0.6210 &  0.6265 &  0.6383 &  0.6313 &  0.6309 &  0.6383 &  0.6209 &  0.6618 \\
\bottomrule
\end{tabular}
}

\begin{figure}
\centering
\includegraphics[width=\columnwidth, keepaspectratio]{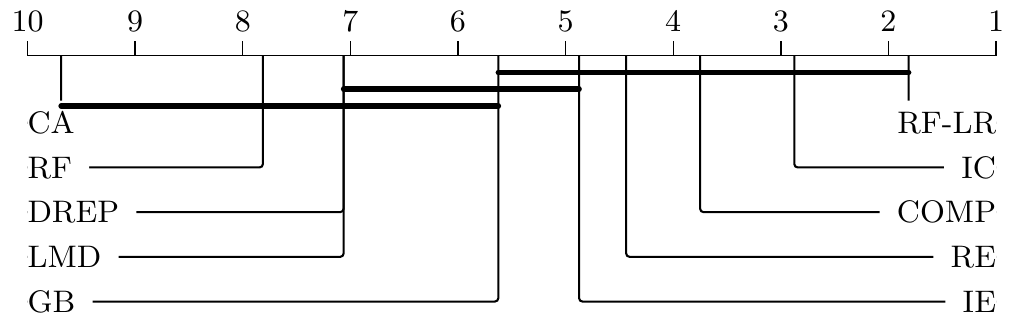}
\caption{Critical Difference Diagram for the normalized area under the Pareto front for different methods over multiple datasets. More to the right (lower rank) is better. Methods in connected cliques are statistically similar.}
\label{fig:cd_auc}
\end{figure}

\section{Revisiting Ensemble Pruning with a Bagging Classifier }

For space reasons, the paper focuses on Random Forest classifier. Here we will repeat our experiment with a Bagging Classifier implemented in Scikit-Learn \cite{Pedregosa/etal/2001}. As before, we either use a 5-fold cross validation or the given test/train split. For reference, recall our experimental protocol: Oshiro et al. showed in \cite{oshiro/etal/2012} that the prediction of a RF stabilizes between $128$ and $256$ trees in the ensemble and adding more trees to the ensemble does not yield significantly better results. Hence, we train the `base' Random Forests with $M = 256$ trees. To control the individual errors of trees we set the maximum number of leaf nodes $n_l$ to values between $n_l \in \{64,128,256,512,1024\}$. For ensemble pruning we use RE and compare it against a random selection of trees from the original ensemble (which is the same a training a smaller forest directly). In both cases a sub-ensemble with $K \in \{2,4,8,16,32,64,128,256\}$ members is selected so that for $K=256$ the original RF is recovered. 

\begin{figure}[H]
\begin{minipage}{.49\textwidth}
    \centering
    \includegraphics[width=\textwidth,keepaspectratio]{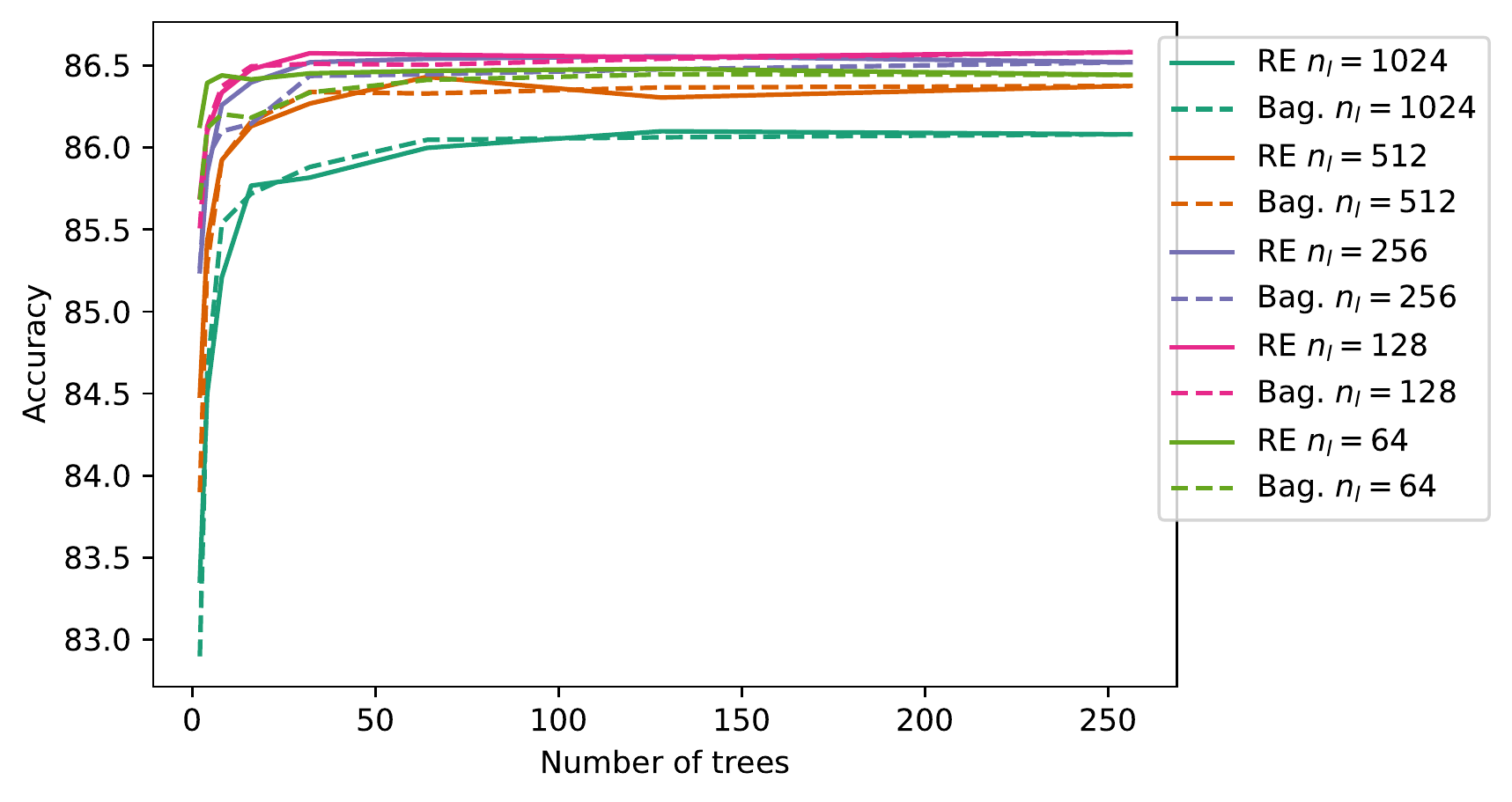}
\end{minipage}\hfill
\begin{minipage}{.49\textwidth}
    \centering 
    \resizebox{\textwidth}{!}{
        \input{figures/BaggingClassifier_adult_table}
    }
\end{minipage}
\caption{(Left) The error over the number of trees in the ensemble on the adult dataset. Dashed lines depict the Random Forest and solid lines are the corresponding pruned ensemble via Reduced Error pruning. (Right) The 5-fold cross-validation accuracy  on the adult dataset. Rounded to the second decimal digit. Larger is better.}
\end{figure}

\begin{figure}[H]
\begin{minipage}{.49\textwidth}
    \centering
    \includegraphics[width=\textwidth,keepaspectratio]{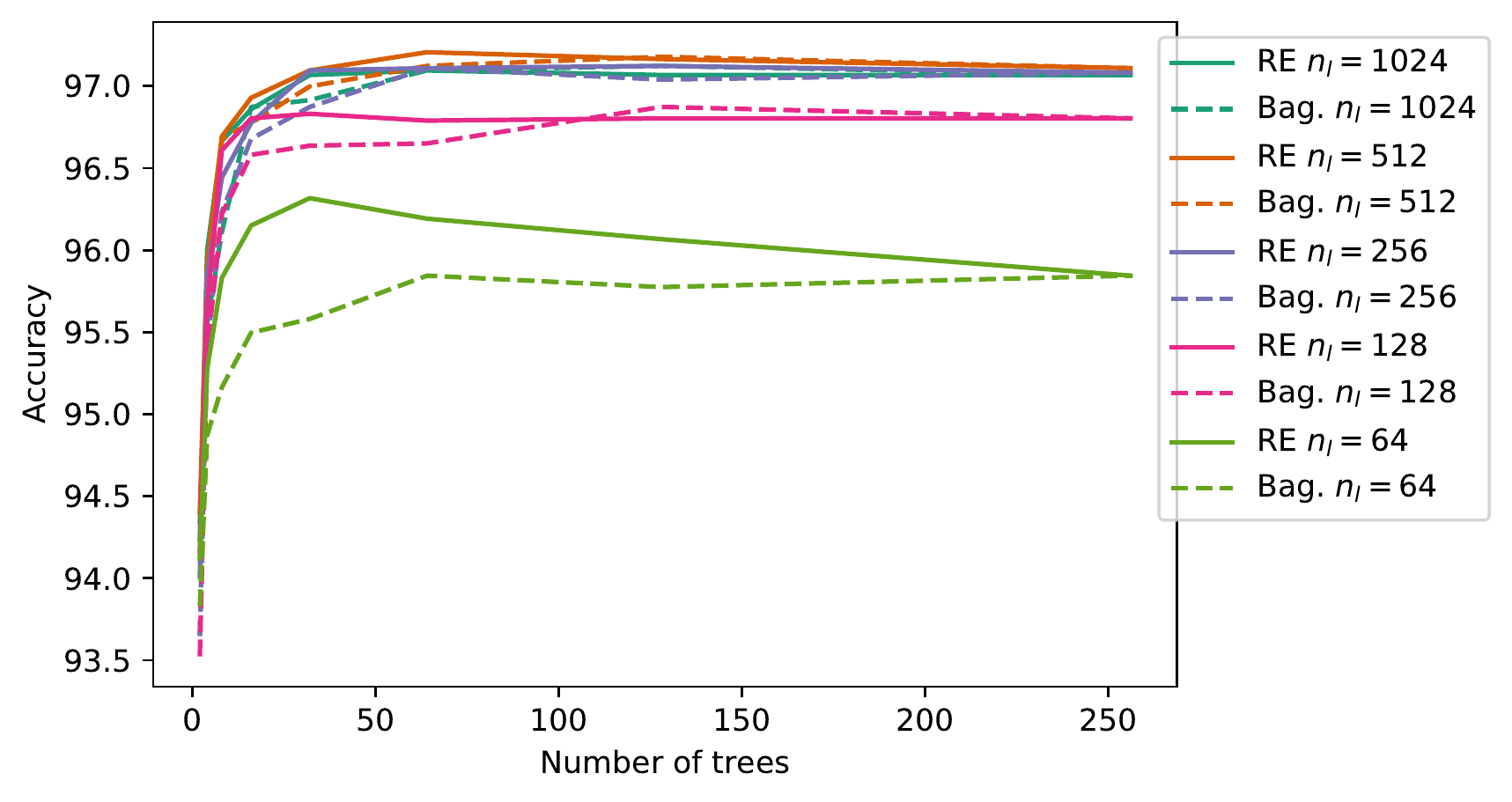}
\end{minipage}\hfill
\begin{minipage}{.49\textwidth}
    \centering 
    \resizebox{\textwidth}{!}{
        \input{figures/BaggingClassifier_anura_table}
    }
\end{minipage}
\caption{(Left) The error over the number of trees in the ensemble on the anura dataset. Dashed lines depict the Random Forest and solid lines are the corresponding pruned ensemble via Reduced Error pruning. (Right) The 5-fold cross-validation accuracy  on the anura dataset. Rounded to the second decimal digit. Larger is better.}
\end{figure}

\begin{figure}[H]
\begin{minipage}{.49\textwidth}
    \centering
    \includegraphics[width=\textwidth,keepaspectratio]{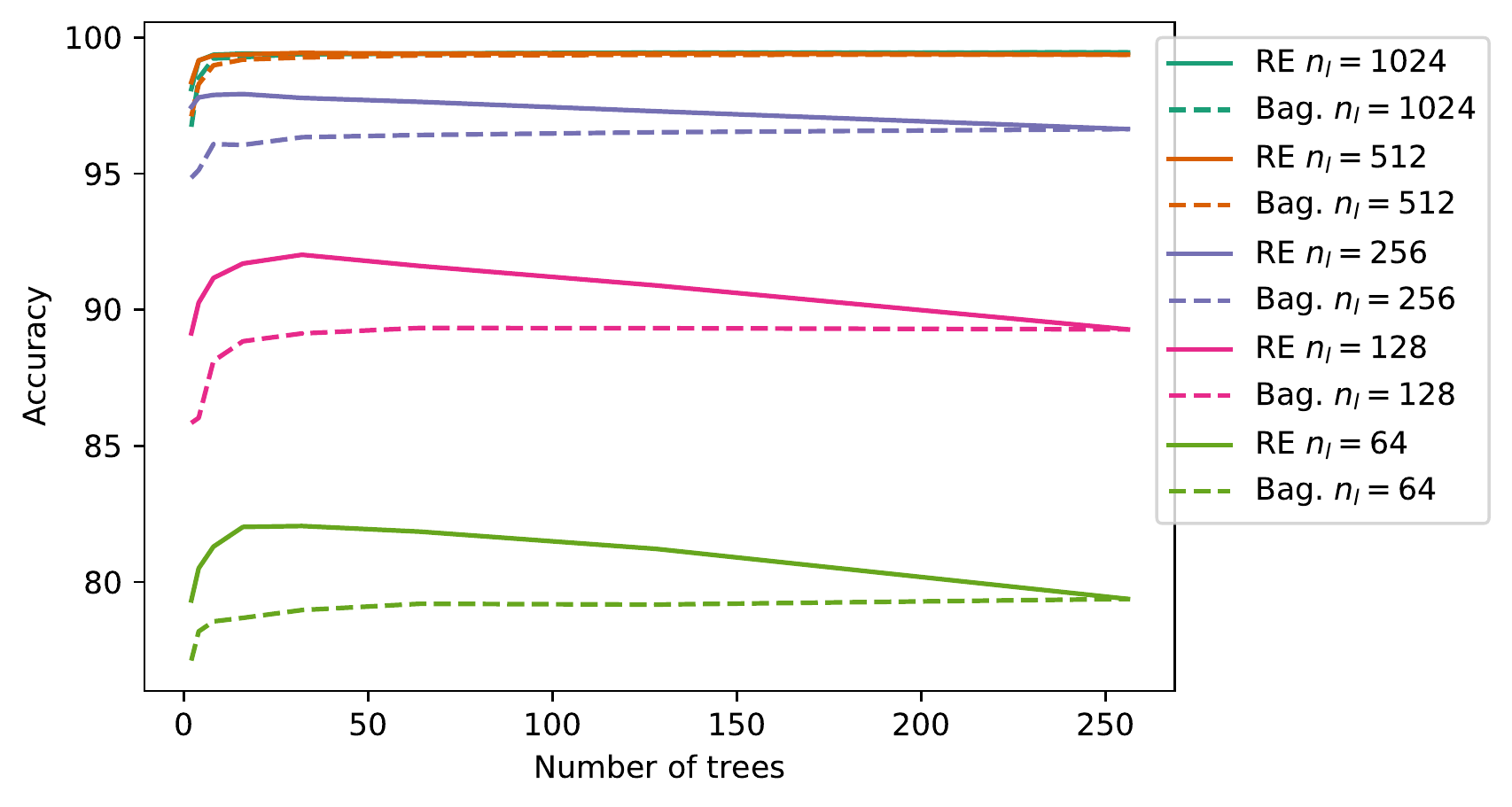}
\end{minipage}\hfill
\begin{minipage}{.49\textwidth}
    \centering 
    \resizebox{\textwidth}{!}{
        \input{figures/BaggingClassifier_avila_table}
    }
\end{minipage}
\caption{(Left) The error over the number of trees in the ensemble on the avila dataset. Dashed lines depict the Random Forest and solid lines are the corresponding pruned ensemble via Reduced Error pruning. (Right) The 5-fold cross-validation accuracy  on the avila dataset. Rounded to the second decimal digit. Larger is better.}
\end{figure}

\begin{figure}[H]
\begin{minipage}{.49\textwidth}
    \centering
    \includegraphics[width=\textwidth,keepaspectratio]{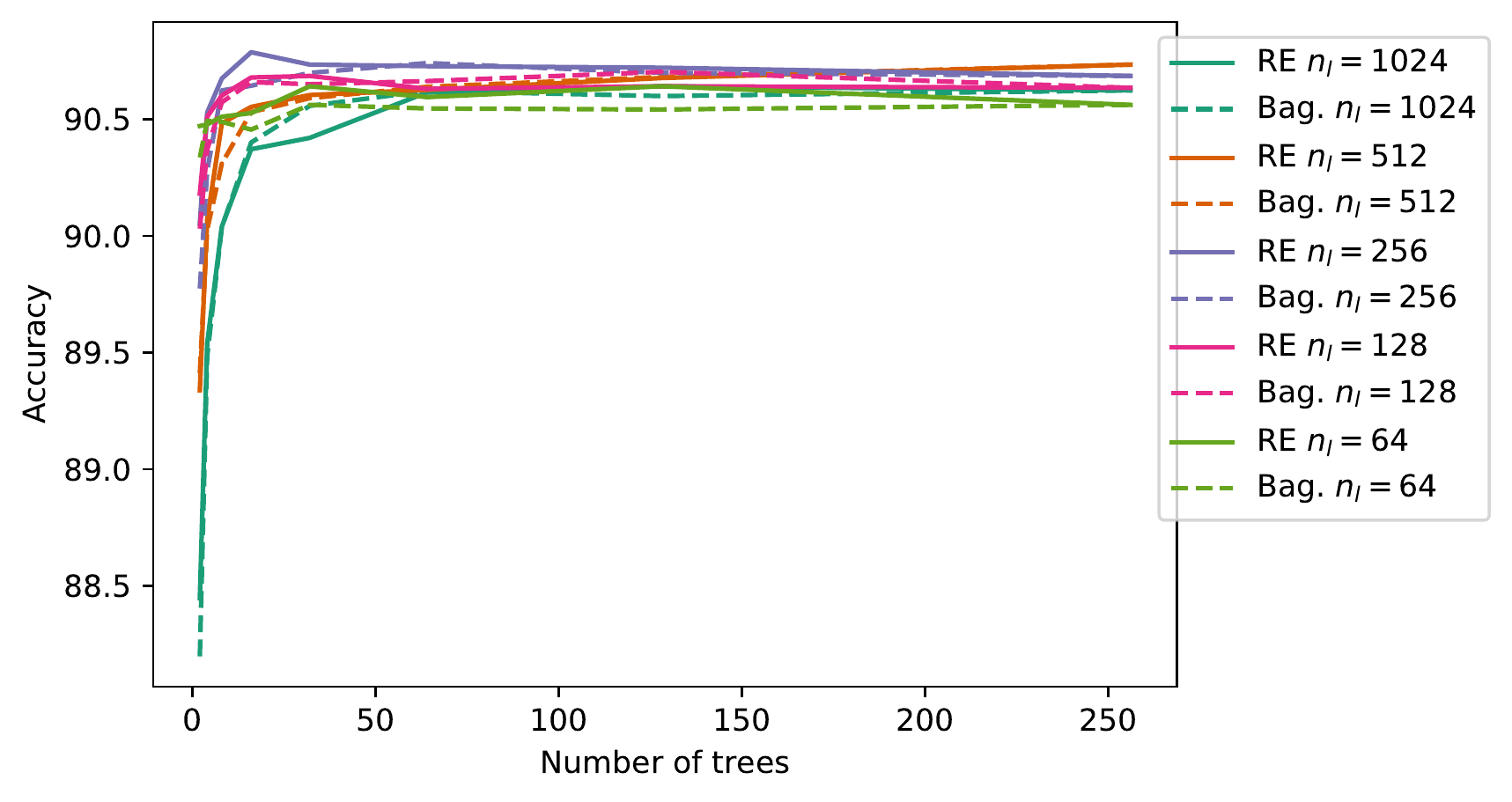}
\end{minipage}\hfill
\begin{minipage}{.49\textwidth}
    \centering 
    \resizebox{\textwidth}{!}{
        \input{figures/BaggingClassifier_bank_table}
    }
\end{minipage}
\caption{(Left) The error over the number of trees in the ensemble on the bank dataset. Dashed lines depict the Random Forest and solid lines are the corresponding pruned ensemble via Reduced Error pruning. (Right) The 5-fold cross-validation accuracy  on the bank dataset. Rounded to the second decimal digit. Larger is better.}
\end{figure}

\begin{figure}[H]
\begin{minipage}{.49\textwidth}
    \centering
    \includegraphics[width=\textwidth,keepaspectratio]{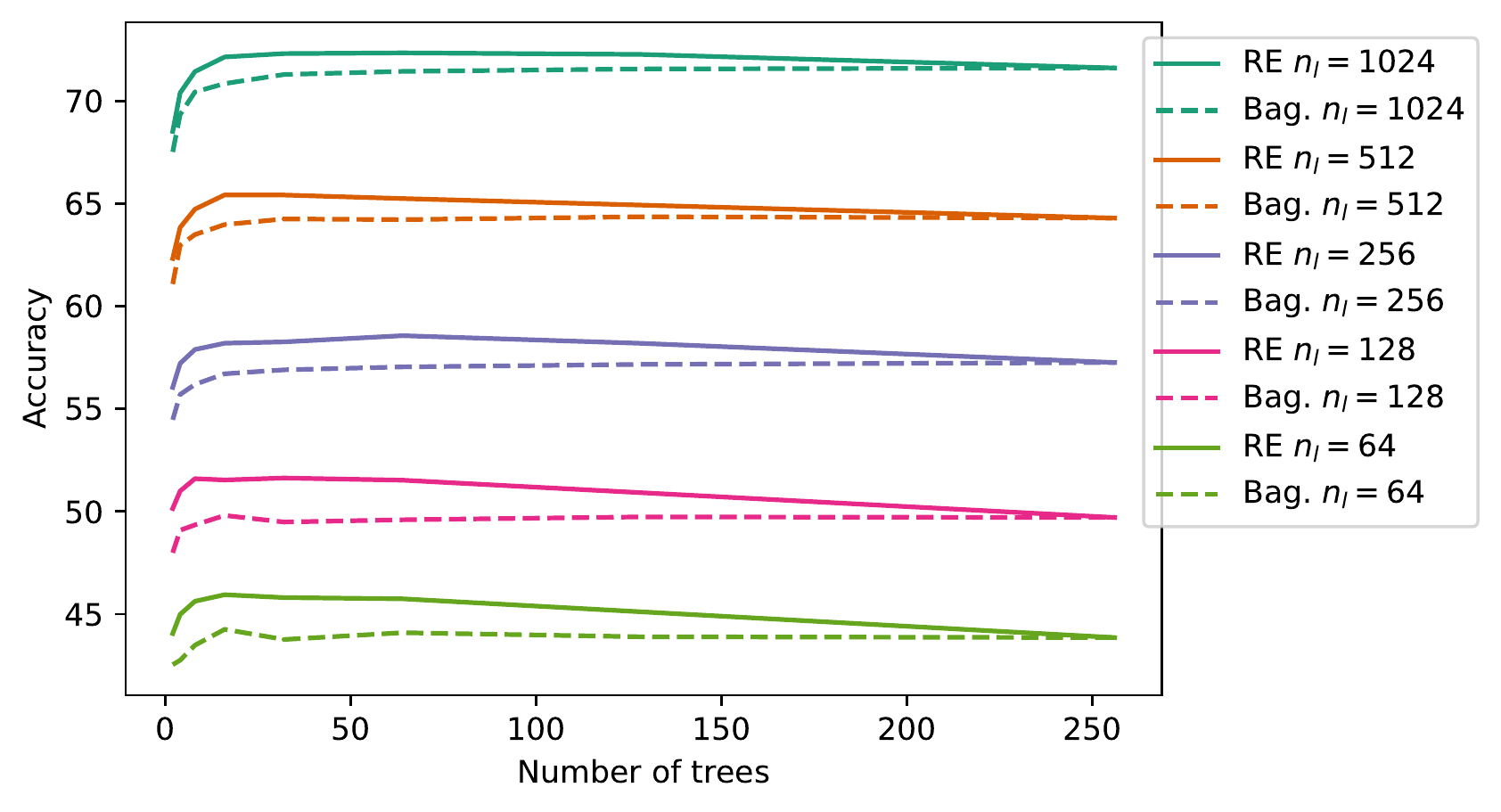}
\end{minipage}\hfill
\begin{minipage}{.49\textwidth}
    \centering 
    \resizebox{\textwidth}{!}{
        \input{figures/BaggingClassifier_chess_table}
    }
\end{minipage}
\caption{(Left) The error over the number of trees in the ensemble on the chess dataset. Dashed lines depict the Random Forest and solid lines are the corresponding pruned ensemble via Reduced Error pruning. (Right) The 5-fold cross-validation accuracy  on the chess dataset. Rounded to the second decimal digit. Larger is better.}
\end{figure}

\begin{figure}[H]
\begin{minipage}{.49\textwidth}
    \centering
    \includegraphics[width=\textwidth,keepaspectratio]{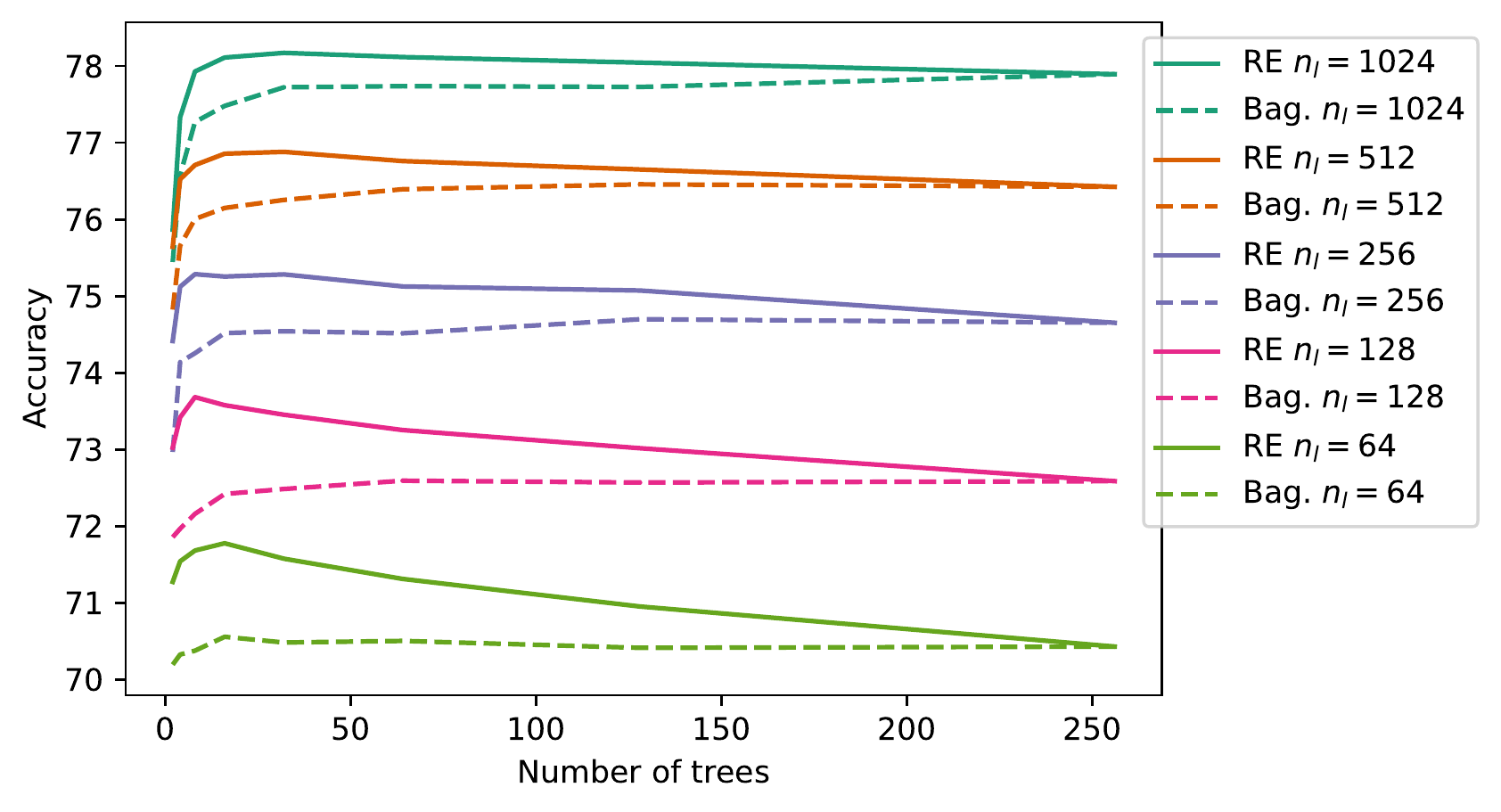}
\end{minipage}\hfill
\begin{minipage}{.49\textwidth}
    \centering 
    \resizebox{\textwidth}{!}{
        \input{figures/BaggingClassifier_connect_table}
    }
\end{minipage}
\caption{(Left) The error over the number of trees in the ensemble on the connect dataset. Dashed lines depict the Random Forest and solid lines are the corresponding pruned ensemble via Reduced Error pruning. (Right) The 5-fold cross-validation accuracy  on the connect dataset. Rounded to the second decimal digit. Larger is better.}
\end{figure}

\begin{figure}[H]
\begin{minipage}{.49\textwidth}
    \centering
    \includegraphics[width=\textwidth,keepaspectratio]{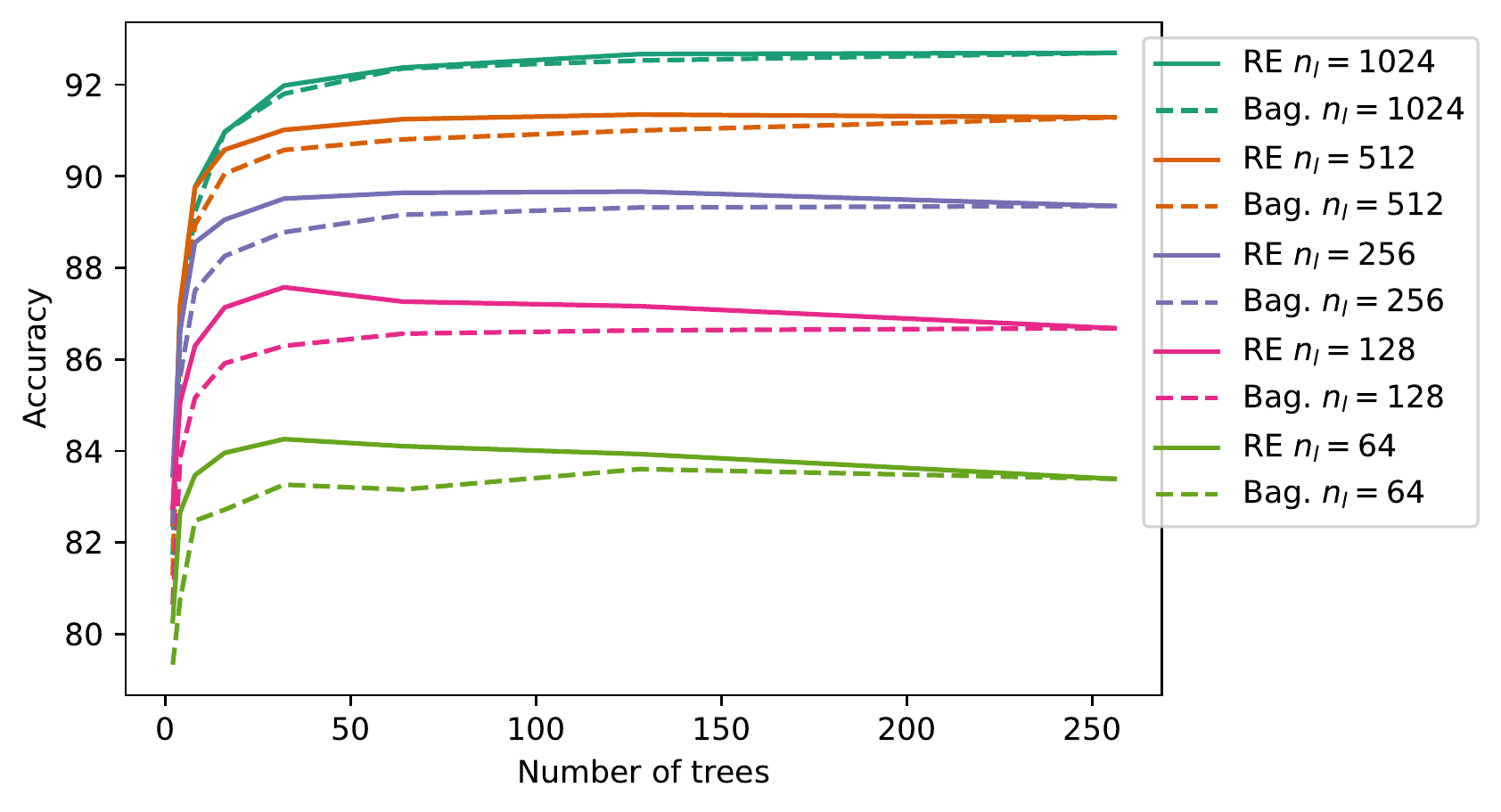}
\end{minipage}\hfill
\begin{minipage}{.49\textwidth}
    \centering 
    \resizebox{\textwidth}{!}{
        \input{figures/BaggingClassifier_eeg_table}
    }
\end{minipage}
\caption{(Left) The error over the number of trees in the ensemble on the eeg dataset. Dashed lines depict the Random Forest and solid lines are the corresponding pruned ensemble via Reduced Error pruning. (Right) The 5-fold cross-validation accuracy  on the eeg dataset. Rounded to the second decimal digit. Larger is better.}
\end{figure}

\begin{figure}[H]
\begin{minipage}{.49\textwidth}
    \centering
    \includegraphics[width=\textwidth,keepaspectratio]{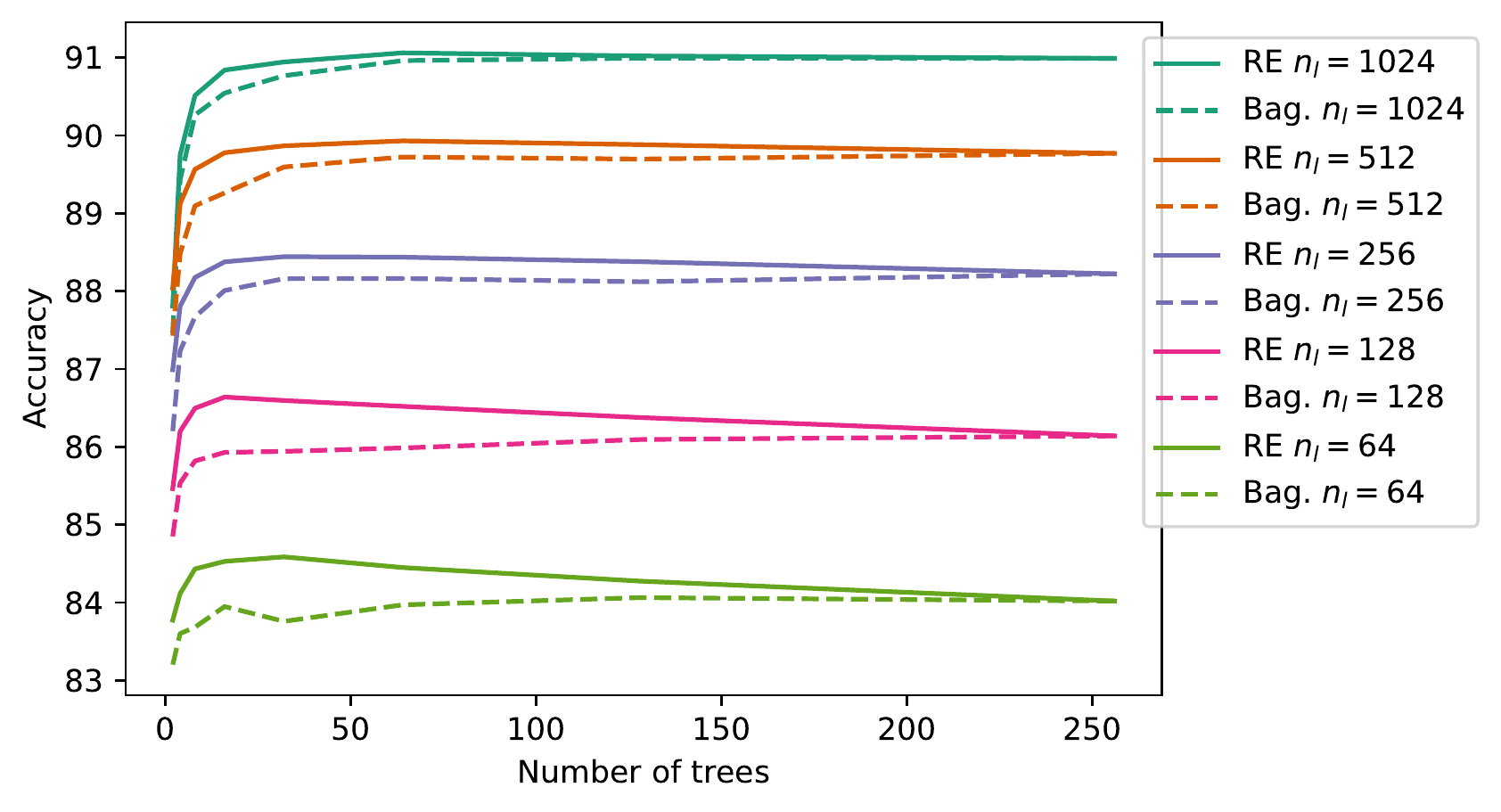}
\end{minipage}\hfill
\begin{minipage}{.49\textwidth}
    \centering 
    \resizebox{\textwidth}{!}{
        \input{figures/BaggingClassifier_elec_table}
    }
\end{minipage}
\caption{(Left) The error over the number of trees in the ensemble on the elec dataset. Dashed lines depict the Random Forest and solid lines are the corresponding pruned ensemble via Reduced Error pruning. (Right) The 5-fold cross-validation accuracy  on the elec dataset. Rounded to the second decimal digit. Larger is better.}
\end{figure}

\begin{figure}[H]
\begin{minipage}{.49\textwidth}
    \centering
    \includegraphics[width=\textwidth,keepaspectratio]{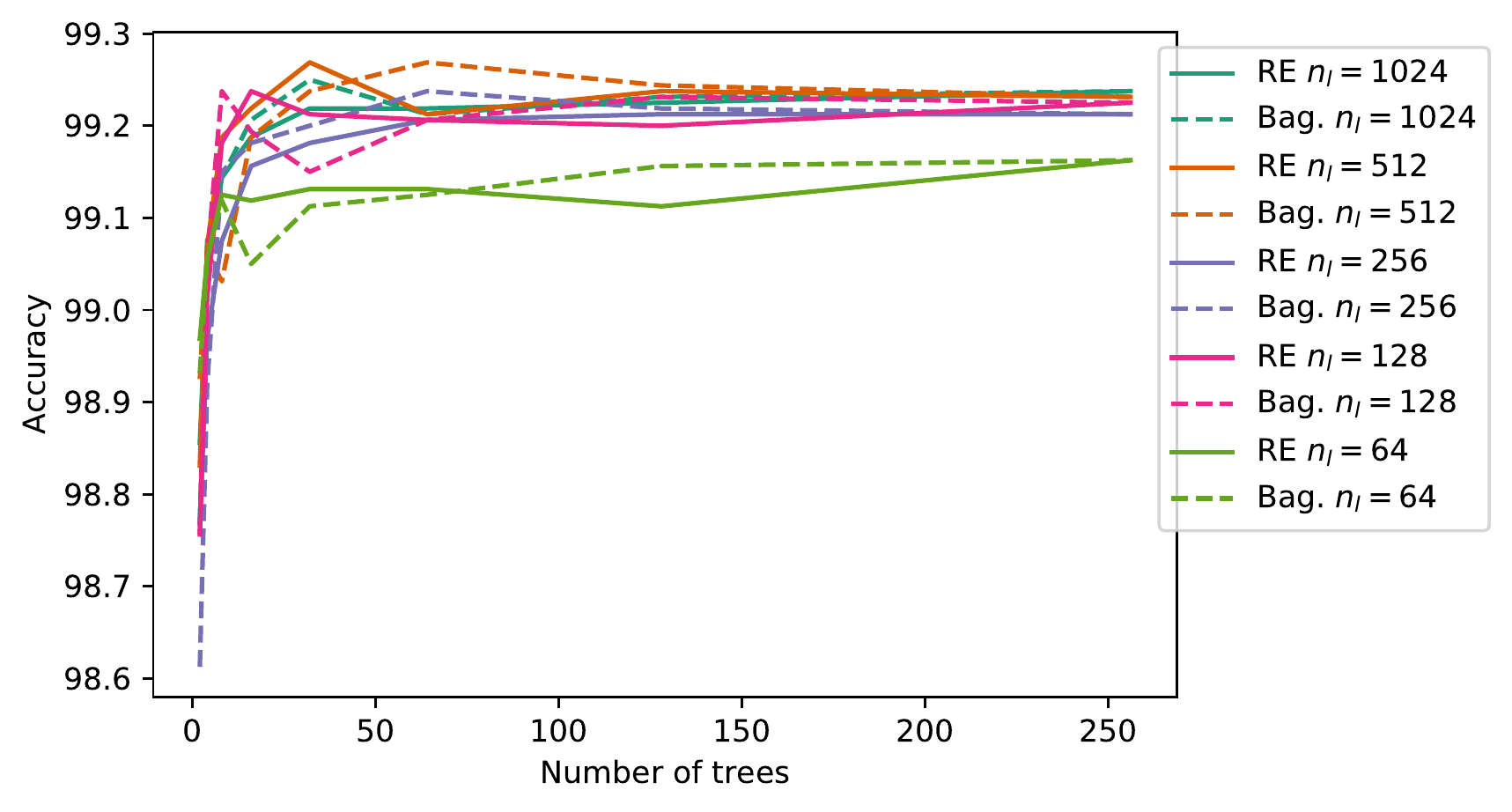}
\end{minipage}\hfill
\begin{minipage}{.49\textwidth}
    \centering 
    \resizebox{\textwidth}{!}{
        \input{figures/BaggingClassifier_ida2016_table}
    }
\end{minipage}
\caption{(Left) The error over the number of trees in the ensemble on the ida2016 dataset. Dashed lines depict the Random Forest and solid lines are the corresponding pruned ensemble via Reduced Error pruning. (Right) The 5-fold cross-validation accuracy  on the ida2016 dataset. Rounded to the second decimal digit. Larger is better. }
\end{figure}

\begin{figure}[H]
\begin{minipage}{.49\textwidth}
    \centering
    \includegraphics[width=\textwidth,keepaspectratio]{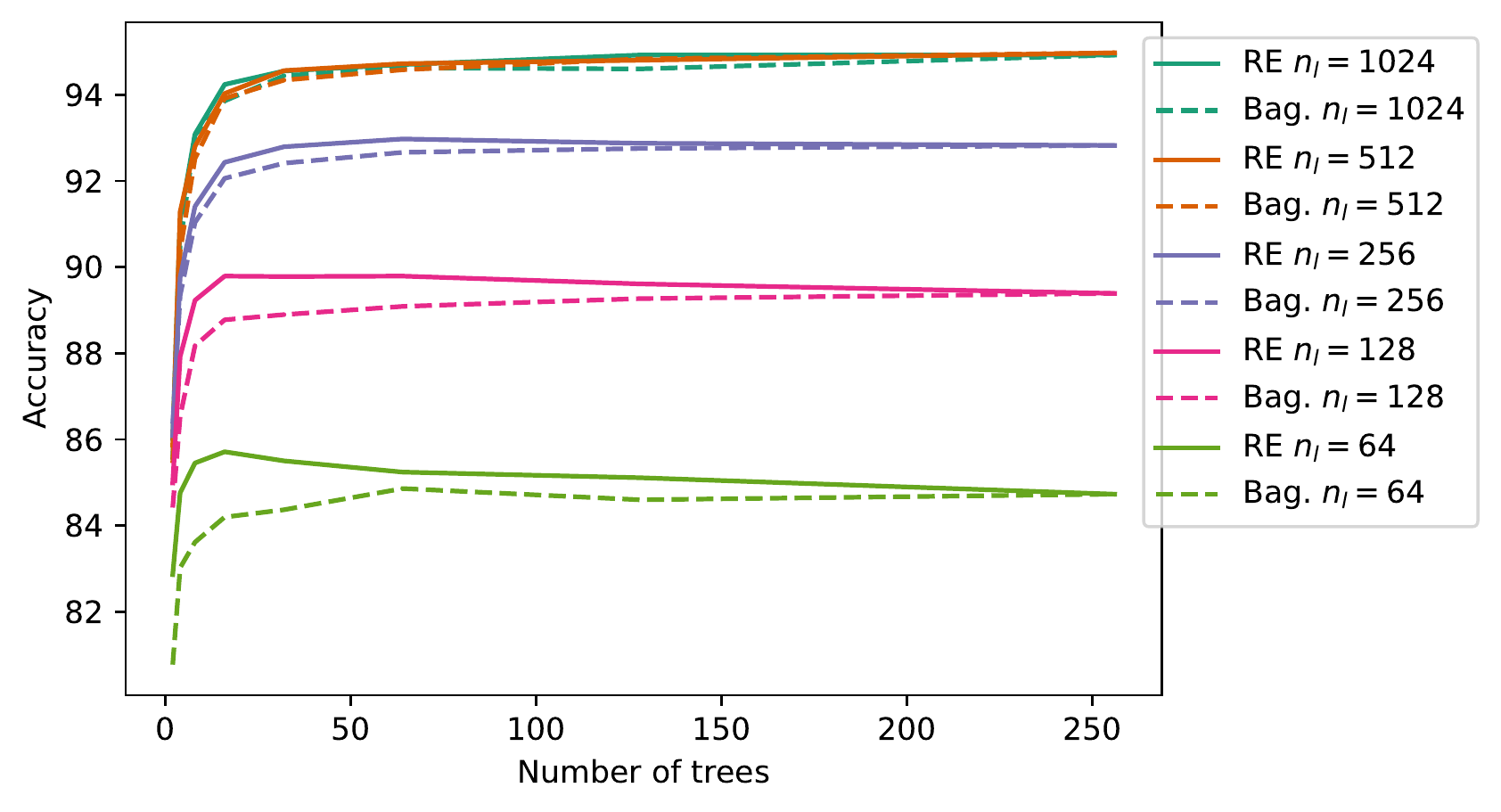}
\end{minipage}\hfill
\begin{minipage}{.49\textwidth}
    \centering 
    \resizebox{\textwidth}{!}{
        \input{figures/BaggingClassifier_japanese-vowels_table}
    }
\end{minipage}
\caption{(Left) The error over the number of trees in the ensemble on the japanese-vowels dataset. Dashed lines depict the Random Forest and solid lines are the corresponding pruned ensemble via Reduced Error pruning. (Right) The 5-fold cross-validation accuracy  on the japanese-vowels dataset. Rounded to the second decimal digit. Larger is better.}
\end{figure}

\begin{figure}[H]
\begin{minipage}{.49\textwidth}
    \centering
    \includegraphics[width=\textwidth,keepaspectratio]{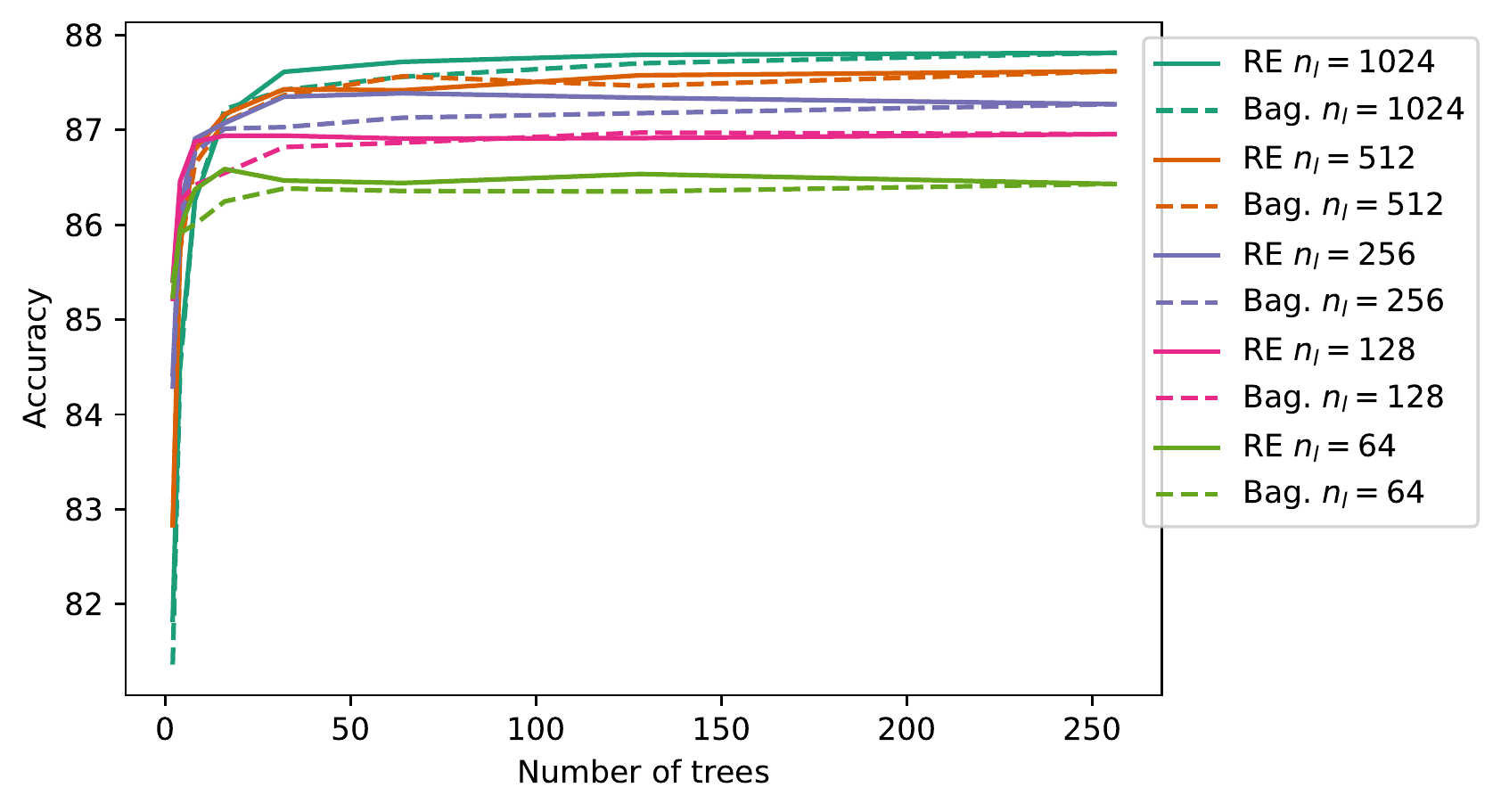}
\end{minipage}\hfill
\begin{minipage}{.49\textwidth}
    \centering 
    \resizebox{\textwidth}{!}{
        \input{figures/BaggingClassifier_magic_table}
    }
\end{minipage}
\caption{(Left) The error over the number of trees in the ensemble on the magic dataset. Dashed lines depict the Random Forest and solid lines are the corresponding pruned ensemble via Reduced Error pruning. (Right) The 5-fold cross-validation accuracy  on the magic dataset. Rounded to the second decimal digit. Larger is better}
\end{figure}

\begin{figure}[H]
\begin{minipage}{.49\textwidth}
    \centering
    \includegraphics[width=\textwidth,keepaspectratio]{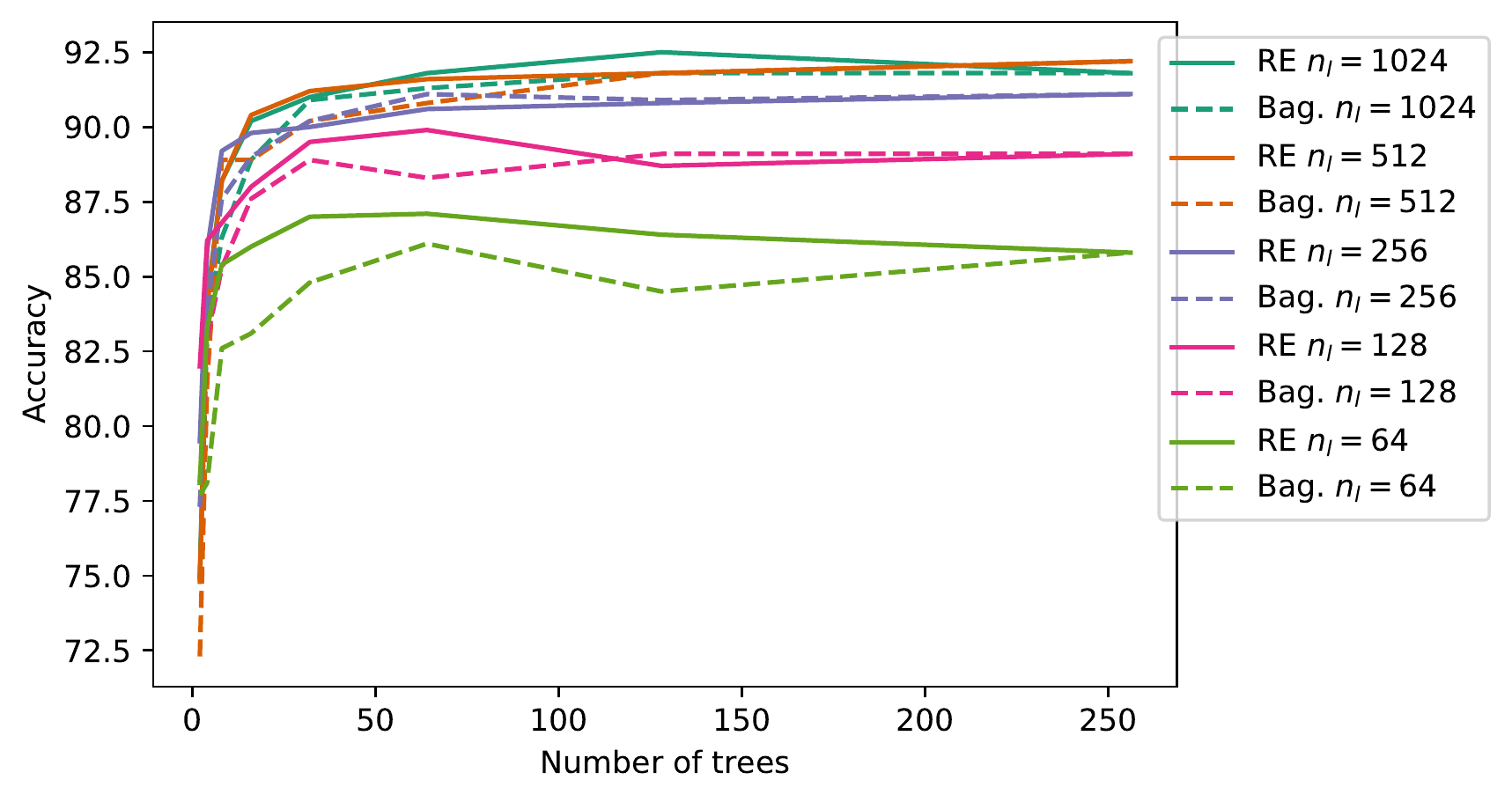}
\end{minipage}\hfill
\begin{minipage}{.49\textwidth}
    \centering 
    \resizebox{\textwidth}{!}{
        \input{figures/BaggingClassifier_mnist_table}
    }
\end{minipage}
\caption{(Left) The error over the number of trees in the ensemble on the mnist dataset. Dashed lines depict the Random Forest and solid lines are the corresponding pruned ensemble via Reduced Error pruning. (Right) The 5-fold cross-validation accuracy  on the mnist dataset. Rounded to the second decimal digit. Larger is better.}
\end{figure}

\begin{figure}[H]
\begin{minipage}{.49\textwidth}
    \centering
    \includegraphics[width=\textwidth,keepaspectratio]{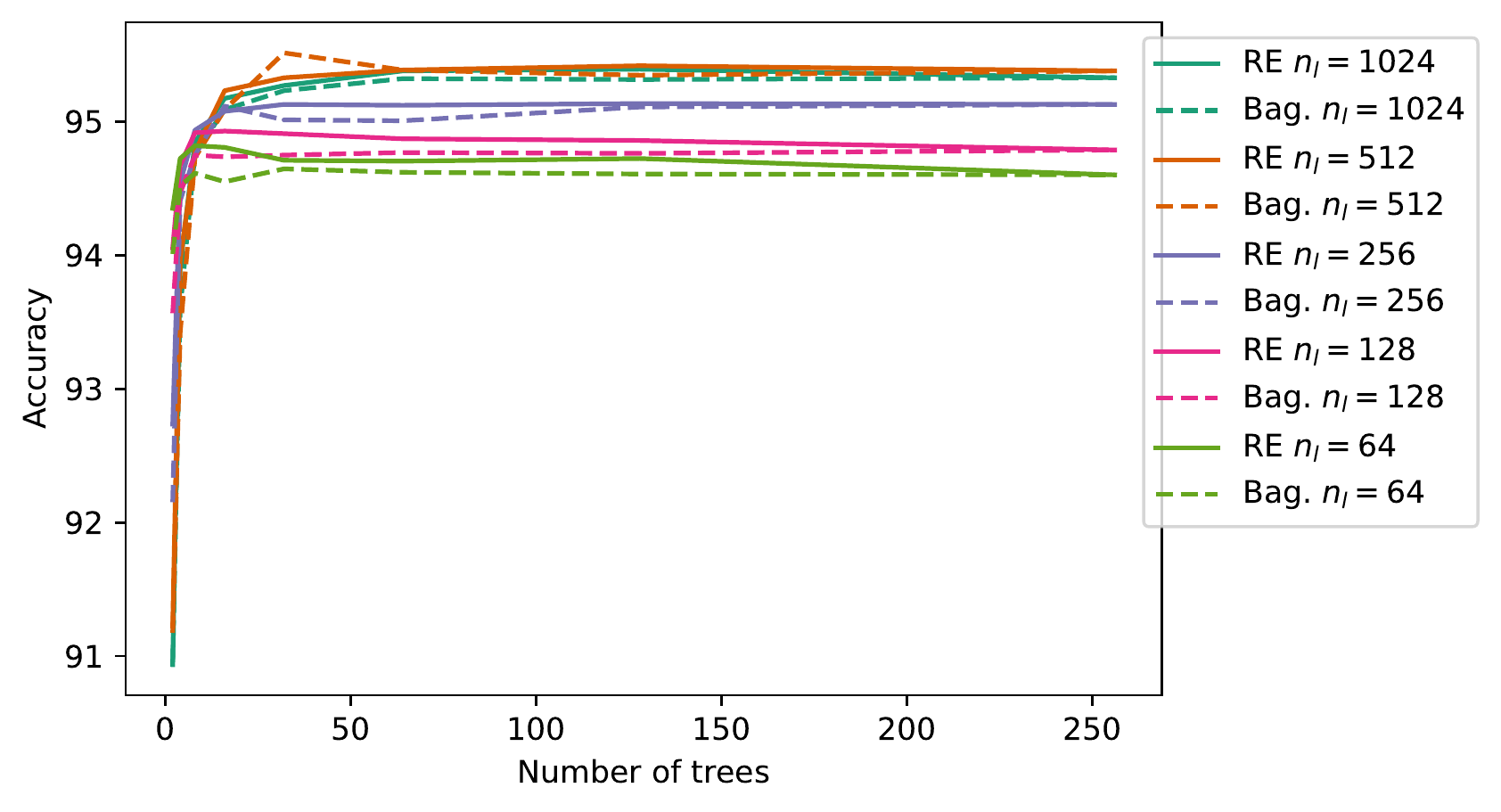}
\end{minipage}\hfill
\begin{minipage}{.49\textwidth}
    \centering 
    \resizebox{\textwidth}{!}{
        \input{figures/BaggingClassifier_mozilla_table}
    }
\end{minipage}
\caption{(Left) The error over the number of trees in the ensemble on the mozilla dataset. Dashed lines depict the Random Forest and solid lines are the corresponding pruned ensemble via Reduced Error pruning. (Right) The 5-fold cross-validation accuracy  on the mozilla dataset. Rounded to the second decimal digit. Larger is better.}
\end{figure}

\begin{figure}[H]
\begin{minipage}{.49\textwidth}
    \centering
    \includegraphics[width=\textwidth,keepaspectratio]{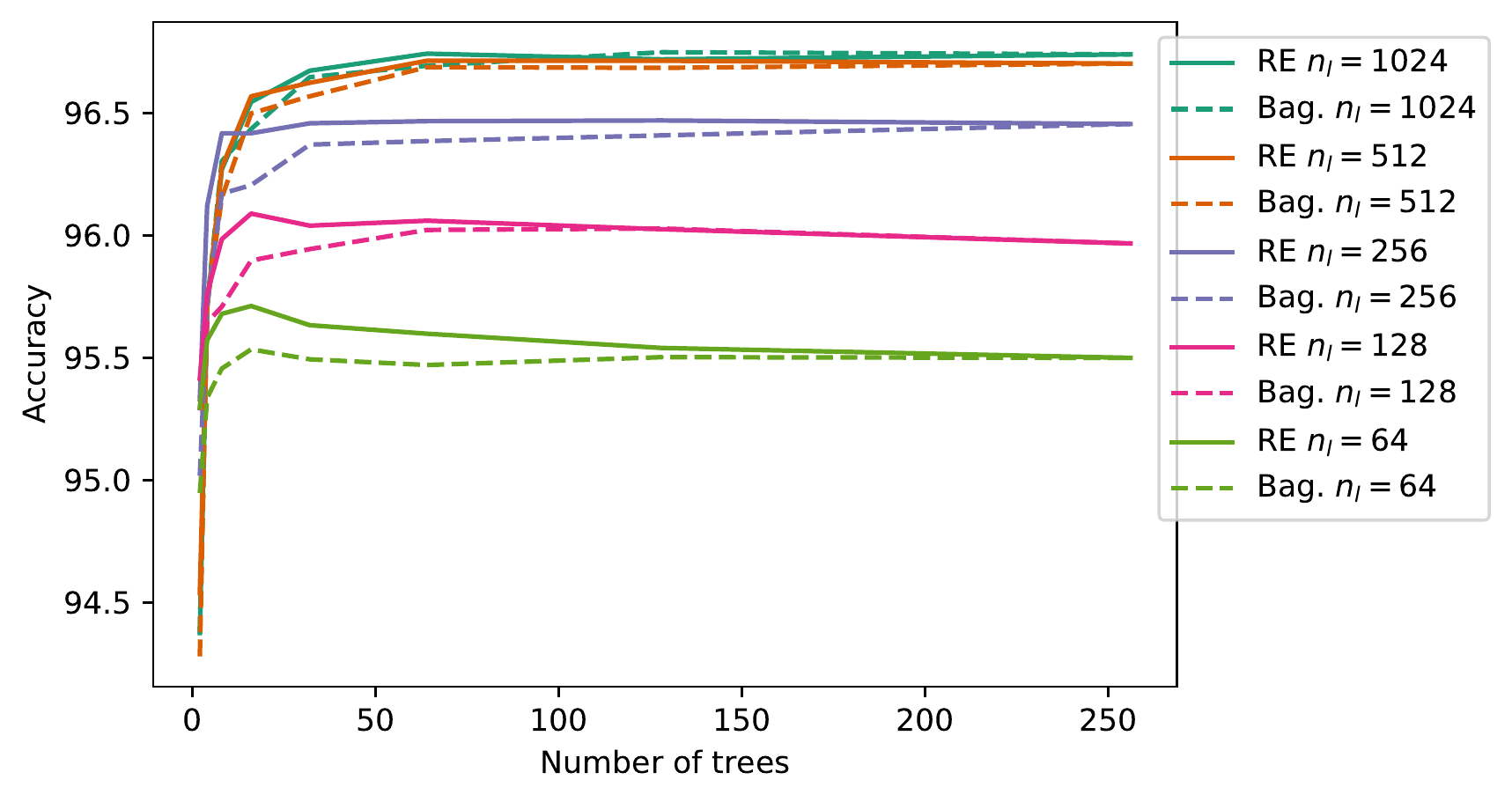}
\end{minipage}\hfill
\begin{minipage}{.49\textwidth}
    \centering 
    \resizebox{\textwidth}{!}{
        \input{figures/BaggingClassifier_nomao_table}
    }
\end{minipage}
\caption{(Left) The error over the number of trees in the ensemble on the nomao dataset. Dashed lines depict the Random Forest and solid lines are the corresponding pruned ensemble via Reduced Error pruning. (Right) The 5-fold cross-validation accuracy  on the nomao dataset. Rounded to the second decimal digit. Larger is better.}
\end{figure}

\begin{figure}[H]
\begin{minipage}{.49\textwidth}
    \centering
    \includegraphics[width=\textwidth,keepaspectratio]{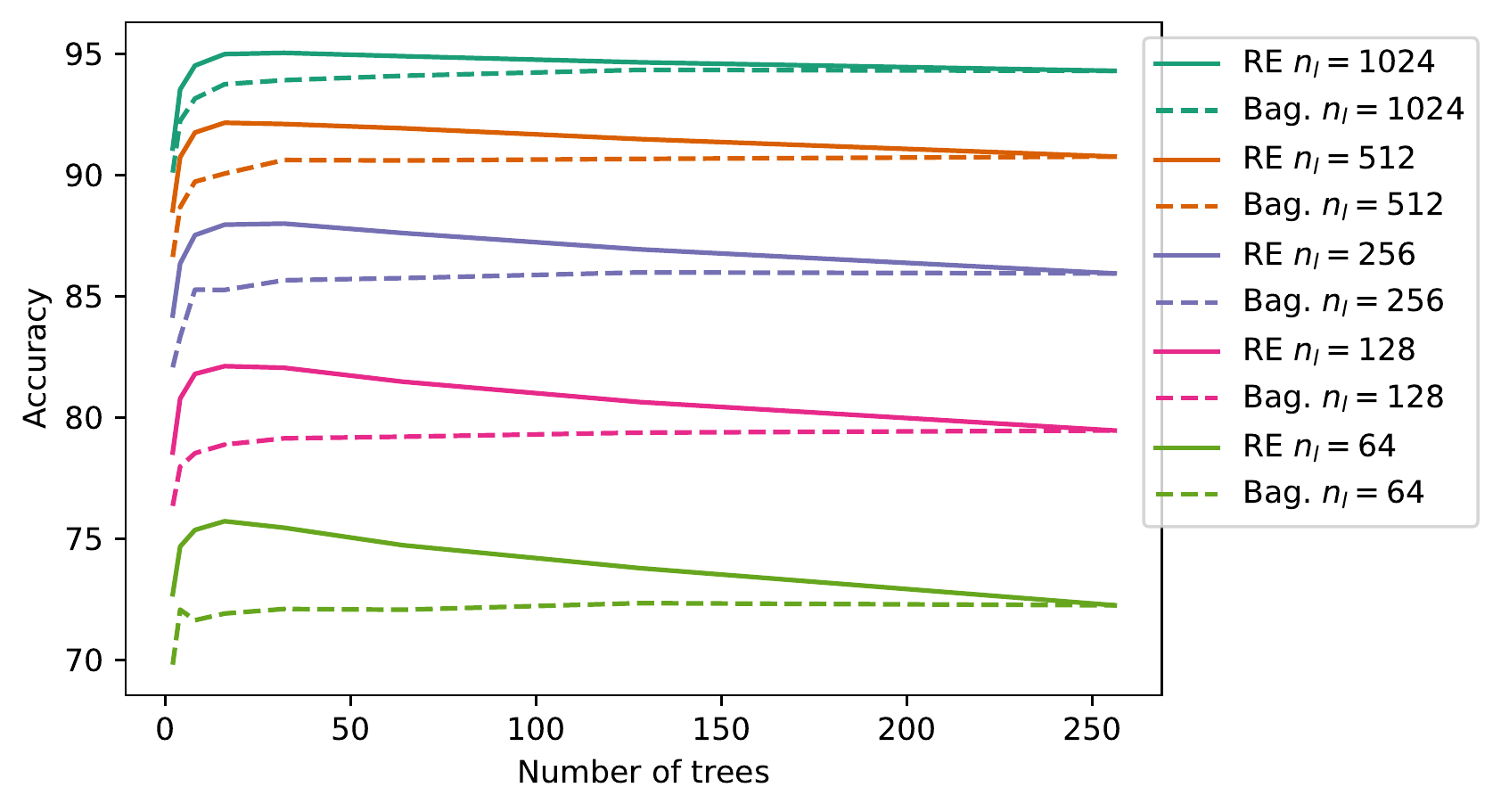}
\end{minipage}\hfill
\begin{minipage}{.49\textwidth}
    \centering 
    \resizebox{\textwidth}{!}{
        \input{figures/BaggingClassifier_postures_table}
    }
\end{minipage}
\caption{(Left) The error over the number of trees in the ensemble on the postures dataset. Dashed lines depict the Random Forest and solid lines are the corresponding pruned ensemble via Reduced Error pruning. (Right) The 5-fold cross-validation accuracy  on the postures dataset. Rounded to the second decimal digit. Larger is better.}
\end{figure}

\begin{figure}[H]
\begin{minipage}{.49\textwidth}
    \centering
    \includegraphics[width=\textwidth,keepaspectratio]{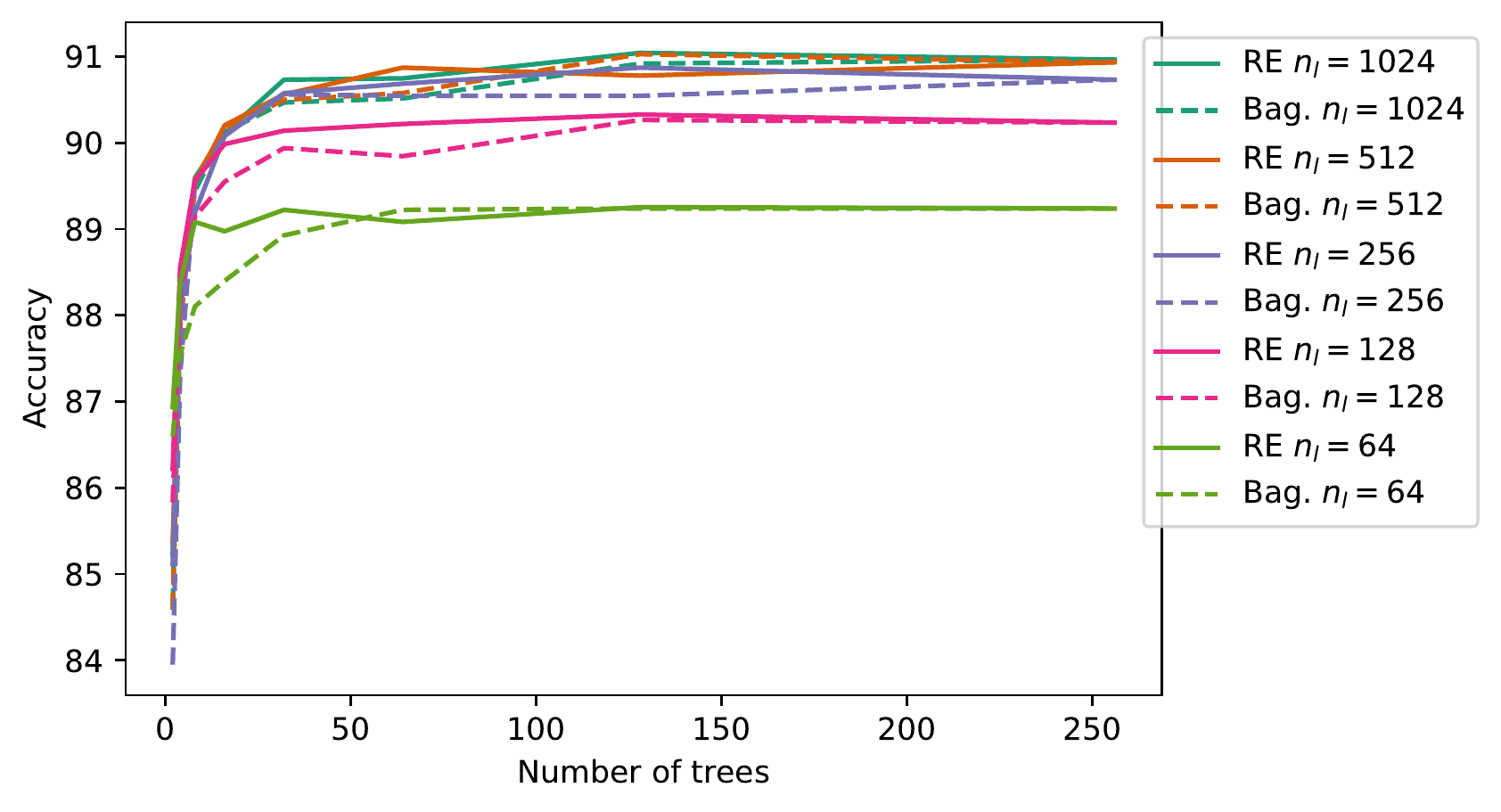}
\end{minipage}\hfill
\begin{minipage}{.49\textwidth}
    \centering 
    \resizebox{\textwidth}{!}{
        \input{figures/BaggingClassifier_satimage_table}
    }
\end{minipage}
\caption{(Left) The error over the number of trees in the ensemble on the satimage dataset. Dashed lines depict the Random Forest and solid lines are the corresponding pruned ensemble via Reduced Error pruning. (Right) The 5-fold cross-validation accuracy  on the satimage dataset. Rounded to the second decimal digit. Larger is better.}
\end{figure}

\subsection{Accuracies under various resource constraints with a BaggingClassifier}

\begin{table}
    \centering
    \resizebox{\textwidth}{!}{
    \input{figures/raw_BaggingClassifier_32}
    }
    \caption{Test accuracies for models with a memory consumption below 32 KB for each method and each dataset averaged over a 5 fold cross validation. Rounded to the third decimal digit. Larger is better. The best method is depicted in bold.}
\end{table}

\begin{table}
    \centering
    \resizebox{\textwidth}{!}{
    \input{figures/raw_BaggingClassifier_64}
    }
    \caption{Test accuracies for models with a memory consumption below 64 KB for each method and each dataset averaged over a 5 fold cross validation. Rounded to the third decimal digit. Larger is better. The best method is depicted in bold.}
\end{table}

\begin{table}
    \centering
    \resizebox{\textwidth}{!}{
    \input{figures/raw_BaggingClassifier_128}
    }
    \caption{Test accuracies for models with a memory consumption below 128 KB for each method and each dataset averaged over a 5 fold cross validation. Rounded to the third decimal digit. Larger is better. The best method is depicted in bold.}
\end{table}

\begin{table}
    \centering
    \resizebox{\textwidth}{!}{
    \input{figures/raw_BaggingClassifier_256}
    }
    \caption{Test accuracies for models with a memory consumption below 256 KB for each method and each dataset averaged over a 5 fold cross validation. Rounded to the third decimal digit. Larger is better. The best method is depicted in bold.}
\end{table}

\subsection{Plotting the Pareto Front For More Datasets with Dedicated Pruning Set}

\begin{figure}[H]
\begin{minipage}{.49\textwidth}
    \centering
    \includegraphics[width=\textwidth,keepaspectratio]{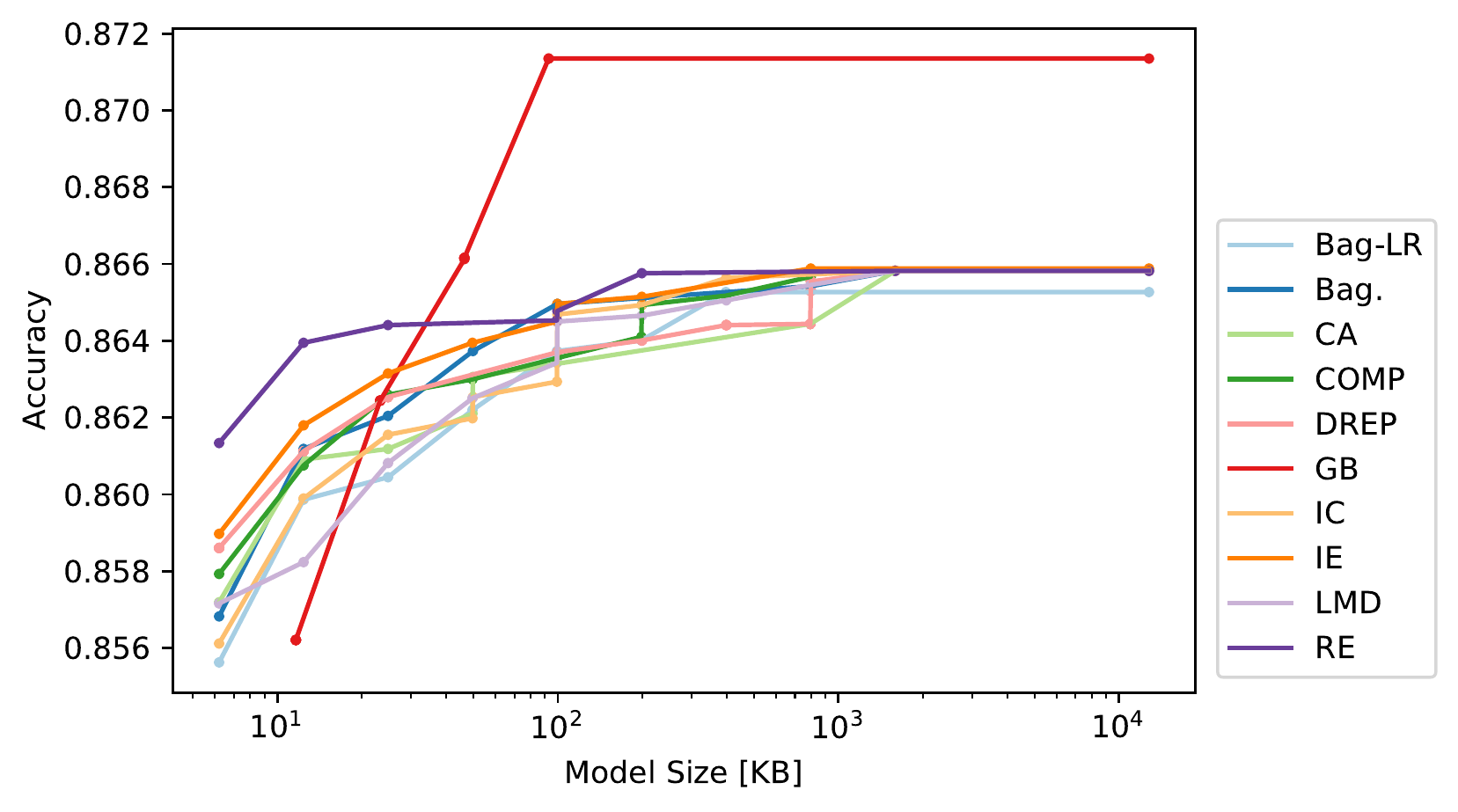}
\end{minipage}\hfill
\begin{minipage}{.49\textwidth}
    \centering 
    \includegraphics[width=\textwidth,keepaspectratio]{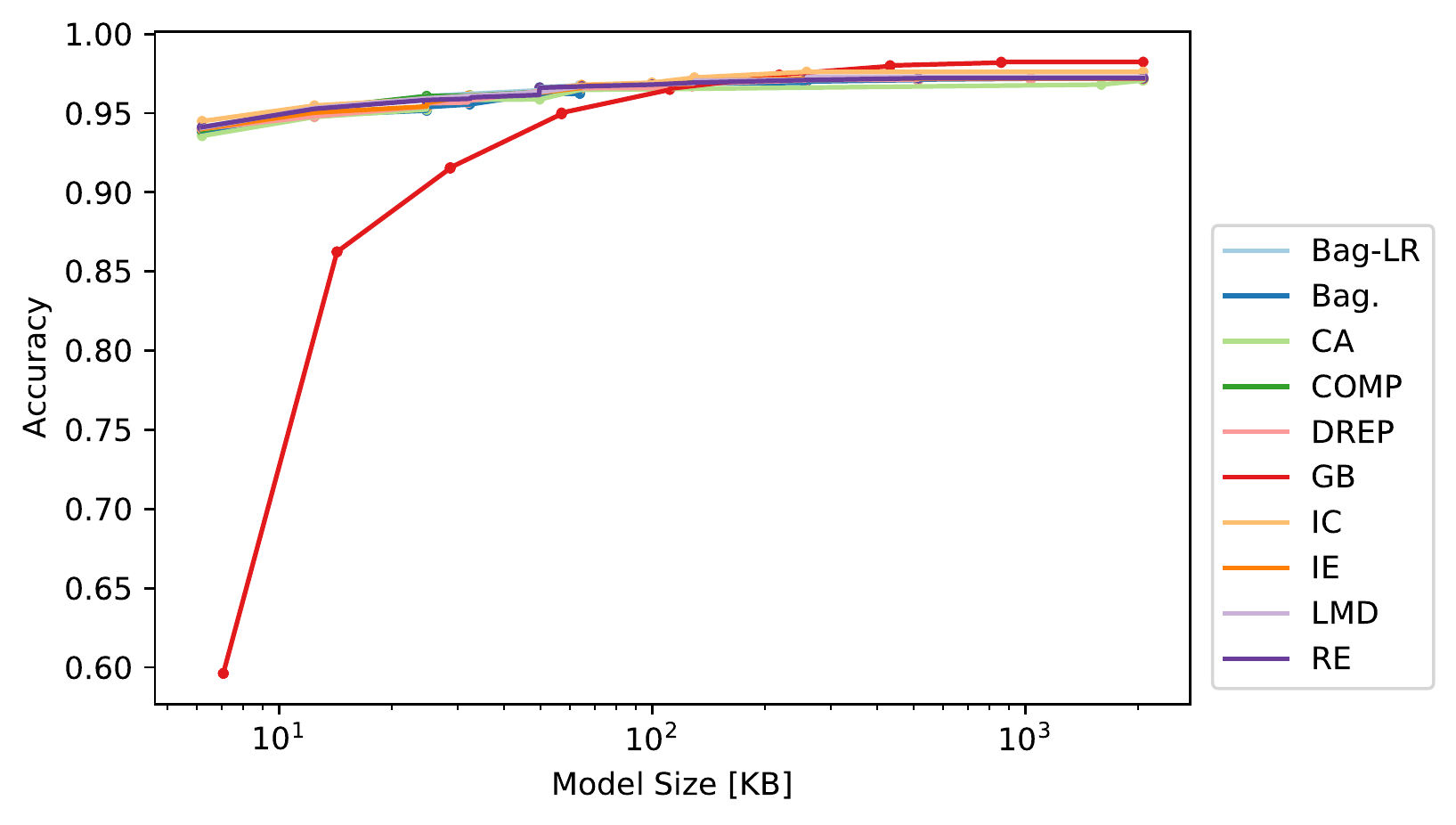}
\end{minipage}
\caption{(left) 5-fold cross-validation accuracy on the adult dataset. (right) 5-fold cross-validation accuracy on the anura dataset.}
\end{figure}

\begin{figure}[H]
\begin{minipage}{.49\textwidth}
    \centering
    \includegraphics[width=\textwidth,keepaspectratio]{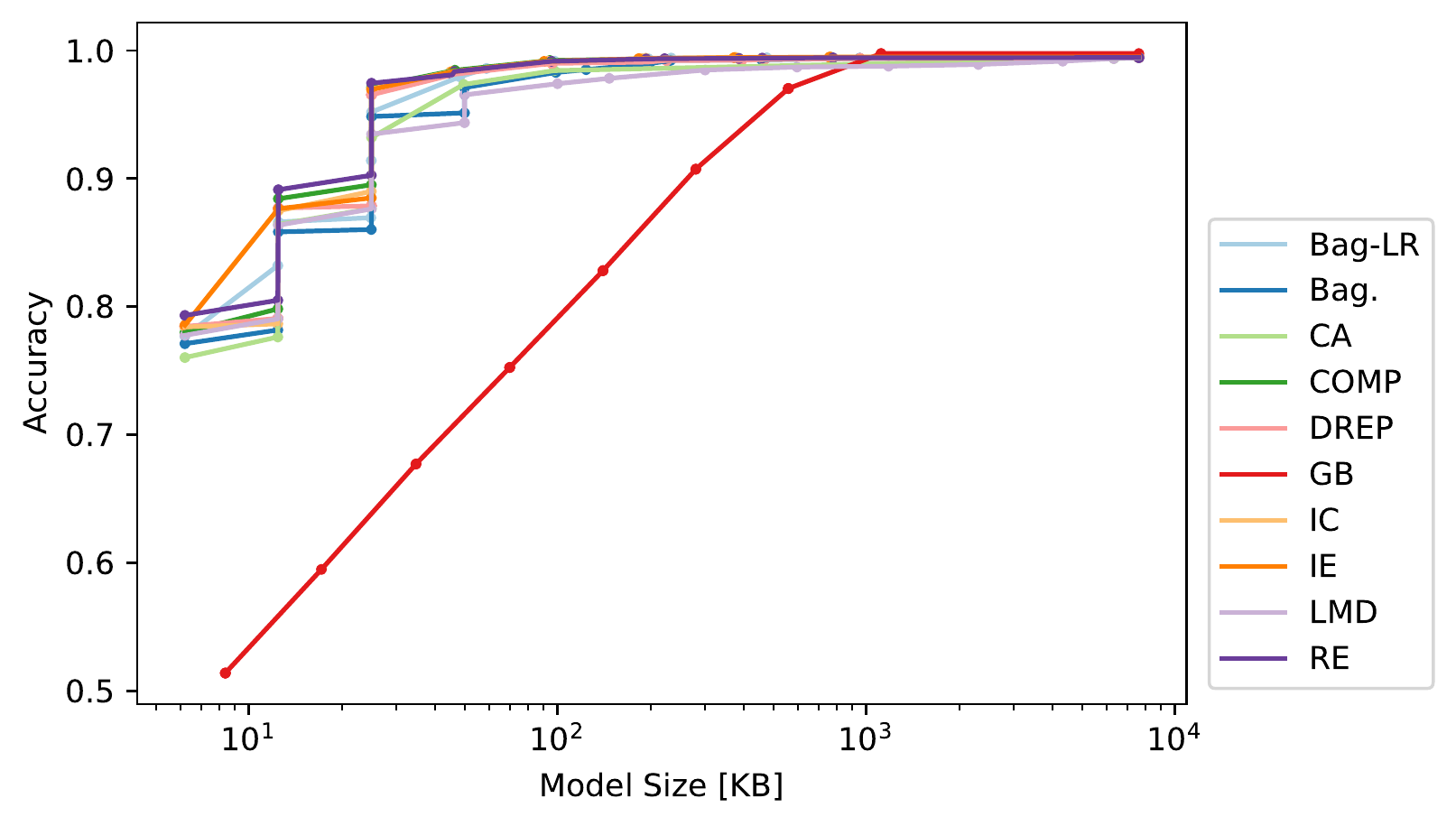}
\end{minipage}\hfill
\begin{minipage}{.49\textwidth}
    \centering 
    \includegraphics[width=\textwidth,keepaspectratio]{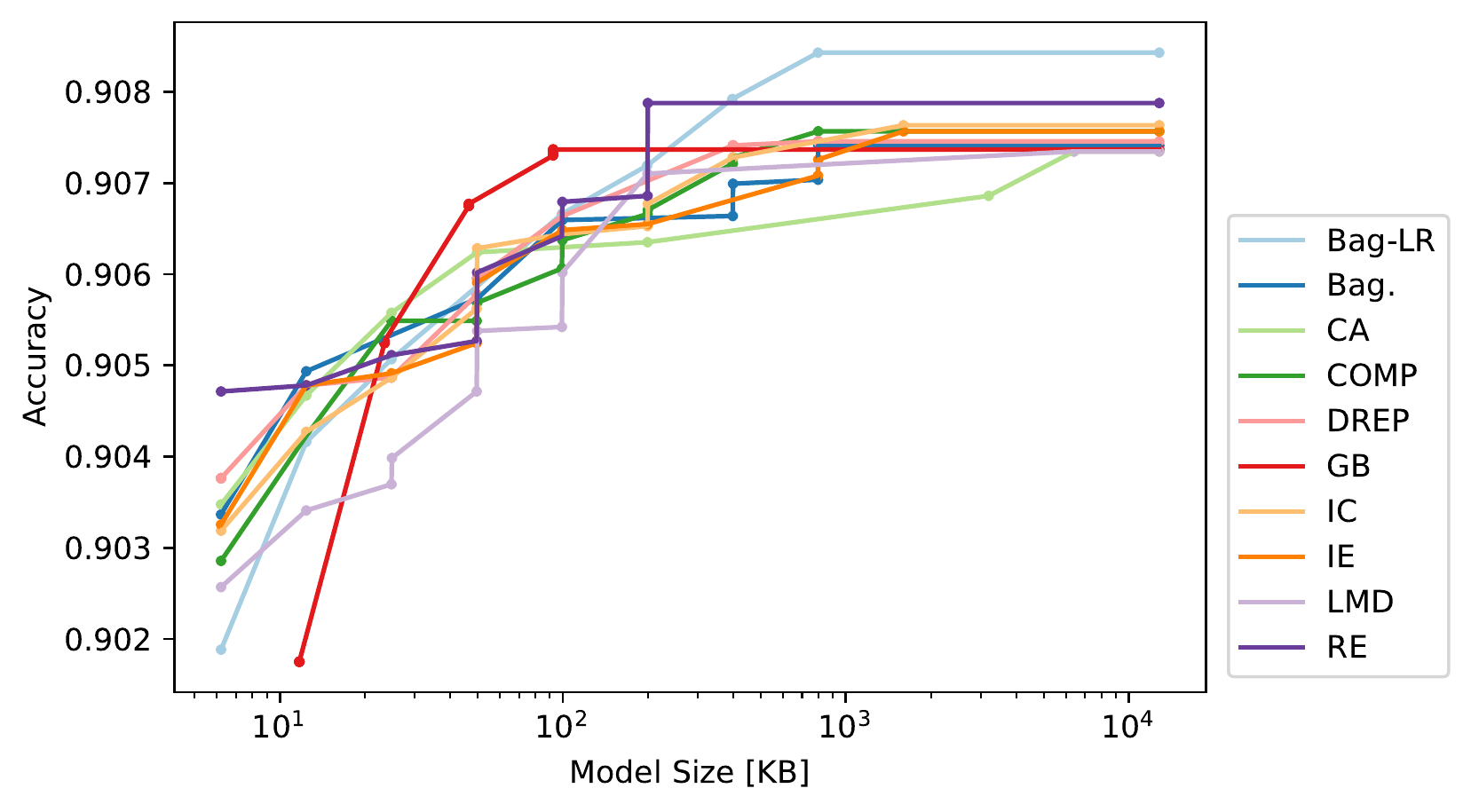}
\end{minipage}
\caption{(left) 5-fold cross-validation accuracy on the avila dataset. (right) 5-fold cross-validation accuracy on the bank dataset.}
\end{figure}

\begin{figure}[H]
\begin{minipage}{.49\textwidth}
    \centering
    \includegraphics[width=\textwidth,keepaspectratio]{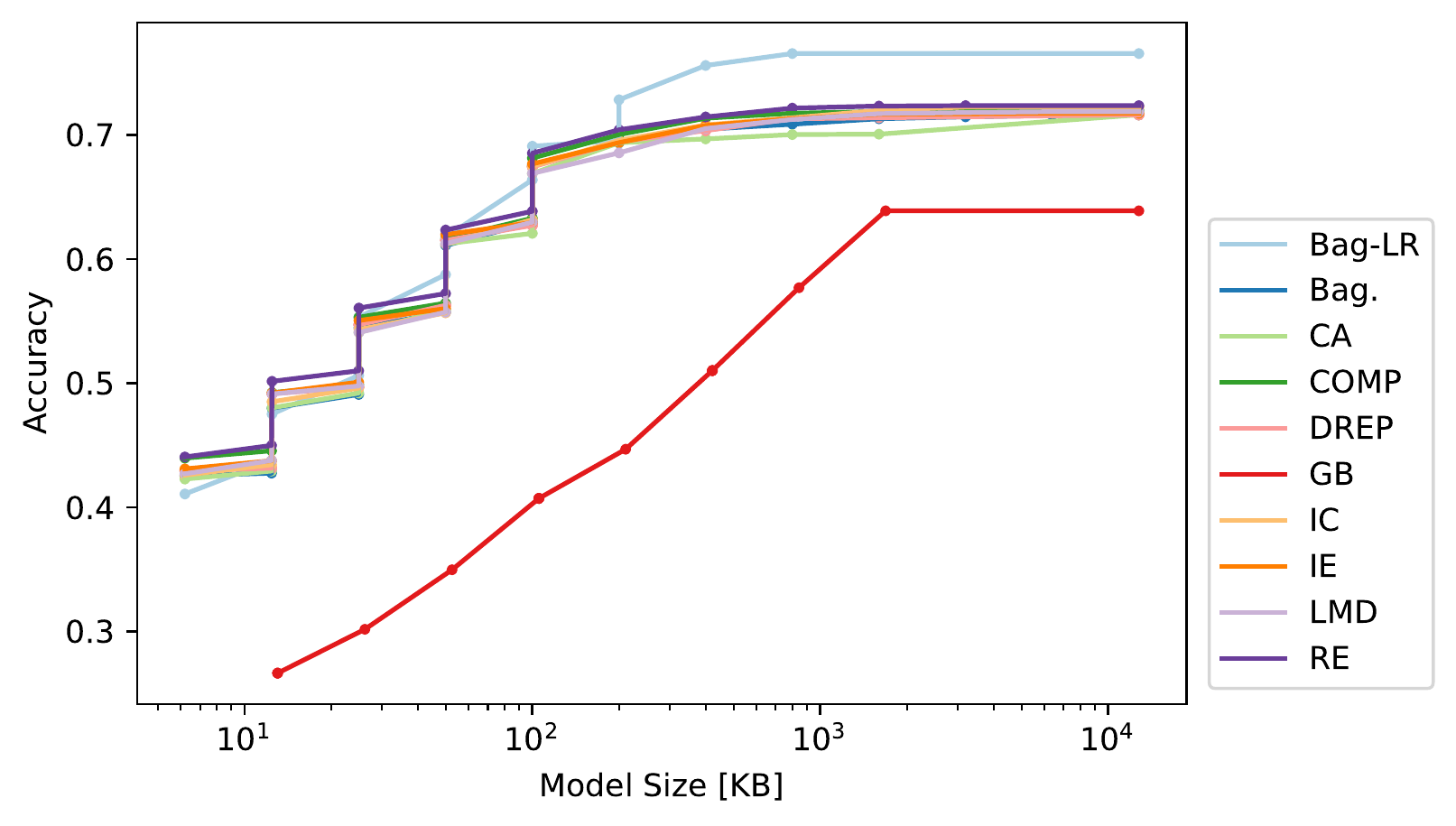}
\end{minipage}\hfill
\begin{minipage}{.49\textwidth}
    \centering 
    \includegraphics[width=\textwidth,keepaspectratio]{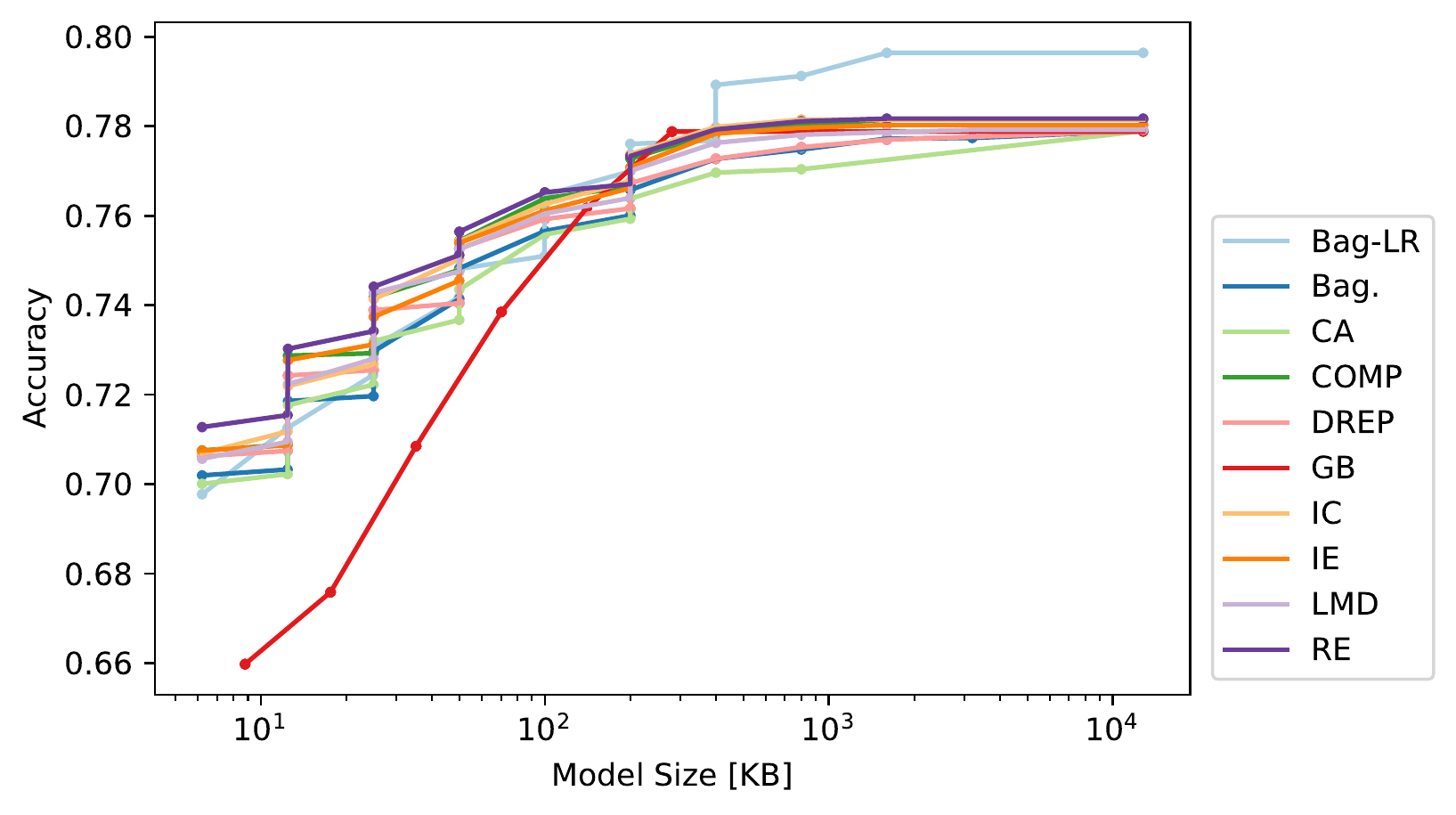}
\end{minipage}
\caption{(left) 5-fold cross-validation accuracy on the chess dataset. (right) 5-fold cross-validation accuracy on the connect dataset.}
\end{figure}

\begin{figure}[H]
\begin{minipage}{.49\textwidth}
    \centering
    \includegraphics[width=\textwidth,keepaspectratio]{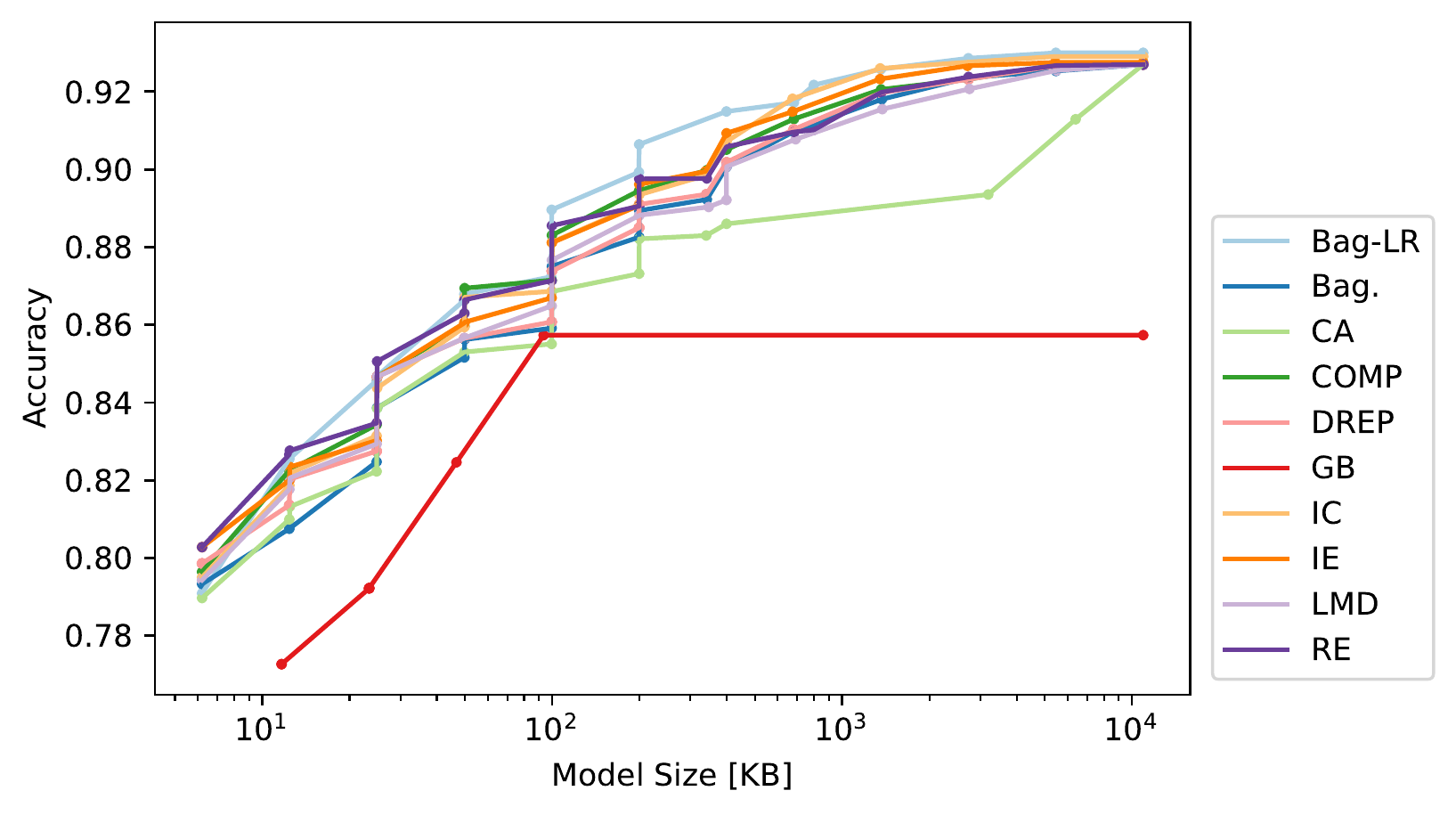}
\end{minipage}\hfill
\begin{minipage}{.49\textwidth}
    \centering 
    \includegraphics[width=\textwidth,keepaspectratio]{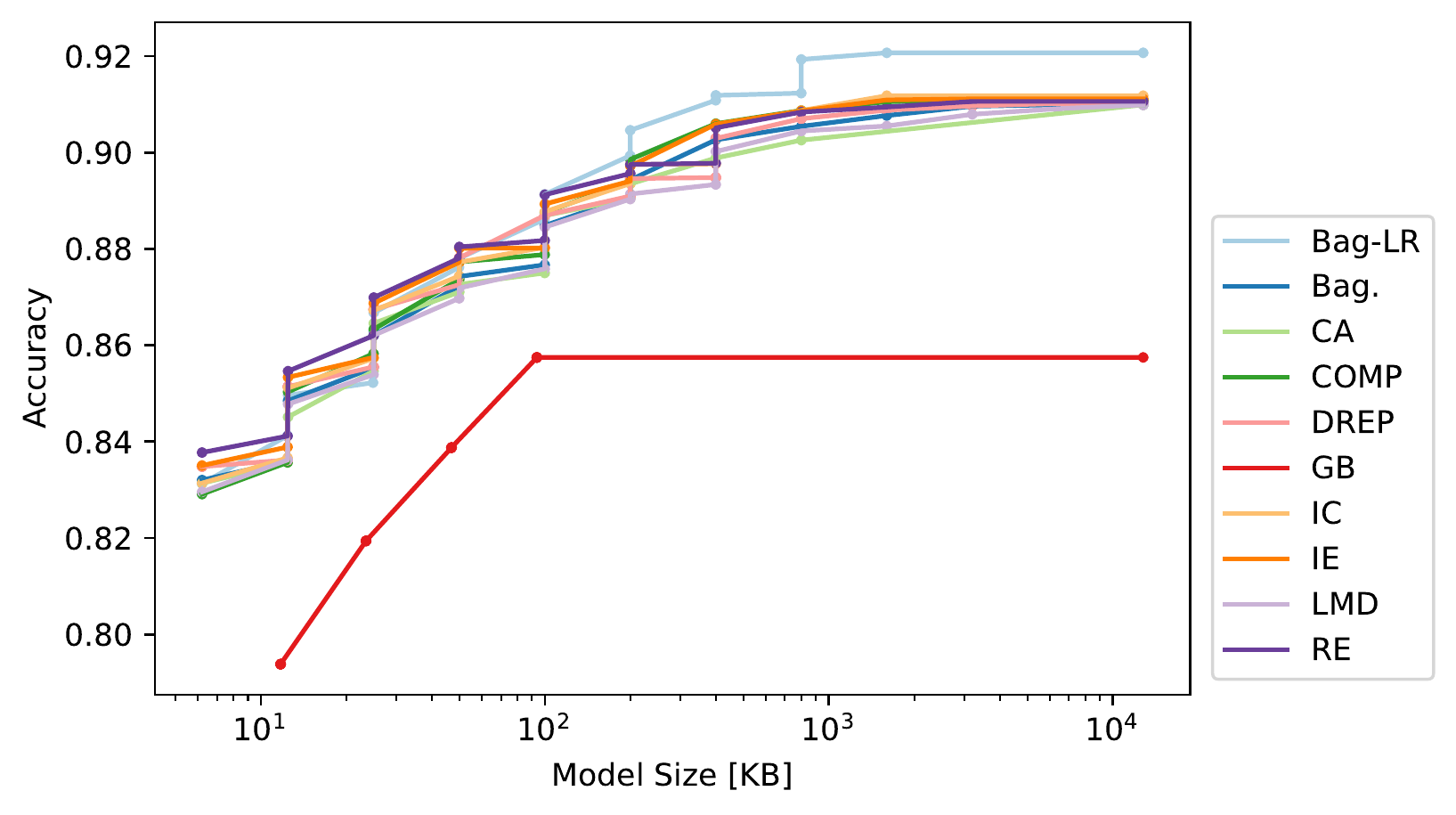}
\end{minipage}
\caption{(left) 5-fold cross-validation accuracy on the eeg dataset. (right) 5-fold cross-validation accuracy on the elec dataset.}
\end{figure}

\begin{figure}[H]
\begin{minipage}{.49\textwidth}
    \centering
    \includegraphics[width=\textwidth,keepaspectratio]{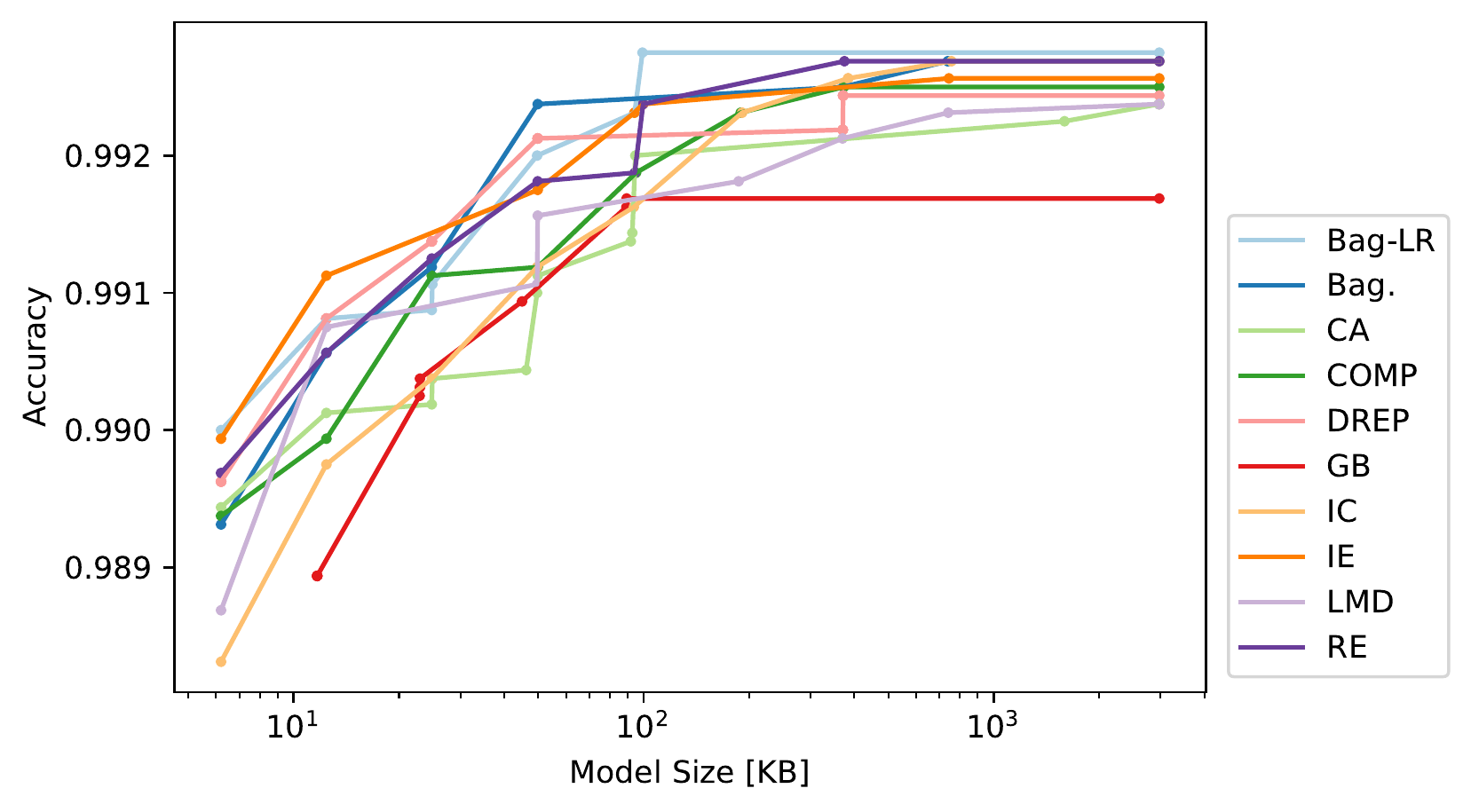}
\end{minipage}\hfill
\begin{minipage}{.49\textwidth}
    \centering 
    \includegraphics[width=\textwidth,keepaspectratio]{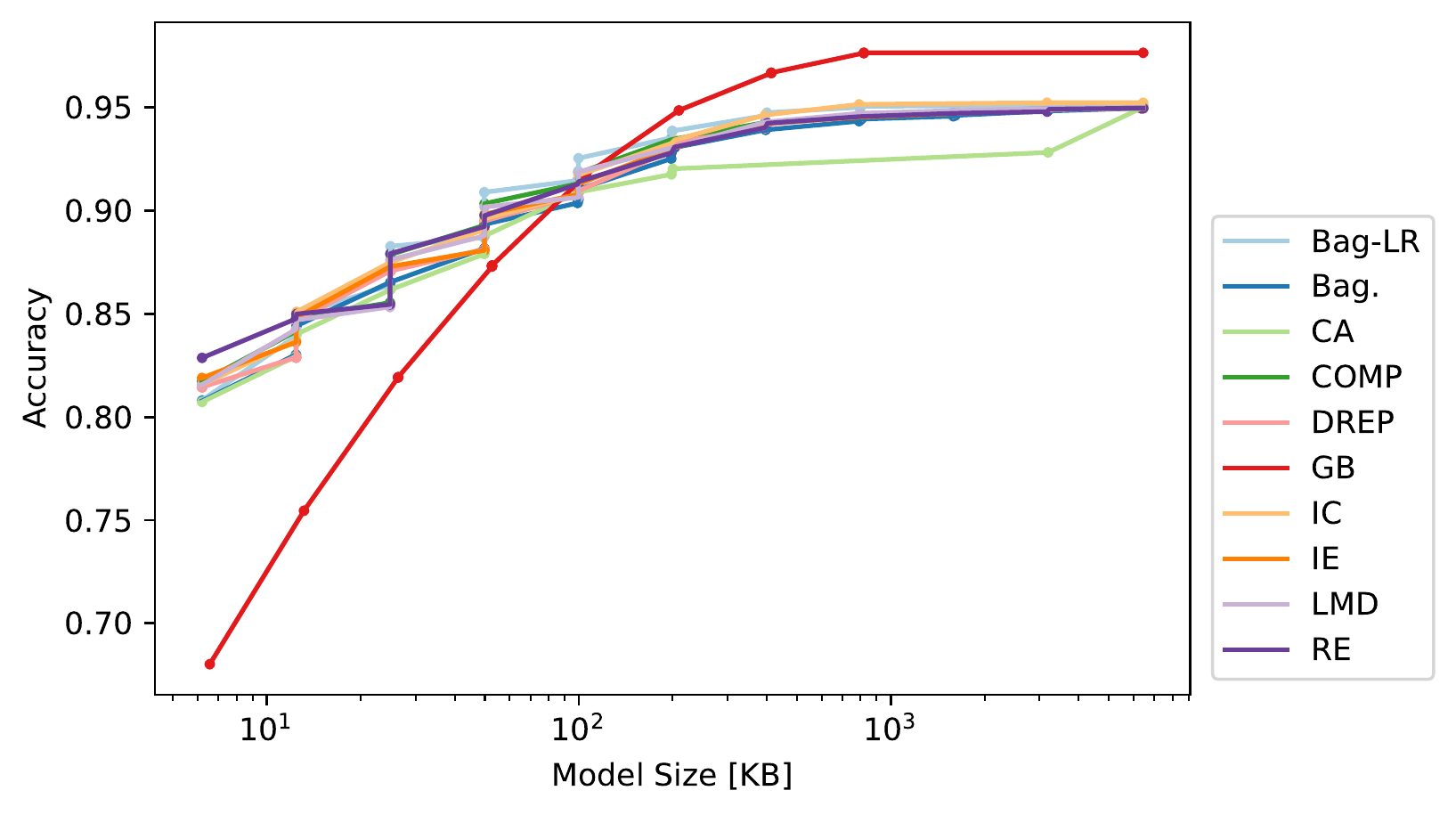}
\end{minipage}
\caption{(left) 5-fold cross-validation accuracy on the ida2016 dataset. (right) 5-fold cross-validation accuracy on the japanese-vowels dataset.}
\end{figure}

\begin{figure}[H]
\begin{minipage}{.49\textwidth}
    \centering
    \includegraphics[width=\textwidth,keepaspectratio]{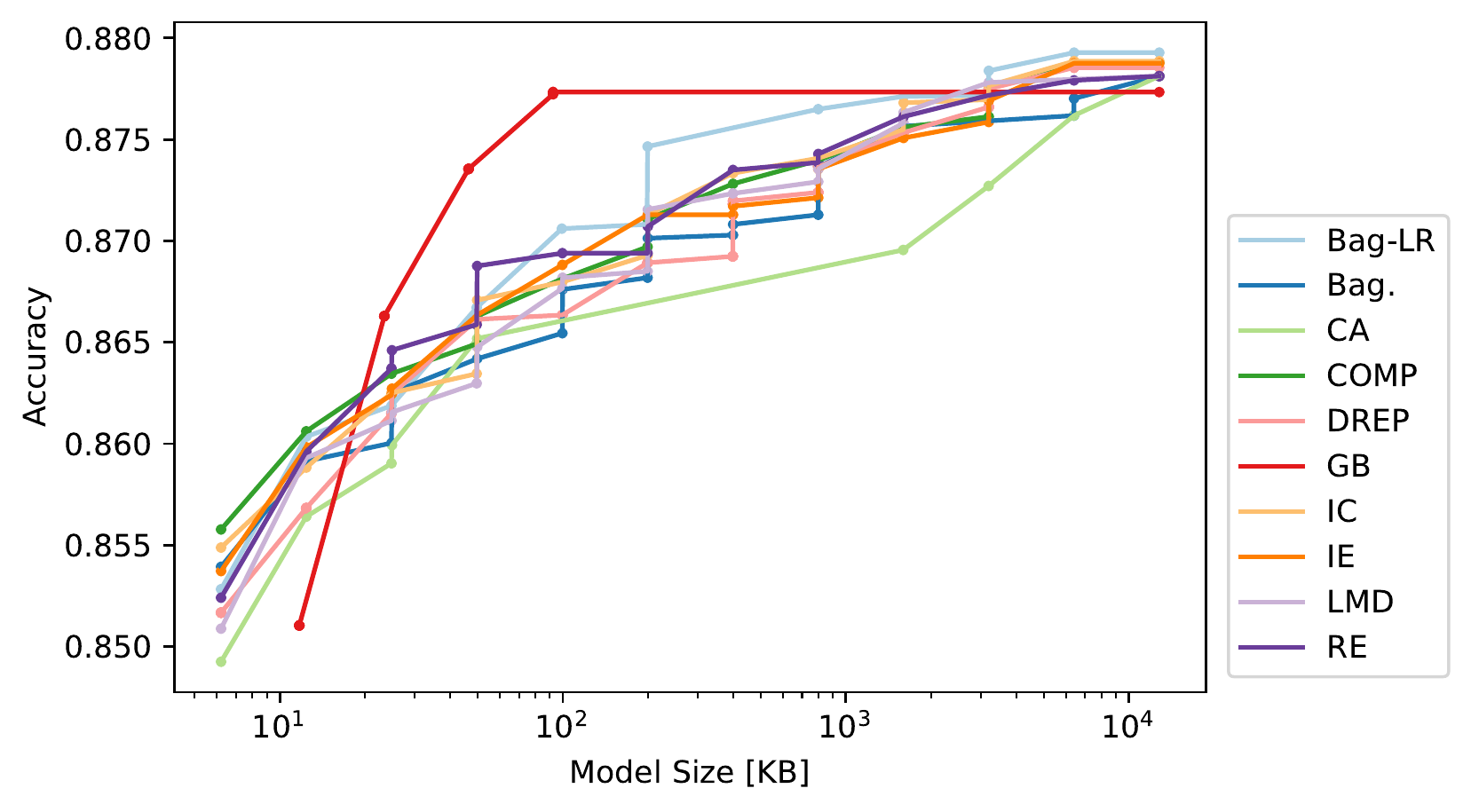}
\end{minipage}\hfill
\begin{minipage}{.49\textwidth}
    \centering 
    \includegraphics[width=\textwidth,keepaspectratio]{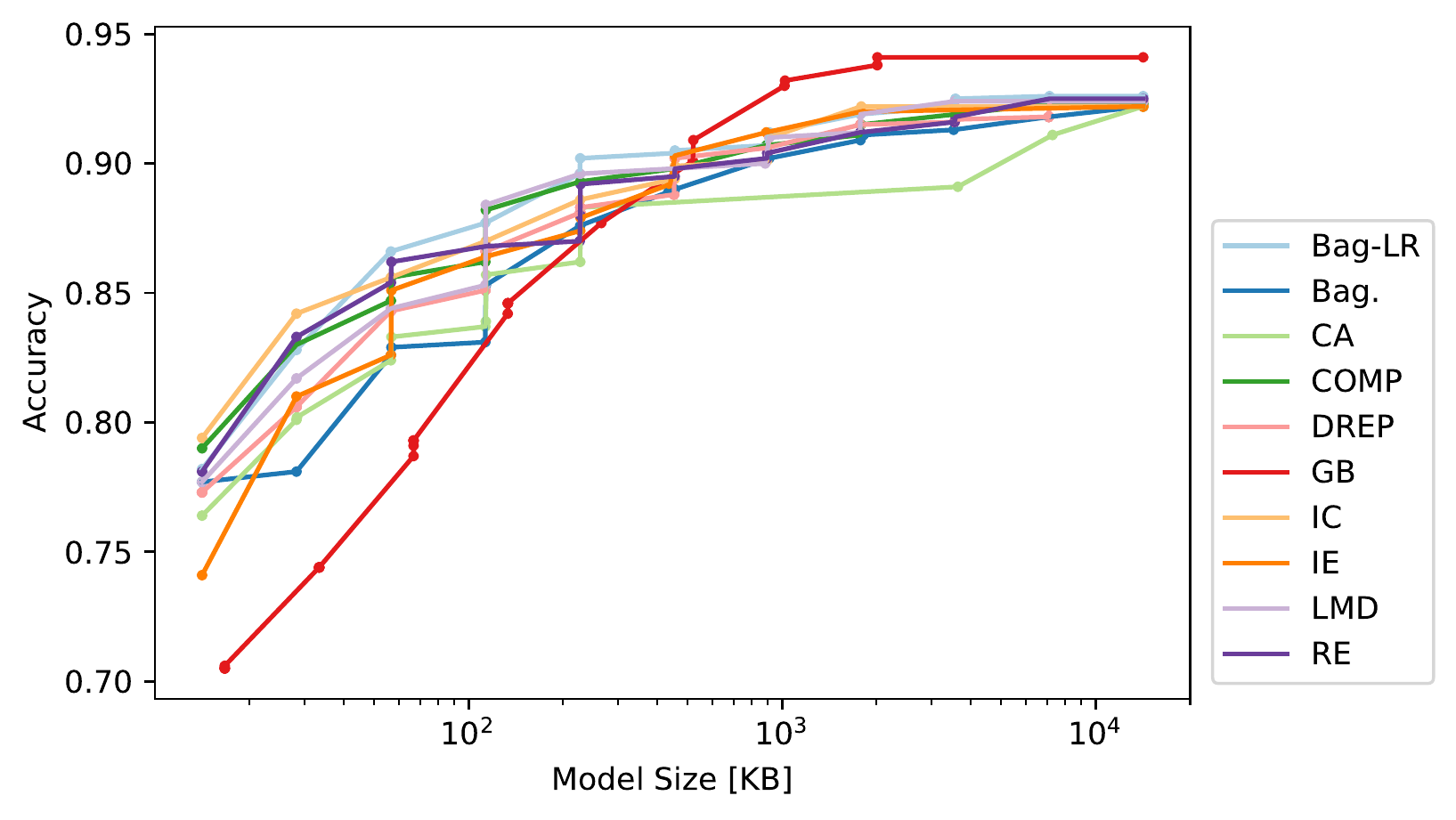}
\end{minipage}
\caption{(left) 5-fold cross-validation accuracy on the magic dataset. (right) 5-fold cross-validation accuracy on the mnist dataset.}
\end{figure}

\begin{figure}[H]
\begin{minipage}{.49\textwidth}
    \centering
    \includegraphics[width=\textwidth,keepaspectratio]{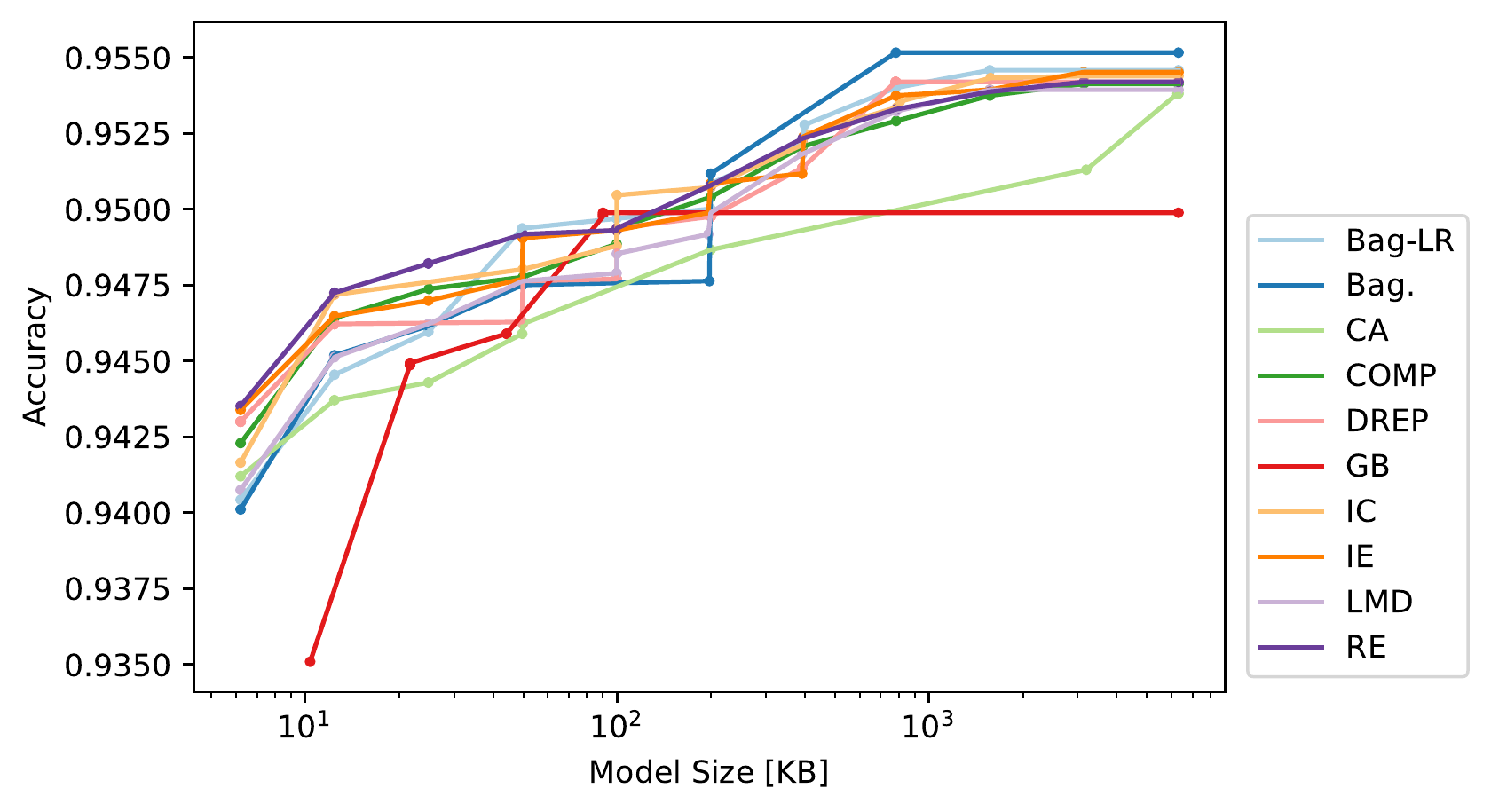}
\end{minipage}\hfill
\begin{minipage}{.49\textwidth}
    \centering 
    \includegraphics[width=\textwidth,keepaspectratio]{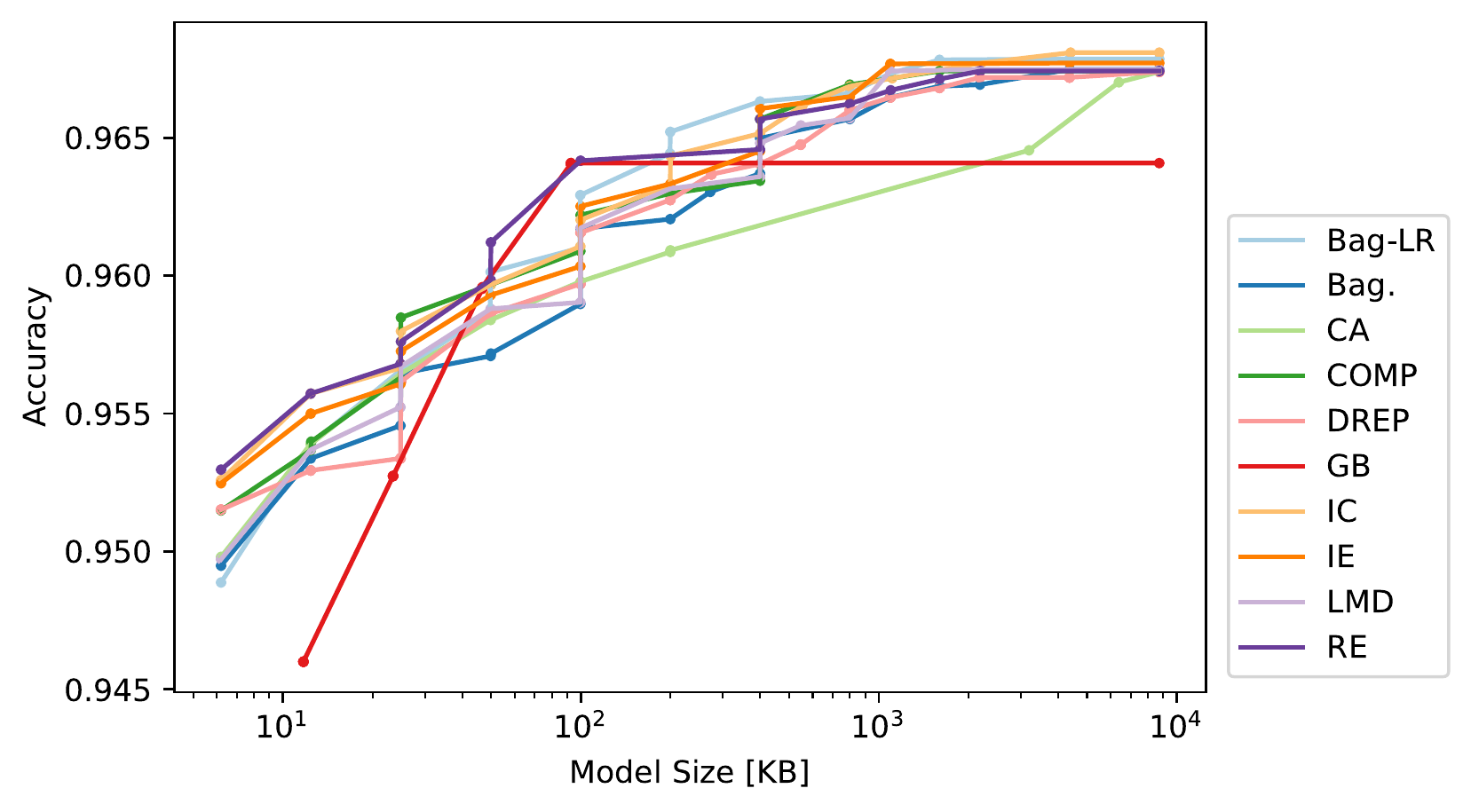}
\end{minipage}
\caption{(left) 5-fold cross-validation accuracy on the mozilla dataset. (right) 5-fold cross-validation accuracy on the nomao dataset.}
\end{figure}

\begin{figure}[H]
\begin{minipage}{.49\textwidth}
    \centering
    \includegraphics[width=\textwidth,keepaspectratio]{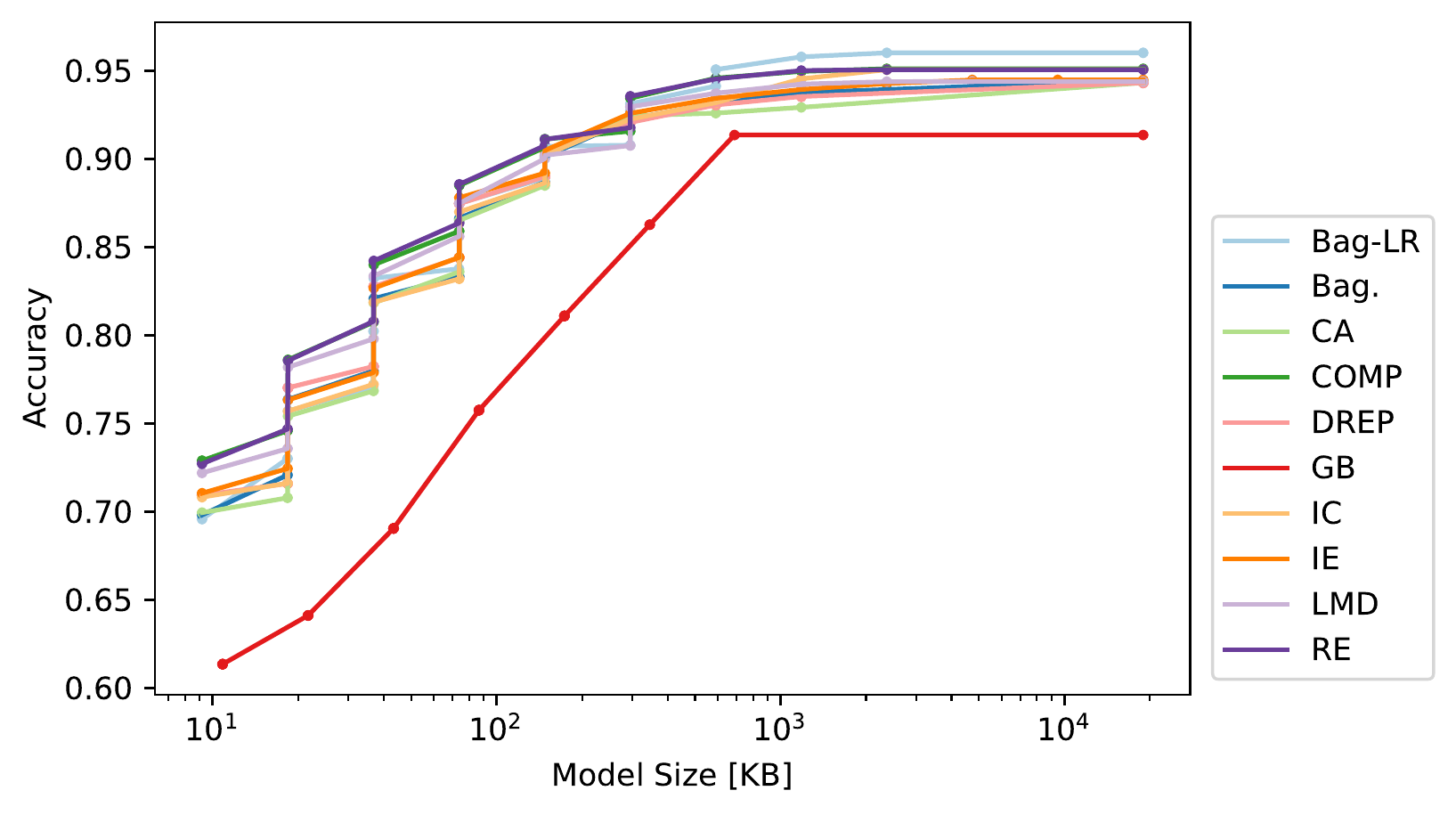}
\end{minipage}\hfill
\begin{minipage}{.49\textwidth}
    \centering 
    \includegraphics[width=\textwidth,keepaspectratio]{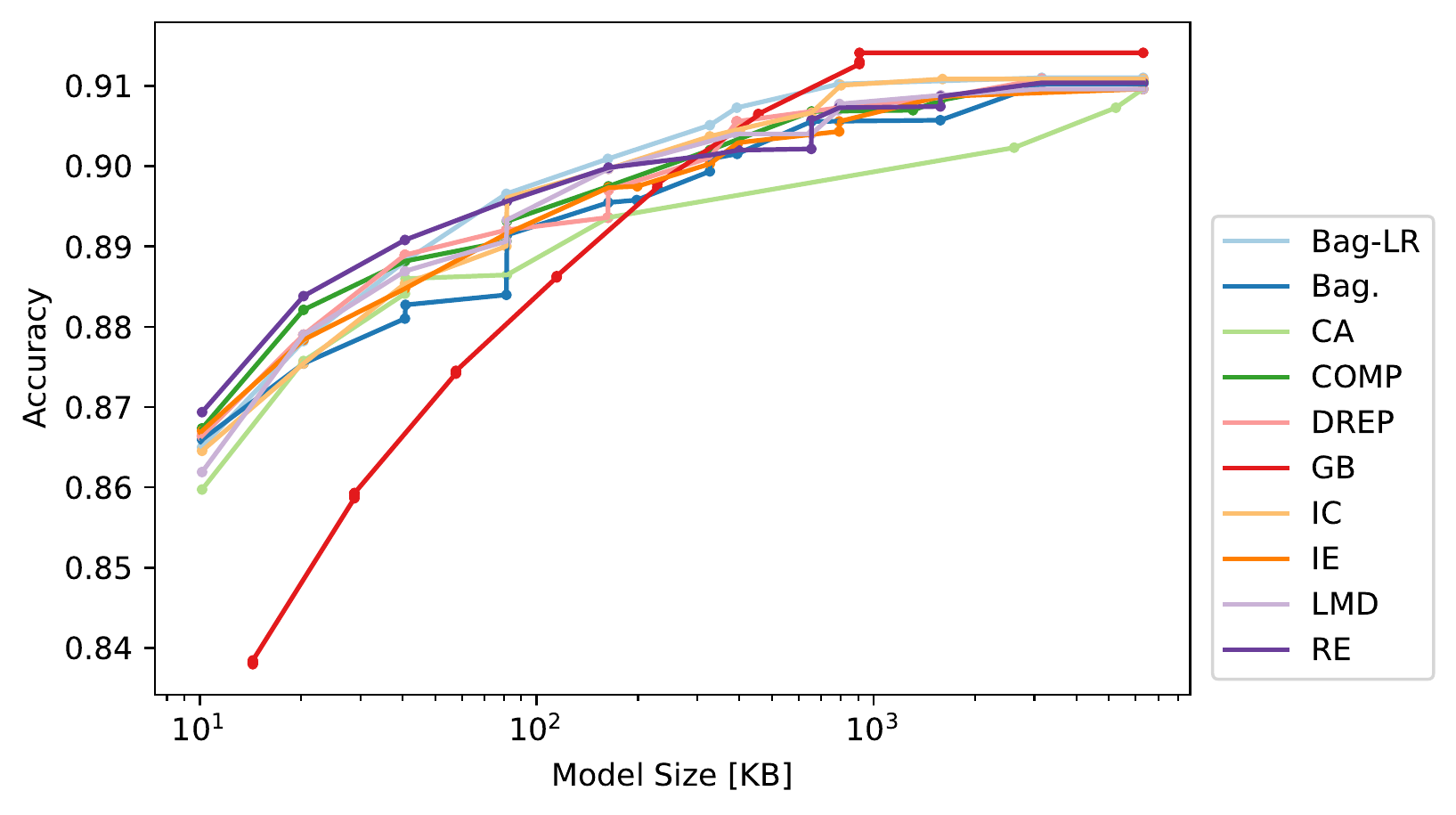}
\end{minipage}
\caption{(left) 5-fold cross-validation accuracy on the postures dataset. (right) 5-fold cross-validation accuracy on the satimage dataset.}
\end{figure}

\subsection{Area Under the Pareto Front with a Bagging Classifier}

\resizebox{\textwidth}{!}{
    \input{figures/aucs_BaggingClassifier}
}

\begin{figure}
\centering
\includegraphics[width=\columnwidth, keepaspectratio]{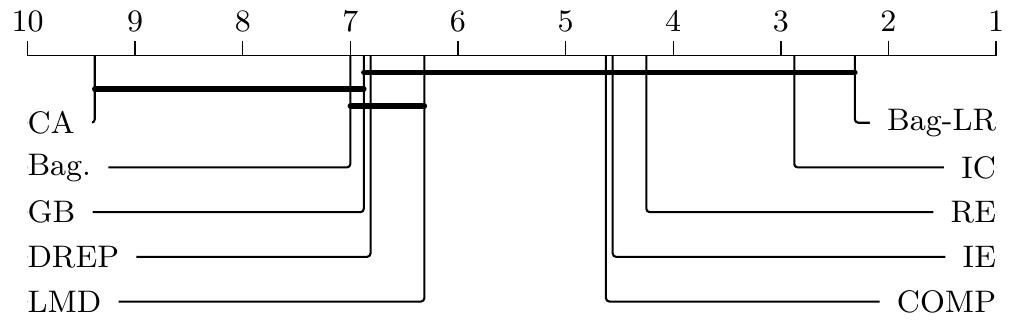}
\caption{Critical Difference Diagram for the normalized area under the Pareto front for different methods over multiple datasets. More to the right (lower rank) is better. Methods in connected cliques are statistically similar.}
\label{fig:cd_auc}
\end{figure}

\section{Revisiting Ensemble Pruning with ExtaTrees Classifier}

For space reasons, the paper focuses on Random Forest classifier. Here we will repeat our experiment with ExtaTreesClassifier implemented in Scikit-Learn \cite{Pedregosa/etal/2001}. As before, we either use a 5-fold cross validation or the given test/train split. For reference, recall our experimental protocol: Oshiro et al. showed in \cite{oshiro/etal/2012} that the prediction of a RF stabilizes between $128$ and $256$ trees in the ensemble and adding more trees to the ensemble does not yield significantly better results. Hence, we train the `base' Random Forests with $M = 256$ trees. To control the individual errors of trees we set the maximum number of leaf nodes $n_l$ to values between $n_l \in \{64,128,256,512,1024\}$. For ensemble pruning we use RE and compare it against a random selection of trees from the original ensemble (which is the same a training a smaller forest directly). In both cases a sub-ensemble with $K \in \{2,4,8,16,32,64,128,256\}$ members is selected so that for $K=256$ the original RF is recovered. 

\begin{figure}[H]
\begin{minipage}{.49\textwidth}
    \centering
    \includegraphics[width=\textwidth,keepaspectratio]{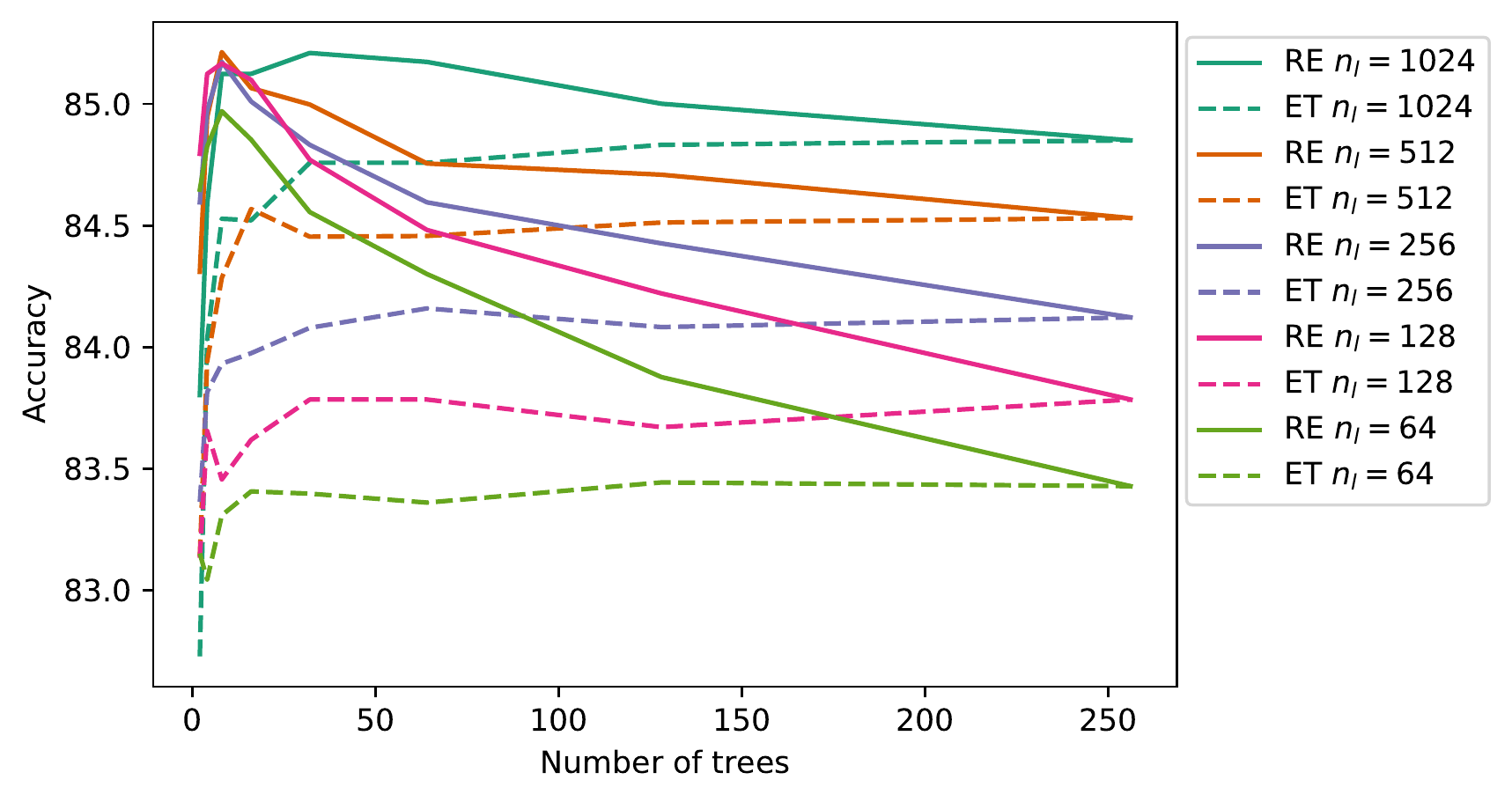}
\end{minipage}\hfill
\begin{minipage}{.49\textwidth}
    \centering 
    \resizebox{\textwidth}{!}{
        \input{figures/ExtraTreesClassifier_adult_table}
    }
\end{minipage}
\caption{(Left) The error over the number of trees in the ensemble on the adult dataset. Dashed lines depict the Random Forest and solid lines are the corresponding pruned ensemble via Reduced Error pruning. (Right) The 5-fold cross-validation accuracy  on the adult dataset. Rounded to the second decimal digit. Larger is better.}
\end{figure}

\begin{figure}[H]
\begin{minipage}{.49\textwidth}
    \centering
    \includegraphics[width=\textwidth,keepaspectratio]{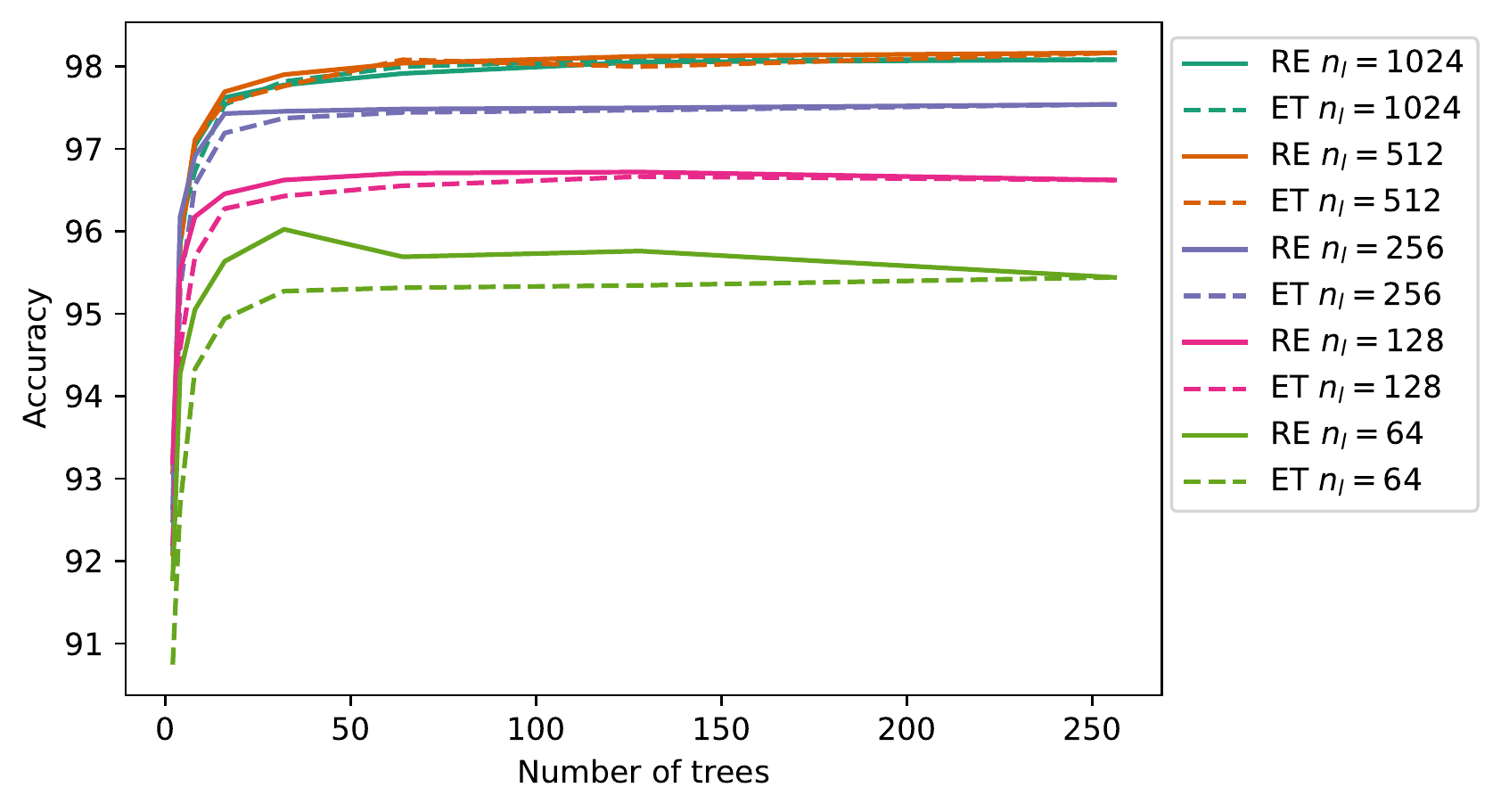}
\end{minipage}\hfill
\begin{minipage}{.49\textwidth}
    \centering 
    \resizebox{\textwidth}{!}{
        \input{figures/ExtraTreesClassifier_anura_table}
    }
\end{minipage}
\caption{(Left) The error over the number of trees in the ensemble on the anura dataset. Dashed lines depict the Random Forest and solid lines are the corresponding pruned ensemble via Reduced Error pruning. (Right) The 5-fold cross-validation accuracy  on the anura dataset. Rounded to the second decimal digit. Larger is better.}
\end{figure}

\begin{figure}[H]
\begin{minipage}{.49\textwidth}
    \centering
    \includegraphics[width=\textwidth,keepaspectratio]{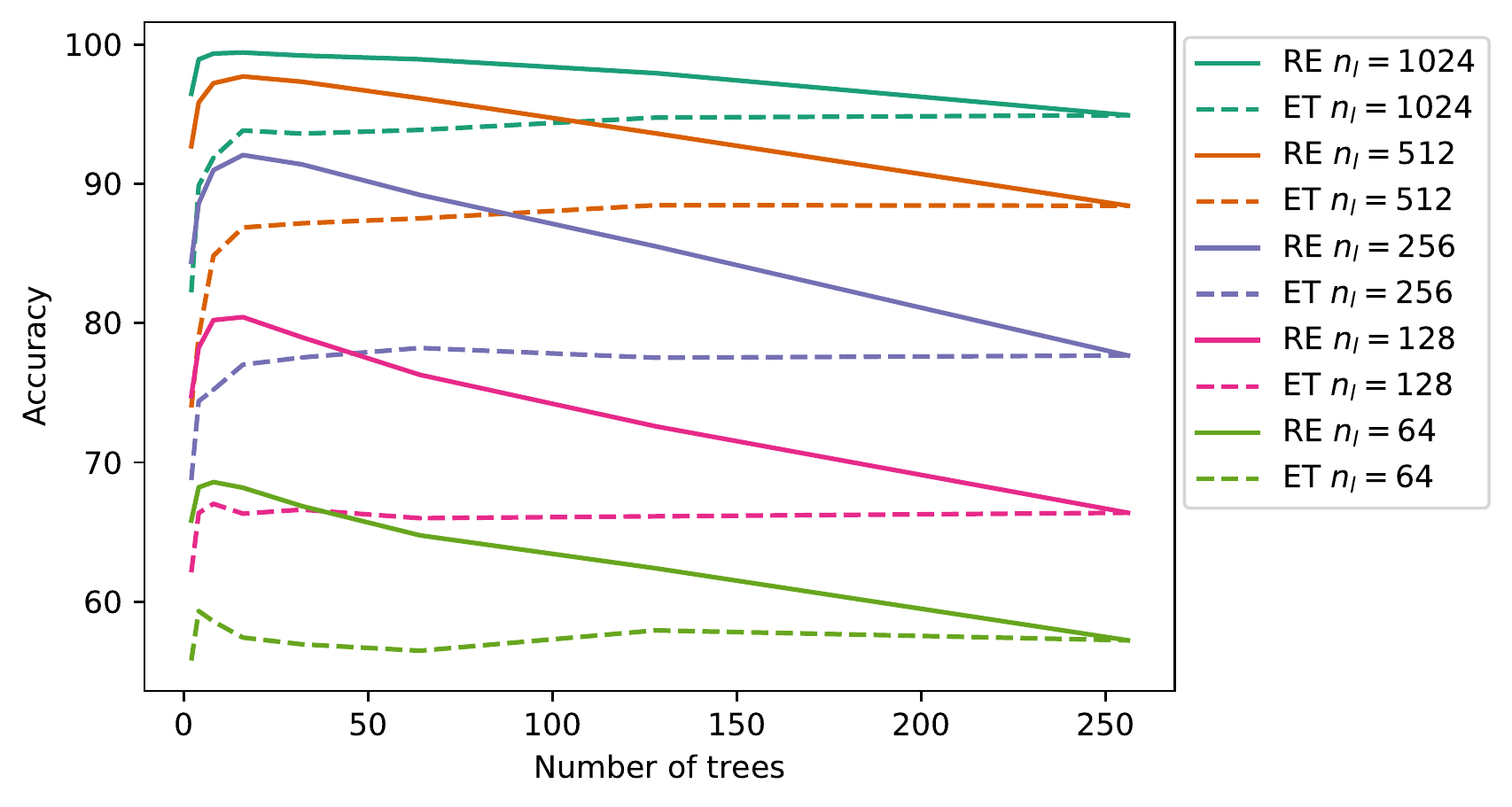}
\end{minipage}\hfill
\begin{minipage}{.49\textwidth}
    \centering 
    \resizebox{\textwidth}{!}{
        \input{figures/ExtraTreesClassifier_avila_table}
    }
\end{minipage}
\caption{(Left) The error over the number of trees in the ensemble on the avila dataset. Dashed lines depict the Random Forest and solid lines are the corresponding pruned ensemble via Reduced Error pruning. (Right) The 5-fold cross-validation accuracy  on the avila dataset. Rounded to the second decimal digit. Larger is better.}
\end{figure}

\begin{figure}[H]
\begin{minipage}{.49\textwidth}
    \centering
    \includegraphics[width=\textwidth,keepaspectratio]{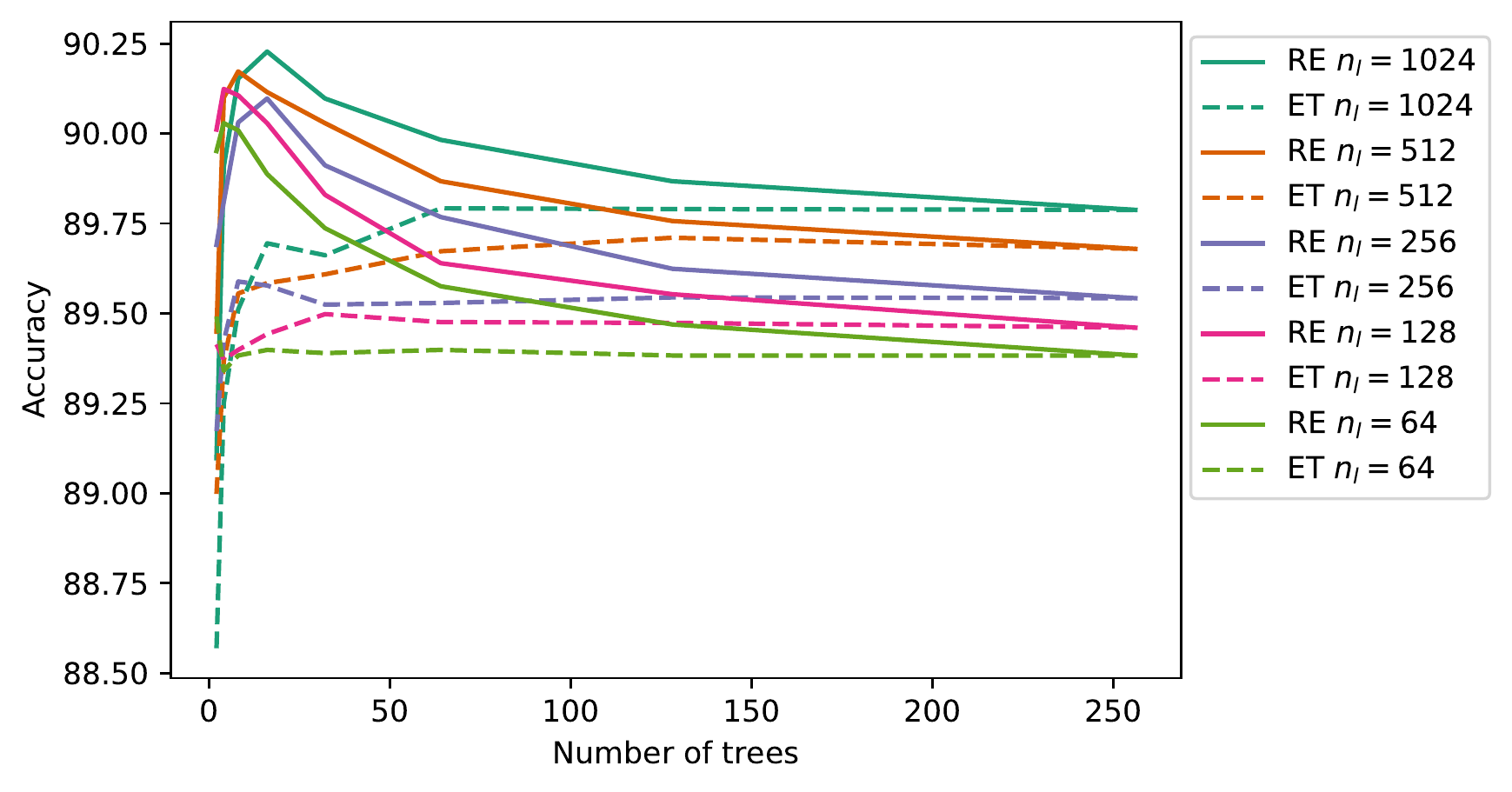}
\end{minipage}\hfill
\begin{minipage}{.49\textwidth}
    \centering 
    \resizebox{\textwidth}{!}{
        \input{figures/ExtraTreesClassifier_bank_table}
    }
\end{minipage}
\caption{(Left) The error over the number of trees in the ensemble on the bank dataset. Dashed lines depict the Random Forest and solid lines are the corresponding pruned ensemble via Reduced Error pruning. (Right) The 5-fold cross-validation accuracy  on the bank dataset. Rounded to the second decimal digit. Larger is better.}
\end{figure}

\begin{figure}[H]
\begin{minipage}{.49\textwidth}
    \centering
    \includegraphics[width=\textwidth,keepaspectratio]{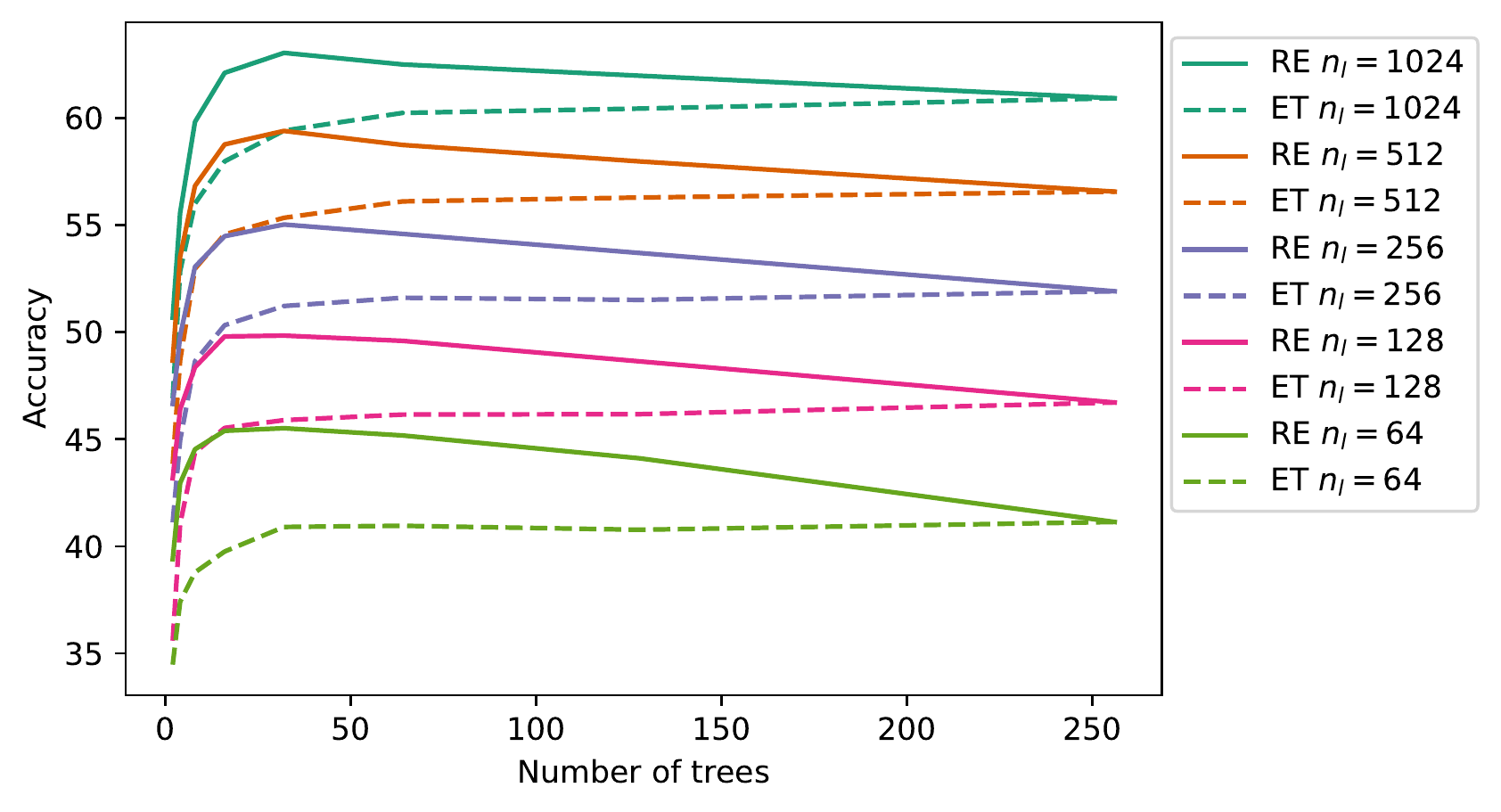}
\end{minipage}\hfill
\begin{minipage}{.49\textwidth}
    \centering 
    \resizebox{\textwidth}{!}{
        \input{figures/ExtraTreesClassifier_chess_table}
    }
\end{minipage}
\caption{(Left) The error over the number of trees in the ensemble on the chess dataset. Dashed lines depict the Random Forest and solid lines are the corresponding pruned ensemble via Reduced Error pruning. (Right) The 5-fold cross-validation accuracy  on the chess dataset. Rounded to the second decimal digit. Larger is better.}
\end{figure}

\begin{figure}[H]
\begin{minipage}{.49\textwidth}
    \centering
    \includegraphics[width=\textwidth,keepaspectratio]{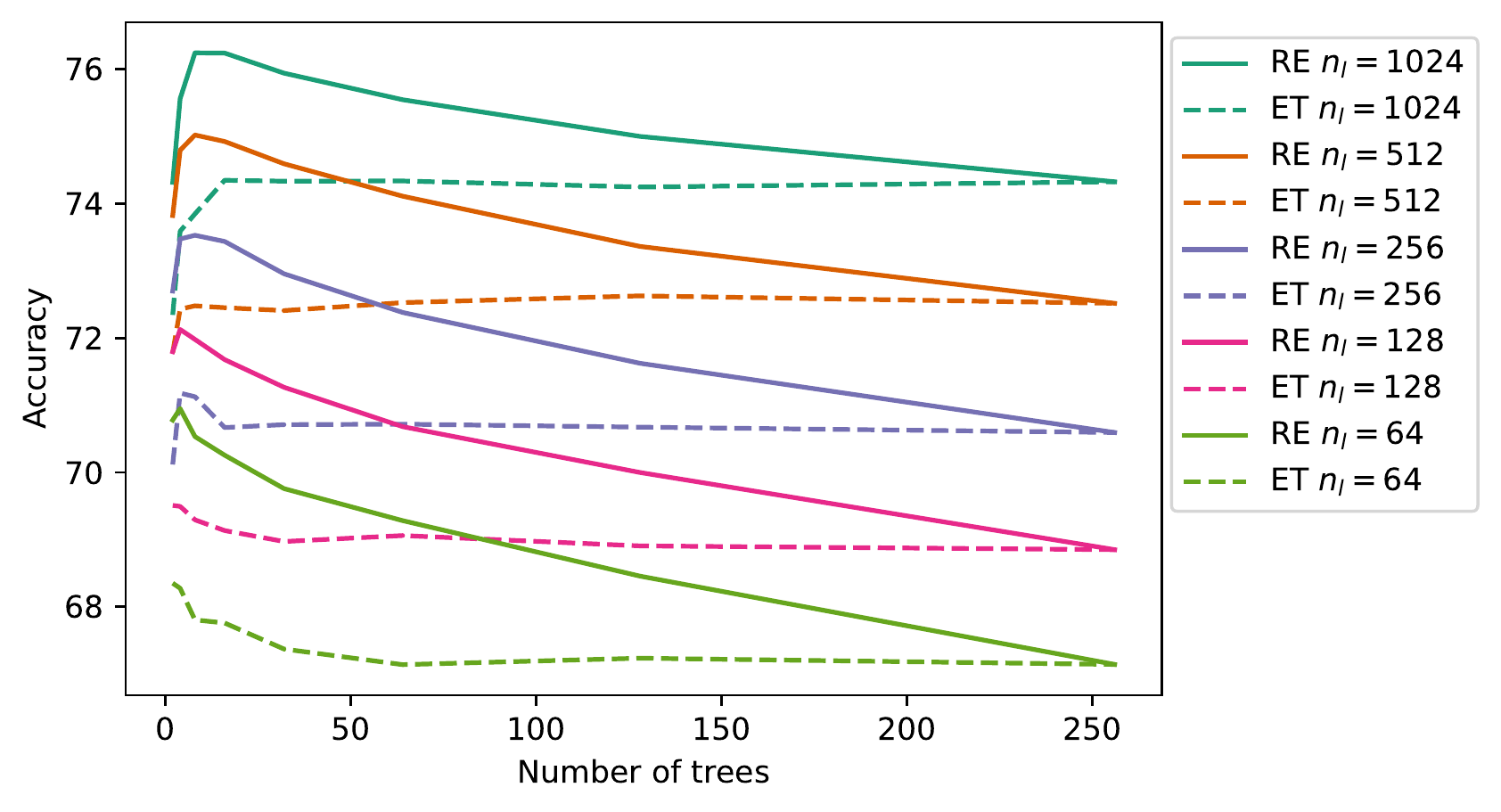}
\end{minipage}\hfill
\begin{minipage}{.49\textwidth}
    \centering 
    \resizebox{\textwidth}{!}{
        \input{figures/ExtraTreesClassifier_connect_table}
    }
\end{minipage}
\caption{(Left) The error over the number of trees in the ensemble on the connect dataset. Dashed lines depict the Random Forest and solid lines are the corresponding pruned ensemble via Reduced Error pruning. (Right) The 5-fold cross-validation accuracy  on the connect dataset. Rounded to the second decimal digit. Larger is better.}
\end{figure}

\begin{figure}[H]
\begin{minipage}{.49\textwidth}
    \centering
    \includegraphics[width=\textwidth,keepaspectratio]{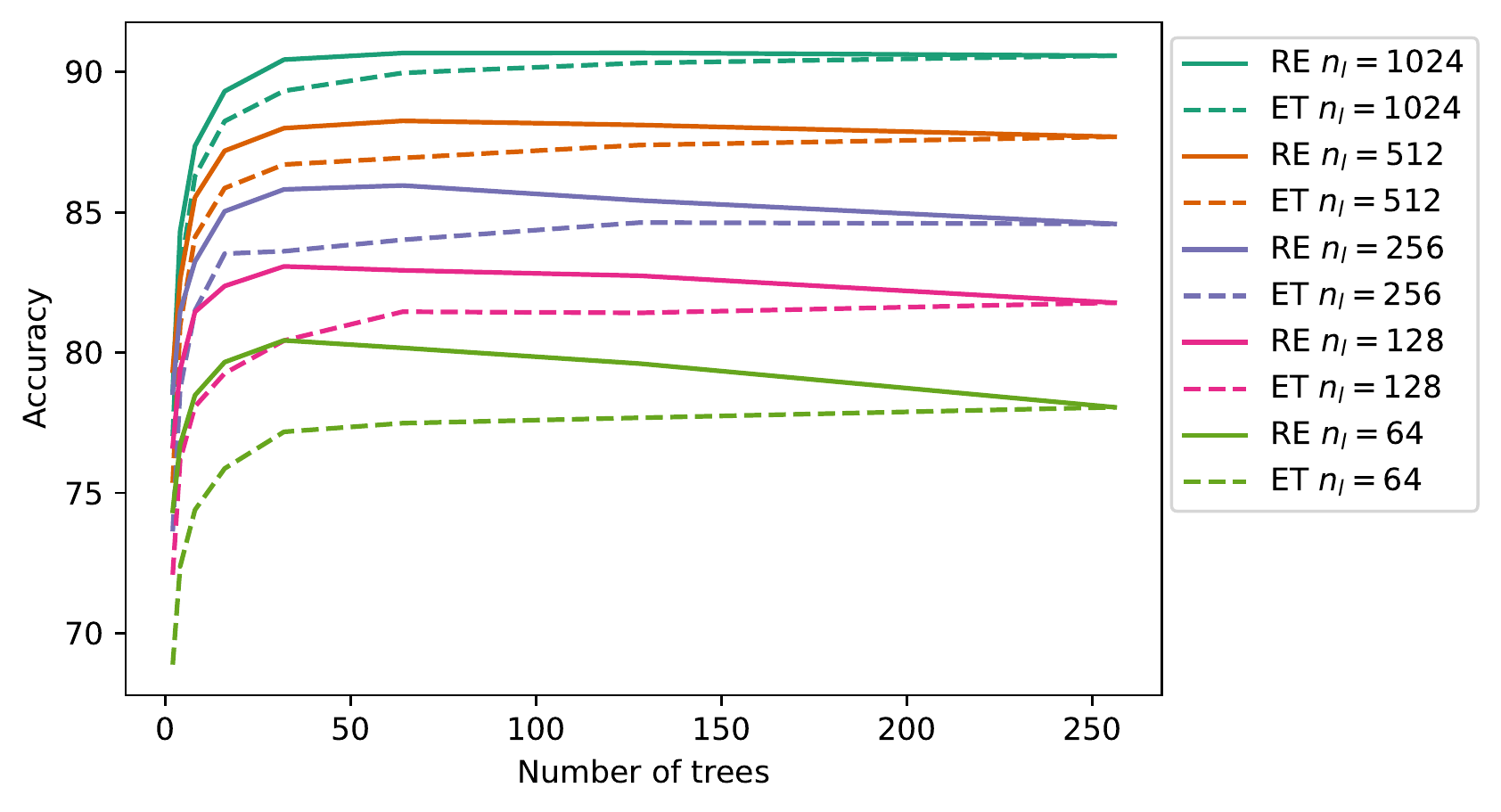}
\end{minipage}\hfill
\begin{minipage}{.49\textwidth}
    \centering 
    \resizebox{\textwidth}{!}{
        \input{figures/ExtraTreesClassifier_eeg_table}
    }
\end{minipage}
\caption{(Left) The error over the number of trees in the ensemble on the eeg dataset. Dashed lines depict the Random Forest and solid lines are the corresponding pruned ensemble via Reduced Error pruning. (Right) The 5-fold cross-validation accuracy  on the eeg dataset. Rounded to the second decimal digit. Larger is better.}
\end{figure}

\begin{figure}[H]
\begin{minipage}{.49\textwidth}
    \centering
    \includegraphics[width=\textwidth,keepaspectratio]{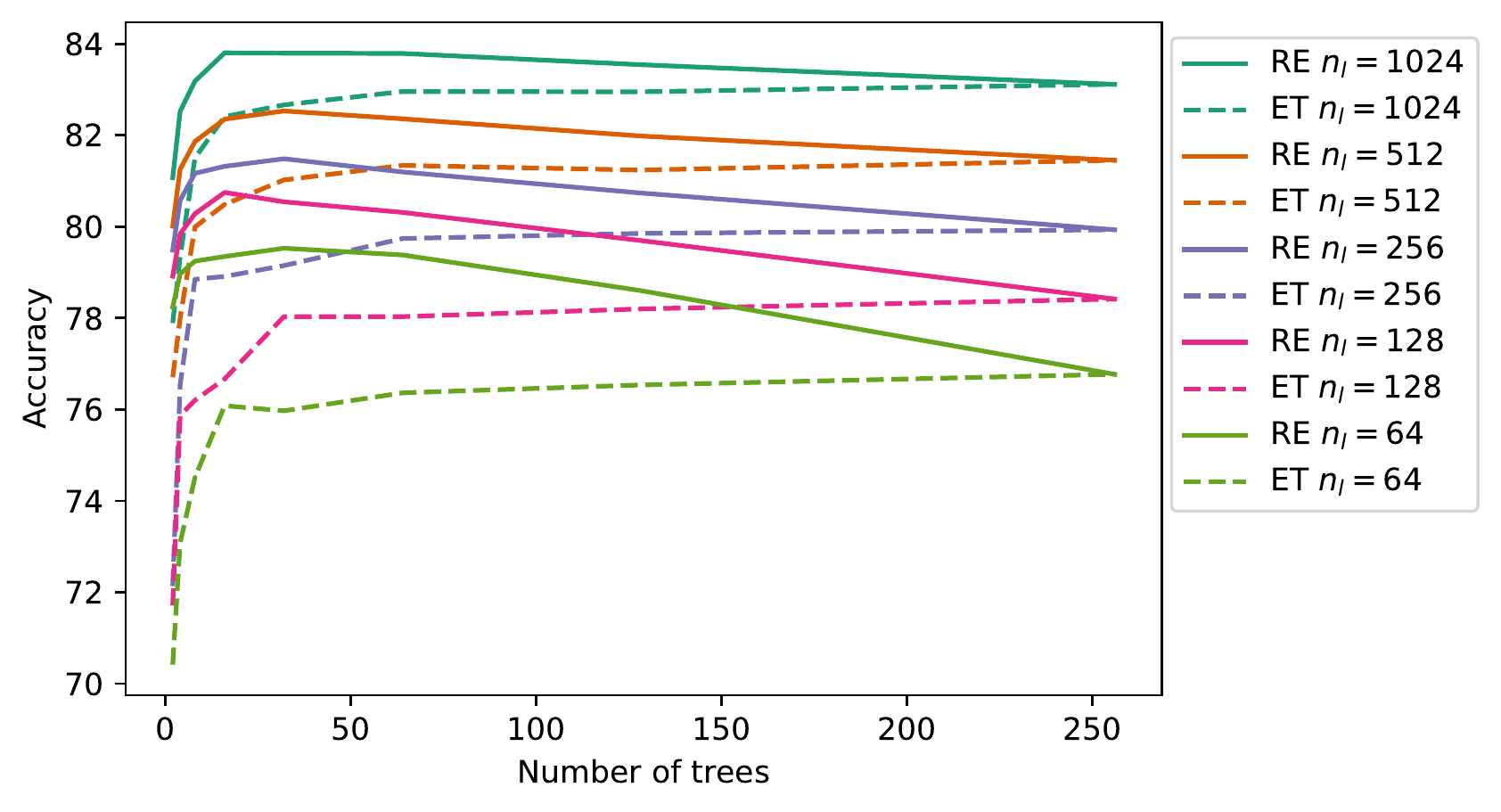}
\end{minipage}\hfill
\begin{minipage}{.49\textwidth}
    \centering 
    \resizebox{\textwidth}{!}{
        \input{figures/ExtraTreesClassifier_elec_table}
    }
\end{minipage}
\caption{(Left) The error over the number of trees in the ensemble on the elec dataset. Dashed lines depict the Random Forest and solid lines are the corresponding pruned ensemble via Reduced Error pruning. (Right) The 5-fold cross-validation accuracy  on the elec dataset. Rounded to the second decimal digit. Larger is better.}
\end{figure}

\begin{figure}[H]
\begin{minipage}{.49\textwidth}
    \centering
    \includegraphics[width=\textwidth,keepaspectratio]{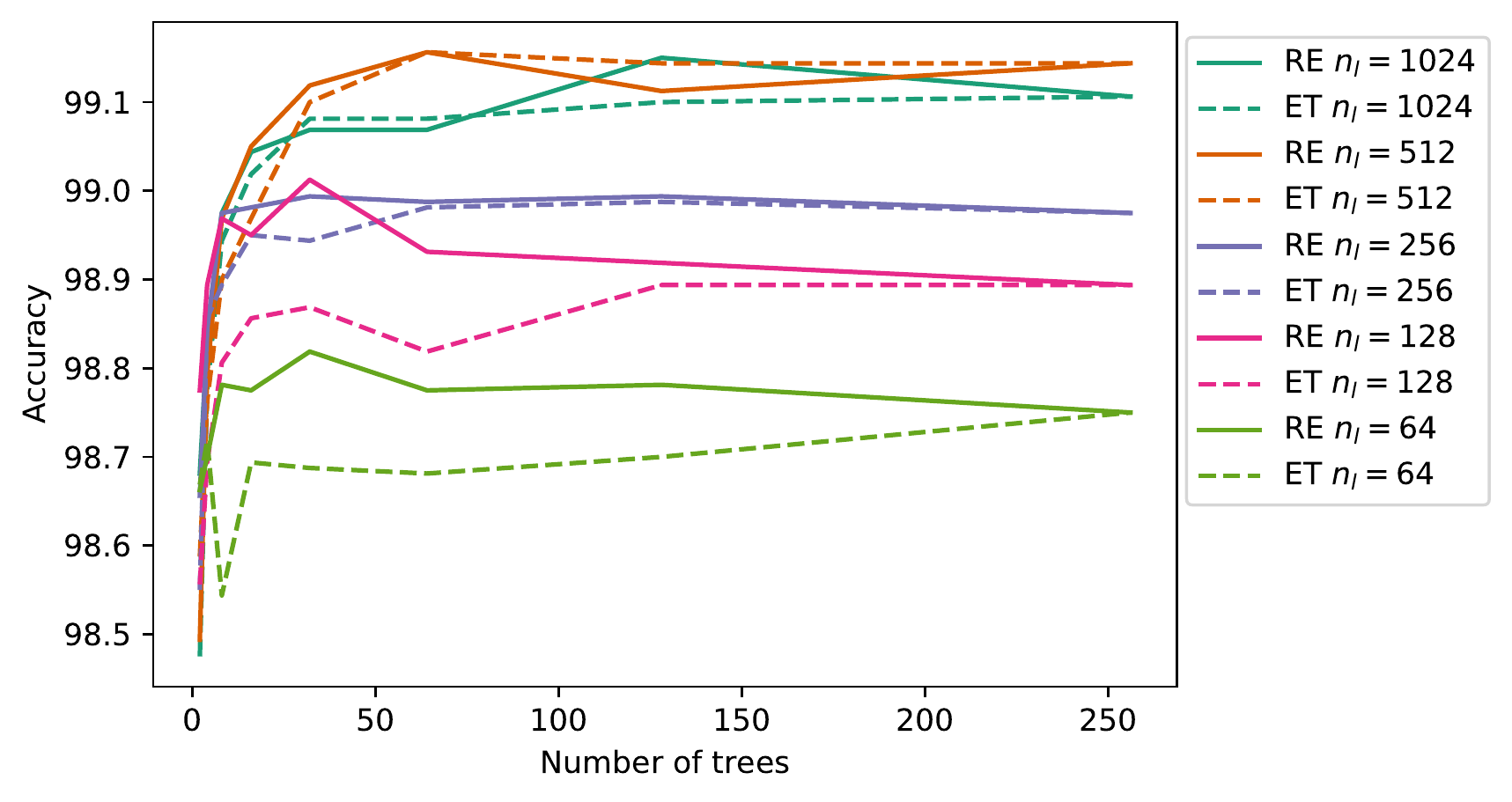}
\end{minipage}\hfill
\begin{minipage}{.49\textwidth}
    \centering 
    \resizebox{\textwidth}{!}{
        \input{figures/ExtraTreesClassifier_ida2016_table}
    }
\end{minipage}
\caption{(Left) The error over the number of trees in the ensemble on the ida2016 dataset. Dashed lines depict the Random Forest and solid lines are the corresponding pruned ensemble via Reduced Error pruning. (Right) The 5-fold cross-validation accuracy  on the ida2016 dataset. Rounded to the second decimal digit. Larger is better. }
\end{figure}

\begin{figure}[H]
\begin{minipage}{.49\textwidth}
    \centering
    \includegraphics[width=\textwidth,keepaspectratio]{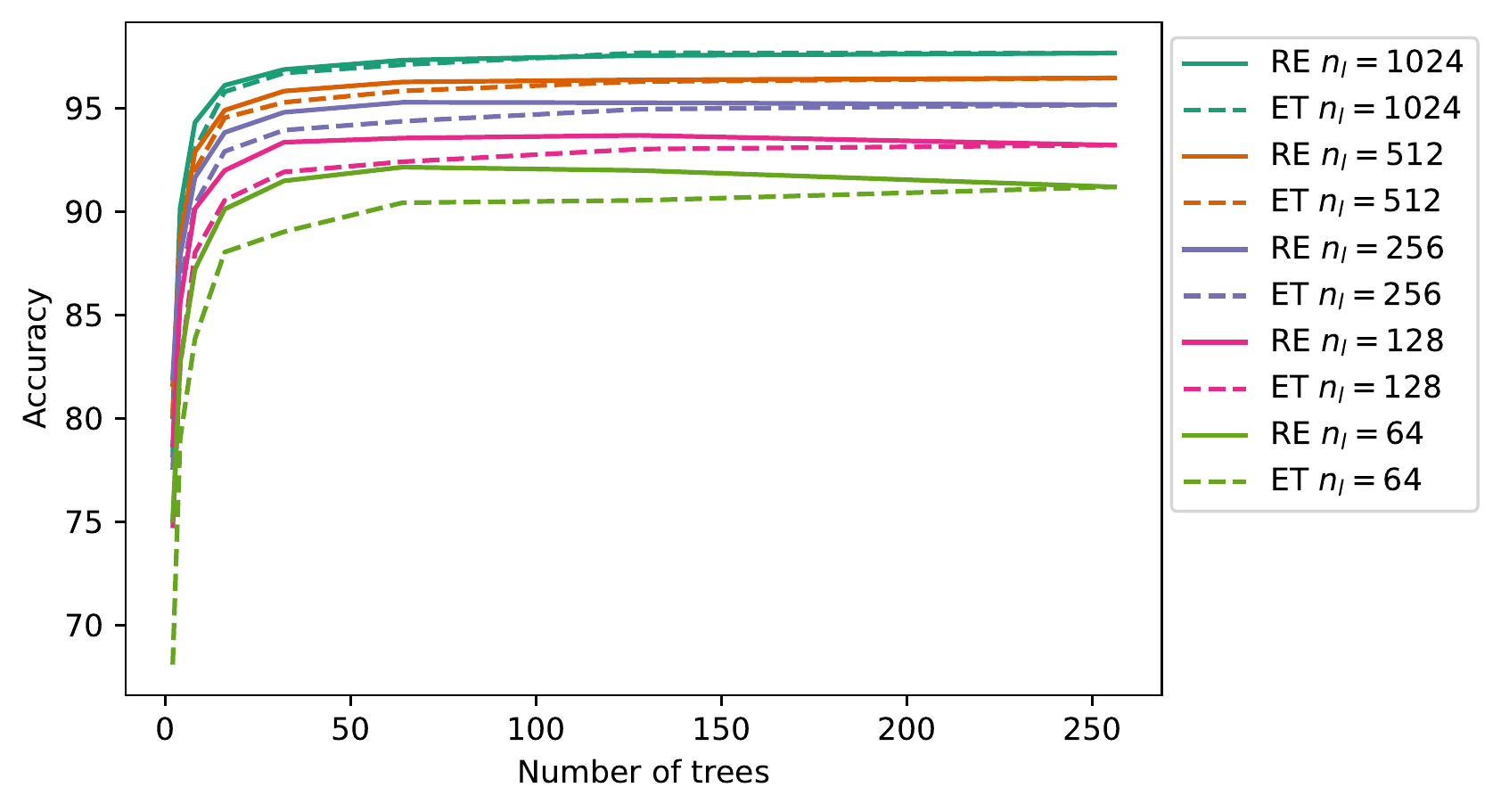}
\end{minipage}\hfill
\begin{minipage}{.49\textwidth}
    \centering 
    \resizebox{\textwidth}{!}{
        \input{figures/ExtraTreesClassifier_japanese-vowels_table}
    }
\end{minipage}
\caption{(Left) The error over the number of trees in the ensemble on the japanese-vowels dataset. Dashed lines depict the Random Forest and solid lines are the corresponding pruned ensemble via Reduced Error pruning. (Right) The 5-fold cross-validation accuracy  on the japanese-vowels dataset. Rounded to the second decimal digit. Larger is better.}
\end{figure}

\begin{figure}[H]
\begin{minipage}{.49\textwidth}
    \centering
    \includegraphics[width=\textwidth,keepaspectratio]{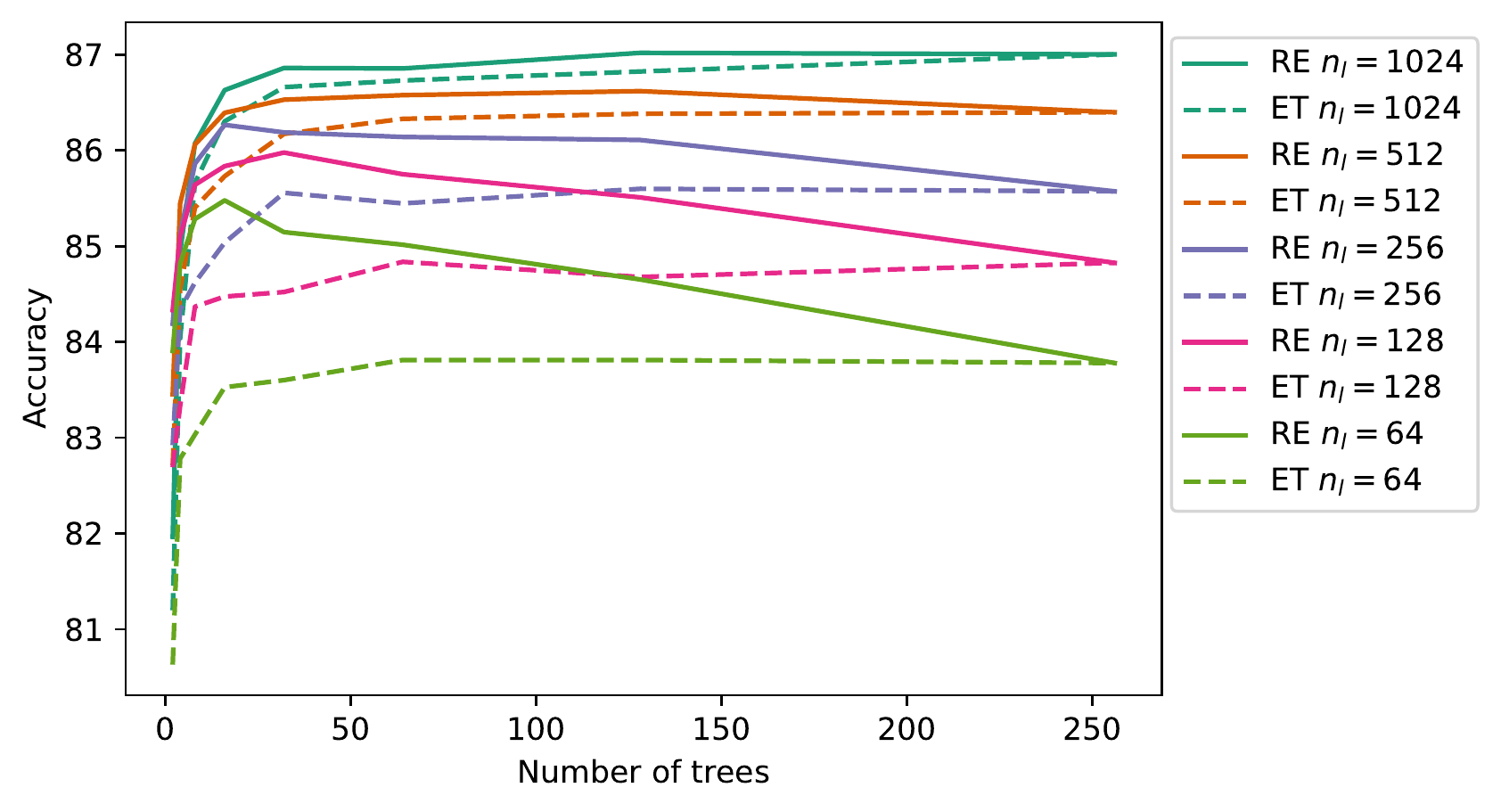}
\end{minipage}\hfill
\begin{minipage}{.49\textwidth}
    \centering 
    \resizebox{\textwidth}{!}{
        \input{figures/ExtraTreesClassifier_magic_table}
    }
\end{minipage}
\caption{(Left) The error over the number of trees in the ensemble on the magic dataset. Dashed lines depict the Random Forest and solid lines are the corresponding pruned ensemble via Reduced Error pruning. (Right) The 5-fold cross-validation accuracy  on the magic dataset. Rounded to the second decimal digit. Larger is better}
\end{figure}

\begin{figure}[H]
\begin{minipage}{.49\textwidth}
    \centering
    \includegraphics[width=\textwidth,keepaspectratio]{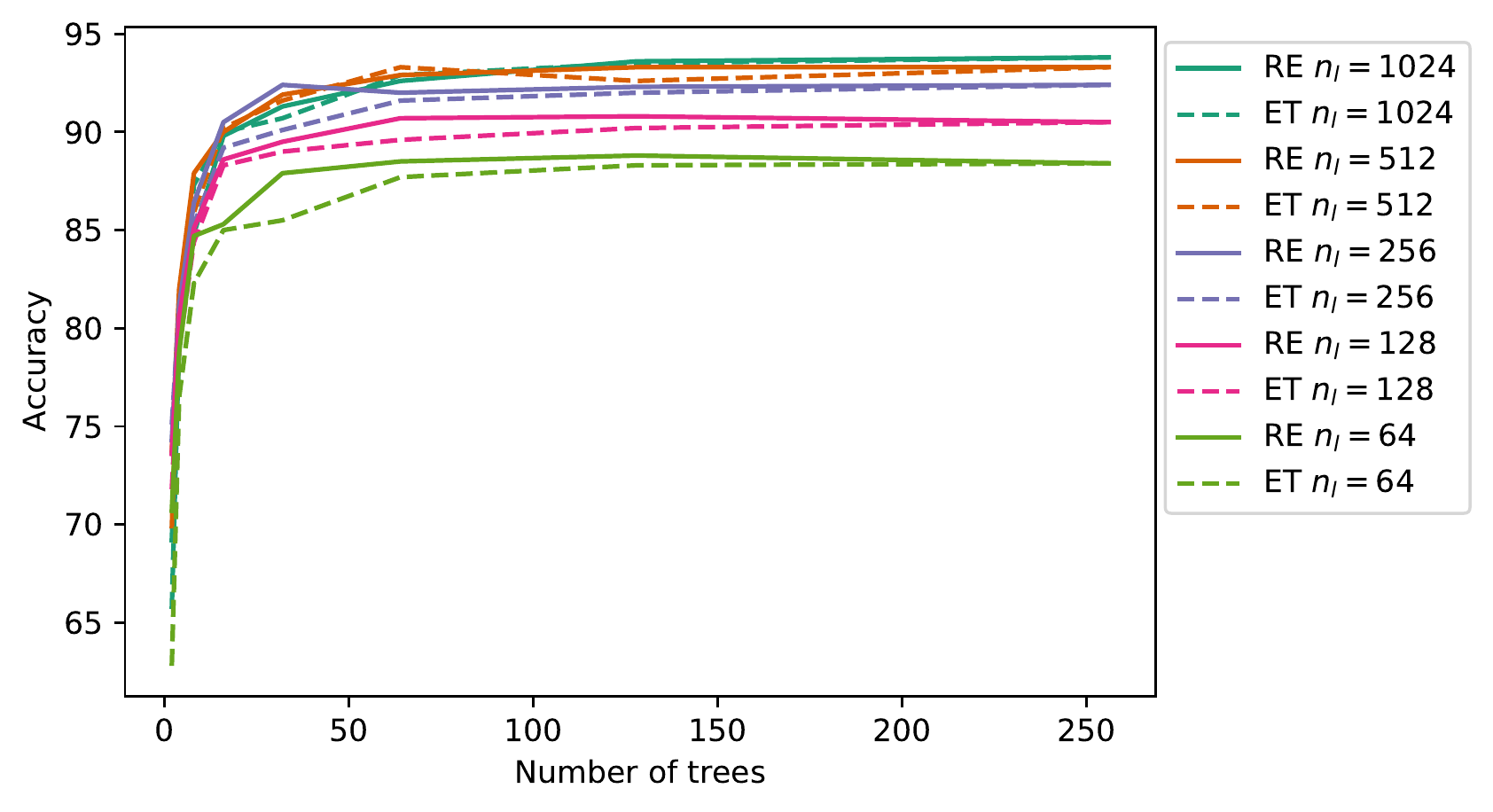}
\end{minipage}\hfill
\begin{minipage}{.49\textwidth}
    \centering 
    \resizebox{\textwidth}{!}{
        \input{figures/ExtraTreesClassifier_mnist_table}
    }
\end{minipage}
\caption{(Left) The error over the number of trees in the ensemble on the mnist dataset. Dashed lines depict the Random Forest and solid lines are the corresponding pruned ensemble via Reduced Error pruning. (Right) The 5-fold cross-validation accuracy  on the mnist dataset. Rounded to the second decimal digit. Larger is better.}
\end{figure}

\begin{figure}[H]
\begin{minipage}{.49\textwidth}
    \centering
    \includegraphics[width=\textwidth,keepaspectratio]{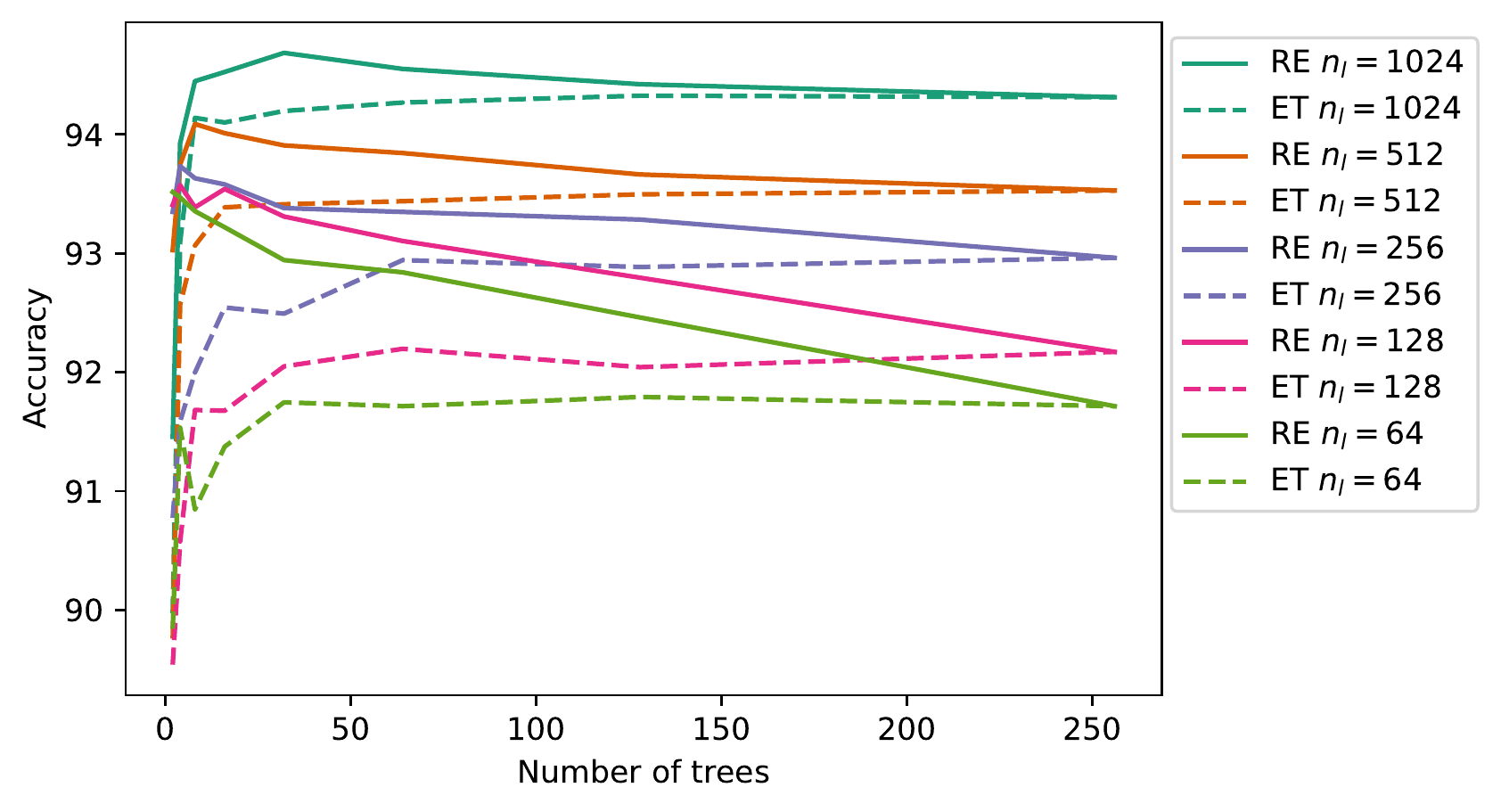}
\end{minipage}\hfill
\begin{minipage}{.49\textwidth}
    \centering 
    \resizebox{\textwidth}{!}{
        \input{figures/ExtraTreesClassifier_mozilla_table}
    }
\end{minipage}
\caption{(Left) The error over the number of trees in the ensemble on the mozilla dataset. Dashed lines depict the Random Forest and solid lines are the corresponding pruned ensemble via Reduced Error pruning. (Right) The 5-fold cross-validation accuracy  on the mozilla dataset. Rounded to the second decimal digit. Larger is better.}
\end{figure}

\begin{figure}[H]
\begin{minipage}{.49\textwidth}
    \centering
    \includegraphics[width=\textwidth,keepaspectratio]{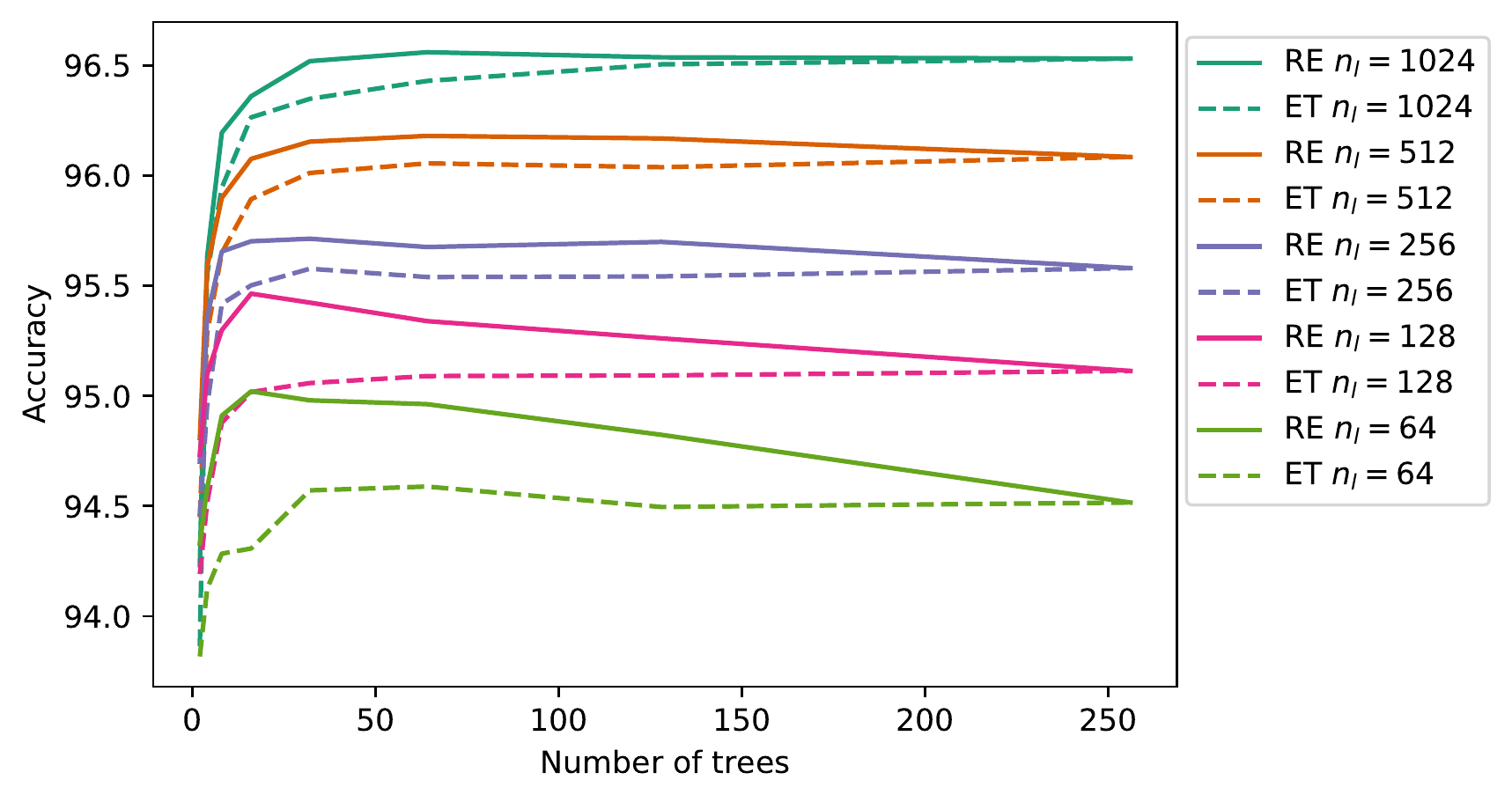}
\end{minipage}\hfill
\begin{minipage}{.49\textwidth}
    \centering 
    \resizebox{\textwidth}{!}{
        \input{figures/ExtraTreesClassifier_nomao_table}
    }
\end{minipage}
\caption{(Left) The error over the number of trees in the ensemble on the nomao dataset. Dashed lines depict the Random Forest and solid lines are the corresponding pruned ensemble via Reduced Error pruning. (Right) The 5-fold cross-validation accuracy  on the nomao dataset. Rounded to the second decimal digit. Larger is better.}
\end{figure}

\begin{figure}[H]
\begin{minipage}{.49\textwidth}
    \centering
    \includegraphics[width=\textwidth,keepaspectratio]{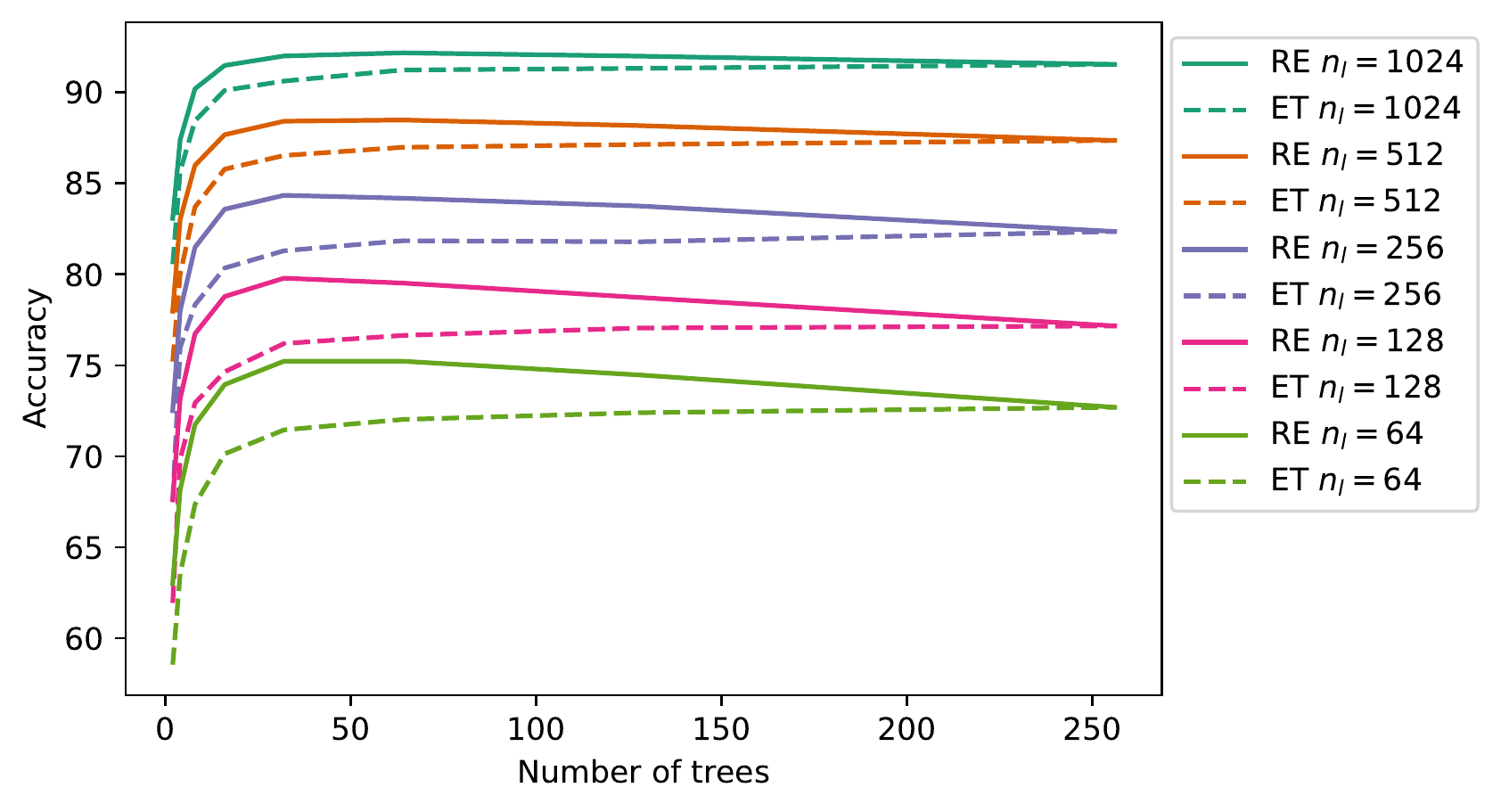}
\end{minipage}\hfill
\begin{minipage}{.49\textwidth}
    \centering 
    \resizebox{\textwidth}{!}{
        \input{figures/ExtraTreesClassifier_postures_table}
    }
\end{minipage}
\caption{(Left) The error over the number of trees in the ensemble on the postures dataset. Dashed lines depict the Random Forest and solid lines are the corresponding pruned ensemble via Reduced Error pruning. (Right) The 5-fold cross-validation accuracy  on the postures dataset. Rounded to the second decimal digit. Larger is better.}
\end{figure}

\begin{figure}[H]
\begin{minipage}{.49\textwidth}
    \centering
    \includegraphics[width=\textwidth,keepaspectratio]{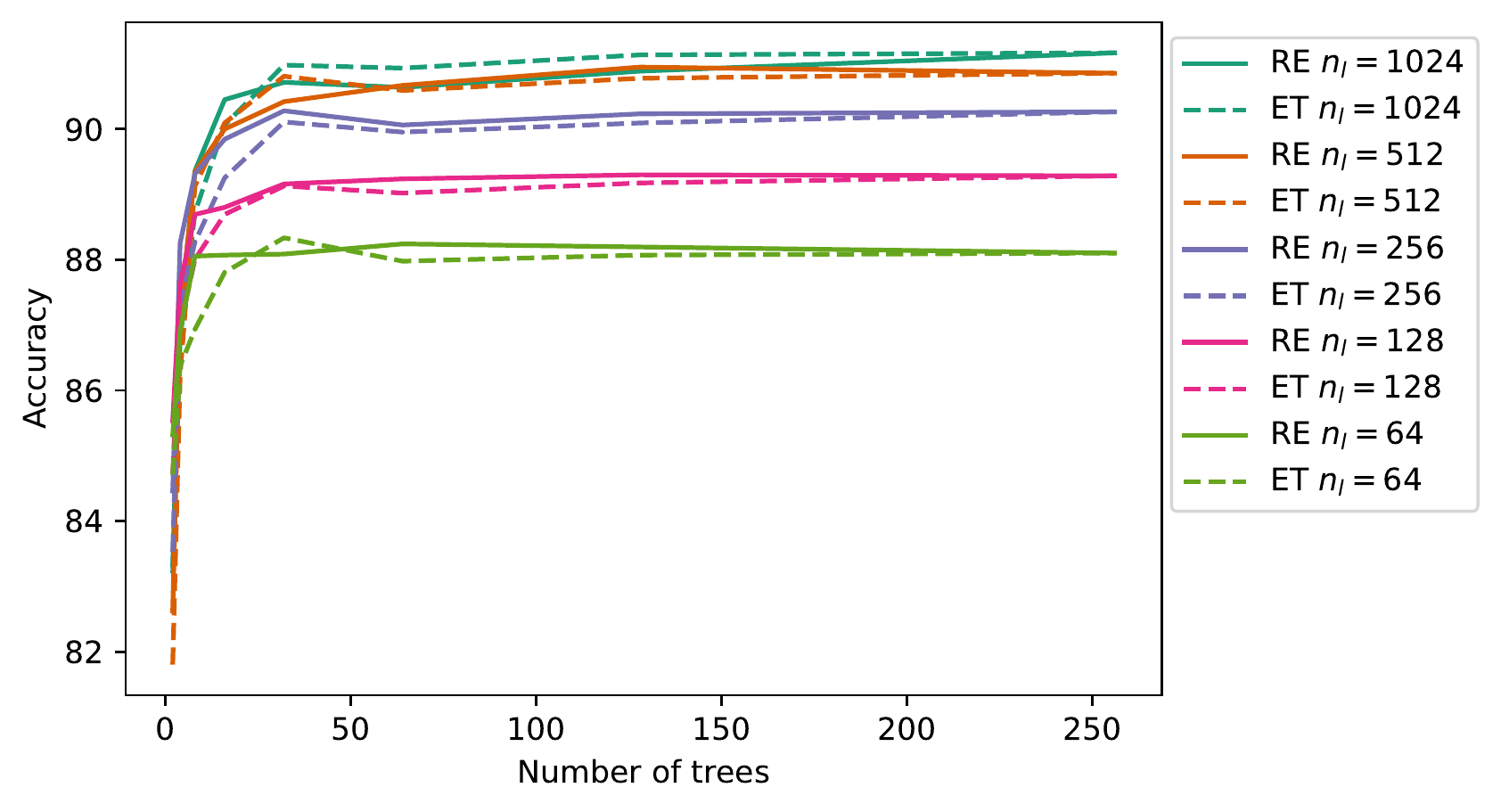}
\end{minipage}\hfill
\begin{minipage}{.49\textwidth}
    \centering 
    \resizebox{\textwidth}{!}{
        \input{figures/ExtraTreesClassifier_satimage_table}
    }
\end{minipage}
\caption{(Left) The error over the number of trees in the ensemble on the satimage dataset. Dashed lines depict the Random Forest and solid lines are the corresponding pruned ensemble via Reduced Error pruning. (Right) The 5-fold cross-validation accuracy  on the satimage dataset. Rounded to the second decimal digit. Larger is better.}
\end{figure}

\subsection{Plotting the Pareto Front For More Datasets with Dedicated Pruning Set}

\begin{figure}[H]
\begin{minipage}{.49\textwidth}
    \centering
    \includegraphics[width=\textwidth,keepaspectratio]{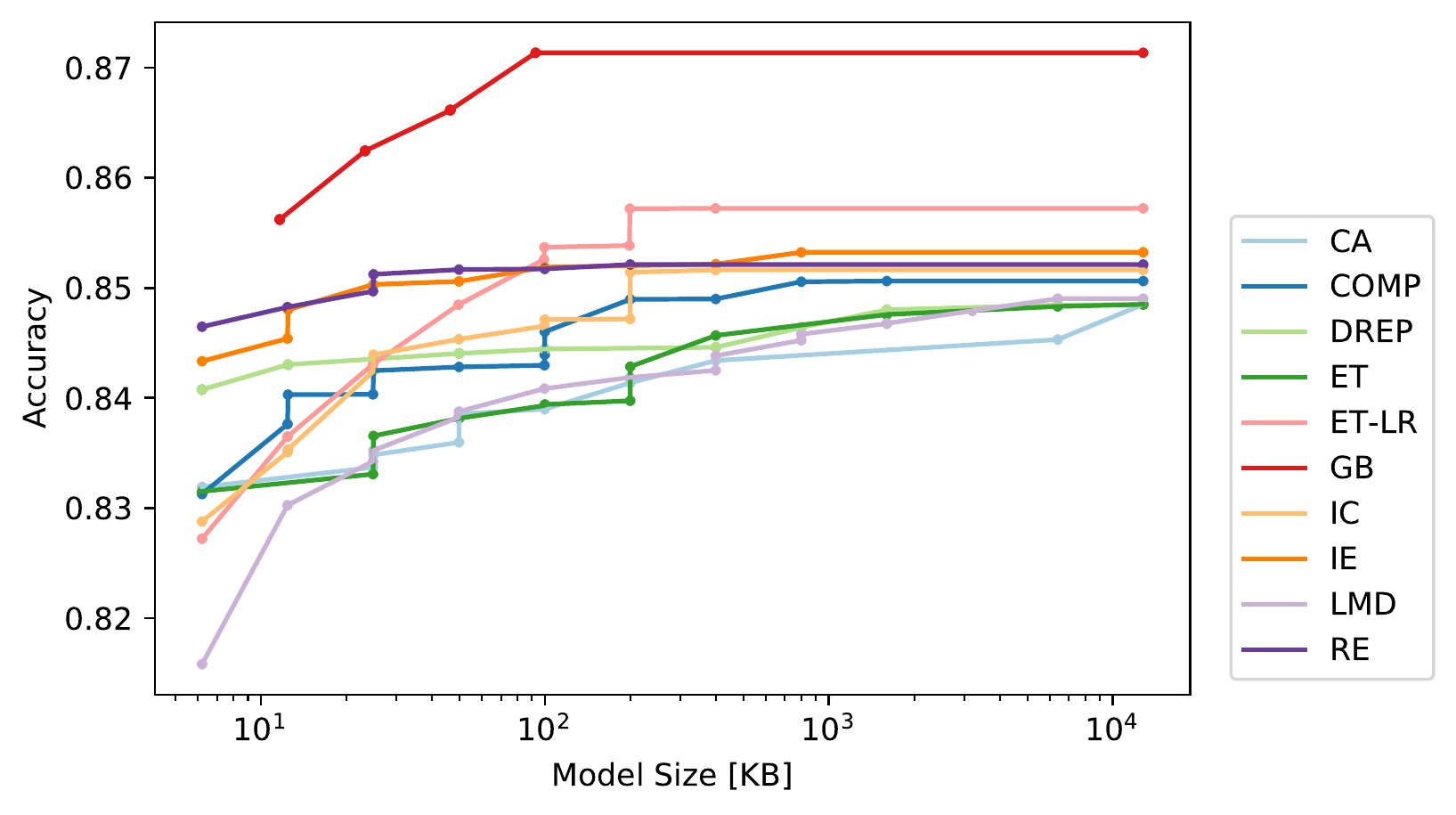}
\end{minipage}\hfill
\begin{minipage}{.49\textwidth}
    \centering 
    \includegraphics[width=\textwidth,keepaspectratio]{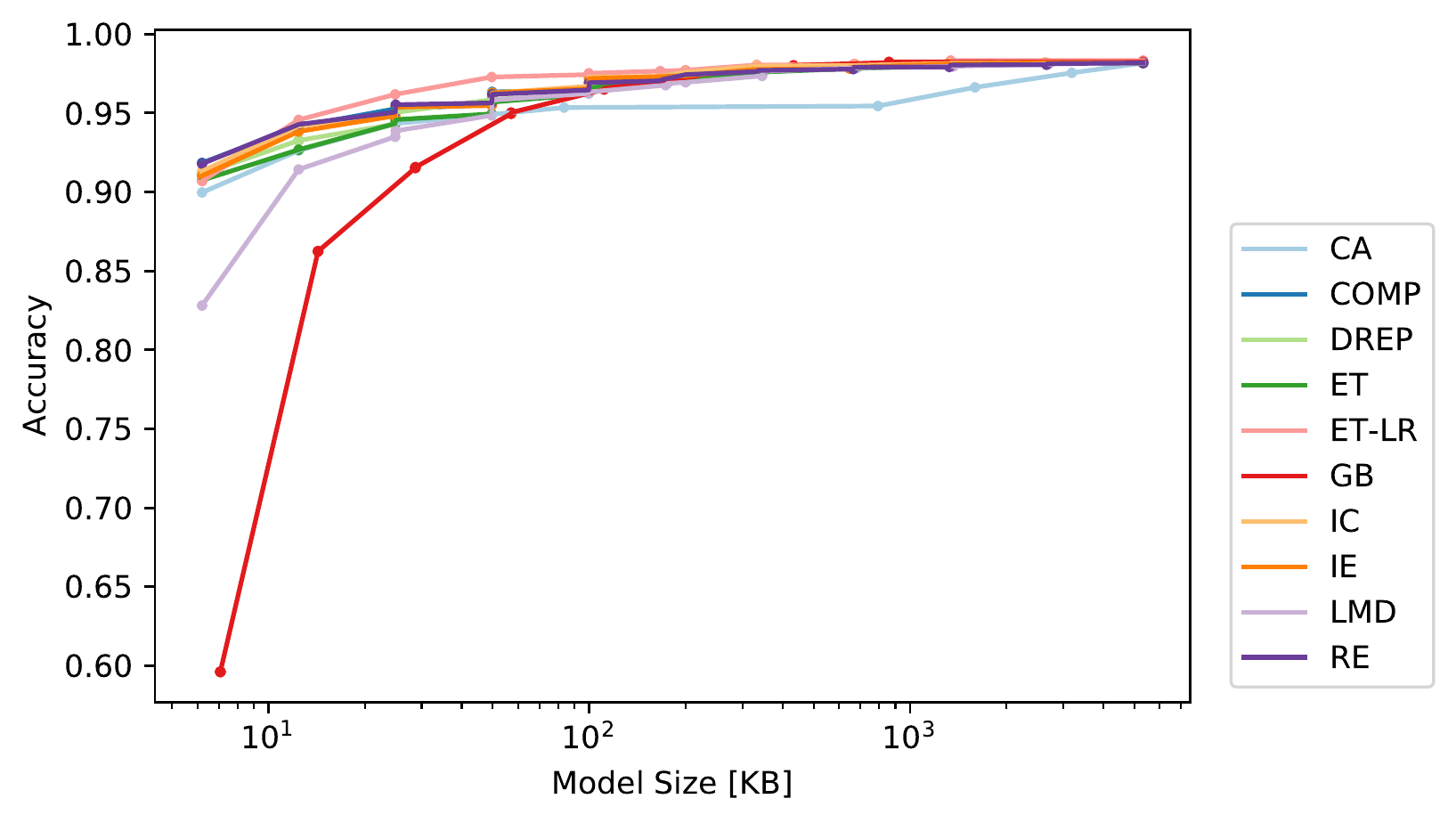}
\end{minipage}
\caption{(left) 5-fold cross-validation accuracy on the adult dataset. (right) 5-fold cross-validation accuracy on the anura dataset.}
\end{figure}

\begin{figure}[H]
\begin{minipage}{.49\textwidth}
    \centering
    \includegraphics[width=\textwidth,keepaspectratio]{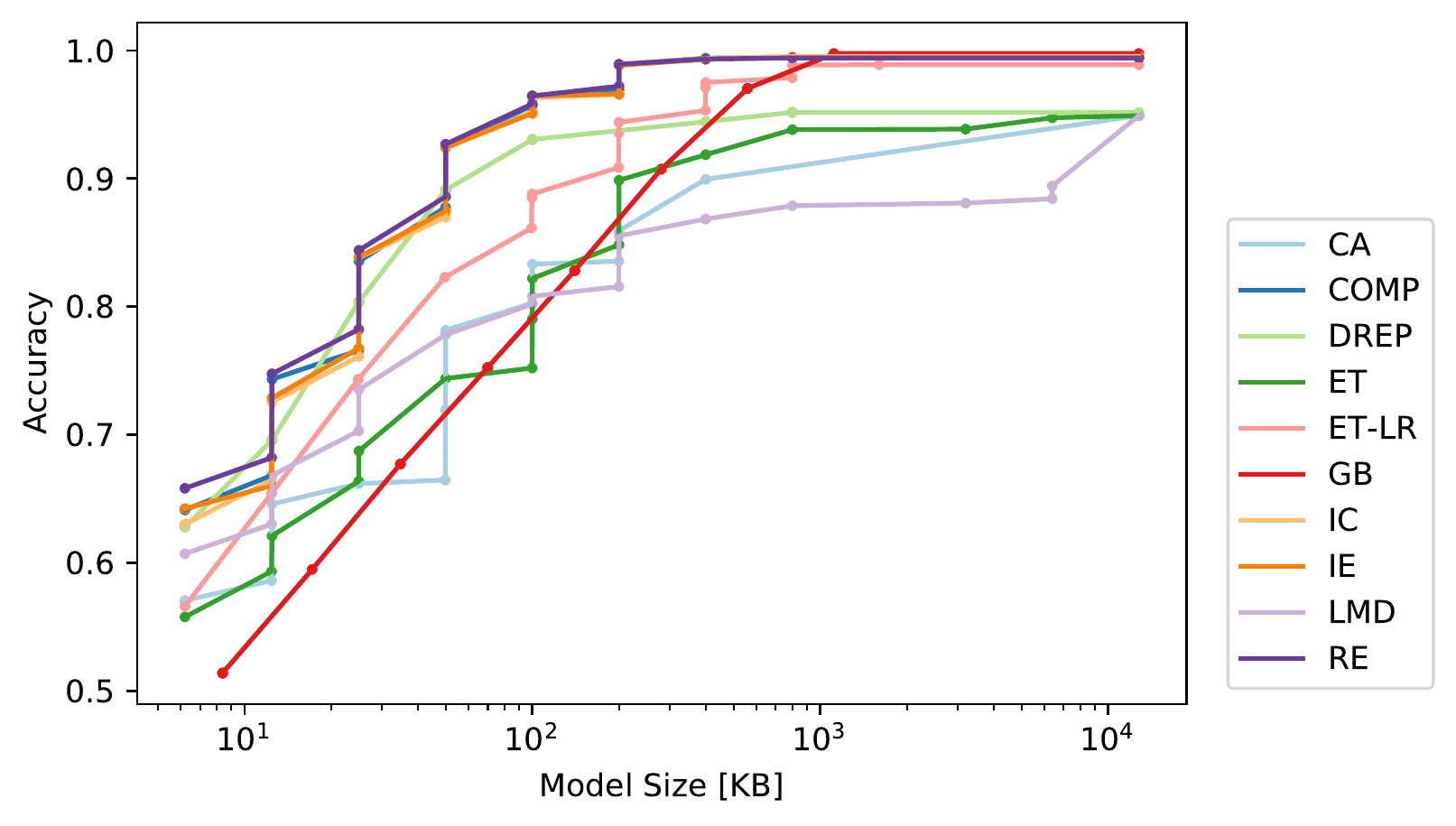}
\end{minipage}\hfill
\begin{minipage}{.49\textwidth}
    \centering 
    \includegraphics[width=\textwidth,keepaspectratio]{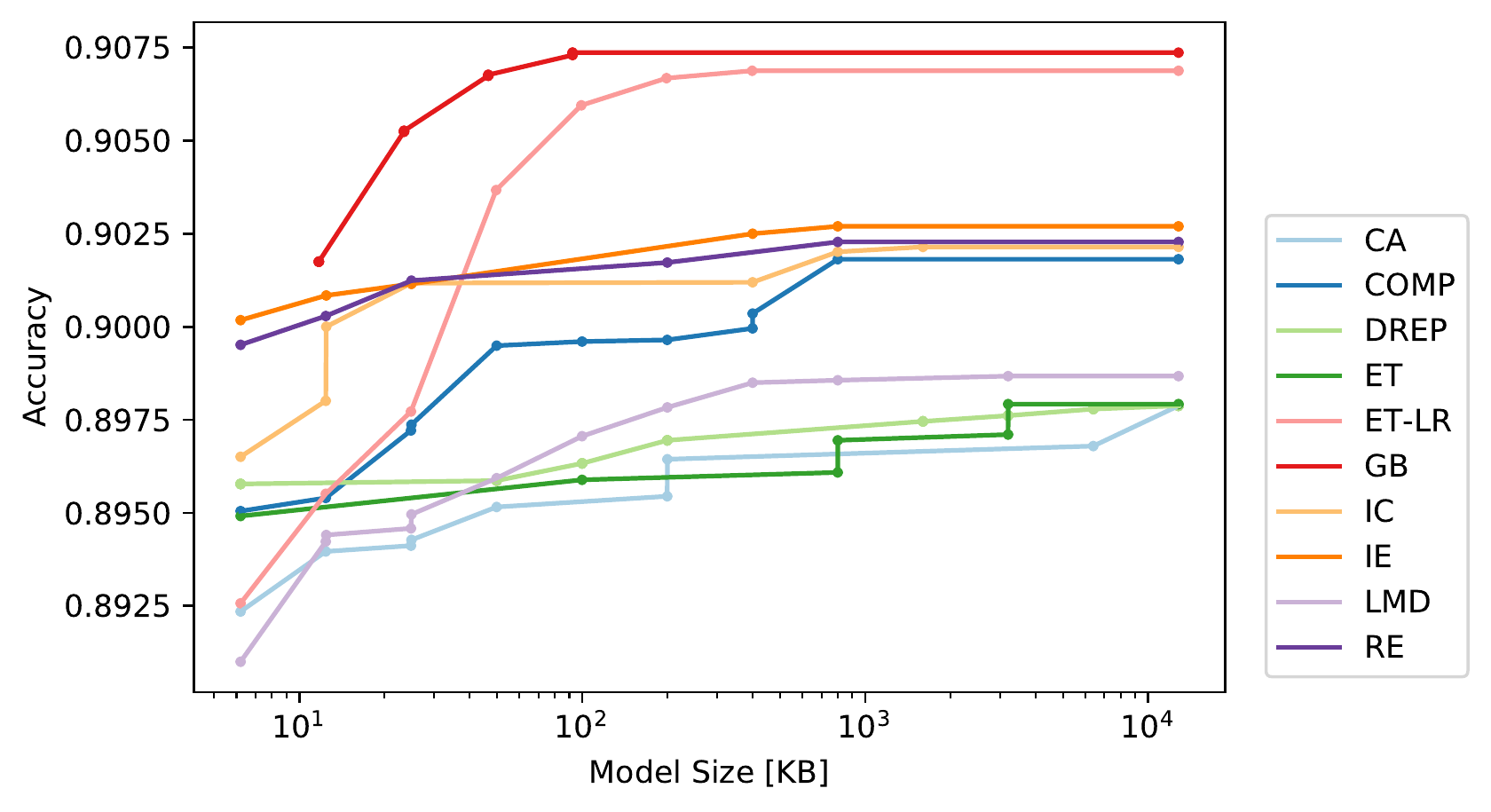}
\end{minipage}
\caption{(left) 5-fold cross-validation accuracy on the avila dataset. (right) 5-fold cross-validation accuracy on the bank dataset.}
\end{figure}

\begin{figure}[H]
\begin{minipage}{.49\textwidth}
    \centering
    \includegraphics[width=\textwidth,keepaspectratio]{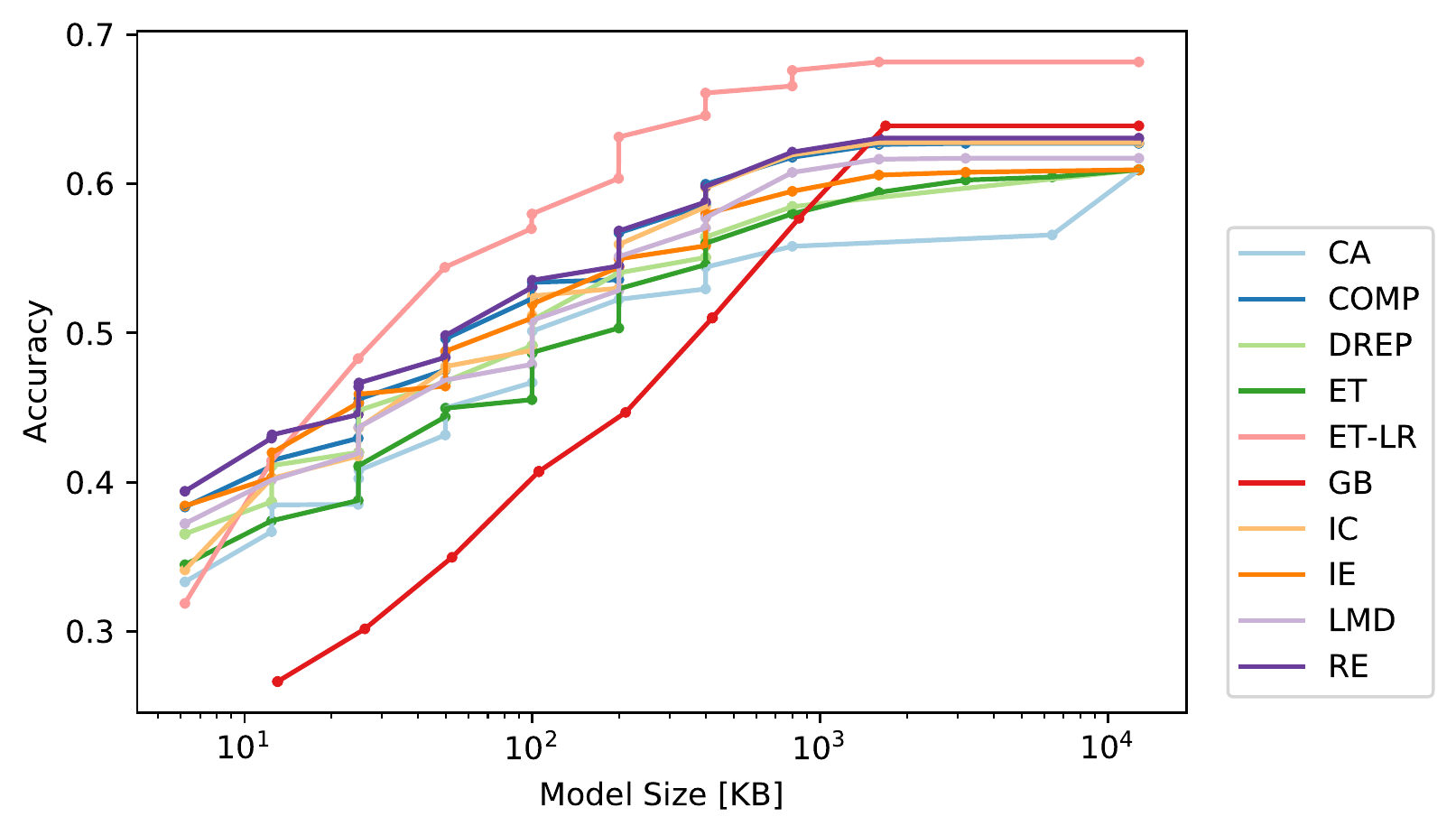}
\end{minipage}\hfill
\begin{minipage}{.49\textwidth}
    \centering 
    \includegraphics[width=\textwidth,keepaspectratio]{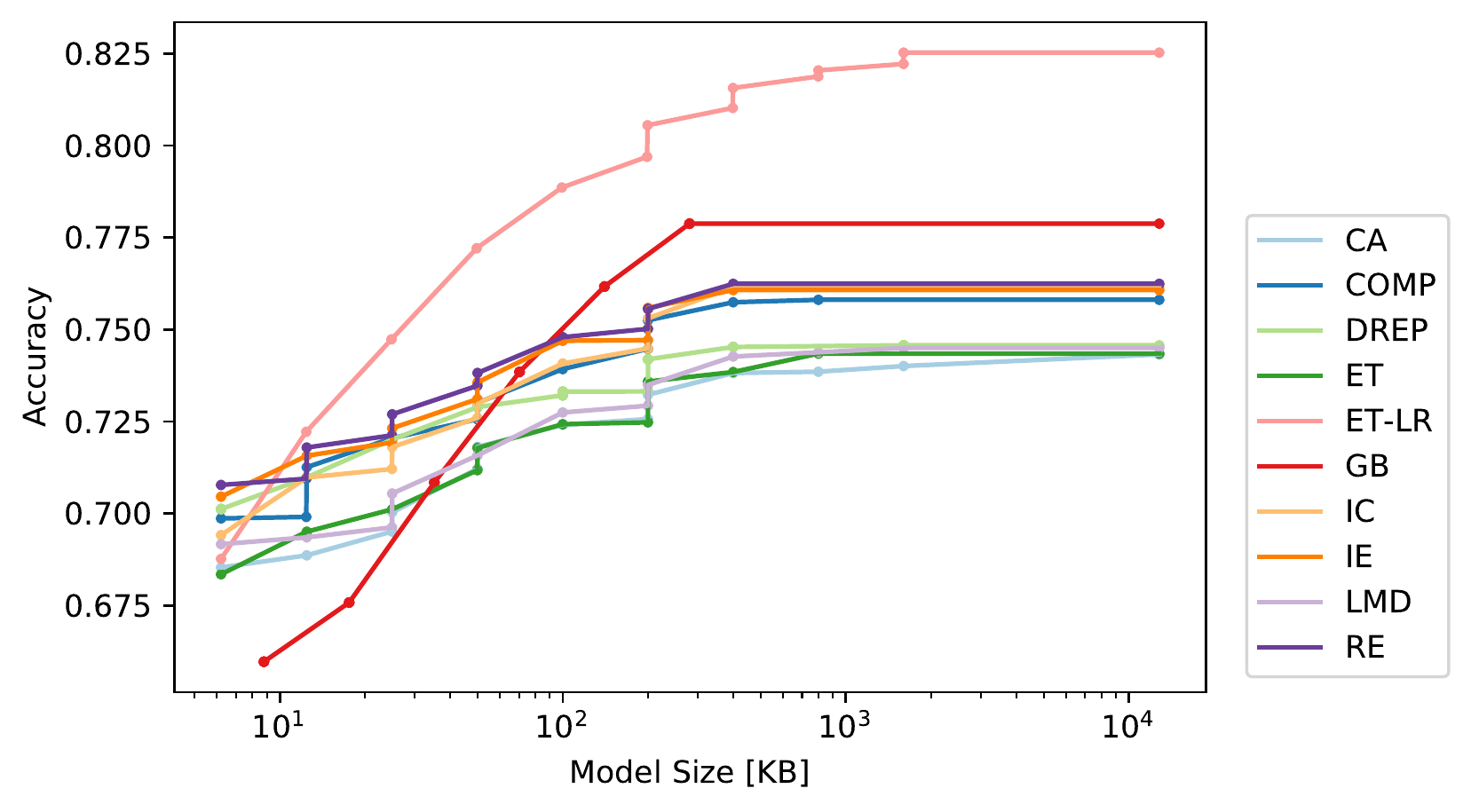}
\end{minipage}
\caption{(left) 5-fold cross-validation accuracy on the chess dataset. (right) 5-fold cross-validation accuracy on the connect dataset.}
\end{figure}

\begin{figure}[H]
\begin{minipage}{.49\textwidth}
    \centering
    \includegraphics[width=\textwidth,keepaspectratio]{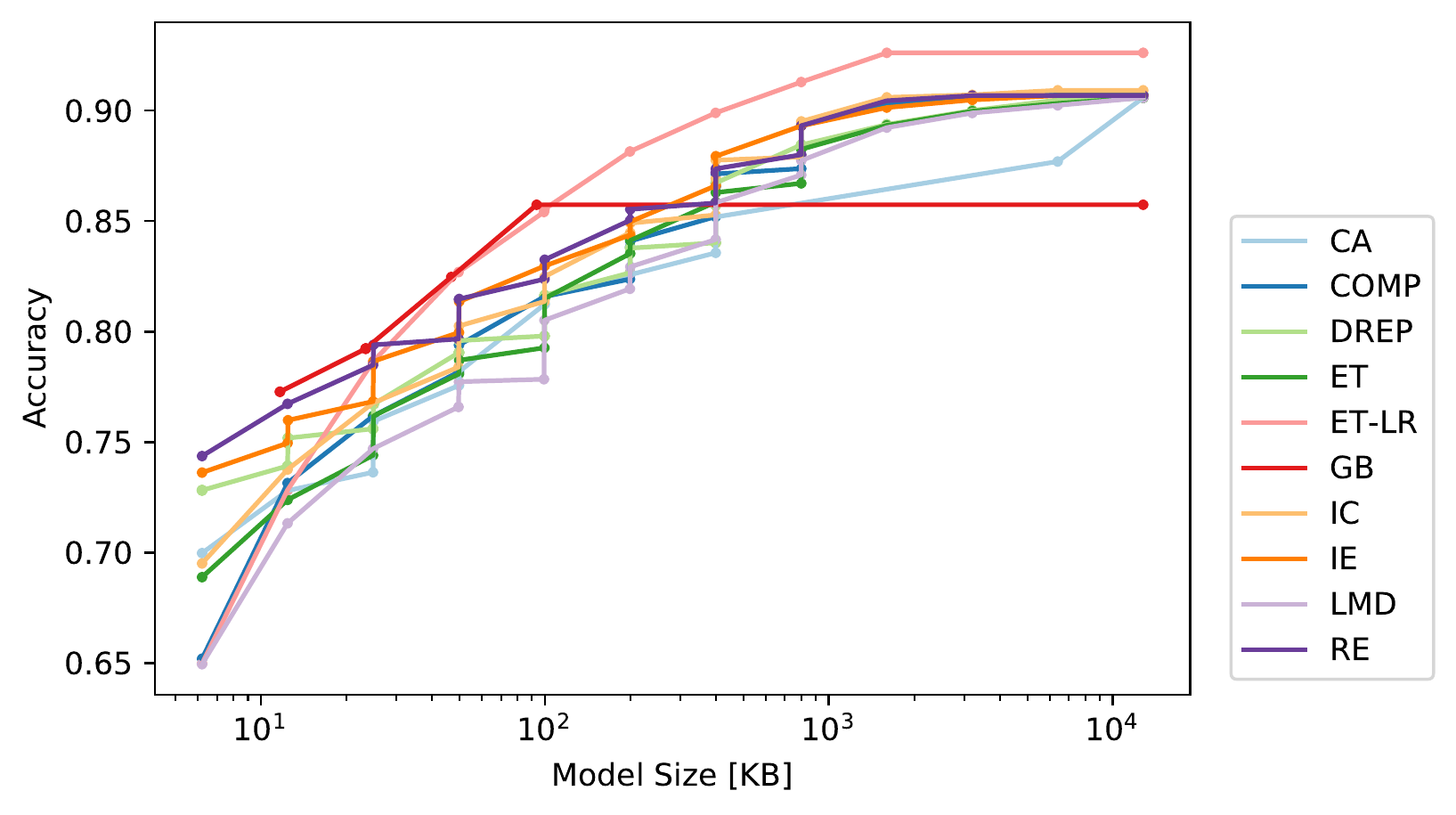}
\end{minipage}\hfill
\begin{minipage}{.49\textwidth}
    \centering 
    \includegraphics[width=\textwidth,keepaspectratio]{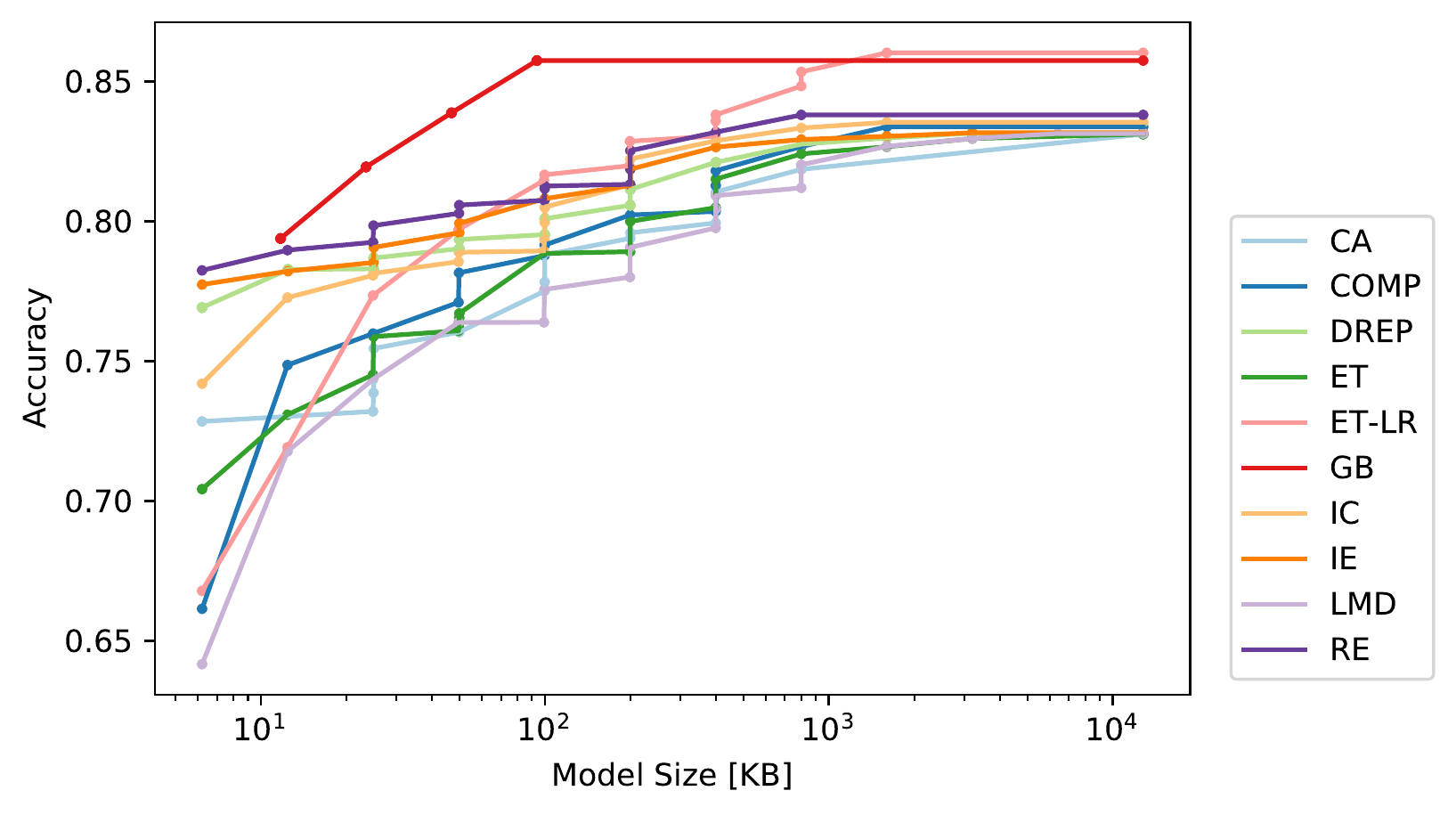}
\end{minipage}
\caption{(left) 5-fold cross-validation accuracy on the eeg dataset. (right) 5-fold cross-validation accuracy on the elec dataset.}
\end{figure}

\begin{figure}[H]
\begin{minipage}{.49\textwidth}
    \centering
    \includegraphics[width=\textwidth,keepaspectratio]{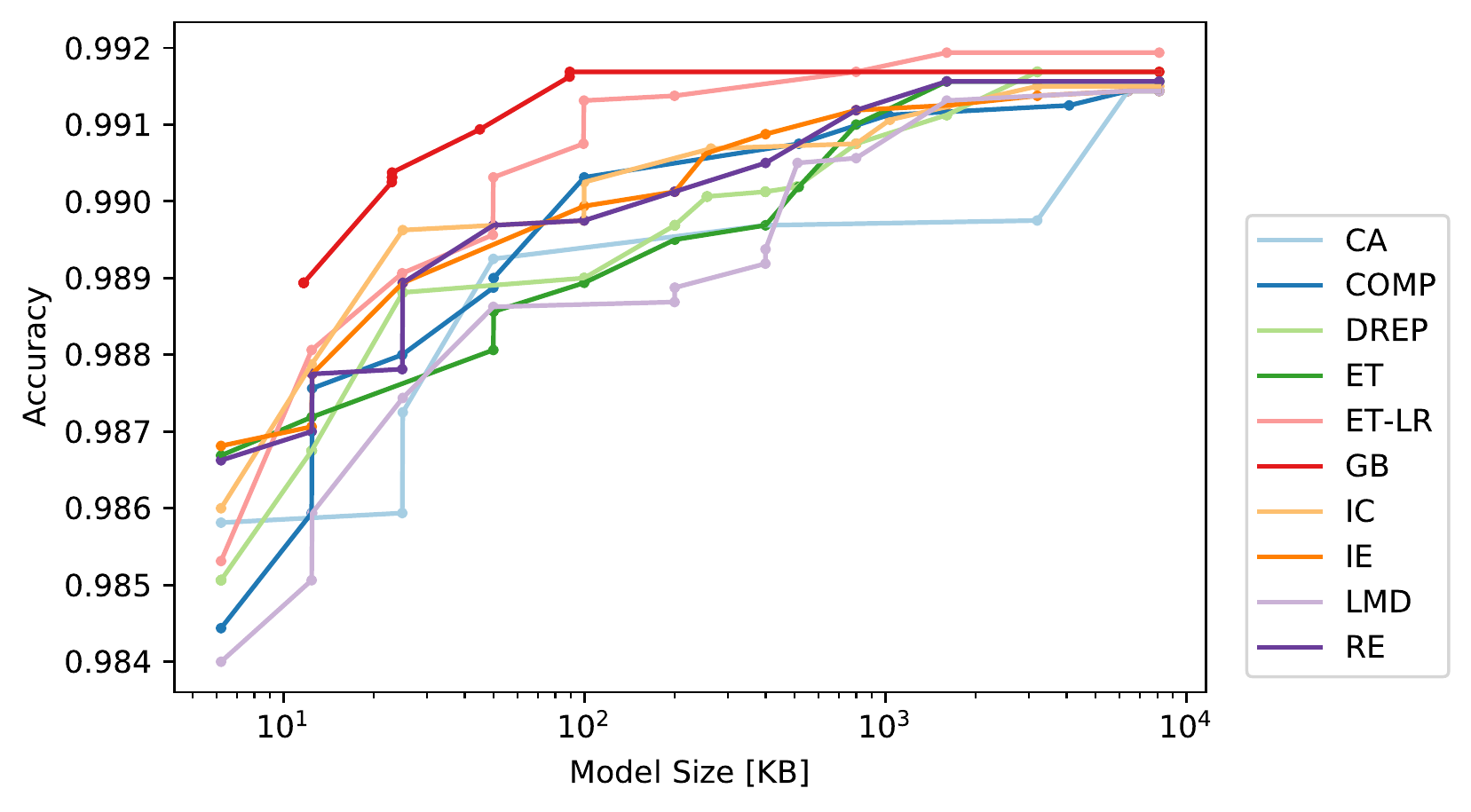}
\end{minipage}\hfill
\begin{minipage}{.49\textwidth}
    \centering 
    \includegraphics[width=\textwidth,keepaspectratio]{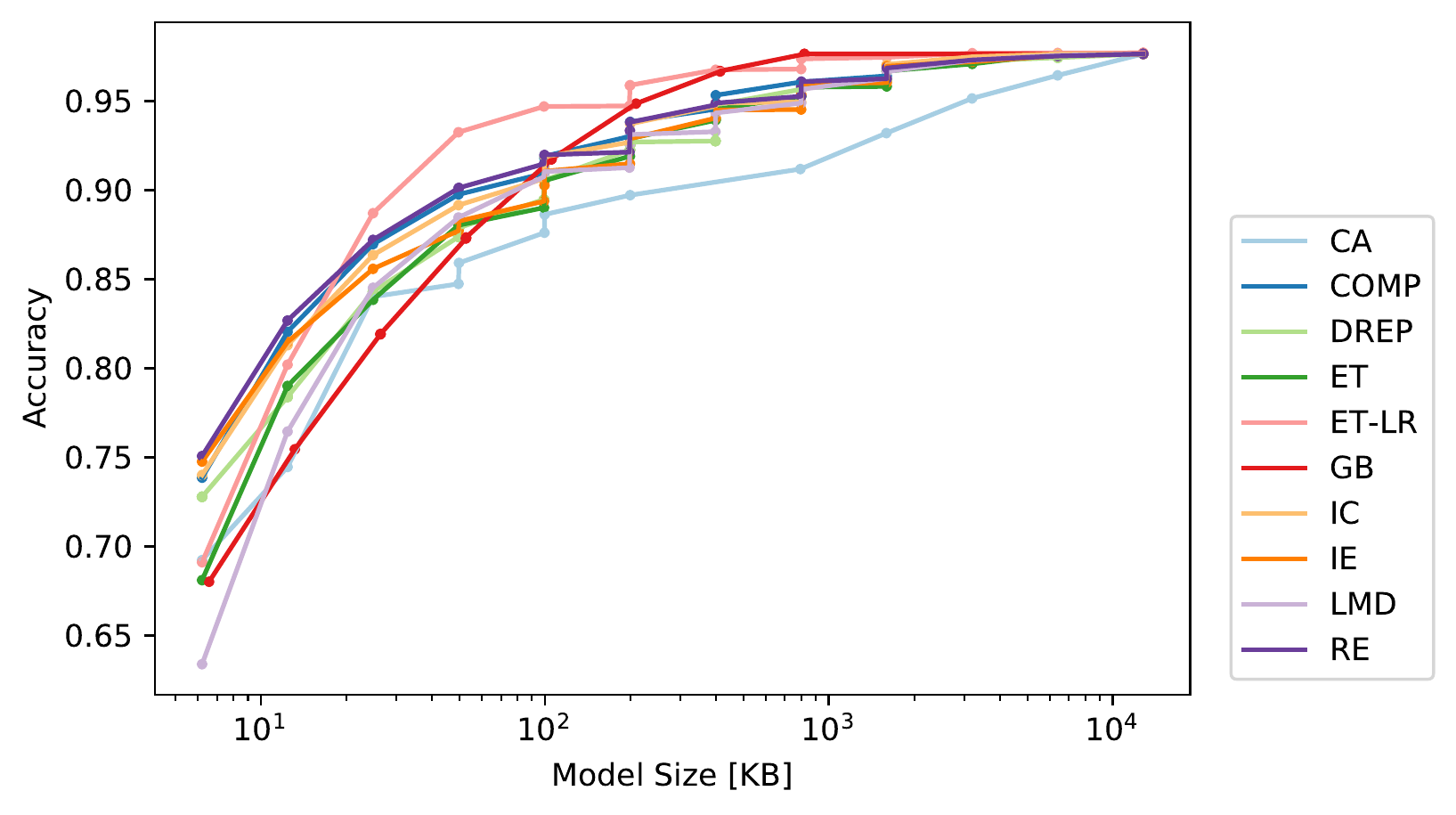}
\end{minipage}
\caption{(left) 5-fold cross-validation accuracy on the ida2016 dataset. (right) 5-fold cross-validation accuracy on the japanese-vowels dataset.}
\end{figure}

\begin{figure}[H]
\begin{minipage}{.49\textwidth}
    \centering
    \includegraphics[width=\textwidth,keepaspectratio]{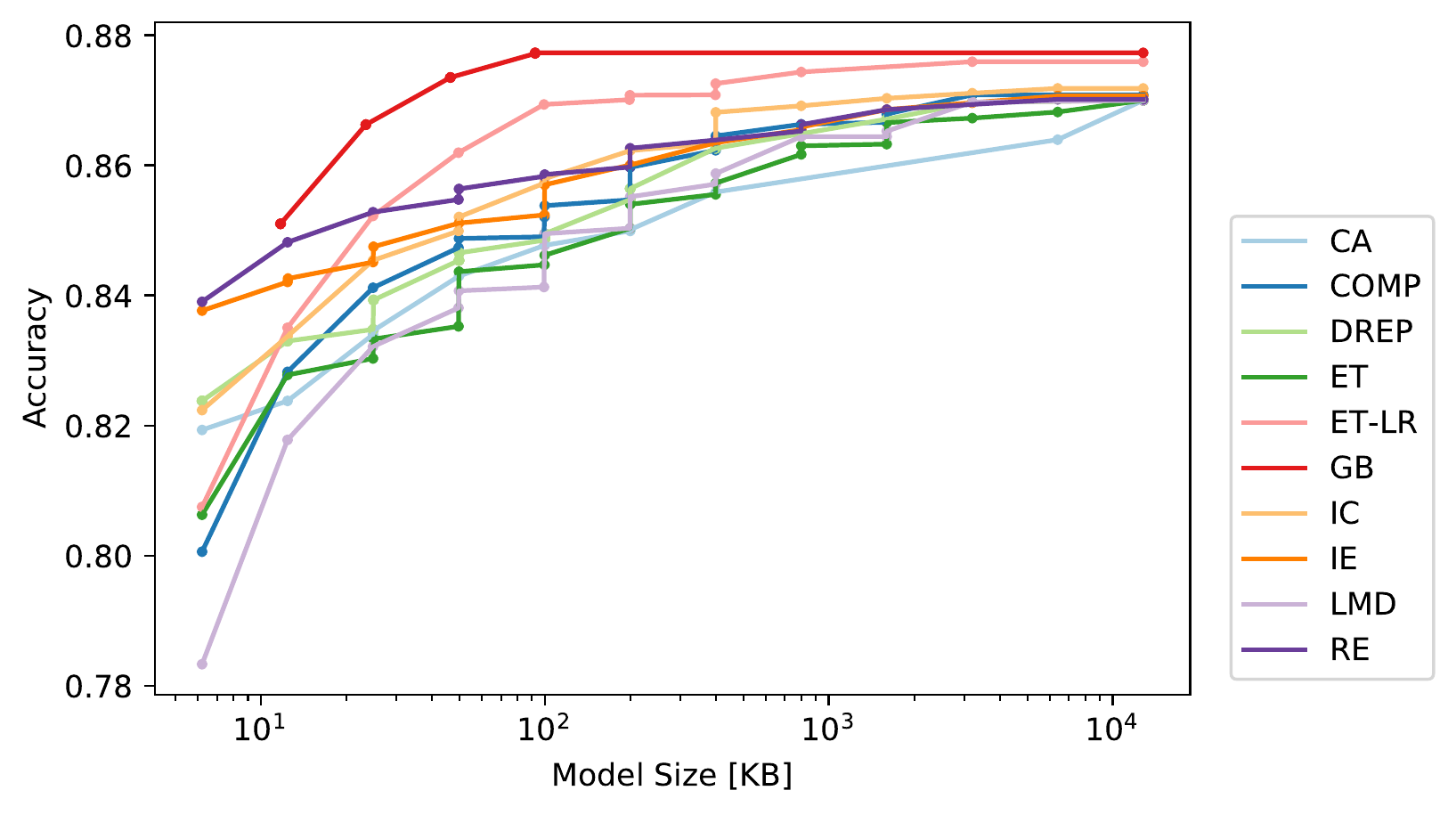}
\end{minipage}\hfill
\begin{minipage}{.49\textwidth}
    \centering 
    \includegraphics[width=\textwidth,keepaspectratio]{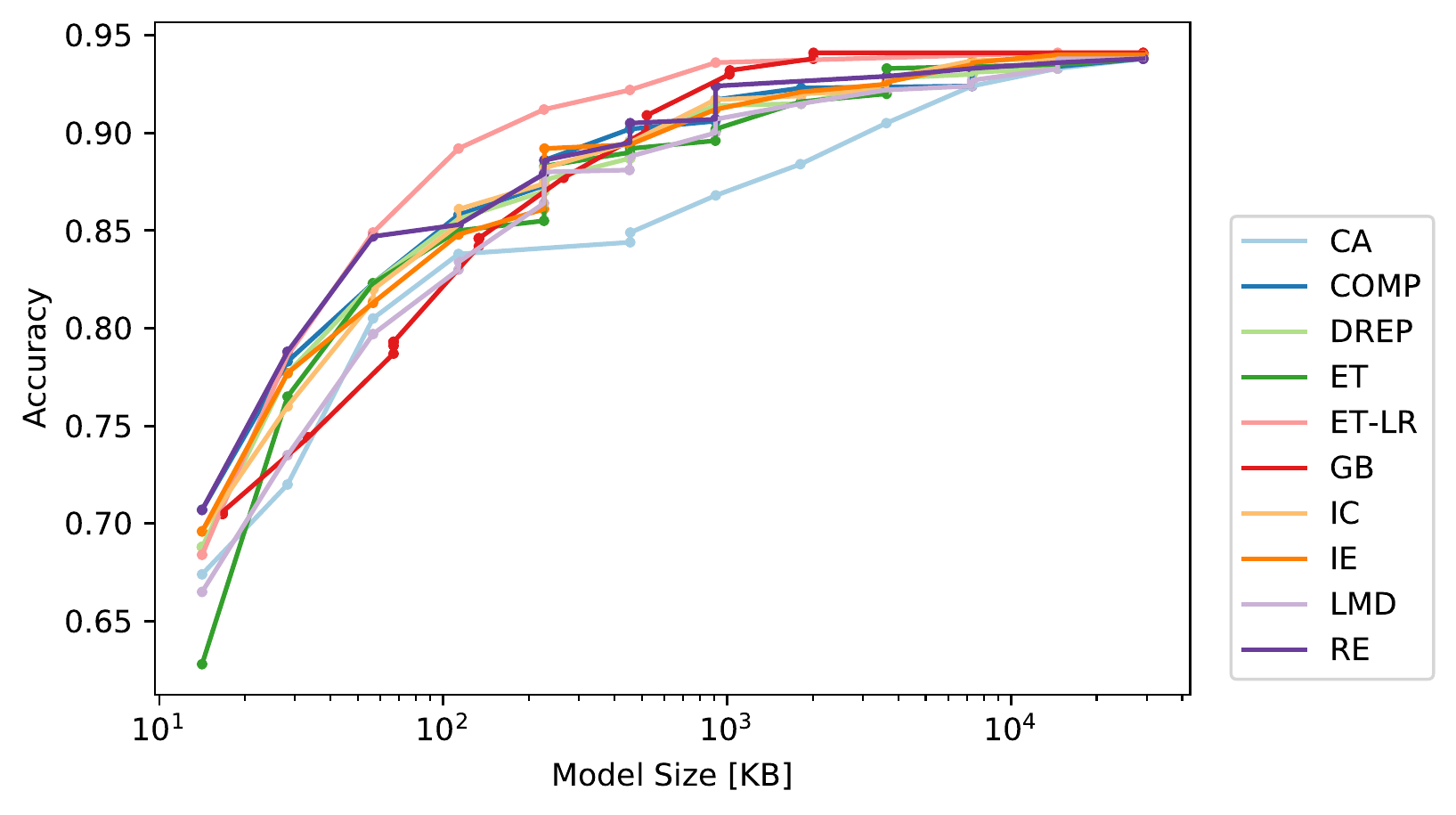}
\end{minipage}
\caption{(left) 5-fold cross-validation accuracy on the magic dataset. (right) 5-fold cross-validation accuracy on the mnist dataset.}
\end{figure}

\begin{figure}[H]
\begin{minipage}{.49\textwidth}
    \centering
    \includegraphics[width=\textwidth,keepaspectratio]{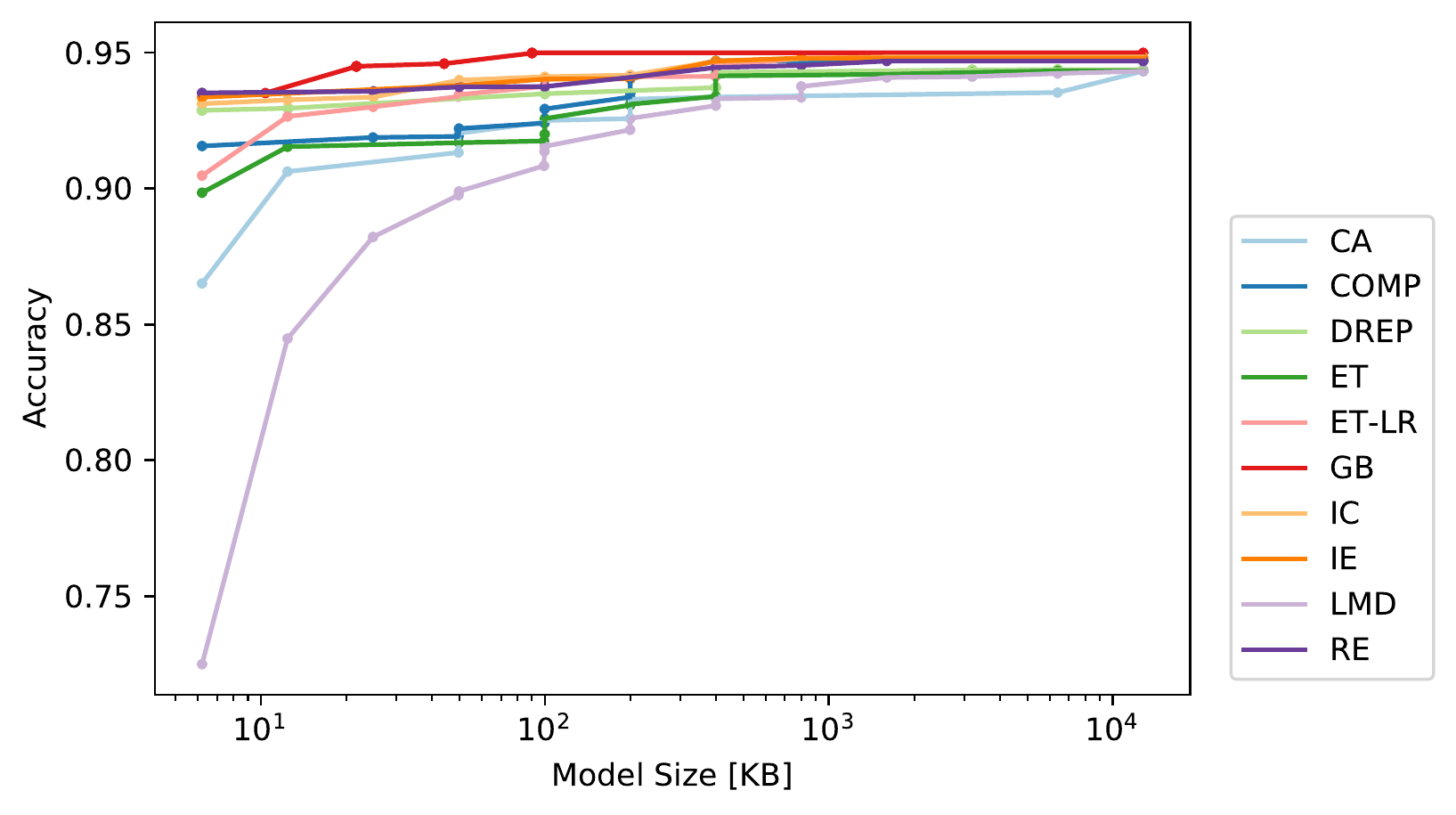}
\end{minipage}\hfill
\begin{minipage}{.49\textwidth}
    \centering 
    \includegraphics[width=\textwidth,keepaspectratio]{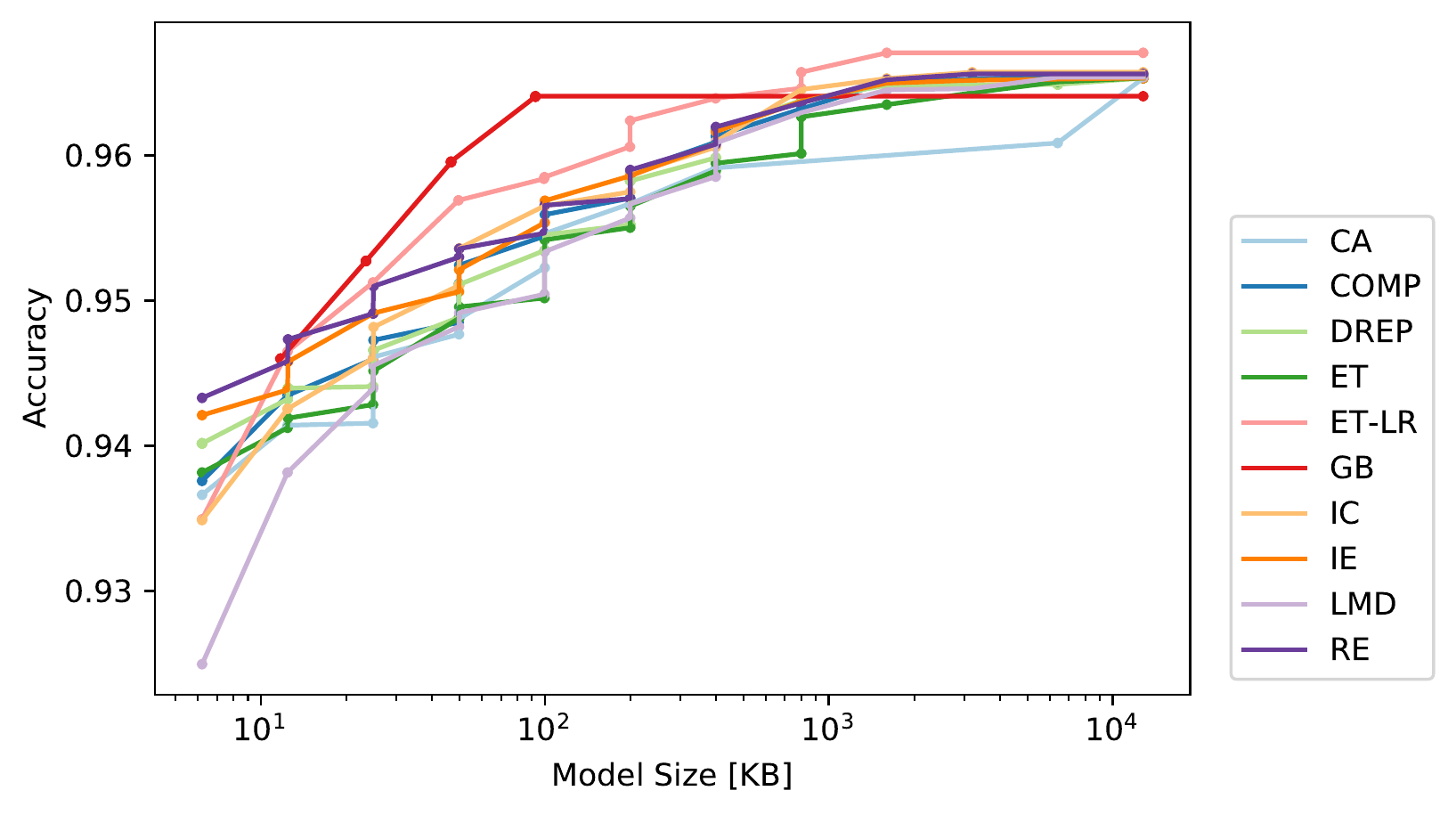}
\end{minipage}
\caption{(left) 5-fold cross-validation accuracy on the mozilla dataset. (right) 5-fold cross-validation accuracy on the nomao dataset.}
\end{figure}

\begin{figure}[H]
\begin{minipage}{.49\textwidth}
    \centering
    \includegraphics[width=\textwidth,keepaspectratio]{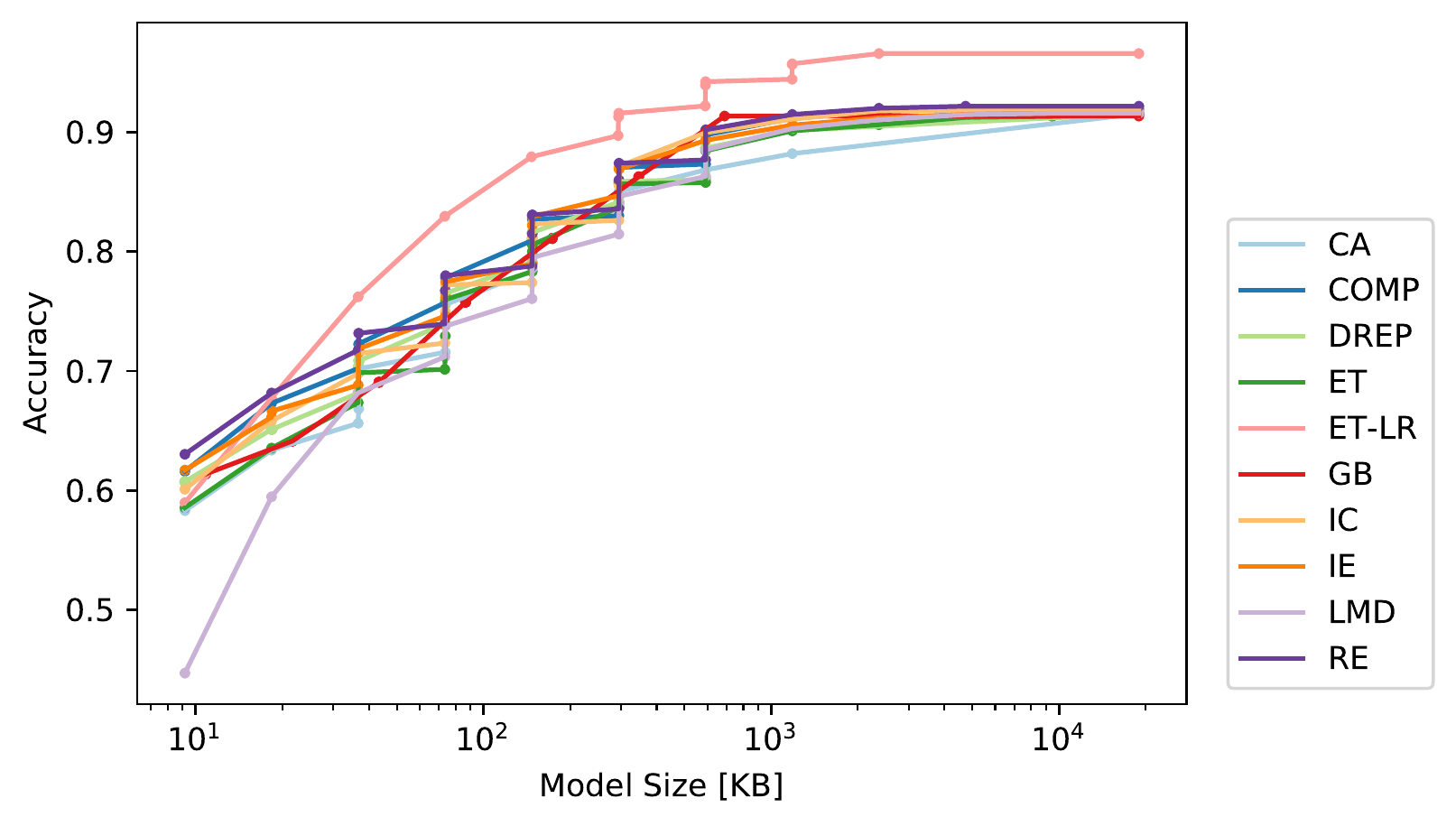}
\end{minipage}\hfill
\begin{minipage}{.49\textwidth}
    \centering 
    \includegraphics[width=\textwidth,keepaspectratio]{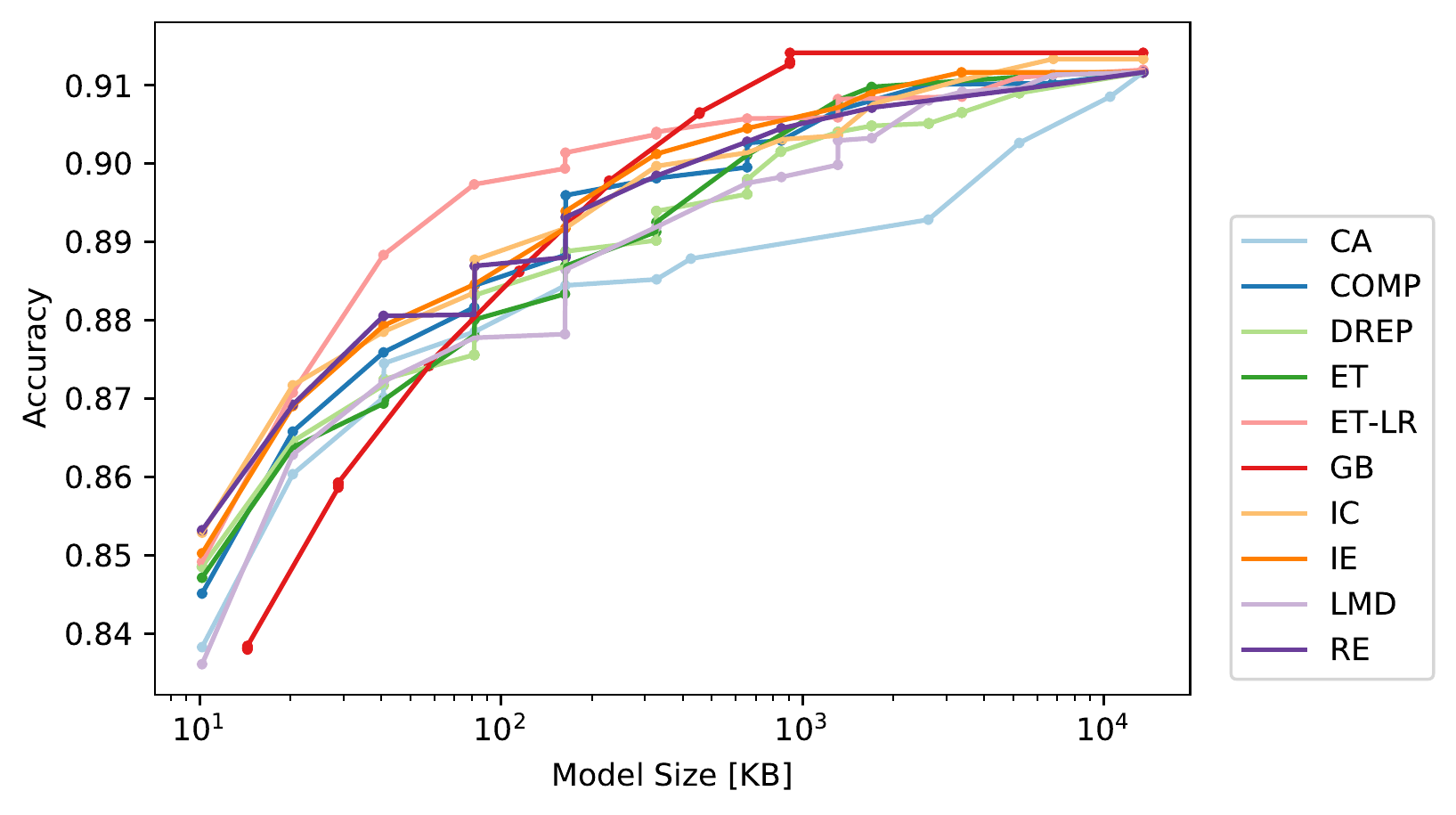}
\end{minipage}
\caption{(left) 5-fold cross-validation accuracy on the postures dataset. (right) 5-fold cross-validation accuracy on the satimage dataset.}
\end{figure}

\subsection{Accuracies under various resource constraints with a ExtraTrees Classifier}

\begin{table}
    \centering
    \resizebox{\textwidth}{!}{
    \input{figures/raw_ExtraTreesClassifier_32}
    }
    \caption{Test accuracies for models with a memory consumption below 32 KB for each method and each dataset averaged over a 5 fold cross validation. Rounded to the third decimal digit. Larger is better. The best method is depicted in bold.}
\end{table}

\begin{table}
    \centering
    \resizebox{\textwidth}{!}{
    \input{figures/raw_ExtraTreesClassifier_64}
    }
    \caption{Test accuracies for models with a memory consumption below 64 KB for each method and each dataset averaged over a 5 fold cross validation. Rounded to the third decimal digit. Larger is better. The best method is depicted in bold.}
\end{table}

\begin{table}
    \centering
    \resizebox{\textwidth}{!}{
    \input{figures/raw_ExtraTreesClassifier_128}
    }
    \caption{Test accuracies for models with a memory consumption below 128 KB for each method and each dataset averaged over a 5 fold cross validation. Rounded to the third decimal digit. Larger is better. The best method is depicted in bold.}
\end{table}

\begin{table}
    \centering
    \resizebox{\textwidth}{!}{
    \input{figures/raw_ExtraTreesClassifier_256}
    }
    \caption{Test accuracies for models with a memory consumption below 256 KB for each method and each dataset averaged over a 5 fold cross validation. Rounded to the third decimal digit. Larger is better. The best method is depicted in bold.}
\end{table}

\subsection{Area Under the Pareto Front with ExtaTrees Classifier}

\resizebox{\textwidth}{!}{
    \input{figures/aucs_ExtraTreesClassifier}
}

\begin{figure}
\centering
\includegraphics[width=\columnwidth, keepaspectratio]{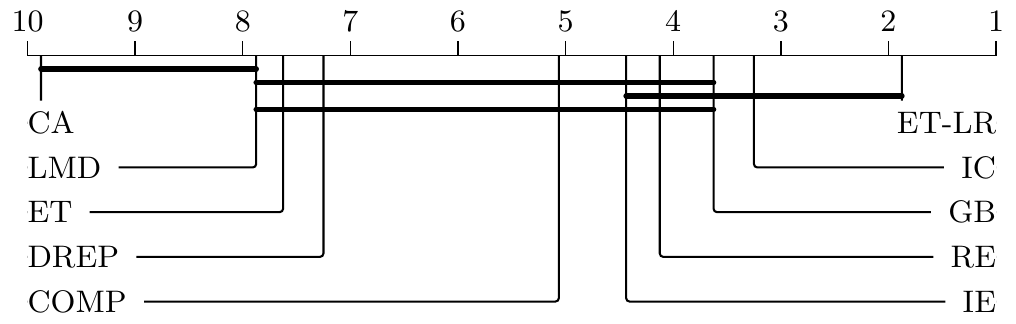}
\caption{Critical Difference Diagram for the normalized area under the Pareto front for different methods over multiple datasets. More to the right (lower rank) is better. Methods in connected cliques are statistically similar.}
\label{fig:cd_auc}
\end{figure}

\bibliography{literature}